\documentclass[11pt]{article}

\usepackage{acl}

\usepackage{times}
\usepackage{latexsym}
\usepackage[T1]{fontenc}
\usepackage[utf8]{inputenc}
\usepackage{microtype}
\usepackage{inconsolata}
\usepackage{graphicx}
\graphicspath{{figures/}}

\usepackage{amsmath}
\usepackage{amssymb}
\usepackage{amsfonts}
\usepackage{bm}
\usepackage{algorithm}
\usepackage{algpseudocode}

\usepackage{booktabs}
\usepackage{multirow}
\usepackage{multicol}
\usepackage{array}

\usepackage{xspace}
\usepackage[table]{xcolor}

\newcommand{\method}{CC-OPD\xspace}

\newcommand{\KL}{\mathrm{KL}}
\newcommand{\studpolicy}{\pi_\theta}
\newcommand{\teacher}{P_T}

\title{Counterfactual Constraint-Conditioned On-Policy Distillation\\
for Multi-Constraint Instruction Following}

\author{
  \textbf{Yanzhao Zheng, Yuanqiang Yu, Tianze Xu, Chao Ma, Zhentao Zhang}\\[-1pt]
  \textbf{Jihuai Zhu, Baohua Dong\thanks{Corresponding author.}, Hangcheng Zhu, Ruohui Huang}\\[2pt]
  Alibaba Group, Hangzhou, China\\
  \texttt{\{zhengyanzhao.zyz,yuyuanqiang.yyq,xutianze.xtz\}@alibaba-inc.com}\\[-1pt]
  \texttt{\{mc524716,zhangzhentao.zzt,zhujihuai.zjh\}@alibaba-inc.com}\\[-1pt]
  \texttt{\{baohua.dbh,linran.lr09,wentong\}@alibaba-inc.com}
}

\begin{document}
\maketitle
\begingroup
\renewcommand{\thefootnote}{}
\footnotetext{Code: \href{https://github.com/zhengyanzhao1997/cc-opd}{\texttt{github.com/zhengyanzhao1997/cc-opd}}}
\endgroup

\begin{abstract}
Multi-constraint instruction following requires a model to respond to a query under many simultaneously active constraints. Even strong instruction-tuned models still routinely violate some of them. Existing approaches either augment supervision with sequence- or token-level RL rewards from external verifiers or learned graders, or use on-policy distillation (OPD) against a single full-context teacher whose probability mass becomes diluted as more constraints become simultaneously active. We propose \method{} (Counterfactual Constraint-Conditioned On-Policy Distillation), which inverts the standard supervision-generation direction in distillation. Rather than enriching the teacher with information beyond what the student sees, \method{} ablates each constraint from the teacher's conditioning in turn, and constructs the per-constraint signal from the resulting per-token probability differentials. The resulting per-token leave-one-out log-likelihood shifts are summed, clipped, and added to the vanilla OPD reward as a token-level shaping term. All shaping terms are obtained from the frozen teacher, without an external verifier during distillation, and the reward equals vanilla OPD wherever the aggregate shift is zero. Across two Qwen model pairs and seven benchmarks, \method{} achieves the highest average among all evaluated student-training methods. A 1.5B student trained with \method{} surpasses its own 7B RL-trained teacher on the MulDimIF benchmark.

\end{abstract}

\section{Introduction}
\label{sec:intro}

\begin{figure*}[t]
  \centering
  \includegraphics[width=0.78\textwidth]{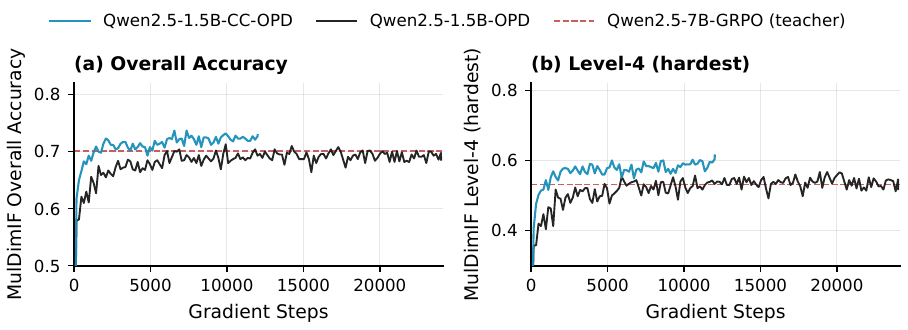}
  \caption{\textbf{Student surpasses its own teacher.} MulDimIF training trajectories on the Qwen2.5-1.5B pair~\citep{qwen2024qwen25} vs.\ gradient steps. \method{} (cyan) crosses the 7B teacher's accuracy ceiling (red dashed) on both \textbf{(left)} the overall score and \textbf{(right)} the hardest Level-4 split, while vanilla sampled-token OPD (black) plateaus below the teacher.}
  \label{fig:teaser}
\end{figure*}

Modern LLM assistants are routinely deployed under user instructions that simultaneously specify multiple requirements such as a target format, a length cap, a style register, content to include or avoid, a safety policy, or a citation convention. Each request bundles these into a single natural-language prompt, and the system is judged by whether the response satisfies all of them at once. Recent benchmarks make clear that this multi-constraint regime is far from solved, and when a prompt carries several simultaneously active constraints, even strong instruction-tuned models routinely violate some of them~\citep{zhou2023ifeval, ifbench2025, muldimif2025}. The harder and more numerous the constraints, the larger this gap becomes, and closing it is now a central post-training challenge for instruction-following LLMs.

Two lines of post-training research are commonly applied to multi-constraint instruction following, but each leaves a structural piece of the constraint signal unused. Rubric- or checklist-based reinforcement learning methods augment training with externally graded constraint signals. They elicit constraints from the prompt, grade each criterion with a verifier or LLM judge, and aggregate criterion-wise scores into a scalar or token-level RL reward~\citep{fang2026ropd, xu2026rtt, gunjal2025rar, viswanathan2025rlcf, zhou2025ruscarl}. On-policy distillation (OPD)~\citep{agarwal2024gkd, gu2024minillm, ko2024distillm, li2026rethinkopd}, in contrast, is a general post-training framework that distills the student directly from a frozen teacher's token distribution. When applied to multi-constraint instruction following, OPD conditions the teacher on the full instruction, which collapses the constraint set into a single joint conditioning context. As more constraints become simultaneously active, the full-context teacher must spread its probability mass across many requirements, and an individual constraint's contribution to the per-token distribution becomes increasingly difficult to disentangle. We call this effect \emph{constraint dilution}. Neither line extracts per-constraint information directly from the teacher.

We propose \textbf{\method{}} (Counterfactual Constraint-Conditioned On-Policy Distillation), which derives constraint-indexed shaping signals through counterfactual teacher scoring while keeping the OPD KL backbone unchanged. The student remains conditioned on the full instruction at both training and deployment. For each student rollout, the frozen teacher scores the same response under the full instruction and under each leave-one-out context $C \setminus \{c_i\}$. The resulting log-likelihood differences yield per-token, per-constraint signals $\delta_i(t)$, which we sum, clip, and add to the vanilla sampled-token OPD reward. Because each contrast is computed before aggregation, the resulting $|C| \times T$ signal retains the identity of individual constraints; wherever the aggregate shift is zero, the reward equals vanilla sampled-token OPD.

A 1.5B student trained with \method{} surpasses its own 7B RL-trained teacher on multi-dimensional constraint following by $+2.8$\,pp on the MulDimIF overall score and $+6.3$\,pp on its hardest difficulty split (Figure~\ref{fig:teaser}). This is a rare outcome for OPD~\citep{li2026rethinkopd}, where students typically recover only part of the teacher--student performance gap rather than crossing it. Across both Qwen settings and seven instruction-following benchmarks, \method{} attains the highest average among all evaluated student-training methods, including recent OPD variants.

Our contributions are threefold. First, we introduce \method{}, a leave-one-out teacher-scoring method that produces a token-level signal for each constraint on student rollouts (\S\ref{sec:method}). Second, we characterize the resulting signal algebraically: the LOO sum assigns cardinality-dependent weights to M\"obius interaction terms, and a zero aggregate shift leaves the vanilla sampled-token OPD reward unchanged (\S\ref{sec:method}; Appendix~\ref{app:mobius}). Third, we evaluate \method{} against SFT, verifier-based RL, vanilla OPD, Teacher-TopK RKL/SRKL, EOPD, ExOPD, and prompt-augmentation variants, together with counterfactual-design and training-efficiency analyses (\S\ref{sec:main_results}).

\begin{figure*}[!t]
  \centering
  \includegraphics[width=\textwidth]{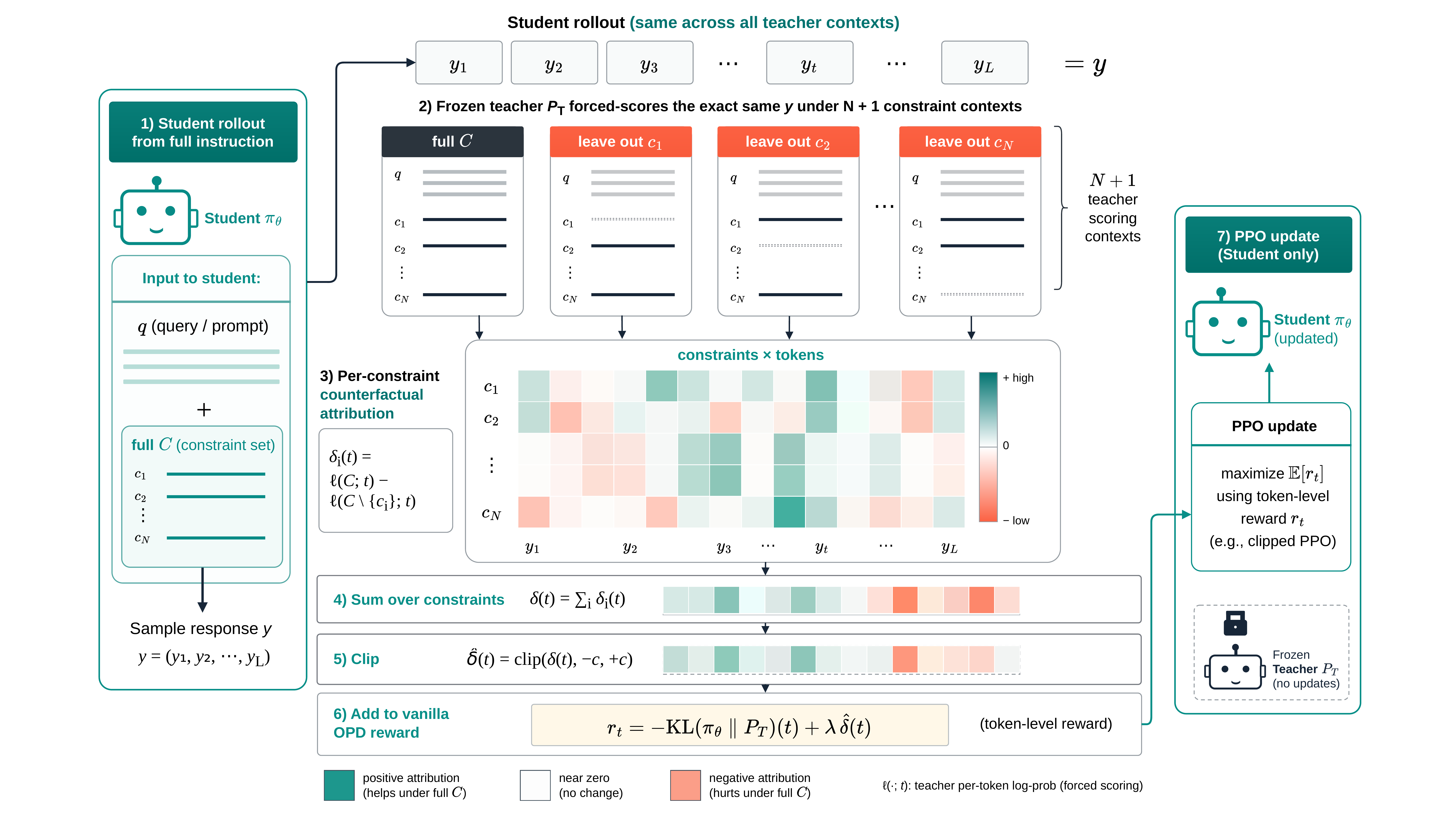}
  \caption{\textbf{\method{} pipeline.} On the same student rollout $y$, the frozen teacher $\teacher$ is forced-scored under $|C|+1$ counterfactual prompts (steps 1--2), namely the full set $C$ and each leave-one-out context $C \setminus \{c_i\}$. The resulting per-constraint shifts $\delta_i(t)$ (step 3) are summed, symmetrically clipped, and additively composed with the vanilla sampled-token OPD reward (steps 4--6) to form the token-level reward $r_t$ used in a standard ratio-clipped PPO update (step 7). Only the prompt varies across teacher contexts, while $y$ is shared.}
  \label{fig:pipeline}
\end{figure*}

%
%
\section{Preliminaries}
\label{sec:prelim}

\subsection{Multi-Constraint Instruction Following}
\label{sec:prelim:task}

We consider \emph{multi-constraint instruction following}, in which a model must produce a response $y$ to a user query $q$ that simultaneously satisfies a set of natural-language constraints $C = \{c_1, \ldots, c_N\}$. Each $c_i$ is a self-contained sentence describing one requirement (length, formatting, taboos, tone, citation format, and similar constraints). The query and constraint set are concatenated into a single natural-language instruction $\mathsf{inst}(q, C)$ in which the constraints appear as inline clauses interleaved with the query. Throughout we use $C$ for the full constraint set, $|C|$ for its cardinality, and $c_i$ for the $i$-th constraint; in realistic scenarios $|C|$ is rarely small, with typical samples carrying several simultaneously active constraints.

\subsection{On-Policy Distillation with Sampled-Token PPO}
\label{sec:prelim:opd}

On-policy distillation~(OPD) distills a frozen teacher $\teacher$ into a student $\studpolicy$ by minimizing a token-level divergence on sequences sampled from the student itself. Given an input $(q, C)$ and a student rollout $y = (y_1, \ldots, y_L) \sim \studpolicy(\cdot \mid q, C)$, the canonical OPD objective is
\begin{equation}
\mathcal{L}_{\mathrm{OPD}} \;=\; \mathbb{E}_{y \sim \studpolicy}\sum_{t=1}^{L}\KL\!\left(\studpolicy(\cdot \mid h_t) \,\big\|\, \teacher(\cdot \mid h_t)\right),
\label{eq:opd_kl}
\end{equation}
where $h_t = (q, C, y_{<t})$ is the shared decoding state at step $t$. Direct minimization of~\eqref{eq:opd_kl} requires materializing the full per-token vocabulary distribution of both models, which is memory-bound at large vocabulary sizes.

We therefore adopt the sampled-token PPO~\citep{schulman2017ppo} formulation of OPD, in which the per-token KL is replaced by its single-sample stochastic estimate at the realized student token,
\begin{equation}
\widehat{\KL}(t) \;=\; \log\studpolicy(y_t \mid h_t) - \log\teacher(y_t \mid h_t),
\label{eq:opd_sample}
\end{equation}
and the resulting scalar $r^{\mathrm{OPD}}_t = -\widehat{\KL}(t)$ is treated as a token-level reward in standard ratio-clipped PPO. Only one teacher and one student log-probability per token need to be computed. Recent work shows that under appropriately tuned training settings the sampled-token form attains end-task performance comparable to top-$k$ OPD variants~\citep{li2026rethinkopd}.

\section{\method{}}
\label{sec:method}

The sampled-token OPD reward of Eq.~\eqref{eq:opd_sample} treats the constraint set $C$ as a single conditioning context, so every constraint contributes to $\teacher$ jointly and the per-token signal cannot tell whether $y_t$ was supported by $c_1$, by $c_7$, or by no constraint at all. As $|C|$ grows, this constraint dilution (\S\ref{sec:intro}) worsens, since the teacher must spread its probability mass across many simultaneously active requirements and individual contributions become difficult to disentangle from the joint signal. \method{} restores per-constraint resolution while keeping the student conditioning unchanged. The student remains conditioned on the full $C$ at both training and deployment, while the frozen teacher is queried $|C|+1$ times on the same student rollout, once under the full instruction and once per constraint with that constraint removed from its natural-language context. The resulting log-likelihood shifts form a dense, per-token, per-constraint signal that is aggregated into a token-level shaping term and added to the vanilla OPD reward (Figure~\ref{fig:pipeline}).

\subsection{Per-Constraint Counterfactual Teacher Scoring}
\label{sec:method:cf}

Let $\mathsf{inst}(q, C)$ denote the natural-language instruction associated with sample $(q, C)$. For each $c_i$, we construct a counterfactual prompt by excising the natural-language span corresponding to $c_i$ from $\mathsf{inst}(q, C)$. The same student rollout $y$ is then forced-scored by the teacher under each of these $|C|+1$ prompts; let
\begin{equation}
\ell(A; t) \;=\; \log\teacher\!\left(y_t \;\big|\; q,\, A,\, y_{<t}\right), \quad A \subseteq C,
\label{eq:teacher_logp}
\end{equation}
denote the resulting log-probability under any constraint subset $A$. The \emph{per-constraint counterfactual signal} for constraint $c_i$ at token $t$ is the leave-one-out log-likelihood shift
\begin{equation}
\delta_i(t) \;=\; \ell(C;\, t) \;-\; \ell\!\left(C\setminus\{c_i\};\, t\right).
\label{eq:delta_i}
\end{equation}
A positive $\delta_i(t)$ means the teacher assigns more probability to the realized student token \emph{because} $c_i$ is in the prompt, identifying a constraint-supporting token. A value near zero indicates that $c_i$ is irrelevant to that token, and a negative value indicates that $c_i$ suppresses $y_t$, either because the token violates $c_i$ or because removing $c_i$ shifts the teacher's prior in an unrelated direction. Because the student rollout $y$ is held fixed across all $|C|+1$ teacher passes, the contrast in~\eqref{eq:delta_i} isolates the marginal effect of $c_i$ from the joint effect of the remaining constraints, without paying for extra rollouts.

\paragraph{Why leave-one-out rather than single-constraint instantiation.}
A more naive design might localize each constraint's contribution by querying the teacher with that constraint in isolation, computing $\delta_i(t) = \ell(\{c_i\};\, t) - \ell(\emptyset;\, t)$. This is conceptually equivalent to constructing $N$ single-constraint teachers and reading off each one's standalone effect. Such a design discards all inter-constraint interactions, since in multi-constraint settings the way the teacher enforces $c_i$ depends on which other constraints are simultaneously active, and these dependencies cannot be recovered from a single-constraint forward pass. We evaluate this alternative empirically in \S\ref{sec:ablation} (LOO vs.\ NAO variants).

\subsection{Counterfactual Reward Shaping}
\label{sec:method:reward}

The signal $\delta_i(t)$ is per-token and per-constraint, but PPO consumes a single scalar advantage per token. We aggregate over constraints by summation,
\begin{equation}
\delta(t) \;=\; \sum_{i=1}^{|C|}\delta_i(t),
\label{eq:delta_agg}
\end{equation}
and apply a symmetric hard clip with threshold $c$ to prevent any single outlier token (typically an end-of-sequence symbol, a rare lexical item, or a numerically pathological context) from dominating the policy gradient,
\begin{equation}
\hat{\delta}(t) \;=\; \mathrm{clip}\!\left(\delta(t),\; -c,\; +c\right).
\label{eq:delta_clip}
\end{equation}
The \method{} token-level reward composes the clipped counterfactual signal additively with the vanilla OPD reward of~\eqref{eq:opd_sample},
\begin{equation}
r_t \;=\; \underbrace{-\widehat{\KL}(t)}_{\text{vanilla OPD}} \;+\; \underbrace{\lambda\,\hat{\delta}(t)}_{\text{\method{} shaping}},
\label{eq:cc_reward}
\end{equation}
and the resulting $r_t$ enters a standard ratio-clipped PPO update. Throughout the paper we use $\lambda = 2.0$ and $c = 5.0$, whose sensitivities are analyzed in Appendix~\ref{app:lambda_clip}. Algorithm~\ref{alg:cc_opd} in Appendix~\ref{app:impl} gives the complete training step as drop-in pseudocode.

\paragraph{Reduction to vanilla OPD at zero aggregate shift.}
When the aggregate counterfactual shift satisfies $\delta(t)=\sum_i\delta_i(t)=0$, Eq.~\eqref{eq:delta_clip} gives $\hat{\delta}(t)=0$, and Eq.~\eqref{eq:cc_reward} reduces exactly to vanilla sampled-token OPD at that token. When $\delta(t)$ is small but nonzero and lies within the clipping range, clipping leaves it unchanged, so \method{} differs from vanilla OPD by the correspondingly small additive term $\lambda\delta(t)$. In all cases, Eq.~\eqref{eq:delta_clip} bounds the absolute shaping contribution by $\lambda c$ per token.

\paragraph{Backbone-agnostic composition.}
Equation~\eqref{eq:cc_reward} writes the \method{} reward as the sum of an OPD KL term and the shaping term $\lambda\hat{\delta}(t)$. Only the first term depends on the form of the OPD KL estimator; the shaping term is constructed entirely from teacher-side forced-scoring log-likelihoods and is independent of the student's KL form. The shaping term can therefore be added to OPD objectives that expose a per-token reward signal; our experiments use sampled-token reverse-KL.

A M\"obius decomposition of~\eqref{eq:delta_i} and an exponential-tilt interpretation of~\eqref{eq:cc_reward} are given in Appendix~\ref{app:mobius}. Token-level reward maps are visualized in Appendix~\ref{app:heatmap}, and alternative aggregations of $\delta_i(t)$ are ablated in Appendix~\ref{app:aggregation}.

\section{Experiments}
\label{sec:experiments}

\subsection{Experimental Setup}
\label{sec:setup}

\paragraph{Model pairs.}
We evaluate \method{} on two student--teacher pairs spanning two model families and two student sizes. The Qwen2.5 pair uses Qwen2.5-1.5B-Instruct~\citep{qwen2024qwen25} as the student and a Qwen2.5-7B teacher obtained by training Qwen2.5-7B-Instruct with GRPO-IR (instruction-level reward; see Baselines below) on the HIR-16K instruction-following corpus~\citep{zhang2025hir}. The Qwen3 pair uses Qwen3-4B-Instruct-2507~\citep{yang2025qwen3} as the student and a same-size teacher obtained by training Qwen3-4B-Instruct with GRPO-IR on the same corpus. Each teacher is GRPO-IR-tuned to obtain a strong instruction-following reference before distillation, following the standard practice of task-adapting the teacher prior to OPD. This setup provides a stringent evaluation because the student is compared against a teacher that has already been adapted to the target instruction-following corpus.

\paragraph{Baselines.}
We compare against several training paradigms, each instantiated for both model families. \textbf{SFT} fine-tunes the student on teacher-generated responses. Two \textbf{GRPO}~\citep{shao2024deepseekmath} variants instantiate rubric-RL: \textbf{GRPO-IR}, matching RL-IR of~\citet{peng2025rl-ir, ifbench2025}, uses an instruction-level reward ($r{=}1$ iff every constraint is satisfied, else $0$), and \textbf{GRPO-CR}, matching RL-CR of~\citet{qi2025rl-cr, ifbench2025}, uses a constraint-level mean reward. Vanilla sampled-token \textbf{OPD} is the direct distillation baseline. Three \textbf{OPD-augmentation} variants (\textbf{OPD-aug-LOO/Rand/SC}) enlarge the OPD training set with leave-one-out, random-subset, and single-constraint counterfactual prompts respectively. We additionally include two teacher-support controls: \textbf{Teacher-TopK RKL} on the teacher's renormalized top-$32$ support, and \textbf{Teacher-TopK SRKL}, our top-$k$ on-policy adaptation of DistiLLM's skewed reverse-KL loss ($\alpha{=}0.1$)~\citep{ko2024distillm} on the same support. The remaining added baselines are entropy-aware \textbf{EOPD} ($k{=}16,\tau{=}0.8,\alpha{=}1.0$)~\citep{jin2026entropyopd}; \textbf{ExOPD}~\citep{yang2026gopd}, for which we evaluate the original sweep settings $\lambda\in\{1.25,1.5\}$ and an additional $\lambda=2.0$ extrapolation setting; and \textbf{RTT-GRPO}~\citep{xu2026rtt}, rubric-based RL using a learned Token-Level Relevance Discriminator and joint response- and token-level advantages. For Qwen3, the GRPO-IR baseline and the teacher share weights but we list them separately in the results table. The GRPO baselines use verifier-derived rewards from rule-based code verifiers or an LLM judge during training. The \method{} update derives its reward from teacher log-probabilities without querying an external grader.

\paragraph{Benchmarks.}
We evaluate on seven instruction-following benchmarks: IFEval~\citep{zhou2023ifeval}, IFBench~\citep{ifbench2025}, MulDimIF~\citep{muldimif2025}, ComplexBench~\citep{wen2024complexbench}, InfoBench~\citep{qin2024infobench}, FollowBench~\citep{jiang2024followbench}, and CFBench~\citep{zhang2025cfbench}. 
We report instruction-level accuracy for all benchmarks, defined as the percentage of model responses that satisfy all constraints specified in the corresponding prompts. Full evaluation details and judge configuration are in Appendix~\ref{app:impl}.

\paragraph{Implementation.}

Vanilla OPD, the three OPD-augmentation variants, ExOPD, and \method{} use sampled-token PPO with $\epsilon=0.2$. Teacher-TopK RKL and Teacher-TopK SRKL apply direct losses to distributions normalized over the teacher's top-$32$ support. EOPD combines the sampled-token reverse-KL OPD reward with an entropy-gated teacher-top-$16$ forward-KL loss. All methods are trained until validation performance saturates, and we report the corresponding saturated checkpoint. \method{} uses $\lambda=2.0$ and $c=5.0$ throughout; IFEval, IFBench, and MulDimIF were used to select these values, while ComplexBench, InfoBench, FollowBench, and CFBench were not used in this selection and serve as transfer checks.
Detailed training budgets, checkpoint selection, hyperparameters, and SFT data construction are provided in Appendix~\ref{app:impl}.

\begin{table*}[!t]
\centering
\footnotesize
\setlength{\tabcolsep}{4pt}
\renewcommand{\arraystretch}{1.05}
\begin{tabular}{l|ccccccc|c}
\toprule
\textbf{Model} & \textbf{IFEval} & \textbf{IFBench} & \textbf{MulDimIF} & \textbf{ComplexB} & \textbf{InfoB} & \textbf{FollowB} & \textbf{CFB} & \textbf{Avg.} \\
\midrule
\multicolumn{9}{l}{\emph{Qwen2.5 family} -- student: Qwen2.5-1.5B-Instruct, teacher: Qwen2.5-7B-Instruct trained with GRPO-IR} \\
\midrule
Qwen2.5-1.5B-Instruct (untrained) & 43.6 & 15.3 & 23.6 & 48.2 & 37.0 & 34.2 & 24.1 & 32.3 \\
Teacher (Qwen2.5-7B + GRPO-IR)  & 80.5 & 32.1 & 70.1 & 66.7 & 52.1 & 51.9 & 51.2 & 57.8 \\
\midrule
SFT                              & 66.7 & 21.6 & 65.4 & 50.8 & 41.9 & 38.2 & 29.6 & 44.9 \\
GRPO-IR                         & 54.3 & 18.6 & 36.0 & 51.1 & 39.9 & 35.0 & 27.8 & 37.5 \\
GRPO-CR                       & 56.6 & 19.2 & 49.0 & 52.3 & 44.1 & 39.6 & 31.4 & 41.7 \\
RTT-GRPO                         & 52.3 & 18.0 & 38.8 & 50.9 & 39.9 & 37.3 & 26.8 & 37.7 \\
OPD                              & 70.7 & 22.8 & 69.5 & 54.4 & 43.0 & 38.7 & 34.4 & 47.6 \\
Teacher-TopK RKL ($k{=}32$)     & 68.0 & 24.2 & 67.5 & 53.5 & 42.9 & 35.7 & 33.8 & 46.5 \\
Teacher-TopK SRKL ($k{=}32,\alpha{=}0.1$) & 66.9 & 23.8 & 68.9 & 53.2 & 44.3 & 37.2 & 32.4 & 46.7 \\
EOPD ($k{=}16,\tau{=}0.8$)      & 69.7 & 24.2 & 70.1 & 54.5 & 43.3 & 37.7 & 33.6 & 47.6 \\
ExOPD ($\lambda{=}1.25$)        & 73.9 & 24.8 & 70.3 & \textbf{55.2} & 43.0 & 39.4 & 35.2 & 48.8 \\
ExOPD ($\lambda{=}1.5$)         & 71.2 & 22.8 & 71.0 & 54.3 & 43.5 & 37.6 & 35.7 & 48.0 \\
ExOPD ($\lambda{=}2.0$)         & 75.0 & 22.4 & 71.3 & 53.8 & 43.7 & 39.1 & 36.5 & 48.9 \\
OPD-aug-SC                      & 72.5 & 25.6 & 69.3 & 54.1 & 42.3 & 39.8 & 33.9 & 48.2 \\
OPD-aug-Rand                     & 72.7 & 24.4 & 69.0 & 53.8 & 41.7 & 39.8 & 34.4 & 48.0 \\
OPD-aug-LOO                      & 72.1 & 23.7 & 69.1 & 54.3 & 42.5 & 39.0 & 33.8 & 47.8 \\
\rowcolor{cyan!15}
\textbf{\method{} (ours)}        & \textbf{76.0} & \textbf{28.2} & \textbf{72.9} & 54.8 & \textbf{46.4} & \textbf{43.8} & \textbf{37.8} & \textbf{51.4} \\
\midrule
\multicolumn{9}{l}{\emph{Qwen3 family} -- student: Qwen3-4B-Instruct-2507, teacher: Qwen3-4B-Instruct trained with GRPO-IR$^{\dagger}$} \\
\midrule
Qwen3-4B-Instruct (untrained)    & 83.8 & 29.9 & 56.9 & 70.6 & 55.3 & 57.8 & 61.2 & 59.4 \\
Teacher (Qwen3-4B + GRPO-IR)$^{\dagger}$ & 83.4 & 32.5 & 74.3 & 71.2 & 57.7 & 58.1 & 62.7 & 62.8 \\
\midrule
SFT                              & 82.1 & 29.7 & 71.7 & 69.7 & 56.3 & 57.5 & 61.9 & 61.3 \\
GRPO-IR$^{\dagger}$             & 83.4 & 32.5 & 74.3 & 71.2 & 57.7 & 58.1 & 62.7 & 62.8 \\
GRPO-CR                       & 83.1 & 32.9 & 73.1 & 70.8 & 55.9 & 58.0 & 62.8 & 62.4 \\
RTT-GRPO                         & 84.7 & 31.3 & 74.6 & \textbf{71.3} & 55.7 & 57.8 & 61.8 & 62.5 \\
OPD                              & 84.8 & 29.8 & 74.0 & 69.3 & 55.8 & 57.4 & 62.0 & 61.9 \\
Teacher-TopK RKL ($k{=}32$)     & 84.7 & 29.9 & 75.2 & 70.9 & 58.4 & 57.8 & 62.7 & 62.8 \\
Teacher-TopK SRKL ($k{=}32,\alpha{=}0.1$) & 84.5 & 31.6 & 73.6 & \textbf{71.3} & 57.8 & 58.1 & 61.8 & 62.7 \\
EOPD ($k{=}16,\tau{=}0.8$)      & 84.3 & 31.0 & 74.2 & 70.8 & 56.7 & 57.8 & 62.2 & 62.4 \\
ExOPD ($\lambda{=}1.25$)        & 85.8 & 31.6 & 74.5 & 70.0 & 57.7 & 55.7 & \textbf{63.4} & 62.7 \\
ExOPD ($\lambda{=}1.5$)         & 84.5 & 30.6 & 73.1 & 69.9 & \textbf{58.9} & 55.4 & 63.1 & 62.2 \\
ExOPD ($\lambda{=}2.0$)         & 82.1 & 28.6 & 72.5 & 69.8 & 54.3 & 56.0 & 63.2 & 60.9 \\
OPD-aug-SC                      & 82.7 & 31.6 & 72.5 & 70.3 & 56.3 & 57.6 & 62.9 & 62.0 \\
OPD-aug-Rand                     & 82.8 & 29.9 & 74.8 & 70.3 & 56.4 & 56.9 & 62.4 & 61.9 \\
OPD-aug-LOO                      & 84.3 & 29.6 & 74.9 & 70.2 & 58.1 & 56.6 & 62.1 & 62.3 \\
\rowcolor{cyan!15}
\textbf{\method{} (ours)}        & \textbf{86.1} & \textbf{37.6} & \textbf{76.2} & 71.0 & 57.4 & \textbf{58.4} & \textbf{63.4} & \textbf{64.3} \\
\midrule
\multicolumn{9}{l}{\emph{Self-Distillation Comparison} -- teacher is replaced by the frozen student initialization weights} \\
\midrule
DR-CFG (Qwen2.5-1.5B)       & 49.2 & \textbf{23.8} & 29.8 & 49.2 & \textbf{40.4} & \textbf{39.2} & \textbf{30.6} & 37.5 \\
self-\method{} (Qwen2.5-1.5B)    & \textbf{56.9} & 19.6 & \textbf{30.0} & \textbf{50.3} & 39.6 & 38.9 & 28.3 & \textbf{37.7} \\
\midrule
DR-CFG (Qwen3-4B)           & 84.7 & 32.8 & 61.3 & 70.0 & 55.9 & 59.9 & \textbf{62.6} & 61.0 \\
self-\method{} (Qwen3-4B)        & \textbf{84.9} & \textbf{36.3} & \textbf{61.5} & \textbf{70.7} & \textbf{57.7} & 59.9 & 62.5 & \textbf{61.9} \\
\bottomrule
\end{tabular}
\caption{
Main results across seven instruction-following benchmarks (\% accuracy / pass rate; higher is better). \textbf{Bold} marks the best score in each column \emph{within its block}, excluding reference rows (Instruct, Teacher, and the identical GRPO-IR$^{\dagger}$ row in the Qwen3 family). The Avg.\ column averages the seven benchmark scores. $^{\dagger}$ For the Qwen3 family, the teacher and the GRPO-IR baseline are the \emph{same} model (Qwen3-4B-Instruct trained with GRPO-IR); we list them separately to keep the baseline grid uniform across families.
}
\label{tab:main_results}
\end{table*}

\begin{figure*}[!t]
  \centering
  \includegraphics[width=\textwidth]{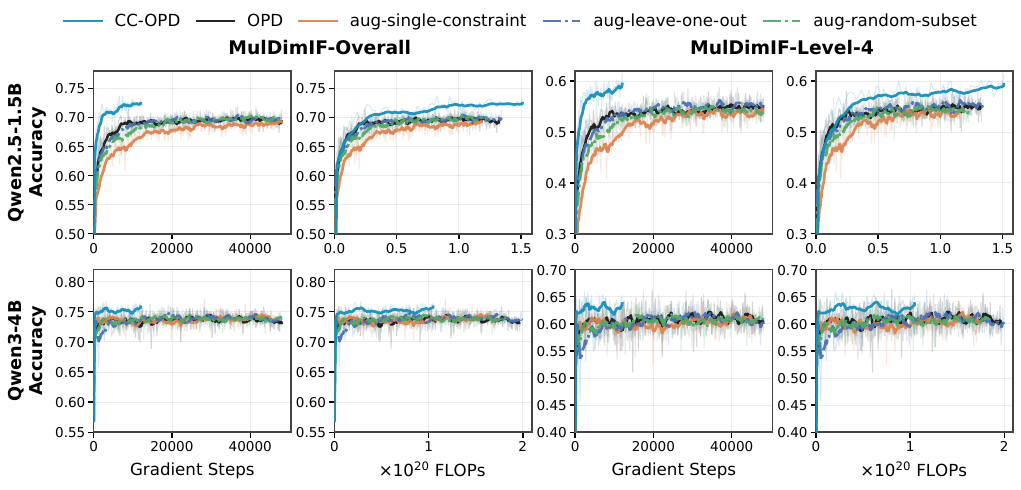}
  \caption{\textbf{MulDimIF training trajectories for five OPD-family methods.}
  Top row: Qwen2.5-1.5B pair; bottom row: Qwen3-4B pair. Each row plots MulDimIF overall accuracy (columns 1--2) and Level-4 accuracy (columns 3--4) against gradient steps (columns 1, 3) and total FLOPs (columns 2, 4). The OPD baseline and three OPD-augmentation baselines are trained to saturation ($4\times$ more gradient steps than \method{}).}
  \label{fig:training_trajectories}
\end{figure*}

\subsection{Main Results}
\label{sec:main_results}

Table~\ref{tab:main_results} reports the seven-benchmark performance of \method{} against all baselines for both model pairs. \method{} achieves the best average score in both families: $51.4$\% on Qwen2.5 ($+2.5$\,pp over the strongest ExOPD setting) and $64.3$\% on Qwen3 ($+1.5$\,pp over Teacher-TopK RKL). It leads on six of seven individual benchmarks for Qwen2.5 and is best or tied on five of seven for Qwen3. Beyond average score, the \method{}-trained Qwen3 student surpasses its own RL-trained teacher (analyzed in \S\ref{sec:surpass_teacher}). The Self-Distillation block at the bottom of Table~\ref{tab:main_results} compares self-\method{} against DR-CFG~\citep{cideron2025drcfg} under the same sampled-token OPD variant, replacing the external teacher with the student's own initialization weights: self-\method{} leads on the Qwen3-4B pair by $+0.9$\,pp on the seven-benchmark average ($61.9$ vs.\ $61.0$) and is close on the Qwen2.5-1.5B pair ($37.7$ vs.\ $37.5$). Appendix~\ref{app:impl:grpo} provides implementation details and per-benchmark self-distillation results.

\subsection{Student Surpasses Teacher}
\label{sec:surpass_teacher}

Table~\ref{tab:main_results} shows that \method{}-trained students cross their teacher's accuracy ceiling under both model pairs on MulDimIF, a rare outcome for OPD. On Qwen2.5, the 1.5B student reaches $72.9$\,\% on MulDimIF overall, $+2.8$\,pp above the 7B teacher's $70.1$\,\%; the gap widens to $+6.3$\,pp on the constraint-saturated Level-$4$ split ($59.4$\,\% vs.\ $53.1$\,\%). On Qwen3, where the student and teacher share the same model size, the student exceeds the teacher on five of seven benchmarks (ComplexBench and InfoBench are within $0.3$\,pp), showing that a teacher--student size gap is not required for teacher-surpassing performance. Figure~\ref{fig:teaser} shows the validation trajectory of the Qwen2.5 student on MulDimIF crossing the teacher ceiling at roughly gradient step $1{,}500$; vanilla sampled-token OPD on the same student asymptotes below it. Among the OPD variants trained on the Qwen2.5 pair, all three ExOPD settings also exceed the $7$B teacher on MulDimIF ($70.3$--$71.3$\,\% vs.\ $70.1$\,\%), while \method{} attains the largest margin ($72.9$\,\%). On the same-size Qwen3 pair, two OPD-augmentation variants also edge past the teacher on MulDimIF ($74.9$ and $74.8$ vs.\ $74.3$), but \method{} attains the largest margin ($76.2$).

\subsection{Gradient-Step Efficiency}
\label{sec:efficiency}

We anchor efficiency on the OPD baseline trajectory and ask, at each smoothed reference accuracy that OPD baseline reaches, how many gradient steps \method{} needs to match it (all accuracy values are EMA-smoothed with weight $0.85$). On the Qwen2.5-1.5B pair, \method{} matches OPD baseline's smoothed MulDimIF peak in $\mathbf{15.2\times}$ fewer gradient steps on the overall metric and $\mathbf{16.6\times}$ fewer on the hardest Level-$4$ split; OPD baseline's trajectory never reaches \method{}'s eventual peak within its full $48{,}000$-step horizon. \method{}'s $4.63\times$ per-step teacher overhead on Qwen2.5 is more than offset by this step reduction: at iso-accuracy \method{} uses $\approx\!30\%$ of OPD baseline's total FLOPs on Qwen2.5 and $\approx\!53\%$ on Qwen3-4B (where the per-step overhead is smaller because student and teacher share size). The three OPD-augmentation baselines fall in between (Figure~\ref{fig:training_trajectories}). Counterfactual subsampling (an additional efficiency lever) is ablated in Appendix~\ref{app:ratio_sweep}.

\begin{table*}[t]
\centering
\footnotesize
\setlength{\tabcolsep}{6pt}
\renewcommand{\arraystretch}{1.1}
\begin{tabular}{l|ccccccc|c}
\toprule
\textbf{Variant} & \textbf{IFEval} & \textbf{IFBench} & \textbf{MulDimIF} & \textbf{ComplexB} & \textbf{InfoB} & \textbf{FollowB} & \textbf{CFB} & \textbf{Avg.} \\
\midrule
\textbf{LOO} (\method{}) & \textbf{76.0} & 28.2          & \textbf{72.9} & \textbf{54.8} & \textbf{46.4} & \textbf{43.8} & 37.8          & \textbf{51.4} \\
LA                       & 75.0          & 27.1          & 71.0          & 53.5          & 45.3          & 43.5          & \textbf{38.0} & 50.5          \\
RSS                      & 75.5          & \textbf{28.8} & 72.1          & 52.9          & 44.5          & 42.6          & 37.3          & 50.5          \\
NAO                      & 71.2          & 26.5          & 66.4          & 54.6          & 45.0          & 41.7          & 35.7          & 48.7          \\
\bottomrule
\end{tabular}
\caption{
Counterfactual variant ablation on the Qwen2.5-1.5B pair (\% accuracy / pass rate; higher is better). All four variants share identical training hyperparameters; they differ only in the structure of the ``with'' and ``without'' constraint subsets used to construct $\delta_i(t)$. \textbf{Bold} marks the best score in each column.
}
\label{tab:ablation_variants}
\end{table*}

\subsection{Counterfactual Variant Ablation}
\label{sec:ablation}

\S\ref{sec:method:cf} defines \method{}'s per-constraint signal $\delta_i(t)$ via the leave-one-out contrast $\ell(C; t) - \ell(C \setminus \{c_i\}; t)$. We now compare this choice against three alternative counterfactual designs that vary the structure of the ``with'' and ``without'' constraint subsets used in the contrast. \textbf{LA} (Leave-All) collapses the contrast to $\ell(C) - \ell(\emptyset)$, producing one scalar shift per token rather than $|C|$ per-constraint shifts. \textbf{RSS} (Random Subset) replaces $C \setminus \{c_i\}$ with $C \setminus S_j$: for each job $j$, it first samples the removal size $|S_j|$ uniformly from $\{1, \dots, |C|-1\}$ and then samples $S_j$ uniformly among subsets of that size, summing the resulting contrasts over $j = 1, \dots, |C|$. \textbf{NAO} (No-context Anchor Only) replaces the full-context ``with'' anchor with singleton contexts, $\ell(\{c_i\}) - \ell(\emptyset)$, summed over $i = 1, \dots, |C|$.
Full derivations and extended signal-level diagnostics for these four variants appear in Appendix~\ref{app:mobius}. Table~\ref{tab:ablation_variants} reports the seven-benchmark performance of all four variants on the Qwen2.5-1.5B pair, under identical training hyperparameters.

The leave-one-out construction obtains the strongest average score (51.4\,\%, $+0.9$\,pp over LA and RSS, $+2.7$\,pp over NAO) and outperforms LA on six of seven individual benchmarks (CFBench: $37.8$ vs.\ $38.0$). The largest cross-benchmark LOO--LA gaps occur on MulDimIF ($+1.9$), ComplexBench ($+1.3$), and InfoBench ($+1.1$\,pp), while the remaining gaps are IFEval $+1.0$, IFBench $+1.1$, FollowBench $+0.3$, and CFBench $-0.2$\,pp.

\noindent\textbf{Signal-level comparison of LOO and LA.}
For a fixed realized token $y_t$, let $f_t(S)$ be its teacher log-probability under constraint subset $S$, and let $m_t(T)$ denote the corresponding M\"obius coefficients~\citep{rota1964mobius}. Then
\begin{equation}
\begin{aligned}
f_t(S) &= \sum_{T\subseteq S}m_t(T), \\
\delta_{\mathrm{LA}}(t) &= \sum_{\emptyset\neq T\subseteq C}m_t(T), \\
\delta_{\mathrm{LOO}}(t) &= \sum_{\emptyset\neq T\subseteq C}|T|\,m_t(T).
\end{aligned}
\label{eq:main_mobius_comparison}
\end{equation}
LA therefore weights every non-empty interaction uniformly, whereas LOO weights an order-$k$ interaction by $k$. Strict additivity across constraints is sufficient for the two signals to agree. Conversely, a nonzero LOO--LA residual certifies higher-order interactions, although a zero residual need not imply additivity because signed terms can cancel.

We examine the raw signals on $15{,}450$ HIR-16K examples with $5$--$10$ constraints, using Qwen2.5-1.5B student rollouts scored by the Qwen2.5-7B teacher. The relative $L_1$ residual is $0.918$, and the two signals disagree in sign on $31.9$\% of tokens. The unsummed $K\!\times\!T$ LOO sensitivity map is also structured: the dominant constraint accounts for $42.6$\% of absolute signal mass at an average token, versus $15.0$\% under uniform attribution, while pairs of constraints have mean top-$20$ affected-token Jaccard overlap $0.269$. The $K\!\times\!T$ LOO map retains these per-constraint sensitivities, while LA provides one full-versus-empty contrast per token. These statistics describe the unclipped teacher log-probability shifts produced by constraint ablation. Downstream benchmark results assess the learned policy after clipping and optimization. Appendix~\ref{app:mobius} gives the full derivations and additional statistics.

\begin{table}[t]
  \centering
  \small
  \begin{tabular*}{\columnwidth}{@{\extracolsep{\fill}}lcccc@{}}
    \toprule
    \textbf{Method} & \textbf{L1} & \textbf{L2} & \textbf{L3} & \textbf{L4} \\
    \midrule
    LA & 86.9 & 74.8 & 65.3 & 57.1 \\
    LOO & 89.2 & 75.3 & 67.6 & 59.4 \\
    $\Delta$ (pp) & $+2.3$ & $+0.6$ & $+2.2$ & $+2.3$ \\
    \bottomrule
  \end{tabular*}
  \caption{LOO and LA accuracy on the four MulDimIF difficulty tiers under the same training step, hyperparameters, deterministic scoring protocol, and constraint vocabulary. Scores are rounded to one decimal; $\Delta$ is computed from unrounded values.}
  \label{tab:loo_la_difficulty}
\end{table}

\paragraph{Controlled comparison across difficulty tiers.}
Table~\ref{tab:loo_la_difficulty} compares LOO and LA within MulDimIF while holding the training step, hyperparameters, deterministic scoring protocol, and constraint vocabulary fixed. LOO improves accuracy on every tier: $+2.3$, $+0.6$, $+2.2$, and $+2.3$\,pp on L1--L4. The gain is positive but not monotonic with difficulty. This pattern is consistent with Eq.~\eqref{eq:main_mobius_comparison}: LA assigns unit weight to each non-empty M\"obius coefficient, whereas LOO weights a coefficient by its interaction order. The identity predicts a difference when higher-order terms do not cancel, but not a steadily increasing performance gap, because the coefficients are signed and the aggregated reward is clipped before PPO. The positive L3/L4 gains show that the LOO advantage persists when more constraints are stacked, while the smaller L2 gap indicates that constraint count alone does not determine the benefit.

\paragraph{Variant-level comparison.}
LA remains a useful, efficient approximation when teacher inference dominates cost (two teacher passes per sample vs.\ $|C|+1$ for LOO and RSS). NAO ($\delta_i(t) = \ell(\{c_i\};\, t) - \ell(\emptyset;\, t)$) instantiates the naive ``$N$ single-constraint teachers'' design rejected in \S\ref{sec:method:cf}. Its clear underperformance ($-2.7$\,pp Avg, $-6.5$\,pp on MulDimIF, $-14.1$\,pp on the MulDimIF L4 tier in Appendix~\ref{app:mobius:crossmodel}) confirms that replacing the full-context anchor with a no-context anchor removes the full-context, per-constraint sensitivity structure used by LOO. These ablations support the full-context leave-one-out anchor as the default reward-shaping design. Sensitivity of the shaping strength $\lambda$ and clip threshold $c$ around our defaults is reported in Appendix~\ref{app:lambda_clip}.

\section{Related Work}
\label{sec:related}

\paragraph{On-policy distillation.}
On-policy distillation~(OPD) trains a student to match a teacher distribution on the student's own rollouts. The framework was introduced by GKD~\citep{agarwal2024gkd} and refined in MiniLLM~\citep{gu2024minillm} and DistiLLM~\citep{ko2024distillm} through reverse-KL and skewed-KL variants, with recent analyses motivating the sampled-token formulation we adopt~\citep{li2026rethinkopd}. \method{} keeps this sampled-token KL backbone unchanged and adds a per-constraint counterfactual shaping term on the same rollout.

\paragraph{Rubric-based RL for instruction following.}
A closely related line treats rubrics or checklists as the reward channel for instruction-following RL. ROPD~\citep{fang2026ropd}, RaR~\citep{gunjal2025rar}, and RLCF~\citep{viswanathan2025rlcf} aggregate criterion-wise verifier or LLM-judge scores into pass-rate, GRPO, or DPO rewards. RTT~\citep{xu2026rtt} trains an external token-level relevance discriminator, and RuscaRL~\citep{zhou2025ruscarl} injects rubrics into the actor's rollout as exploration scaffolding before grading. All convert constraints into externally graded reward signals on the student side. \method{} instead inverts this design. Rather than enriching the supervision channel with external grading, \method{} ablates the teacher's conditioning one constraint at a time and extracts per-constraint, per-token credit from the teacher's leave-one-out forced-scoring shifts.

\paragraph{Contrastive likelihood signals.}
Several methods shape generation through log-probability contrasts at different stages of the pipeline: CFG-LM~\citep{sanchez2024cfg}, CAD~\citep{shi2024cad}, and Contrastive Decoding~\citep{li2023contrastive} reweight logits at decoding time using prompt-vs-no-prompt or expert-vs-amateur contrasts (two forward passes, no parameter updates), and AlignDistil~\citep{zhang2025aligndistil} performs training-time distillation against a DPO-derived log-ratio teacher. Diversity-Rewarded CFG Distillation~\citep{cideron2025drcfg} is a training-time analog in text-to-music generation. It distills a single CFG-augmented teacher (instantiated from the frozen student initialization) into a single-pass student via a single full-prompt-vs-negative-prompt contrast. \method{} decomposes the contrast across each individual constraint via leave-one-out rather than collapsing $C$ into one prompt-vs-no-prompt contrast.

\section{Conclusion}
\label{sec:conclusion}

We presented \method{}, a counterfactual constraint-conditioned on-policy distillation method that extracts per-constraint signal from a frozen teacher's leave-one-out forced scoring on the same student rollout. Across two Qwen settings and seven instruction-following benchmarks, \method{} attains the highest average among the evaluated student-training methods in both settings. It also yields the largest student-over-teacher MulDimIF margin among the tested OPD variants. More broadly, our results reframe constraint-conditioned supervision from an \emph{additive} paradigm, in which the supervision channel is enriched with external rubrics or graders, to an \emph{ablative} one, in which the same frozen teacher is queried across counterfactual conditioning subsets, exposing per-constraint supervision that no single full-context forward pass produces.

\section*{Limitations}

\method{} is trained on HIR-16K and evaluated on two Qwen-family student--teacher pairs (Qwen2.5-1.5B/7B and Qwen3-4B/4B). Generalization to other instruction-following corpora, to non-Qwen model families, and to multi-turn or system-prompt-steerable settings remains untested.

The method assumes a \emph{white-box} teacher available for forced scoring: per-token log-probability access under multiple counterfactual prompts is required, so teachers exposed only through generation APIs are outside the scope of the current construction.

Our experiments evaluate \method{} only with the sampled-token reverse-KL OPD objective. Its effectiveness with other OPD objectives has not been tested.

The counterfactual prompt construction further assumes that each sample's constraints can be cleanly excised from the natural-language instruction via byte-aligned span deletion; non-inline constraint formats such as structured fields or programmatic templates may require a different counterfactual prompt construction.

Each sample requires $|C|+1$ teacher forward passes, which increases per-step computation and can increase memory use and latency relative to single-pass OPD. Section~\ref{sec:efficiency} reports lower total FLOPs at matched accuracy on the two evaluated model pairs. The additional per-step overhead remains relevant for memory- or throughput-constrained training systems.

Finally, the self-distillation comparison in Table~\ref{tab:main_results} shows a clear average gain for self-\method{} on the Qwen3-4B pair but only a small average difference from DR-CFG on the Qwen2.5-1.5B pair. With only these two model settings, the comparison is insufficient to establish whether the usefulness of the LOO signal changes systematically with teacher scale. Characterizing this relationship across a wider range of teacher sizes is an open direction.

\bibliography{references}

\appendix

\section{Implementation and Evaluation Details}
\label{app:impl}

This appendix details the hyperparameters, datasets, baselines, and evaluation protocol referenced by the main paper. Algorithm~\ref{alg:cc_opd} gives the complete \method{} training step in a form that can be dropped into an existing sampled-token OPD trainer with no change to the rollout or PPO modules.
\subsection{Training Hyperparameters for OPD-Family Methods}
\label{app:impl:opd}

The base configuration in Table~\ref{tab:hyperparams_opd}, implemented in the \texttt{verl}~\citep{sheng2024hybridflow} framework, is used by \method{}, the vanilla OPD baseline, and the three OPD-augmentation variants. ExOPD, Teacher-TopK RKL, and Teacher-TopK SRKL use the same training corpus, optimizer, batching, rollout settings, and hardware as \method{}, differing only in their training losses and the corresponding loss-specific hyperparameters reported in \S\ref{sec:setup} and Table~\ref{tab:main_results}. ExOPD retains the sampled-token PPO update, whereas Teacher-TopK RKL and Teacher-TopK SRKL optimize direct losses over the teacher's normalized top-$32$ support; PPO-specific reward and advantage entries therefore do not apply to the two Teacher-TopK controls. Both the Qwen2.5-1.5B and Qwen3-4B student pairs use identical settings; per-step compute differs because the model sizes differ across pairs. PPO advantage estimation uses the OPD branch (no critic, no GAE), and the actor is updated for a single PPO epoch per mini-batch. Each verl-step samples $128$ prompts $\times$ $8$ rollouts $= 1{,}024$ trajectories and triggers $8$ gradient updates (mini-batch $128$, micro-batch $4$ per GPU); we therefore report results on the gradient-step axis using the conversion $1\text{ verl-step} = 8\text{ gradient steps}$. \method{} is trained for $1{,}500$ verl-steps; the vanilla OPD baseline and the three OPD-augmentation variants are trained for $6{,}000$ verl-steps. We report each method at its saturated checkpoint: step $1{,}500$ for \method{} and step $3{,}000$ for the four baselines. This asymmetric training horizon is a saturation check rather than a budget asymmetry that favors \method{}: \method{} plateaus on validation MulDimIF by step $1{,}500$ while the baselines plateau by step $3{,}000$, and we train the baselines through step $6{,}000$ to confirm no further improvement (the full $48{,}000$-gradient-step baseline trajectory is shown in \S\ref{sec:efficiency}). These methods use $32$ GPUs (96 GB per GPU) in a single Ray cluster.

\begin{algorithm}[t]
  \small
  \caption{One \method{} training step.}
  \label{alg:cc_opd}
  \begin{algorithmic}[1]
  \Require Student $\studpolicy$, frozen teacher $\teacher$, mini-batch $B$, shaping strength $\lambda$, clip threshold $c$
  \For{$(q, C, \mathsf{inst}(q,C)) \in B$}
    \State $y \sim \studpolicy(\cdot \mid q, C)$; cache $\log\studpolicy(y_t \mid q, C, y_{<t})$, $t = 1,\dots,L$ \Comment{student rollout + log-probs}
    \State $\ell(C; t) \gets \log\teacher(y_t \mid q, C, y_{<t})$, $t = 1,\dots,L$ \Comment{full-context teacher pass}
    \State $\widehat{\KL}(t) \gets \log\studpolicy(y_t \mid q, C, y_{<t}) - \ell(C; t)$, $t = 1,\dots,L$ \Comment{Eq.~\eqref{eq:opd_sample}}
    \For{$i = 1, \dots, |C|$}
      \State $q'_i \gets \texttt{inline\_remove}(\mathsf{inst}(q,C), c_i)$
      \State $\ell(C\!\setminus\!\{c_i\}; t) \gets \log\teacher(y_t \mid q'_i, y_{<t})$
      \State $\delta_i(t) \gets \ell(C; t) - \ell(C\!\setminus\!\{c_i\}; t)$ \Comment{Eq.~\eqref{eq:delta_i}}
    \EndFor
    \State $\hat{\delta}(t) \gets \mathrm{clip}\!\left(\textstyle\sum_{i}\delta_i(t),\; -c,\; +c\right)$ \Comment{Eq.~\eqref{eq:delta_agg}--\eqref{eq:delta_clip}}
    \State $r_t \gets -\widehat{\KL}(t) + \lambda\,\hat{\delta}(t)$ \Comment{Eq.~\eqref{eq:cc_reward}}
  \EndFor
  \State Update $\studpolicy$ with ratio-clipped PPO on $\{r_t, y, q, C\}_{B}$
  \end{algorithmic}
\end{algorithm}

\begin{table}[t]
  \centering
  \small
  \setlength{\tabcolsep}{4pt}
  \begin{tabular}{ll}
    \toprule
    \textbf{Hyperparameter} & \textbf{Value} \\
    \midrule
    \multicolumn{2}{l}{\emph{PPO}} \\
    Clip ratio $\epsilon$           & $0.2$ \\
    PPO epochs / mini-batch          & $1$ \\
    \midrule
    \multicolumn{2}{l}{\emph{Optimizer}} \\
    Optimizer                       & AdamW \\
    Actor LR                        & $2\!\times\!10^{-6}$ \\
    Betas $(\beta_1, \beta_2)$      & $(0.9, 0.999)$ \\
    Weight decay                    & $0.01$ \\
    LR schedule                     & constant \\
    \midrule
    \multicolumn{2}{l}{\emph{Batching}} \\
    Prompts per verl-step           & $128$ \\
    Rollouts per prompt             & $8$ \\
    PPO mini-batch                  & $128$ \\
    Micro-batch per GPU             & $4$ \\
    \midrule
    \multicolumn{2}{l}{\emph{Rollout (vLLM)}} \\
    Temperature / top-$p$ / top-$k$ & $1.0$ / $1.0$ / $-1$ \\
    Max prompt / response length    & $2048$ / $1024$ \\
    \midrule
    \multicolumn{2}{l}{\emph{\method{}-specific}} \\
    Shaping strength $\lambda$      & $2.0$ \\
    Clip threshold $c$     & $5.0$ \\
    Constraint subsample ratio $\rho$ & $1.00$ \\
    \bottomrule
  \end{tabular}
  \caption{Base training configuration for the OPD-family methods. The final \method{}-specific block applies only to \method{}. ExOPD and the two Teacher-TopK controls share the optimizer, batching, rollout, and hardware settings but use the objective-specific losses described in \S\ref{sec:setup}. PPO-specific entries do not apply to the direct-loss Teacher-TopK controls.}
  \label{tab:hyperparams_opd}
\end{table}

\subsection{HIR-16K Training Corpus}
\label{app:impl:hir}

We train all compared methods, including the GRPO baselines, on HIR-16K~\citep{zhang2025hir}, an instruction-following corpus containing $16{,}968$ samples with multi-constraint specifications. Each sample is a tuple $(q, C, \mathcal{V})$ where $q$ is a base prompt, $C = \{c_1, \dots, c_N\}$ is its constraint set, and $\mathcal{V}$ is the constraint-verification metadata. Constraint cardinality $|C|$ ranges from $5$ to $17$ (mean $7.25$, median $7$). Constraints fall into three verifier categories: rule-based atomic constraints checkable with deterministic code ($\texttt{ifeval\_like}$, $6{,}998$ samples), table-structured rule constraints ($\texttt{constraint\_table}$, $3{,}608$ samples), and free-text constraints requiring an LLM judge ($\texttt{llm\_checker}$, $6{,}362$ samples).

\paragraph{Exact-match audit.} After Unicode NFKC, case, and whitespace normalization, no evaluation prompt exactly matches either a HIR base prompt or a full rendered training instruction.

\subsection{Teacher Training and GRPO Baselines}
\label{app:impl:grpo}

\paragraph{GRPO-IR teachers.} The Qwen2.5-7B and Qwen3-4B teachers are obtained by GRPO~\citep{shao2024deepseekmath} fine-tuning of the corresponding Instruct base with an instruction-level accuracy reward (\textbf{GRPO-IR}, matching RL-IR of~\citet{peng2025rl-ir, ifbench2025}): $r = 1$ iff every constraint $c_i \in C$ is satisfied, else $r = 0$. The \textbf{GRPO-CR} baseline (matching RL-CR of~\citet{qi2025rl-cr, ifbench2025}) replaces this with a constraint-level mean reward $r = \frac{1}{|C|}\sum_i \mathbf{1}[c_i \text{ satisfied}]$. The training implementation strictly follows \citet{xu2026rtt}. Per-constraint satisfaction is determined by combining the original benchmark verifiers: rule-based constraints (\texttt{ifeval\_like} and \texttt{constraint\_table} subsets) are checked by the IFEval-style code verifier, and free-text constraints (\texttt{llm\_checker} subset) are scored by DeepSeek-V3.1\footnote{\url{https://huggingface.co/deepseek-ai/DeepSeek-V3.1}} as the unified judge. The shared GRPO actor configuration is summarized in Table~\ref{tab:hyperparams_grpo}. GRPO uses no KL penalty; group-normalized advantages and asymmetric PPO clipping are applied as in the standard GRPO recipe.

\paragraph{RTT-GRPO.} RTT-GRPO follows the original implementation and method-specific training hyperparameters of \citet{xu2026rtt} without modification, while using the model pairs, HIR-16K training corpus, and evaluation protocol of this paper.

\begin{table}[t]
  \centering
  \small
  \setlength{\tabcolsep}{4pt}
  \begin{tabular}{ll}
    \toprule
    \textbf{Hyperparameter} & \textbf{Value} \\
    \midrule
    Group size (rollouts per prompt) & $8$ \\
    PPO epochs                       & $1$ \\
    PPO clip (low / high)            & $0.2$ / $0.27$ \\
    rollout batch size               & $64$ \\
    n rollout                        & $8$ \\
    ppo batch size                   & $256$ \\
    Advantage normalization          & per-group mean and std \\
    KL coefficient                   & $0$ (no reference) \\
    Reward whitening                 & off \\
    Actor LR                         & $1\!\times\!10^{-6}$ \\
    Warmup steps                     & $20$ \\
    Weight decay                     & $0$ \\
    Rollout temp.\ / top-$p$ / top-$k$ & $0.99$ / $0.99$ / $100$ \\
    Prompt / response length         & $2048$ / $2048$ \\
    Max training steps               & $500$ \\
    Precision                        & bf16 \\
    \bottomrule
  \end{tabular}
  \caption{GRPO hyperparameters shared between GRPO-IR and GRPO-CR. The two recipes differ only in the reward aggregation function.}
  \label{tab:hyperparams_grpo}
\end{table}

\paragraph{Teacher checkpoint convergence.}
For the Qwen2.5-7B GRPO-IR teacher, 25 validation evaluations cover steps 20--500. At steps 420, 440, 460, 480, and 500, the valid-sample strict averages over IFEval, IFBench, and MulDimIF are $0.6124$, $0.6088$, $0.6040$, $0.6102$, and $0.6090$, respectively (mean $0.6089$); the late-stage slopes on all three benchmarks are approximately zero. The Qwen3-4B GRPO-IR teacher is likewise evaluated 25 times from steps 20--500. Its final checkpoint is only $0.0055$, $0.0034$, and $0.0092$ below the respective IFEval, IFBench, and MulDimIF trajectory peaks. These trajectories indicate a plateau or near-saturation under the evaluated recipes, but do not rule out improvement from a different GRPO recipe.

\paragraph{SFT baseline.} The SFT baseline fine-tunes each student on greedy teacher-generated responses to all HIR-16K prompts. We use AdamW with learning rate $10^{-5}$, batch size $128$, three epochs, and FSDP-2 sharding across $32$ GPUs.

\paragraph{Self-distillation baselines.} In the Self-Distillation Comparison block of Table~\ref{tab:main_results}, the external teacher is replaced by the student's own frozen initialization weights and the same PPO machinery is used otherwise. \textbf{DR-CFG} reproduces the core CFG-distillation objective of \citet{cideron2025drcfg}, who form a guided teacher distribution by linearly combining conditional and unconditional logits with guidance strength $\gamma=3$ and minimize the KL from the student to a softmax over the guided logits. Because our framework operates on a sampled-token OPD estimator rather than the full-vocabulary KL objective of the original paper, we adapt this objective to a sampled-token estimator. At each realized student token we compute the conditional-vs-unconditional log-probability difference under the frozen self-initialization, $\log\teacher(y_t \mid q, C, y_{<t}) - \log\teacher(y_t \mid q, y_{<t})$, scale it by $\gamma - 1 = 2$, and add the symmetrically clipped result to the vanilla sampled-token OPD reward. This adaptation matches the log-ratio direction and ignores the full-vocabulary normalizer term that drops out of the sampled estimator. The diversity-reward component of the original paper is omitted because it targets text-to-music generation and is not directly applicable in our text-instruction domain. \textbf{self-\method{}} applies \method{}'s reward shaping (Eq.~\eqref{eq:cc_reward}) to the self-distillation setting by replacing the external teacher in Eq.~\eqref{eq:delta_i} with the frozen self-initialization.

\paragraph{Per-benchmark breakdown.}
On Qwen3-4B, self-\method{} averages $61.9$ across the seven benchmarks, compared with $61.0$ for DR-CFG. It scores higher on IFEval, IFBench, MulDimIF, ComplexBench, and InfoBench, ties on FollowBench, and is $0.1$\,pp lower on CFBench. On Qwen2.5-1.5B, the corresponding averages are $37.7$ and $37.5$. Self-\method{} is $7.7$\,pp higher on IFEval and also scores higher on MulDimIF and ComplexBench; DR-CFG scores higher on the other four benchmarks.

\subsection{OPD Data-Augmentation Baselines}
\label{app:impl:opdaug}

The three OPD-augmentation baselines train vanilla sampled-token OPD on a HIR-16K corpus that has been enlarged by synthetically rewriting each row's constraint set. From a row $(q, C, y_\text{teacher})$ with $|C|=N$, each augmentation emits $N$ additional rows according to a different rule:
\begin{itemize}\setlength\itemsep{0pt}
  \item \textbf{OPD-aug-LOO}: the $i$-th new row keeps $C \setminus \{c_i\}$ (one constraint removed at a time).
  \item \textbf{OPD-aug-Rand}: each new row samples a subset of size uniformly drawn from $1$ to $N-1$ and keeps the corresponding random complement.
  \item \textbf{OPD-aug-SC}: the $i$-th new row keeps only $\{c_i\}$ (one constraint at a time).
\end{itemize}
Augmented rows are concatenated with the original $16{,}968$ rows; the training objective and PPO configuration are identical to vanilla sampled-token OPD with no counterfactual reward shaping. Because the augmented corpus is roughly $1 + \bar{N} \approx 8\times$ larger (where $\bar{N}\!\approx\!7.25$ is the mean constraint count), these baselines need substantially more gradient updates to traverse the data; we train them to $6{,}000$ verl-steps and report the step-$3{,}000$ checkpoint where the validation curve has plateaued.

\subsection{Counterfactual Variant Ablations}
\label{app:impl:variants}

\S\ref{sec:ablation} compared three alternative counterfactual schemes to the main LOO construction of \method{}. All four variants share $\lambda=2.0$, clip threshold $c=5.0$, and constraint subsample ratio $\rho=1.00$; they differ only in which counterfactual prompts $A$ are constructed and how the per-token signals $\delta(t) = \ell_C(t) - \ell_A(t)$ are aggregated. The same realized student rollout $y$ is shared across all teacher forward passes within a variant; only the prompt context varies.
\begin{itemize}\setlength\itemsep{0pt}
  \item \textbf{LOO} (main): for each constraint $c_i$, rescore the teacher under $A_i = C \setminus \{c_i\}$. Per-token signal $\delta_i(t) = \ell_C(t) - \ell_{C\setminus\{c_i\}}(t)$; aggregate by $\delta(t) = \sum_i \delta_i(t)$.
  \item \textbf{LA (leave-all)}: a single full-vs-empty contrast. Rescore the teacher under the empty constraint context $A = \emptyset$; $\delta(t) = \ell_C(t) - \ell_\emptyset(t)$.
  \item \textbf{RSS (random subset)}: for each of $|C|$ jobs, sample a removed-subset size uniformly from $1$ to $|C|-1$, draw the removed indices without replacement, and rescore under $A_j = C \setminus S_j$. Aggregate by $\delta(t) = \sum_j \delta_j(t)$.
  \item \textbf{NAO (No-context Anchor Only)}: for each $c_i$, rescore the teacher both under the singleton $A^{+}_i = \{c_i\}$ and under the empty context $A^{0} = \emptyset$ (the latter is shared across all $i$). Per-token signal $\delta_i(t) = \ell_{\{c_i\}}(t) - \ell_\emptyset(t)$; aggregate by $\delta(t) = \sum_i \delta_i(t)$.
\end{itemize}
LA isolates the effect of constraint presence as a single bulk contrast. RSS preserves the multi-constraint structure of LOO but blurs the attribution to any individual constraint. NAO replaces the full-context anchor with a no-context anchor, asking how each singleton conditioning shifts likelihood relative to no conditioning at all. LOO is the only variant whose contrasts vary by exactly one named constraint.

\subsection{Evaluation Protocol}
\label{app:impl:eval}

\paragraph{Inference.} For every evaluated model we generate responses with vLLM using $T=0.6$, top-$p\!=\!1.0$, $\text{max\_tokens}\!=\!4096$, and $n=3$ samples per prompt. The reported number is the arithmetic mean across the three samples. Samples whose judge call fails to parse on all three attempts are excluded from scoring.

\paragraph{Rule-based scoring.} IFEval, IFBench, and MulDimIF are scored by deterministic verifiers from their official evaluation scripts. We report strict prompt accuracy for IFEval and IFBench and overall accuracy averaged over the four difficulty levels for MulDimIF.

\paragraph{LLM-judged scoring.} ComplexBench, InfoBench, FollowBench, and CFBench rely on an LLM judge for each atomic criterion. The original judges specified by these benchmarks (legacy \texttt{gpt-4-1106} and \texttt{gpt-4-0314}) have been deprecated by OpenAI; to keep scoring consistent across all evaluated models and to avoid drift from per-benchmark judge changes, we adopt a unified contemporary judge: \texttt{gpt-5.4-mini-0317} for ComplexBench, InfoBench, and FollowBench, and \texttt{gpt-4o-0513} for CFBench (whose original judge is still active). All judge prompt templates and aggregation rules follow the official scorer of each benchmark.

\section{Structural Relationship between LOO Sum and LA: a M\"obius Decomposition}
\label{app:mobius}

Section~\ref{sec:ablation} summarizes the structural difference between the per-constraint leave-one-out aggregation $\sum_i \delta_{\text{LOO},i}$ and the single leave-all contrast $\delta_{\text{LA}} = \log P_T(y_t \mid q, C, y_{<t}) - \log P_T(y_t \mid q, \emptyset, y_{<t})$. Both reduce to $1\!\times\!T$ token-level vectors before entering the OPD update, but they apply different weights to the same underlying set-function interactions. This appendix gives the full derivation, extends it to NAO and RSS, and reports additional diagnostics and caveats. Strict additivity of the teacher's constraint response is sufficient for LOO and LA to agree; equality at a single token does not imply additivity because signed higher-order terms can cancel.

\subsection{Setup: the teacher log-probability as a set function}
\label{app:mobius:setup}

For a token $y_t$ in a realized student rollout, fix the query $q$, the prefix $y_{<t}$, and consider only how the teacher's log-probability of $y_t$ depends on the conditioning constraint subset $S \subseteq C$. Define
\begin{equation}
  f_t(S) \;:=\; \log P_T\bigl(y_t \,\big|\, q,\, S,\, y_{<t}\bigr).
  \label{eq:fS}
\end{equation}
Then $f_t$ is a real-valued function on the Boolean lattice $2^C$ ordered by inclusion. The two estimators we compare are
\begin{align}
  \delta_{\text{LA}}(t)  &\;=\; f_t(C) - f_t(\emptyset), \label{eq:LAdef}\\
  \delta_{\text{LOO}}(t) &\;=\; \sum_{i=1}^N \Bigl[\, f_t(C) - f_t\bigl(C \!\setminus\! \{c_i\}\bigr) \Bigr]. \label{eq:LOOdef}
\end{align}
We suppress the token index $t$ in the rest of this section; all derivations are pointwise per-token.

\subsection{M\"obius decomposition of \texorpdfstring{$f$}{f}}
\label{app:mobius:decomp}

The M\"obius inversion theorem for the Boolean lattice~\citep{rota1964mobius} states that any real-valued function $f$ on $2^C$ admits a unique decomposition
\begin{equation}
  \begin{aligned}
    f(S) \;&=\; \sum_{T \subseteq S} m(T), \\
    m(T) \;&=\; \sum_{U \subseteq T} (-1)^{|T| - |U|} \, f(U).
  \end{aligned}
  \label{eq:mobius_inversion}
\end{equation}
The coefficient $m(T)$ has a natural interpretation as the order-$|T|$ \emph{irreducible interaction} of the constraints in $T$:
\begin{itemize}\setlength\itemsep{0pt}
  \item $m(\emptyset) = f(\emptyset)$, the unconditioned log-probability;
  \item $m(\{c_i\}) = f(\{c_i\}) - f(\emptyset)$, the first-order main effect of constraint $c_i$ alone;
  \item $m(\{c_i, c_j\}) = f(\{c_i, c_j\}) - f(\{c_i\}) - f(\{c_j\}) + f(\emptyset)$, the second-order pairwise interaction between $c_i$ and $c_j$ that is not explained by either single-constraint effect;
  \item more generally, $m(T)$ captures the joint effect of the constraints in $T$ that no strict subset can account for.
\end{itemize}
The teacher's conditioning function is \emph{additive in constraints} if and only if $m(T) = 0$ for every $|T| \geq 2$. In that case, each constraint contributes a fixed log-probability shift independent of the others, and no multi-constraint interaction exists.

\subsection{LA and LOO sum as different weighted aggregations}
\label{app:mobius:lavsloo}

Applying Eq.~\eqref{eq:mobius_inversion} to LA:
\begin{equation}
  \begin{aligned}
    \delta_{\text{LA}} \;&=\; f(C) - f(\emptyset) \\
    &=\; \sum_{T \subseteq C} m(T) - m(\emptyset) \\
    &=\; \sum_{\emptyset \neq T \subseteq C} m(T).
  \end{aligned}
  \label{eq:LA_mobius}
\end{equation}
LA is thus the \emph{uniformly weighted} sum over all non-empty M\"obius coefficients. Applying the inversion to each leave-one-out term:
\begin{equation}
  \delta_{\text{LOO},i}
  \;=\; f(C) - f(C \!\setminus\! \{c_i\})
  \;=\; \sum_{T \subseteq C,\, c_i \in T} m(T).
  \label{eq:LOOi_mobius}
\end{equation}
Each $\delta_{\text{LOO},i}$ collects exactly those M\"obius coefficients whose subset $T$ contains $c_i$. Summing over $i$:
\begin{equation}
  \begin{aligned}
    \delta_{\text{LOO}}
    \;&=\; \sum_{i=1}^N \delta_{\text{LOO},i} \\
    &=\; \sum_{T \subseteq C} m(T) \cdot |\{i : c_i \in T\}| \\
    &=\; \sum_{\emptyset \neq T \subseteq C} |T| \cdot m(T).
  \end{aligned}
  \label{eq:LOO_mobius}
\end{equation}
LOO sum is therefore the \emph{cardinality-weighted} sum over the same non-empty M\"obius coefficients. The two estimators are aligned on each subset $T$ but weight it differently: LA assigns weight $1$, while LOO sum assigns weight $|T|$.

The residual is a clean signature of the higher-order interactions:
\begin{equation}
  \begin{aligned}
    \delta_{\text{LA}} - \delta_{\text{LOO}}
    \;&=\; \sum_{\emptyset \neq T \subseteq C} \bigl(1 - |T|\bigr)\, m(T) \\
    &=\; -\sum_{|T| \geq 2} \bigl(|T| - 1\bigr) \cdot m(T).
  \end{aligned}
  \label{eq:residual}
\end{equation}
Three properties follow immediately. (i) Singleton main effects $m(\{c_i\})$ contribute equally to both LA and LOO sum and therefore drop out of Eq.~\eqref{eq:residual}. (ii) A nonzero residual certifies that at least one higher-order coefficient is nonzero, whereas a zero residual does not rule out such interactions because signed terms can cancel. (iii) Coefficient by coefficient, LOO assigns an order-$k$ interaction $k$ times the weight assigned by LA; this does not imply that the aggregate residual must grow monotonically with interaction order because the coefficients are signed and the production reward is clipped.

\paragraph{NAO and RSS in the same framework.} The remaining two variants from \S\ref{sec:ablation} also admit clean M\"obius expansions. NAO computes $\delta_{\text{NAO}} = \sum_i [f(\{c_i\}) - f(\emptyset)] = \sum_{|T|=1} m(T)$, retaining only the order-$1$ main effects and discarding every higher-order interaction. This is the most aggressive truncation among the four variants and is therefore expected---but not guaranteed---to lose useful signal when higher-order interactions matter; the policy-level ablation in \S\ref{sec:ablation} tests that expectation.

For RSS, let $S_j$ denote the subset removed by job $j$, and let $k=|T|$ for a non-empty M\"obius subset $T$. A job includes $m(T)$ exactly when $T\cap S_j\neq\emptyset$. Under our sampling rule---first drawing $|S_j|$ uniformly from $\{1,\ldots,N-1\}$, then drawing a subset uniformly at that size---the inclusion probability is
\begin{equation}
  \begin{aligned}
    p_{N,k}
    &= 1-\frac{1}{N-1}\sum_{s=1}^{N-1}
        \frac{\binom{N-k}{s}}{\binom{N}{s}} \\
    &= \frac{Nk-1}{(N-1)(k+1)}.
  \end{aligned}
  \label{eq:rss_hit_probability}
\end{equation}
Because RSS sums $N$ independent jobs, the cumulative coefficient weight satisfies $W_T^{\mathrm{RSS}}\sim\mathrm{Binomial}(N,p_{N,k})$, with
\begin{equation}
  \mathbb{E}[W_T^{\mathrm{RSS}}]
  = w_{N,k}
  = \frac{N(Nk-1)}{(N-1)(k+1)}.
  \label{eq:rss_expected_weight}
\end{equation}
Hence, before clipping,
\begin{equation}
  \mathbb{E}[\delta_{\mathrm{RSS}}]
  = \sum_{\emptyset\neq T\subseteq C} w_{N,|T|}\,m(T).
  \label{eq:rss_expected_signal}
\end{equation}
The expected weight increases with interaction order but with diminishing increments: $w_{N,1}=N/2$, $w_{N,N}=N$, and $w_{N,k+1}-w_{N,k}=N(N+1)/[(N-1)(k+1)(k+2)]$. Relative to its singleton weight, RSS therefore increases from $1$ to $2$, rather than linearly with $k$. Importantly, the unnormalized $N$-job RSS sum does not interpolate in absolute scale between LA and LOO: $w_{N,k}\geq k$. RSS thus changes both interaction-order weighting and pre-clip reward scale, while also introducing subset-sampling variance. The full four-variant taxonomy is therefore: LA (uniform weight $1$), RSS (the expected weight in Eq.~\eqref{eq:rss_expected_weight}, with sampling variance), LOO (cardinality weight $|T|$ deterministically), and NAO (order-$1$ truncation).

\subsection{Empirical validation: the additivity hypothesis is violated}
\label{app:mobius:emp}

We measure the empirical residual at scale by forced-scoring the frozen teacher on the same student rollouts under all $N{+}2$ contexts $\{C\} \cup \{C \!\setminus\! \{c_i\}\}_{i=1}^{N} \cup \{\emptyset\}$. On $15{,}450$ HIR-16K samples with $|C| \in [5, 10]$ scored against a Qwen2.5-1.5B-Instruct student rollout under a Qwen2.5-7B teacher, Table~\ref{tab:mobius_residual} reports per-token statistics that summarize the gap.

\begin{table}[t]
  \centering
  \small
  \setlength{\tabcolsep}{4pt}
  \begin{tabular}{l r}
    \toprule
    \textbf{Metric} & \textbf{Value} \\
    \midrule
    Sample size (HIR-16K, $|C| \in [5, 10]$)         & $15{,}450$ \\
    Mean constraint count $\bar{N}$                       & $6.63$ \\
    \midrule
    $\|\delta_{\text{LA}} - \delta_{\text{LOO}}\|_1 \,/\, \|\delta_{\text{LA}}\|_1$ & $0.918$ \\
    Sign agreement $\mathrm{Pr}[\mathrm{sgn}(\delta_{\text{LA}}) = \mathrm{sgn}(\delta_{\text{LOO}})]$ & $0.681$ \\
    Signed Pearson correlation $\rho(\delta_{\text{LA}}, \delta_{\text{LOO}})$ & $0.665$ \\
    Top-$20$ token Jaccard overlap                    & $0.452$ \\
    $L_2$-norm ratio $\|\delta_{\text{LOO}}\|_2 \,/\, \|\delta_{\text{LA}}\|_2$ & $1.137$ \\
    \bottomrule
  \end{tabular}
  \caption{Per-token comparison of the LA and LOO sum estimators on $15{,}450$ HIR-16K samples. The relative $L_1$ residual measures the aggregate magnitude of $\sum_{|T| \geq 2} (|T|-1) m(T)$ relative to LA; sign agreement measures the fraction of tokens where the two estimators provide shaping signals in the same direction. All statistics are computed before clipping.}
  \label{tab:mobius_residual}
\end{table}

If the teacher response were strictly additive, Eq.~\eqref{eq:residual} would vanish. The observed relative $L_1$ residual of $0.918$ is nonzero and therefore rejects strict additivity for these measured signals; it quantifies the aggregate weighted residual, not the magnitudes of individual M\"obius coefficients. The signed Pearson correlation of $0.665$ and the $31.9$\% sign disagreement further show that LA and LOO differ in both scale and direction, while the $L_2$-norm ratio indicates that LOO's aggregate signal has a $13.7$\% larger norm in this sample. Symmetric clipping preserves the sign of nonzero values by construction, so clipping behavior is not treated as independent evidence about whether the residual is driven by outliers.

\subsection{What LOO retains beyond LA: the \texorpdfstring{$K\!\times\!T$}{KxT} sensitivity map}
\label{app:mobius:kt}

\begin{figure*}[!t]
  \centering
  \includegraphics[width=\textwidth]{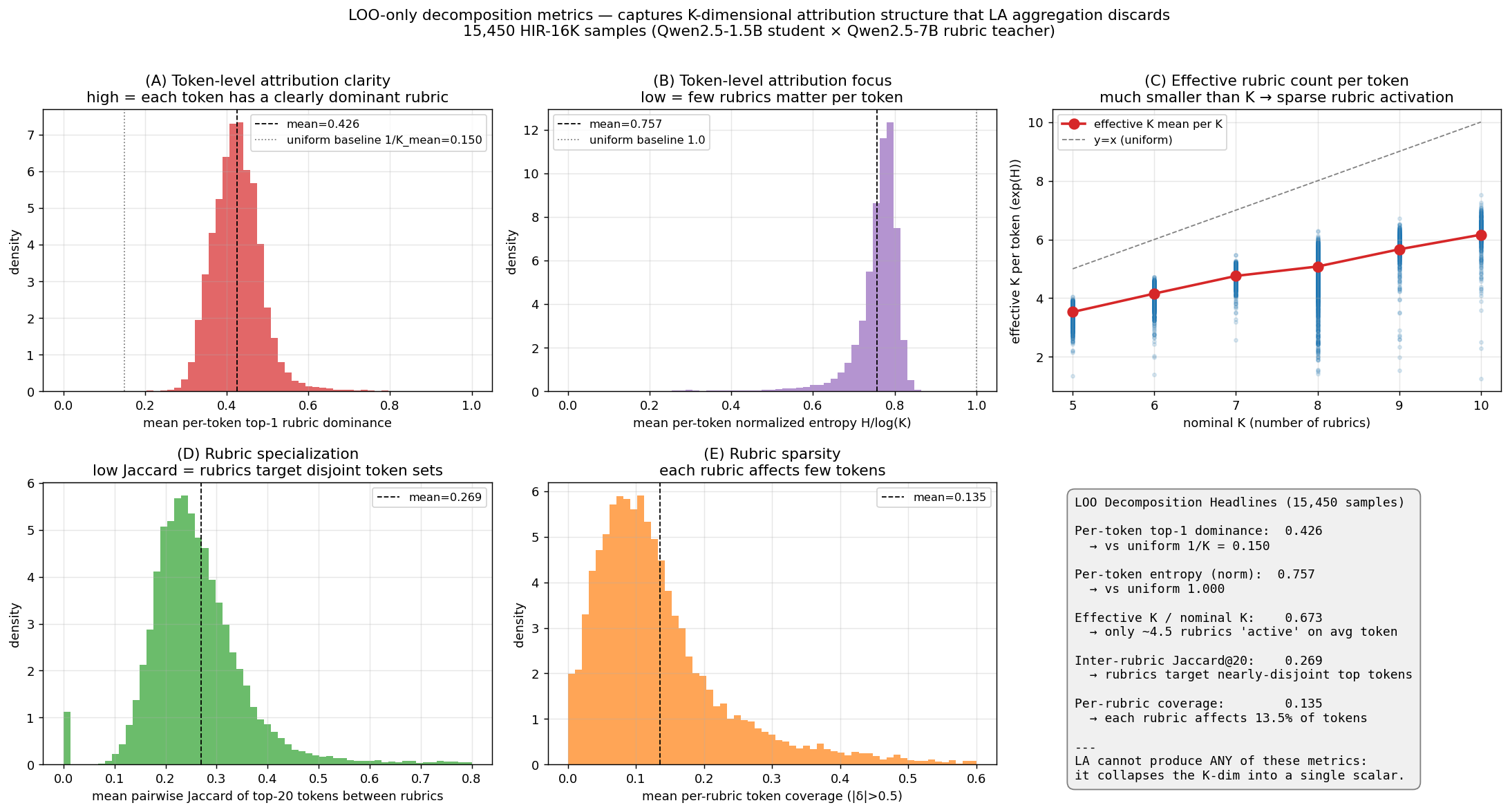}
  \caption{\textbf{Per-constraint sensitivity structure exposed by LOO.} Five per-token statistics of the $K\!\times\!T$ LOO sensitivity matrix on $15{,}450$ HIR-16K samples (Qwen2.5-1.5B student under Qwen2.5-7B teacher). Panels (A--B): histograms of per-token top-$1$ constraint dominance and per-token normalized entropy. Panel (C): effective constraint count per token versus nominal $K$. Panel (D): inter-constraint top-$20$ token Jaccard, measuring how distinctly different constraints localize. Panel (E): per-constraint token coverage at $|\delta| > 0.5$. The summary panel quantifies how each metric departs from the uniform-attribution baseline.}
  \label{fig:loo_decomp}
\end{figure*}

While both $\delta_{\text{LA}}$ and $\delta_{\text{LOO}}$ are $1\!\times\!T$ vectors after summation, LOO first constructs an \emph{unsummed} $K\!\times\!T$ matrix $\delta_{\text{LOO},i}(t)$. By Eq.~\eqref{eq:LOOi_mobius}, its $i$-th row collects the M\"obius coefficients of every subset that contains $c_i$, forming a per-token, per-constraint finite-difference sensitivity map. LA's single full-vs-empty contrast cannot recover this map: with only the two probe contexts $\{C, \emptyset\}$, LA produces one scalar at each token and collapses the entire $K$-dimensional identity structure.

Figure~\ref{fig:loo_decomp} reports five summary statistics of the per-token sensitivity map across the same $15{,}450$ samples. The map is concentrated rather than diffuse. On average $42.6$\% of the per-constraint signal mass at any single token comes from one dominant constraint (vs.\ $15.0$\% expected under uniform attribution), and an inverse-participation-ratio estimate of ``active constraints per token'' gives $4.46$ against a nominal mean of $6.63$ ($67$\% participation). The mean Jaccard similarity between pairs of per-constraint top-$20$ token sets is $0.269$. These five metrics jointly describe a structured, sparse sensitivity map, and none of them can be computed from LA alone. This row-wise concentration and the multi-constraint interactions measured in \S\ref{app:mobius:emp} characterize complementary aspects of the signal: the former describes how sensitivity is distributed across constraints at a token, whereas the latter measures non-additivity across conditioning subsets. While the $K\!\times\!T$ matrix is summed to a scalar before entering PPO, this per-constraint structure is what enables LOO's summed reward to weight higher-order interactions by $|T|$ (\S\ref{app:mobius:lavsloo}), an option LA's full-vs-empty contrast does not have.

\subsection{Cross-model comparison of counterfactual variants}
\label{app:mobius:crossmodel}

\begin{table*}[!t]
\centering
\footnotesize
\setlength{\tabcolsep}{6pt}
\renewcommand{\arraystretch}{1.1}
\begin{tabular}{l|cccc|cccc}
\toprule
& \multicolumn{4}{c|}{\textbf{Qwen2.5-1.5B}} & \multicolumn{4}{c}{\textbf{Qwen3-4B}} \\
\cmidrule(lr){2-5} \cmidrule(lr){6-9}
\textbf{Variant} & \textbf{IFEval} & \textbf{IFBench} & \textbf{MulDimIF All} & \textbf{MulDimIF L4} & \textbf{IFEval} & \textbf{IFBench} & \textbf{MulDim All} & \textbf{MulDim L4} \\
\midrule
\textbf{LOO} & \textbf{76.0} & 28.2          & \textbf{72.9} & \textbf{59.4} & \textbf{86.1} & 37.6          & \textbf{76.2} & \textbf{63.8} \\
LA                       & 75.0          & 27.1          & 71.0          & 57.1          & 85.4          & 34.2          & 74.6          & 62.0          \\
RSS                      & 75.5          & \textbf{28.8} & 72.1          & 57.7          & 85.6          & \textbf{38.0} & 76.0          & 62.4          \\
NAO                      & 71.2          & 26.5          & 66.4          & 45.3          & 84.2          & 30.2          & 71.4          & 53.4          \\
\bottomrule
\end{tabular}
\caption{Cross-model variant comparison on rule-based benchmarks. The four counterfactual variants (LOO/LA/RSS/NAO) are evaluated on IFEval, IFBench, and the MulDimIF Overall (MulDim All) and Level-$4$ (MulDim L4) splits for both model pairs. Within each pair, all variants share identical hyperparameters ($\lambda{=}2$, $c{=}5$, $\rho{=}1.0$, $1{,}500$ verl-steps); only the counterfactual structure differs. \textbf{Bold} marks the best score per column. The Qwen2.5 IFEval/IFBench/MulDimIF-Overall columns are a subset of Table~\ref{tab:ablation_variants}; the Qwen2.5 LOO/LA Level-$4$ entries reproduce Table~\ref{tab:loo_la_difficulty}, while the RSS/NAO Level-$4$ entries and all Qwen3 columns are reported here.}
\label{tab:variant_crossmodel_rule}
\end{table*}

Table~\ref{tab:variant_crossmodel_rule} compares the four counterfactual variants on IFEval, IFBench, and the MulDimIF Overall and Level-$4$ splits for both model pairs. For each pair, LOO has the highest score in three of the four cells (IFEval, MulDimIF Overall, and MulDimIF Level-$4$), while RSS has the highest score on IFBench.

\subsection{Theory--empirics bridge and limitations}
\label{app:mobius:bridge}

Taken together, the derivation and signal measurements show that LA and LOO apply different weights to the same interaction coefficients and that, relative to LA, LOO retains a per-constraint sensitivity map before summation. Token-level case studies illustrating these properties on individual HIR-16K samples are presented in Appendix~\ref{app:heatmap}. This analysis characterizes the \emph{reward geometry} presented to the optimizer, while the policy-level consequences are measured separately in \S\ref{sec:ablation}. The full M\"obius expansion has $2^N$ terms and is intractable to enumerate beyond very small constraint sets, so LA and LOO are best understood as two tractable projections of an exponentially large interaction lattice rather than as reconstructions of the lattice itself.

\paragraph{Reconciling structural and policy-level views.} The $+0.9$\,pp Avg gap between LOO and LA in Table~\ref{tab:ablation_variants} is much smaller than the $91.8$\,\% relative $L_1$ residual and $31.9$\,\% token-level sign disagreement. The two quantities describe different stages of the pipeline: reward geometry does not uniquely determine the policy reached by stochastic, clipped PPO, and benchmark-level metrics compress many token-level differences into a single response score. The evidence shows that LOO provides a structured per-constraint sensitivity map that LA cannot recover from one full-vs-empty contrast; the magnitude of the resulting policy-level gap remains an empirical outcome of optimization and evaluation. Determining when this signal-level distinction yields a larger policy-level gap remains an empirical question.

\subsection{Idealized implicit target distribution}
\label{app:mobius:target}

Beyond its M\"obius decomposition, the \method{} reward shape admits an idealized tokenwise interpretation. For a fixed decoding state $h_t$ and candidate vocabulary token $v$, define $\delta_i(v,t)$ by replacing the realized token $y_t$ in Eq.~\eqref{eq:delta_i} with $v$, and let $\Delta_c(v,t)=\mathrm{clip}(\sum_i\delta_i(v,t),-c,c)$. Maximizing the full-expectation objective $\lambda\,\mathbb{E}_{v\sim q}[\Delta_c(v,t)]-\KL(q\,\|\,\teacher)$ over $q$ yields
\begin{equation}
  q^*(v \mid h_t) \;\propto\; \teacher(v \mid h_t)
  \exp\!\bigl(\lambda\Delta_c(v,t)\bigr),
  \label{eq:implicit_target}
\end{equation}
the unique exponential-tilt solution. Its multiplicative form can be viewed as a product-of-experts-style target~\citep{hinton2002poe} driven by teacher-side finite differences rather than an external reward model. The M\"obius decomposition of \S\ref{app:mobius:lavsloo} shows that the unclipped sum inside $\Delta_c$ cardinality-weights the interaction coefficients; replacing it with LA would assign unit rather than cardinality weights and would not construct the pre-aggregation $K\!\times\!T$ sensitivity map. Equation~\eqref{eq:implicit_target} describes the idealized full-expectation objective; our experiments instead optimize the sampled-token reward with ratio-clipped PPO.

\section{Token-Level Reward Heatmap Case Studies}
\label{app:heatmap}

This appendix presents three representative HIR-16K samples that visualize, at the individual-token level, the structural difference between LOO and LA characterized in Appendix~\ref{app:mobius}. Each case shows two complementary views. \textbf{Section A} of each case lists \emph{per-constraint attribution snippets}. For the highest-magnitude constraints on that sample, we render short continuous-text excerpts where the per-constraint leave-one-out shift $\delta_{\text{LOO},i}(t) = \log P_T(y_t \mid q, C, y_{<t}) - \log P_T(y_t \mid q, C \!\setminus\! \{c_i\}, y_{<t})$ is largest in absolute value. Each token has a colored background reflecting its $\delta_{\text{LOO},i}(t)$ value. \textcolor[HTML]{cc0000}{Red} indicates that the leave-one-out context decreases the teacher log-probability of this token relative to the full constraint set (i.e., constraint $c_i$ \emph{boosts} this token), \textcolor[HTML]{0000cc}{blue} indicates the opposite, and intensity scales with $|\delta_{\text{LOO},i}(t)|$ capped at $\pm 8.0$. \textbf{Section B} of each case shows the same response colored \emph{twice} using the two $1\!\times\!T$ aggregated signals, namely the LOO sum $\sum_i \delta_{\text{LOO},i}(t)$ (red header) and the LA contrast $\delta_{\text{LA}}(t) = \log P_T(y_t \mid q, C, y_{<t}) - \log P_T(y_t \mid q, \emptyset, y_{<t})$ (blue header). Direct visual comparison of the two colored renderings shows where the two estimators agree on important tokens and, more often, where they diverge.

The three cases were selected by an offline LLM-verified screening procedure described at the end of this appendix. Together they illustrate three qualitatively distinct ways in which LOO's per-constraint signal supplies information that LA cannot recover. To accommodate the full-width tabular layout and the flowing color-coded responses, the case content below is set in a single-column block that may span multiple pages.

\onecolumn

\subsection{Case 1: Literal keyword and format-token alignment}
\label{app:heatmap:case1}

In the first case, the constraint set requires the response to include specific keywords (\textquotedbl{}Hardware Components\textquotedbl{}, \textquotedbl{}Components\textquotedbl{}, \textquotedbl{}Software Components\textquotedbl{}) and to format the answer as a numbered list with colon-separated descriptions. The per-constraint attribution snippets in Section~A below show that $\delta_{\text{LOO},i}$ for the keyword constraint peaks literally on the tokens \texttt{Hardware} and \texttt{Components} ($\delta = +11.25$ and $+8.57$ respectively), and that $\delta_{\text{LOO},i}$ for the numbered-list constraint peaks on the colon-newline token \texttt{:\textbackslash{}n} ($\delta = +19.99$) and on the digit \texttt{1} ($\delta = +16.54$). In Section~B, LOO sum mirrors these per-constraint peaks. The heaviest red tokens fall on \texttt{Hardware}, \texttt{Components}, \texttt{:\textbackslash{}n}, and \texttt{1}, while the LA rendering of the same response highlights instead a cluster of generic discourse markers (\texttt{or}, \texttt{essential}, sentence-final periods) and does not surface any of the four literal constraint-relevant tokens within its top-$15$. This case exemplifies the cleanest mode of disagreement. When a constraint demands a specific lexical or format token, the LOO signal localizes on that token, whereas LA, lacking access to per-constraint contrasts, settles on a less localized aggregate.

\medskip

\begingroup
\noindent\textbf{\large Sample 15400} \quad $K{=}8$ \quad $T{=}480$ \quad LOO/LA top-20 Jaccard$=0.18$\par\nopagebreak\medskip\nopagebreak
\noindent\textbf{Section A: Per-constraint attribution snippets} ($\textcolor[HTML]{cc0000}{\blacksquare}{=}+\delta$, $\textcolor[HTML]{0000cc}{\blacksquare}{=}-\delta$, intensity $\propto |\delta|$, capped at $\pm8.0$)\par\nopagebreak\smallskip\nopagebreak
\setlength{\fboxsep}{0.6pt}
\setlength{\tabcolsep}{4pt}
\renewcommand{\arraystretch}{1.15}
\begin{tabular}{@{}p{0.10\textwidth}@{\hspace{4pt}}p{0.85\textwidth}@{}}
\toprule
\textbf{C1}\\[1pt]{\scriptsize $|\delta|_{\max}{=}11.6$} & \textit{\scriptsize ``Confine word counts to 35 per sentence.''}\\[2pt]
{\scriptsize\textcolor{gray}{\dots}}\colorbox[HTML]{FFFFFF}{\textcolor{black}{\scriptsize\ttfamily\strut  data}}\colorbox[HTML]{F8F8FF}{\textcolor{black}{\scriptsize\ttfamily\strut  between}}\colorbox[HTML]{FDFDFF}{\textcolor{black}{\scriptsize\ttfamily\strut  different}}\colorbox[HTML]{FFFFFF}{\textcolor{black}{\scriptsize\ttfamily\strut  parts}}\colorbox[HTML]{FFFFFF}{\textcolor{black}{\scriptsize\ttfamily\strut  of}}\colorbox[HTML]{FFFFFF}{\textcolor{black}{\scriptsize\ttfamily\strut  the}}\colorbox[HTML]{F5F5FF}{\textcolor{black}{\scriptsize\ttfamily\strut  computer}}\colorbox[HTML]{FF4040}{\textcolor{white}{\scriptsize\ttfamily\strut .\textcolor{gray}{$\hookleftarrow$}}}\colorbox[HTML]{FFFFFF}{\textcolor{black}{\scriptsize\ttfamily\strut 2}}\colorbox[HTML]{FFFFFF}{\textcolor{black}{\scriptsize\ttfamily\strut .}}\colorbox[HTML]{FFFFFF}{\textcolor{black}{\scriptsize\ttfamily\strut  Random}}\colorbox[HTML]{FFFFFF}{\textcolor{black}{\scriptsize\ttfamily\strut  Access}}\colorbox[HTML]{FFFFFF}{\textcolor{black}{\scriptsize\ttfamily\strut  Memory}}\colorbox[HTML]{FFFFFF}{\textcolor{black}{\scriptsize\ttfamily\strut  (}}\colorbox[HTML]{FFFFFF}{\textcolor{black}{\scriptsize\ttfamily\strut RAM}}{\scriptsize\textcolor{gray}{\dots}}\\[1pt]
{\scriptsize\textcolor{gray}{\dots}}\colorbox[HTML]{FCFCFF}{\textcolor{black}{\scriptsize\ttfamily\strut  the}}\colorbox[HTML]{FFFFFF}{\textcolor{black}{\scriptsize\ttfamily\strut  overall}}\colorbox[HTML]{FFFCFC}{\textcolor{black}{\scriptsize\ttfamily\strut  functionality}}\colorbox[HTML]{F6F6FF}{\textcolor{black}{\scriptsize\ttfamily\strut  of}}\colorbox[HTML]{FFFFFF}{\textcolor{black}{\scriptsize\ttfamily\strut  the}}\colorbox[HTML]{FFFEFE}{\textcolor{black}{\scriptsize\ttfamily\strut  computer}}\colorbox[HTML]{FFCACA}{\textcolor{black}{\scriptsize\ttfamily\strut .}}\colorbox[HTML]{FF4040}{\textcolor{white}{\scriptsize\ttfamily\strut \textless{}\textbar{}im\_end\textbar{}\textgreater{}}}\\[1pt]\\
\addlinespace[2pt]
\textbf{C2}\\[1pt]{\scriptsize $|\delta|_{\max}{=}11.2$} & \textit{\scriptsize ``Be sure to mention \textquotedbl{}Hardware Components\textquotedbl{}, \textquotedbl{}Components\textquotedbl{}, \textquotedbl{}Software Components\textquotedbl{} in your response to highlight the key elements and add depth.''}\\[2pt]
\colorbox[HTML]{FF4040}{\textcolor{white}{\scriptsize\ttfamily\strut Hardware}}\colorbox[HTML]{FF4040}{\textcolor{white}{\scriptsize\ttfamily\strut  Components}}\colorbox[HTML]{FBFBFF}{\textcolor{black}{\scriptsize\ttfamily\strut :\textcolor{gray}{$\hookleftarrow$}}}\colorbox[HTML]{FFFFFF}{\textcolor{black}{\scriptsize\ttfamily\strut 1}}\colorbox[HTML]{FFFFFF}{\textcolor{black}{\scriptsize\ttfamily\strut .}}\colorbox[HTML]{F8F8FF}{\textcolor{black}{\scriptsize\ttfamily\strut  Central}}\colorbox[HTML]{FFFFFF}{\textcolor{black}{\scriptsize\ttfamily\strut  Processing}}\colorbox[HTML]{FFFFFF}{\textcolor{black}{\scriptsize\ttfamily\strut  Unit}}{\scriptsize\textcolor{gray}{\dots}}\\[1pt]
{\scriptsize\textcolor{gray}{\dots}}\colorbox[HTML]{FFFDFD}{\textcolor{black}{\scriptsize\ttfamily\strut  the}}\colorbox[HTML]{FFFFFF}{\textcolor{black}{\scriptsize\ttfamily\strut  overall}}\colorbox[HTML]{FFFDFD}{\textcolor{black}{\scriptsize\ttfamily\strut  functionality}}\colorbox[HTML]{FEFEFF}{\textcolor{black}{\scriptsize\ttfamily\strut  of}}\colorbox[HTML]{FFFFFF}{\textcolor{black}{\scriptsize\ttfamily\strut  the}}\colorbox[HTML]{FFFFFF}{\textcolor{black}{\scriptsize\ttfamily\strut  computer}}\colorbox[HTML]{FFFEFE}{\textcolor{black}{\scriptsize\ttfamily\strut .}}\colorbox[HTML]{FFA0A0}{\textcolor{black}{\scriptsize\ttfamily\strut \textless{}\textbar{}im\_end\textbar{}\textgreater{}}}\\[1pt]\\
\addlinespace[2pt]
\textbf{C3}\\[1pt]{\scriptsize $|\delta|_{\max}{=}20.0$} & \textit{\scriptsize ``Individual entries within the lists should be numbered (1., 2., 3., etc.) and followed by a colon and a brief description, as seen in the g…''}\\[2pt]
\colorbox[HTML]{5D5DFF}{\textcolor{white}{\scriptsize\ttfamily\strut Hardware}}\colorbox[HTML]{FFF3F3}{\textcolor{black}{\scriptsize\ttfamily\strut  Components}}\colorbox[HTML]{FF4040}{\textcolor{white}{\scriptsize\ttfamily\strut :\textcolor{gray}{$\hookleftarrow$}}}\colorbox[HTML]{FF4040}{\textcolor{white}{\scriptsize\ttfamily\strut 1}}\colorbox[HTML]{FFFFFF}{\textcolor{black}{\scriptsize\ttfamily\strut .}}\colorbox[HTML]{FFF7F7}{\textcolor{black}{\scriptsize\ttfamily\strut  Central}}\colorbox[HTML]{FFFFFF}{\textcolor{black}{\scriptsize\ttfamily\strut  Processing}}\colorbox[HTML]{FFFFFF}{\textcolor{black}{\scriptsize\ttfamily\strut  Unit}}\colorbox[HTML]{FFFFFF}{\textcolor{black}{\scriptsize\ttfamily\strut  (}}\colorbox[HTML]{FFFFFF}{\textcolor{black}{\scriptsize\ttfamily\strut CPU}}{\scriptsize\textcolor{gray}{\dots}}\\[1pt]
{\scriptsize\textcolor{gray}{\dots}}\colorbox[HTML]{FBFBFF}{\textcolor{black}{\scriptsize\ttfamily\strut  the}}\colorbox[HTML]{FFFFFF}{\textcolor{black}{\scriptsize\ttfamily\strut  overall}}\colorbox[HTML]{FCFCFF}{\textcolor{black}{\scriptsize\ttfamily\strut  functionality}}\colorbox[HTML]{FDFDFF}{\textcolor{black}{\scriptsize\ttfamily\strut  of}}\colorbox[HTML]{FFFFFF}{\textcolor{black}{\scriptsize\ttfamily\strut  the}}\colorbox[HTML]{FFFFFF}{\textcolor{black}{\scriptsize\ttfamily\strut  computer}}\colorbox[HTML]{FFFBFB}{\textcolor{black}{\scriptsize\ttfamily\strut .}}\colorbox[HTML]{8787FF}{\textcolor{black}{\scriptsize\ttfamily\strut \textless{}\textbar{}im\_end\textbar{}\textgreater{}}}\\[1pt]\\
\addlinespace[2pt]
\bottomrule
\end{tabular}
\par\medskip
\noindent\textbf{Section B: Full response} (same tokens shown twice, colored once by LOO sum, once by LA) $-$ vmax$=\pm8.0$\par\nopagebreak\smallskip\nopagebreak
\noindent\textbf{\textcolor[HTML]{c41e3a}{LOO sum}} {\scriptsize $|\delta|_{\max}{=}20.3$}: \\[2pt]
\sloppy
\colorbox[HTML]{FFDDDD}{\textcolor{black}{\tiny\ttfamily\strut Hardware}}\hskip 0pt plus 1pt\relax\allowbreak{}\colorbox[HTML]{FF4040}{\textcolor{white}{\tiny\ttfamily\strut  Components}}\hskip 0pt plus 1pt\relax\allowbreak{}\colorbox[HTML]{FF4040}{\textcolor{white}{\tiny\ttfamily\strut :\textcolor{gray}{$\hookleftarrow$}}}\hskip 0pt plus 1pt\relax\allowbreak{}\colorbox[HTML]{FF4040}{\textcolor{white}{\tiny\ttfamily\strut 1}}\hskip 0pt plus 1pt\relax\allowbreak{}\colorbox[HTML]{FFFFFF}{\textcolor{black}{\tiny\ttfamily\strut .}}\hskip 0pt plus 1pt\relax\allowbreak{}\colorbox[HTML]{FFFAFA}{\textcolor{black}{\tiny\ttfamily\strut  Central}}\hskip 0pt plus 1pt\relax\allowbreak{}\colorbox[HTML]{FFFFFF}{\textcolor{black}{\tiny\ttfamily\strut  Processing}}\hskip 0pt plus 1pt\relax\allowbreak{}\colorbox[HTML]{FFFFFF}{\textcolor{black}{\tiny\ttfamily\strut  Unit}}\hskip 0pt plus 1pt\relax\allowbreak{}\colorbox[HTML]{FFFFFF}{\textcolor{black}{\tiny\ttfamily\strut  (}}\hskip 0pt plus 1pt\relax\allowbreak{}\colorbox[HTML]{FFFFFF}{\textcolor{black}{\tiny\ttfamily\strut CPU}}\hskip 0pt plus 1pt\relax\allowbreak{}\colorbox[HTML]{FFA3A3}{\textcolor{black}{\tiny\ttfamily\strut ):}}\hskip 0pt plus 1pt\relax\allowbreak{}\colorbox[HTML]{FFEDED}{\textcolor{black}{\tiny\ttfamily\strut  The}}\hskip 0pt plus 1pt\relax\allowbreak{}\colorbox[HTML]{FFEDED}{\textcolor{black}{\tiny\ttfamily\strut  CPU}}\hskip 0pt plus 1pt\relax\allowbreak{}\colorbox[HTML]{DBDBFF}{\textcolor{black}{\tiny\ttfamily\strut ,}}\hskip 0pt plus 1pt\relax\allowbreak{}\colorbox[HTML]{9292FF}{\textcolor{black}{\tiny\ttfamily\strut  or}}\hskip 0pt plus 1pt\relax\allowbreak{}\colorbox[HTML]{FFCACA}{\textcolor{black}{\tiny\ttfamily\strut  the}}\hskip 0pt plus 1pt\relax\allowbreak{}\colorbox[HTML]{FFFCFC}{\textcolor{black}{\tiny\ttfamily\strut  brain}}\hskip 0pt plus 1pt\relax\allowbreak{}\colorbox[HTML]{FFFBFB}{\textcolor{black}{\tiny\ttfamily\strut  of}}\hskip 0pt plus 1pt\relax\allowbreak{}\colorbox[HTML]{FFFFFF}{\textcolor{black}{\tiny\ttfamily\strut  the}}\hskip 0pt plus 1pt\relax\allowbreak{}\colorbox[HTML]{FFFFFF}{\textcolor{black}{\tiny\ttfamily\strut  computer}}\hskip 0pt plus 1pt\relax\allowbreak{}\colorbox[HTML]{FFFFFF}{\textcolor{black}{\tiny\ttfamily\strut ,}}\hskip 0pt plus 1pt\relax\allowbreak{}\colorbox[HTML]{FFC6C6}{\textcolor{black}{\tiny\ttfamily\strut  is}}\hskip 0pt plus 1pt\relax\allowbreak{}\colorbox[HTML]{E7E7FF}{\textcolor{black}{\tiny\ttfamily\strut  the}}\hskip 0pt plus 1pt\relax\allowbreak{}\colorbox[HTML]{FFE2E2}{\textcolor{black}{\tiny\ttfamily\strut  component}}\hskip 0pt plus 1pt\relax\allowbreak{}\colorbox[HTML]{FDFDFF}{\textcolor{black}{\tiny\ttfamily\strut  that}}\hskip 0pt plus 1pt\relax\allowbreak{}\colorbox[HTML]{FFF9F9}{\textcolor{black}{\tiny\ttfamily\strut  executes}}\hskip 0pt plus 1pt\relax\allowbreak{}\colorbox[HTML]{FFFFFF}{\textcolor{black}{\tiny\ttfamily\strut  instructions}}\hskip 0pt plus 1pt\relax\allowbreak{}\colorbox[HTML]{FFEEEE}{\textcolor{black}{\tiny\ttfamily\strut ,}}\hskip 0pt plus 1pt\relax\allowbreak{}\colorbox[HTML]{FFF2F2}{\textcolor{black}{\tiny\ttfamily\strut  performs}}\hskip 0pt plus 1pt\relax\allowbreak{}\colorbox[HTML]{FFFEFE}{\textcolor{black}{\tiny\ttfamily\strut  calculations}}\hskip 0pt plus 1pt\relax\allowbreak{}\colorbox[HTML]{FFFFFF}{\textcolor{black}{\tiny\ttfamily\strut ,}}\hskip 0pt plus 1pt\relax\allowbreak{}\colorbox[HTML]{FFFFFF}{\textcolor{black}{\tiny\ttfamily\strut  and}}\hskip 0pt plus 1pt\relax\allowbreak{}\colorbox[HTML]{F5F5FF}{\textcolor{black}{\tiny\ttfamily\strut  controls}}\hskip 0pt plus 1pt\relax\allowbreak{}\colorbox[HTML]{FCFCFF}{\textcolor{black}{\tiny\ttfamily\strut  the}}\hskip 0pt plus 1pt\relax\allowbreak{}\colorbox[HTML]{FFECEC}{\textcolor{black}{\tiny\ttfamily\strut  flow}}\hskip 0pt plus 1pt\relax\allowbreak{}\colorbox[HTML]{FFFFFF}{\textcolor{black}{\tiny\ttfamily\strut  of}}\hskip 0pt plus 1pt\relax\allowbreak{}\colorbox[HTML]{FEFEFF}{\textcolor{black}{\tiny\ttfamily\strut  data}}\hskip 0pt plus 1pt\relax\allowbreak{}\colorbox[HTML]{ECECFF}{\textcolor{black}{\tiny\ttfamily\strut  between}}\hskip 0pt plus 1pt\relax\allowbreak{}\colorbox[HTML]{FFDEDE}{\textcolor{black}{\tiny\ttfamily\strut  different}}\hskip 0pt plus 1pt\relax\allowbreak{}\colorbox[HTML]{FFF8F8}{\textcolor{black}{\tiny\ttfamily\strut  parts}}\hskip 0pt plus 1pt\relax\allowbreak{}\colorbox[HTML]{FFFDFD}{\textcolor{black}{\tiny\ttfamily\strut  of}}\hskip 0pt plus 1pt\relax\allowbreak{}\colorbox[HTML]{FFFFFF}{\textcolor{black}{\tiny\ttfamily\strut  the}}\hskip 0pt plus 1pt\relax\allowbreak{}\colorbox[HTML]{E7E7FF}{\textcolor{black}{\tiny\ttfamily\strut  computer}}\hskip 0pt plus 1pt\relax\allowbreak{}\colorbox[HTML]{FF4040}{\textcolor{white}{\tiny\ttfamily\strut .\textcolor{gray}{$\hookleftarrow$}}}\hskip 0pt plus 1pt\relax\allowbreak{}\colorbox[HTML]{FFFFFF}{\textcolor{black}{\tiny\ttfamily\strut 2}}\hskip 0pt plus 1pt\relax\allowbreak{}\colorbox[HTML]{FFFFFF}{\textcolor{black}{\tiny\ttfamily\strut .}}\hskip 0pt plus 1pt\relax\allowbreak{}\colorbox[HTML]{FFFDFD}{\textcolor{black}{\tiny\ttfamily\strut  Random}}\hskip 0pt plus 1pt\relax\allowbreak{}\colorbox[HTML]{FFFFFF}{\textcolor{black}{\tiny\ttfamily\strut  Access}}\hskip 0pt plus 1pt\relax\allowbreak{}\colorbox[HTML]{FFFFFF}{\textcolor{black}{\tiny\ttfamily\strut  Memory}}\hskip 0pt plus 1pt\relax\allowbreak{}\colorbox[HTML]{FFFFFF}{\textcolor{black}{\tiny\ttfamily\strut  (}}\hskip 0pt plus 1pt\relax\allowbreak{}\colorbox[HTML]{FFFFFF}{\textcolor{black}{\tiny\ttfamily\strut RAM}}\hskip 0pt plus 1pt\relax\allowbreak{}\colorbox[HTML]{FFFFFF}{\textcolor{black}{\tiny\ttfamily\strut ):}}\hskip 0pt plus 1pt\relax\allowbreak{}\colorbox[HTML]{FFFDFD}{\textcolor{black}{\tiny\ttfamily\strut  RAM}}\hskip 0pt plus 1pt\relax\allowbreak{}\colorbox[HTML]{F8F8FF}{\textcolor{black}{\tiny\ttfamily\strut ,}}\hskip 0pt plus 1pt\relax\allowbreak{}\colorbox[HTML]{D9D9FF}{\textcolor{black}{\tiny\ttfamily\strut  or}}\hskip 0pt plus 1pt\relax\allowbreak{}\colorbox[HTML]{FFE5E5}{\textcolor{black}{\tiny\ttfamily\strut  the}}\hskip 0pt plus 1pt\relax\allowbreak{}\colorbox[HTML]{FFEAEA}{\textcolor{black}{\tiny\ttfamily\strut  working}}\hskip 0pt plus 1pt\relax\allowbreak{}\colorbox[HTML]{FFFFFF}{\textcolor{black}{\tiny\ttfamily\strut  memory}}\hskip 0pt plus 1pt\relax\allowbreak{}\colorbox[HTML]{FFFCFC}{\textcolor{black}{\tiny\ttfamily\strut ,}}\hskip 0pt plus 1pt\relax\allowbreak{}\colorbox[HTML]{F9F9FF}{\textcolor{black}{\tiny\ttfamily\strut  is}}\hskip 0pt plus 1pt\relax\allowbreak{}\colorbox[HTML]{FF8888}{\textcolor{black}{\tiny\ttfamily\strut  essential}}\hskip 0pt plus 1pt\relax\allowbreak{}\colorbox[HTML]{FFFDFD}{\textcolor{black}{\tiny\ttfamily\strut  for}}\hskip 0pt plus 1pt\relax\allowbreak{}\colorbox[HTML]{FFF8F8}{\textcolor{black}{\tiny\ttfamily\strut  storing}}\hskip 0pt plus 1pt\relax\allowbreak{}\colorbox[HTML]{FFFEFE}{\textcolor{black}{\tiny\ttfamily\strut  data}}\hskip 0pt plus 1pt\relax\allowbreak{}\colorbox[HTML]{FFF8F8}{\textcolor{black}{\tiny\ttfamily\strut  temporarily}}\hskip 0pt plus 1pt\relax\allowbreak{}\colorbox[HTML]{EEEEFF}{\textcolor{black}{\tiny\ttfamily\strut  while}}\hskip 0pt plus 1pt\relax\allowbreak{}\colorbox[HTML]{FCFCFF}{\textcolor{black}{\tiny\ttfamily\strut  the}}\hskip 0pt plus 1pt\relax\allowbreak{}\colorbox[HTML]{F2F2FF}{\textcolor{black}{\tiny\ttfamily\strut  computer}}\hskip 0pt plus 1pt\relax\allowbreak{}\colorbox[HTML]{FFF8F8}{\textcolor{black}{\tiny\ttfamily\strut  is}}\hskip 0pt plus 1pt\relax\allowbreak{}\colorbox[HTML]{FBFBFF}{\textcolor{black}{\tiny\ttfamily\strut  in}}\hskip 0pt plus 1pt\relax\allowbreak{}\colorbox[HTML]{F5F5FF}{\textcolor{black}{\tiny\ttfamily\strut  use}}\hskip 0pt plus 1pt\relax\allowbreak{}\colorbox[HTML]{5353FF}{\textcolor{white}{\tiny\ttfamily\strut .}}\hskip 0pt plus 1pt\relax\allowbreak{}\colorbox[HTML]{FFBEBE}{\textcolor{black}{\tiny\ttfamily\strut  A}}\hskip 0pt plus 1pt\relax\allowbreak{}\colorbox[HTML]{FFE2E2}{\textcolor{black}{\tiny\ttfamily\strut  larger}}\hskip 0pt plus 1pt\relax\allowbreak{}\colorbox[HTML]{FDFDFF}{\textcolor{black}{\tiny\ttfamily\strut  amount}}\hskip 0pt plus 1pt\relax\allowbreak{}\colorbox[HTML]{FFFFFF}{\textcolor{black}{\tiny\ttfamily\strut  of}}\hskip 0pt plus 1pt\relax\allowbreak{}\colorbox[HTML]{FFFFFF}{\textcolor{black}{\tiny\ttfamily\strut  RAM}}\hskip 0pt plus 1pt\relax\allowbreak{}\colorbox[HTML]{FFFCFC}{\textcolor{black}{\tiny\ttfamily\strut  can}}\hskip 0pt plus 1pt\relax\allowbreak{}\colorbox[HTML]{FFBCBC}{\textcolor{black}{\tiny\ttfamily\strut  significantly}}\hskip 0pt plus 1pt\relax\allowbreak{}\colorbox[HTML]{FFF6F6}{\textcolor{black}{\tiny\ttfamily\strut  enhance}}\hskip 0pt plus 1pt\relax\allowbreak{}\colorbox[HTML]{D8D8FF}{\textcolor{black}{\tiny\ttfamily\strut  the}}\hskip 0pt plus 1pt\relax\allowbreak{}\colorbox[HTML]{FFFCFC}{\textcolor{black}{\tiny\ttfamily\strut  computer}}\hskip 0pt plus 1pt\relax\allowbreak{}\colorbox[HTML]{FDFDFF}{\textcolor{black}{\tiny\ttfamily\strut 's}}\hskip 0pt plus 1pt\relax\allowbreak{}\colorbox[HTML]{D0D0FF}{\textcolor{black}{\tiny\ttfamily\strut  performance}}\hskip 0pt plus 1pt\relax\allowbreak{}\colorbox[HTML]{FFEFEF}{\textcolor{black}{\tiny\ttfamily\strut  by}}\hskip 0pt plus 1pt\relax\allowbreak{}\colorbox[HTML]{FFF9F9}{\textcolor{black}{\tiny\ttfamily\strut  allowing}}\hskip 0pt plus 1pt\relax\allowbreak{}\colorbox[HTML]{F0F0FF}{\textcolor{black}{\tiny\ttfamily\strut  it}}\hskip 0pt plus 1pt\relax\allowbreak{}\colorbox[HTML]{FFFFFF}{\textcolor{black}{\tiny\ttfamily\strut  to}}\hskip 0pt plus 1pt\relax\allowbreak{}\colorbox[HTML]{FFF8F8}{\textcolor{black}{\tiny\ttfamily\strut  handle}}\hskip 0pt plus 1pt\relax\allowbreak{}\colorbox[HTML]{FCFCFF}{\textcolor{black}{\tiny\ttfamily\strut  more}}\hskip 0pt plus 1pt\relax\allowbreak{}\colorbox[HTML]{F3F3FF}{\textcolor{black}{\tiny\ttfamily\strut  data}}\hskip 0pt plus 1pt\relax\allowbreak{}\colorbox[HTML]{FFF4F4}{\textcolor{black}{\tiny\ttfamily\strut  at}}\hskip 0pt plus 1pt\relax\allowbreak{}\colorbox[HTML]{FFFEFE}{\textcolor{black}{\tiny\ttfamily\strut  once}}\hskip 0pt plus 1pt\relax\allowbreak{}\colorbox[HTML]{FFCECE}{\textcolor{black}{\tiny\ttfamily\strut .\textcolor{gray}{$\hookleftarrow$}}}\hskip 0pt plus 1pt\relax\allowbreak{}\colorbox[HTML]{FFFCFC}{\textcolor{black}{\tiny\ttfamily\strut 3}}\hskip 0pt plus 1pt\relax\allowbreak{}\colorbox[HTML]{FFFFFF}{\textcolor{black}{\tiny\ttfamily\strut .}}\hskip 0pt plus 1pt\relax\allowbreak{}\colorbox[HTML]{FFD8D8}{\textcolor{black}{\tiny\ttfamily\strut  Hard}}\hskip 0pt plus 1pt\relax\allowbreak{}\colorbox[HTML]{FFFFFF}{\textcolor{black}{\tiny\ttfamily\strut  Disk}}\hskip 0pt plus 1pt\relax\allowbreak{}\colorbox[HTML]{FFFFFF}{\textcolor{black}{\tiny\ttfamily\strut  Drive}}\hskip 0pt plus 1pt\relax\allowbreak{}\colorbox[HTML]{FFFEFE}{\textcolor{black}{\tiny\ttfamily\strut  (}}\hskip 0pt plus 1pt\relax\allowbreak{}\colorbox[HTML]{FFFFFF}{\textcolor{black}{\tiny\ttfamily\strut H}}\hskip 0pt plus 1pt\relax\allowbreak{}\colorbox[HTML]{FFFFFF}{\textcolor{black}{\tiny\ttfamily\strut DD}}\hskip 0pt plus 1pt\relax\allowbreak{}\colorbox[HTML]{FEFEFF}{\textcolor{black}{\tiny\ttfamily\strut )}}\hskip 0pt plus 1pt\relax\allowbreak{}\colorbox[HTML]{F8F8FF}{\textcolor{black}{\tiny\ttfamily\strut  or}}\hskip 0pt plus 1pt\relax\allowbreak{}\colorbox[HTML]{FFFFFF}{\textcolor{black}{\tiny\ttfamily\strut  Solid}}\hskip 0pt plus 1pt\relax\allowbreak{}\colorbox[HTML]{FFFFFF}{\textcolor{black}{\tiny\ttfamily\strut  State}}\hskip 0pt plus 1pt\relax\allowbreak{}\colorbox[HTML]{FFFFFF}{\textcolor{black}{\tiny\ttfamily\strut  Drive}}\hskip 0pt plus 1pt\relax\allowbreak{}\colorbox[HTML]{FFFFFF}{\textcolor{black}{\tiny\ttfamily\strut  (}}\hskip 0pt plus 1pt\relax\allowbreak{}\colorbox[HTML]{FFFFFF}{\textcolor{black}{\tiny\ttfamily\strut SS}}\hskip 0pt plus 1pt\relax\allowbreak{}\colorbox[HTML]{FFFFFF}{\textcolor{black}{\tiny\ttfamily\strut D}}\hskip 0pt plus 1pt\relax\allowbreak{}\colorbox[HTML]{FFFFFF}{\textcolor{black}{\tiny\ttfamily\strut ):}}\hskip 0pt plus 1pt\relax\allowbreak{}\colorbox[HTML]{D4D4FF}{\textcolor{black}{\tiny\ttfamily\strut  The}}\hskip 0pt plus 1pt\relax\allowbreak{}\colorbox[HTML]{FFE6E6}{\textcolor{black}{\tiny\ttfamily\strut  HDD}}\hskip 0pt plus 1pt\relax\allowbreak{}\colorbox[HTML]{FFFDFD}{\textcolor{black}{\tiny\ttfamily\strut  or}}\hskip 0pt plus 1pt\relax\allowbreak{}\colorbox[HTML]{FFFFFF}{\textcolor{black}{\tiny\ttfamily\strut  SSD}}\hskip 0pt plus 1pt\relax\allowbreak{}\colorbox[HTML]{EAEAFF}{\textcolor{black}{\tiny\ttfamily\strut  is}}\hskip 0pt plus 1pt\relax\allowbreak{}\colorbox[HTML]{FFD5D5}{\textcolor{black}{\tiny\ttfamily\strut  used}}\hskip 0pt plus 1pt\relax\allowbreak{}\colorbox[HTML]{FAFAFF}{\textcolor{black}{\tiny\ttfamily\strut  to}}\hskip 0pt plus 1pt\relax\allowbreak{}\colorbox[HTML]{FFFFFF}{\textcolor{black}{\tiny\ttfamily\strut  store}}\hskip 0pt plus 1pt\relax\allowbreak{}\colorbox[HTML]{F0F0FF}{\textcolor{black}{\tiny\ttfamily\strut  the}}\hskip 0pt plus 1pt\relax\allowbreak{}\colorbox[HTML]{FFFAFA}{\textcolor{black}{\tiny\ttfamily\strut  computer}}\hskip 0pt plus 1pt\relax\allowbreak{}\colorbox[HTML]{FFFFFF}{\textcolor{black}{\tiny\ttfamily\strut 's}}\hskip 0pt plus 1pt\relax\allowbreak{}\colorbox[HTML]{F8F8FF}{\textcolor{black}{\tiny\ttfamily\strut  operating}}\hskip 0pt plus 1pt\relax\allowbreak{}\colorbox[HTML]{FFFFFF}{\textcolor{black}{\tiny\ttfamily\strut  system}}\hskip 0pt plus 1pt\relax\allowbreak{}\colorbox[HTML]{FFFFFF}{\textcolor{black}{\tiny\ttfamily\strut ,}}\hskip 0pt plus 1pt\relax\allowbreak{}\colorbox[HTML]{FFFDFD}{\textcolor{black}{\tiny\ttfamily\strut  applications}}\hskip 0pt plus 1pt\relax\allowbreak{}\colorbox[HTML]{FFFFFF}{\textcolor{black}{\tiny\ttfamily\strut ,}}\hskip 0pt plus 1pt\relax\allowbreak{}\colorbox[HTML]{FFFFFF}{\textcolor{black}{\tiny\ttfamily\strut  and}}\hskip 0pt plus 1pt\relax\allowbreak{}\colorbox[HTML]{FFFFFF}{\textcolor{black}{\tiny\ttfamily\strut  user}}\hskip 0pt plus 1pt\relax\allowbreak{}\colorbox[HTML]{FFFFFF}{\textcolor{black}{\tiny\ttfamily\strut  data}}\hskip 0pt plus 1pt\relax\allowbreak{}\colorbox[HTML]{FCFCFF}{\textcolor{black}{\tiny\ttfamily\strut .}}\hskip 0pt plus 1pt\relax\allowbreak{}\colorbox[HTML]{FFB0B0}{\textcolor{black}{\tiny\ttfamily\strut  A}}\hskip 0pt plus 1pt\relax\allowbreak{}\colorbox[HTML]{FFFDFD}{\textcolor{black}{\tiny\ttfamily\strut  faster}}\hskip 0pt plus 1pt\relax\allowbreak{}\colorbox[HTML]{FFF4F4}{\textcolor{black}{\tiny\ttfamily\strut  drive}}\hskip 0pt plus 1pt\relax\allowbreak{}\colorbox[HTML]{FFEBEB}{\textcolor{black}{\tiny\ttfamily\strut  can}}\hskip 0pt plus 1pt\relax\allowbreak{}\colorbox[HTML]{F9F9FF}{\textcolor{black}{\tiny\ttfamily\strut  improve}}\hskip 0pt plus 1pt\relax\allowbreak{}\colorbox[HTML]{FFE8E8}{\textcolor{black}{\tiny\ttfamily\strut  the}}\hskip 0pt plus 1pt\relax\allowbreak{}\colorbox[HTML]{FFFBFB}{\textcolor{black}{\tiny\ttfamily\strut  computer}}\hskip 0pt plus 1pt\relax\allowbreak{}\colorbox[HTML]{FFFFFF}{\textcolor{black}{\tiny\ttfamily\strut 's}}\hskip 0pt plus 1pt\relax\allowbreak{}\colorbox[HTML]{FFDEDE}{\textcolor{black}{\tiny\ttfamily\strut  overall}}\hskip 0pt plus 1pt\relax\allowbreak{}\colorbox[HTML]{F2F2FF}{\textcolor{black}{\tiny\ttfamily\strut  performance}}\hskip 0pt plus 1pt\relax\allowbreak{}\colorbox[HTML]{FFF3F3}{\textcolor{black}{\tiny\ttfamily\strut .\textcolor{gray}{$\hookleftarrow$}}}\hskip 0pt plus 1pt\relax\allowbreak{}\colorbox[HTML]{FFCFCF}{\textcolor{black}{\tiny\ttfamily\strut 4}}\hskip 0pt plus 1pt\relax\allowbreak{}\colorbox[HTML]{FFFFFF}{\textcolor{black}{\tiny\ttfamily\strut .}}\hskip 0pt plus 1pt\relax\allowbreak{}\colorbox[HTML]{FFEEEE}{\textcolor{black}{\tiny\ttfamily\strut  Graphics}}\hskip 0pt plus 1pt\relax\allowbreak{}\colorbox[HTML]{FFFFFF}{\textcolor{black}{\tiny\ttfamily\strut  Processing}}\hskip 0pt plus 1pt\relax\allowbreak{}\colorbox[HTML]{FFFFFF}{\textcolor{black}{\tiny\ttfamily\strut  Unit}}\hskip 0pt plus 1pt\relax\allowbreak{}\colorbox[HTML]{FFFFFF}{\textcolor{black}{\tiny\ttfamily\strut  (}}\hskip 0pt plus 1pt\relax\allowbreak{}\colorbox[HTML]{FFFFFF}{\textcolor{black}{\tiny\ttfamily\strut GPU}}\hskip 0pt plus 1pt\relax\allowbreak{}\colorbox[HTML]{FFFFFF}{\textcolor{black}{\tiny\ttfamily\strut ):}}\hskip 0pt plus 1pt\relax\allowbreak{}\colorbox[HTML]{FFFEFE}{\textcolor{black}{\tiny\ttfamily\strut  The}}\hskip 0pt plus 1pt\relax\allowbreak{}\colorbox[HTML]{FFFFFF}{\textcolor{black}{\tiny\ttfamily\strut  GPU}}\hskip 0pt plus 1pt\relax\allowbreak{}\colorbox[HTML]{FFF4F4}{\textcolor{black}{\tiny\ttfamily\strut  is}}\hskip 0pt plus 1pt\relax\allowbreak{}\colorbox[HTML]{FFA1A1}{\textcolor{black}{\tiny\ttfamily\strut  responsible}}\hskip 0pt plus 1pt\relax\allowbreak{}\colorbox[HTML]{FFFFFF}{\textcolor{black}{\tiny\ttfamily\strut  for}}\hskip 0pt plus 1pt\relax\allowbreak{}\colorbox[HTML]{FFFEFE}{\textcolor{black}{\tiny\ttfamily\strut  rendering}}\hskip 0pt plus 1pt\relax\allowbreak{}\colorbox[HTML]{FFDEDE}{\textcolor{black}{\tiny\ttfamily\strut  graphical}}\hskip 0pt plus 1pt\relax\allowbreak{}\colorbox[HTML]{FFF8F8}{\textcolor{black}{\tiny\ttfamily\strut  content}}\hskip 0pt plus 1pt\relax\allowbreak{}\colorbox[HTML]{FFF9F9}{\textcolor{black}{\tiny\ttfamily\strut  and}}\hskip 0pt plus 1pt\relax\allowbreak{}\colorbox[HTML]{FCFCFF}{\textcolor{black}{\tiny\ttfamily\strut  is}}\hskip 0pt plus 1pt\relax\allowbreak{}\colorbox[HTML]{FFFAFA}{\textcolor{black}{\tiny\ttfamily\strut  crucial}}\hskip 0pt plus 1pt\relax\allowbreak{}\colorbox[HTML]{FFFFFF}{\textcolor{black}{\tiny\ttfamily\strut  for}}\hskip 0pt plus 1pt\relax\allowbreak{}\colorbox[HTML]{FCFCFF}{\textcolor{black}{\tiny\ttfamily\strut  computer}}\hskip 0pt plus 1pt\relax\allowbreak{}\colorbox[HTML]{FFEAEA}{\textcolor{black}{\tiny\ttfamily\strut -generated}}\hskip 0pt plus 1pt\relax\allowbreak{}\colorbox[HTML]{FBFBFF}{\textcolor{black}{\tiny\ttfamily\strut  graphics}}\hskip 0pt plus 1pt\relax\allowbreak{}\colorbox[HTML]{FFF5F5}{\textcolor{black}{\tiny\ttfamily\strut  and}}\hskip 0pt plus 1pt\relax\allowbreak{}\colorbox[HTML]{FFFAFA}{\textcolor{black}{\tiny\ttfamily\strut  video}}\hskip 0pt plus 1pt\relax\allowbreak{}\colorbox[HTML]{E1E1FF}{\textcolor{black}{\tiny\ttfamily\strut  games}}\hskip 0pt plus 1pt\relax\allowbreak{}\colorbox[HTML]{FCFCFF}{\textcolor{black}{\tiny\ttfamily\strut .}}\hskip 0pt plus 1pt\relax\allowbreak{}\colorbox[HTML]{FFFCFC}{\textcolor{black}{\tiny\ttfamily\strut  A}}\hskip 0pt plus 1pt\relax\allowbreak{}\colorbox[HTML]{FFFDFD}{\textcolor{black}{\tiny\ttfamily\strut  strong}}\hskip 0pt plus 1pt\relax\allowbreak{}\colorbox[HTML]{FFFFFF}{\textcolor{black}{\tiny\ttfamily\strut  GPU}}\hskip 0pt plus 1pt\relax\allowbreak{}\colorbox[HTML]{FFFEFE}{\textcolor{black}{\tiny\ttfamily\strut  can}}\hskip 0pt plus 1pt\relax\allowbreak{}\colorbox[HTML]{FFEDED}{\textcolor{black}{\tiny\ttfamily\strut  greatly}}\hskip 0pt plus 1pt\relax\allowbreak{}\colorbox[HTML]{FFFFFF}{\textcolor{black}{\tiny\ttfamily\strut  enhance}}\hskip 0pt plus 1pt\relax\allowbreak{}\colorbox[HTML]{FFE9E9}{\textcolor{black}{\tiny\ttfamily\strut  the}}\hskip 0pt plus 1pt\relax\allowbreak{}\colorbox[HTML]{FFEFEF}{\textcolor{black}{\tiny\ttfamily\strut  computer}}\hskip 0pt plus 1pt\relax\allowbreak{}\colorbox[HTML]{FFFEFE}{\textcolor{black}{\tiny\ttfamily\strut 's}}\hskip 0pt plus 1pt\relax\allowbreak{}\colorbox[HTML]{E8E8FF}{\textcolor{black}{\tiny\ttfamily\strut  gaming}}\hskip 0pt plus 1pt\relax\allowbreak{}\colorbox[HTML]{EFEFFF}{\textcolor{black}{\tiny\ttfamily\strut  performance}}\hskip 0pt plus 1pt\relax\allowbreak{}\colorbox[HTML]{FFFEFE}{\textcolor{black}{\tiny\ttfamily\strut .\textcolor{gray}{$\hookleftarrow$}}}\hskip 0pt plus 1pt\relax\allowbreak{}\colorbox[HTML]{FFD9D9}{\textcolor{black}{\tiny\ttfamily\strut 5}}\hskip 0pt plus 1pt\relax\allowbreak{}\colorbox[HTML]{FFFFFF}{\textcolor{black}{\tiny\ttfamily\strut .}}\hskip 0pt plus 1pt\relax\allowbreak{}\colorbox[HTML]{F0F0FF}{\textcolor{black}{\tiny\ttfamily\strut  Mother}}\hskip 0pt plus 1pt\relax\allowbreak{}\colorbox[HTML]{FFFFFF}{\textcolor{black}{\tiny\ttfamily\strut board}}\hskip 0pt plus 1pt\relax\allowbreak{}\colorbox[HTML]{FFFFFF}{\textcolor{black}{\tiny\ttfamily\strut :}}\hskip 0pt plus 1pt\relax\allowbreak{}\colorbox[HTML]{F9F9FF}{\textcolor{black}{\tiny\ttfamily\strut  The}}\hskip 0pt plus 1pt\relax\allowbreak{}\colorbox[HTML]{FFFFFF}{\textcolor{black}{\tiny\ttfamily\strut  motherboard}}\hskip 0pt plus 1pt\relax\allowbreak{}\colorbox[HTML]{F7F7FF}{\textcolor{black}{\tiny\ttfamily\strut  is}}\hskip 0pt plus 1pt\relax\allowbreak{}\colorbox[HTML]{FFFAFA}{\textcolor{black}{\tiny\ttfamily\strut  the}}\hskip 0pt plus 1pt\relax\allowbreak{}\colorbox[HTML]{FFF8F8}{\textcolor{black}{\tiny\ttfamily\strut  backbone}}\hskip 0pt plus 1pt\relax\allowbreak{}\colorbox[HTML]{FFFDFD}{\textcolor{black}{\tiny\ttfamily\strut  of}}\hskip 0pt plus 1pt\relax\allowbreak{}\colorbox[HTML]{FFFDFD}{\textcolor{black}{\tiny\ttfamily\strut  the}}\hskip 0pt plus 1pt\relax\allowbreak{}\colorbox[HTML]{FFFFFF}{\textcolor{black}{\tiny\ttfamily\strut  computer}}\hskip 0pt plus 1pt\relax\allowbreak{}\colorbox[HTML]{FFFEFE}{\textcolor{black}{\tiny\ttfamily\strut ,}}\hskip 0pt plus 1pt\relax\allowbreak{}\colorbox[HTML]{FFFEFE}{\textcolor{black}{\tiny\ttfamily\strut  connecting}}\hskip 0pt plus 1pt\relax\allowbreak{}\colorbox[HTML]{FCFCFF}{\textcolor{black}{\tiny\ttfamily\strut  all}}\hskip 0pt plus 1pt\relax\allowbreak{}\colorbox[HTML]{FFFEFE}{\textcolor{black}{\tiny\ttfamily\strut  the}}\hskip 0pt plus 1pt\relax\allowbreak{}\colorbox[HTML]{F9F9FF}{\textcolor{black}{\tiny\ttfamily\strut  hardware}}\hskip 0pt plus 1pt\relax\allowbreak{}\colorbox[HTML]{FFFFFF}{\textcolor{black}{\tiny\ttfamily\strut  components}}\hskip 0pt plus 1pt\relax\allowbreak{}\colorbox[HTML]{FFFEFE}{\textcolor{black}{\tiny\ttfamily\strut .}}\hskip 0pt plus 1pt\relax\allowbreak{}\colorbox[HTML]{F9F9FF}{\textcolor{black}{\tiny\ttfamily\strut  It}}\hskip 0pt plus 1pt\relax\allowbreak{}\colorbox[HTML]{FFE1E1}{\textcolor{black}{\tiny\ttfamily\strut  facilitates}}\hskip 0pt plus 1pt\relax\allowbreak{}\colorbox[HTML]{FFF1F1}{\textcolor{black}{\tiny\ttfamily\strut  communication}}\hskip 0pt plus 1pt\relax\allowbreak{}\colorbox[HTML]{FFFAFA}{\textcolor{black}{\tiny\ttfamily\strut  between}}\hskip 0pt plus 1pt\relax\allowbreak{}\colorbox[HTML]{FDFDFF}{\textcolor{black}{\tiny\ttfamily\strut  the}}\hskip 0pt plus 1pt\relax\allowbreak{}\colorbox[HTML]{FEFEFF}{\textcolor{black}{\tiny\ttfamily\strut  CPU}}\hskip 0pt plus 1pt\relax\allowbreak{}\colorbox[HTML]{FEFEFF}{\textcolor{black}{\tiny\ttfamily\strut ,}}\hskip 0pt plus 1pt\relax\allowbreak{}\colorbox[HTML]{FFFDFD}{\textcolor{black}{\tiny\ttfamily\strut  RAM}}\hskip 0pt plus 1pt\relax\allowbreak{}\colorbox[HTML]{FFFFFF}{\textcolor{black}{\tiny\ttfamily\strut ,}}\hskip 0pt plus 1pt\relax\allowbreak{}\colorbox[HTML]{FFF0F0}{\textcolor{black}{\tiny\ttfamily\strut  and}}\hskip 0pt plus 1pt\relax\allowbreak{}\colorbox[HTML]{FDFDFF}{\textcolor{black}{\tiny\ttfamily\strut  other}}\hskip 0pt plus 1pt\relax\allowbreak{}\colorbox[HTML]{F8F8FF}{\textcolor{black}{\tiny\ttfamily\strut  peripherals}}\hskip 0pt plus 1pt\relax\allowbreak{}\colorbox[HTML]{CCCCFF}{\textcolor{black}{\tiny\ttfamily\strut .\textcolor{gray}{$\hookleftarrow$}}}\hskip 0pt plus 1pt\relax\allowbreak{}\colorbox[HTML]{F6F6FF}{\textcolor{black}{\tiny\ttfamily\strut Software}}\hskip 0pt plus 1pt\relax\allowbreak{}\colorbox[HTML]{FFFFFF}{\textcolor{black}{\tiny\ttfamily\strut  Components}}\hskip 0pt plus 1pt\relax\allowbreak{}\colorbox[HTML]{FFFFFF}{\textcolor{black}{\tiny\ttfamily\strut :\textcolor{gray}{$\hookleftarrow$}}}\hskip 0pt plus 1pt\relax\allowbreak{}\colorbox[HTML]{FDFDFF}{\textcolor{black}{\tiny\ttfamily\strut 1}}\hskip 0pt plus 1pt\relax\allowbreak{}\colorbox[HTML]{FFFFFF}{\textcolor{black}{\tiny\ttfamily\strut .}}\hskip 0pt plus 1pt\relax\allowbreak{}\colorbox[HTML]{FFFFFF}{\textcolor{black}{\tiny\ttfamily\strut  Operating}}\hskip 0pt plus 1pt\relax\allowbreak{}\colorbox[HTML]{FFFFFF}{\textcolor{black}{\tiny\ttfamily\strut  System}}\hskip 0pt plus 1pt\relax\allowbreak{}\colorbox[HTML]{FFFFFF}{\textcolor{black}{\tiny\ttfamily\strut  (}}\hskip 0pt plus 1pt\relax\allowbreak{}\colorbox[HTML]{FFFFFF}{\textcolor{black}{\tiny\ttfamily\strut OS}}\hskip 0pt plus 1pt\relax\allowbreak{}\colorbox[HTML]{FFFFFF}{\textcolor{black}{\tiny\ttfamily\strut ):}}\hskip 0pt plus 1pt\relax\allowbreak{}\colorbox[HTML]{FFFFFF}{\textcolor{black}{\tiny\ttfamily\strut  The}}\hskip 0pt plus 1pt\relax\allowbreak{}\colorbox[HTML]{FFFFFF}{\textcolor{black}{\tiny\ttfamily\strut  OS}}\hskip 0pt plus 1pt\relax\allowbreak{}\colorbox[HTML]{F2F2FF}{\textcolor{black}{\tiny\ttfamily\strut ,}}\hskip 0pt plus 1pt\relax\allowbreak{}\colorbox[HTML]{EAEAFF}{\textcolor{black}{\tiny\ttfamily\strut  such}}\hskip 0pt plus 1pt\relax\allowbreak{}\colorbox[HTML]{FFFFFF}{\textcolor{black}{\tiny\ttfamily\strut  as}}\hskip 0pt plus 1pt\relax\allowbreak{}\colorbox[HTML]{FFFFFF}{\textcolor{black}{\tiny\ttfamily\strut  Windows}}\hskip 0pt plus 1pt\relax\allowbreak{}\colorbox[HTML]{CFCFFF}{\textcolor{black}{\tiny\ttfamily\strut ,}}\hskip 0pt plus 1pt\relax\allowbreak{}\colorbox[HTML]{FEFEFF}{\textcolor{black}{\tiny\ttfamily\strut  macOS}}\hskip 0pt plus 1pt\relax\allowbreak{}\colorbox[HTML]{FFFFFF}{\textcolor{black}{\tiny\ttfamily\strut ,}}\hskip 0pt plus 1pt\relax\allowbreak{}\colorbox[HTML]{FFFFFF}{\textcolor{black}{\tiny\ttfamily\strut  or}}\hskip 0pt plus 1pt\relax\allowbreak{}\colorbox[HTML]{FFFFFF}{\textcolor{black}{\tiny\ttfamily\strut  Linux}}\hskip 0pt plus 1pt\relax\allowbreak{}\colorbox[HTML]{FFFFFF}{\textcolor{black}{\tiny\ttfamily\strut ,}}\hskip 0pt plus 1pt\relax\allowbreak{}\colorbox[HTML]{F6F6FF}{\textcolor{black}{\tiny\ttfamily\strut  is}}\hskip 0pt plus 1pt\relax\allowbreak{}\colorbox[HTML]{FFFDFD}{\textcolor{black}{\tiny\ttfamily\strut  the}}\hskip 0pt plus 1pt\relax\allowbreak{}\colorbox[HTML]{FFFDFD}{\textcolor{black}{\tiny\ttfamily\strut  software}}\hskip 0pt plus 1pt\relax\allowbreak{}\colorbox[HTML]{FFFDFD}{\textcolor{black}{\tiny\ttfamily\strut  that}}\hskip 0pt plus 1pt\relax\allowbreak{}\colorbox[HTML]{FFFFFF}{\textcolor{black}{\tiny\ttfamily\strut  manages}}\hskip 0pt plus 1pt\relax\allowbreak{}\colorbox[HTML]{FFF7F7}{\textcolor{black}{\tiny\ttfamily\strut  the}}\hskip 0pt plus 1pt\relax\allowbreak{}\colorbox[HTML]{F5F5FF}{\textcolor{black}{\tiny\ttfamily\strut  computer}}\hskip 0pt plus 1pt\relax\allowbreak{}\colorbox[HTML]{FFFEFE}{\textcolor{black}{\tiny\ttfamily\strut 's}}\hskip 0pt plus 1pt\relax\allowbreak{}\colorbox[HTML]{FDFDFF}{\textcolor{black}{\tiny\ttfamily\strut  hardware}}\hskip 0pt plus 1pt\relax\allowbreak{}\colorbox[HTML]{FFFAFA}{\textcolor{black}{\tiny\ttfamily\strut  resources}}\hskip 0pt plus 1pt\relax\allowbreak{}\colorbox[HTML]{FFFEFE}{\textcolor{black}{\tiny\ttfamily\strut  and}}\hskip 0pt plus 1pt\relax\allowbreak{}\colorbox[HTML]{FEFEFF}{\textcolor{black}{\tiny\ttfamily\strut  provides}}\hskip 0pt plus 1pt\relax\allowbreak{}\colorbox[HTML]{FFFEFE}{\textcolor{black}{\tiny\ttfamily\strut  a}}\hskip 0pt plus 1pt\relax\allowbreak{}\colorbox[HTML]{FFF7F7}{\textcolor{black}{\tiny\ttfamily\strut  user}}\hskip 0pt plus 1pt\relax\allowbreak{}\colorbox[HTML]{FFFFFF}{\textcolor{black}{\tiny\ttfamily\strut  interface}}\hskip 0pt plus 1pt\relax\allowbreak{}\colorbox[HTML]{E1E1FF}{\textcolor{black}{\tiny\ttfamily\strut  for}}\hskip 0pt plus 1pt\relax\allowbreak{}\colorbox[HTML]{F8F8FF}{\textcolor{black}{\tiny\ttfamily\strut  interacting}}\hskip 0pt plus 1pt\relax\allowbreak{}\colorbox[HTML]{FFFFFF}{\textcolor{black}{\tiny\ttfamily\strut  with}}\hskip 0pt plus 1pt\relax\allowbreak{}\colorbox[HTML]{FFFEFE}{\textcolor{black}{\tiny\ttfamily\strut  the}}\hskip 0pt plus 1pt\relax\allowbreak{}\colorbox[HTML]{FFF2F2}{\textcolor{black}{\tiny\ttfamily\strut  computer}}\hskip 0pt plus 1pt\relax\allowbreak{}\colorbox[HTML]{F6F6FF}{\textcolor{black}{\tiny\ttfamily\strut .\textcolor{gray}{$\hookleftarrow$}}}\hskip 0pt plus 1pt\relax\allowbreak{}\colorbox[HTML]{FFFFFF}{\textcolor{black}{\tiny\ttfamily\strut 2}}\hskip 0pt plus 1pt\relax\allowbreak{}\colorbox[HTML]{FFFFFF}{\textcolor{black}{\tiny\ttfamily\strut .}}\hskip 0pt plus 1pt\relax\allowbreak{}\colorbox[HTML]{F7F7FF}{\textcolor{black}{\tiny\ttfamily\strut  Applications}}\hskip 0pt plus 1pt\relax\allowbreak{}\colorbox[HTML]{FFFEFE}{\textcolor{black}{\tiny\ttfamily\strut :}}\hskip 0pt plus 1pt\relax\allowbreak{}\colorbox[HTML]{FFF9F9}{\textcolor{black}{\tiny\ttfamily\strut  Applications}}\hskip 0pt plus 1pt\relax\allowbreak{}\colorbox[HTML]{FFF6F6}{\textcolor{black}{\tiny\ttfamily\strut ,}}\hskip 0pt plus 1pt\relax\allowbreak{}\colorbox[HTML]{FFE1E1}{\textcolor{black}{\tiny\ttfamily\strut  such}}\hskip 0pt plus 1pt\relax\allowbreak{}\colorbox[HTML]{FFFFFF}{\textcolor{black}{\tiny\ttfamily\strut  as}}\hskip 0pt plus 1pt\relax\allowbreak{}\colorbox[HTML]{FFFCFC}{\textcolor{black}{\tiny\ttfamily\strut  word}}\hskip 0pt plus 1pt\relax\allowbreak{}\colorbox[HTML]{FFFFFF}{\textcolor{black}{\tiny\ttfamily\strut  processors}}\hskip 0pt plus 1pt\relax\allowbreak{}\colorbox[HTML]{FFFDFD}{\textcolor{black}{\tiny\ttfamily\strut ,}}\hskip 0pt plus 1pt\relax\allowbreak{}\colorbox[HTML]{F4F4FF}{\textcolor{black}{\tiny\ttfamily\strut  web}}\hskip 0pt plus 1pt\relax\allowbreak{}\colorbox[HTML]{FFFFFF}{\textcolor{black}{\tiny\ttfamily\strut  browsers}}\hskip 0pt plus 1pt\relax\allowbreak{}\colorbox[HTML]{FFFFFF}{\textcolor{black}{\tiny\ttfamily\strut ,}}\hskip 0pt plus 1pt\relax\allowbreak{}\colorbox[HTML]{FFFFFF}{\textcolor{black}{\tiny\ttfamily\strut  and}}\hskip 0pt plus 1pt\relax\allowbreak{}\colorbox[HTML]{E9E9FF}{\textcolor{black}{\tiny\ttfamily\strut  games}}\hskip 0pt plus 1pt\relax\allowbreak{}\colorbox[HTML]{FFFFFF}{\textcolor{black}{\tiny\ttfamily\strut ,}}\hskip 0pt plus 1pt\relax\allowbreak{}\colorbox[HTML]{FBFBFF}{\textcolor{black}{\tiny\ttfamily\strut  run}}\hskip 0pt plus 1pt\relax\allowbreak{}\colorbox[HTML]{FFFFFF}{\textcolor{black}{\tiny\ttfamily\strut  on}}\hskip 0pt plus 1pt\relax\allowbreak{}\colorbox[HTML]{FEFEFF}{\textcolor{black}{\tiny\ttfamily\strut  top}}\hskip 0pt plus 1pt\relax\allowbreak{}\colorbox[HTML]{FFFFFF}{\textcolor{black}{\tiny\ttfamily\strut  of}}\hskip 0pt plus 1pt\relax\allowbreak{}\colorbox[HTML]{FFFFFF}{\textcolor{black}{\tiny\ttfamily\strut  the}}\hskip 0pt plus 1pt\relax\allowbreak{}\colorbox[HTML]{FFFBFB}{\textcolor{black}{\tiny\ttfamily\strut  OS}}\hskip 0pt plus 1pt\relax\allowbreak{}\colorbox[HTML]{FFF3F3}{\textcolor{black}{\tiny\ttfamily\strut  and}}\hskip 0pt plus 1pt\relax\allowbreak{}\colorbox[HTML]{FEFEFF}{\textcolor{black}{\tiny\ttfamily\strut  provide}}\hskip 0pt plus 1pt\relax\allowbreak{}\colorbox[HTML]{FDFDFF}{\textcolor{black}{\tiny\ttfamily\strut  specific}}\hskip 0pt plus 1pt\relax\allowbreak{}\colorbox[HTML]{E5E5FF}{\textcolor{black}{\tiny\ttfamily\strut  functionality}}\hskip 0pt plus 1pt\relax\allowbreak{}\colorbox[HTML]{FCFCFF}{\textcolor{black}{\tiny\ttfamily\strut  to}}\hskip 0pt plus 1pt\relax\allowbreak{}\colorbox[HTML]{FEFEFF}{\textcolor{black}{\tiny\ttfamily\strut  the}}\hskip 0pt plus 1pt\relax\allowbreak{}\colorbox[HTML]{FFFFFF}{\textcolor{black}{\tiny\ttfamily\strut  user}}\hskip 0pt plus 1pt\relax\allowbreak{}\colorbox[HTML]{E2E2FF}{\textcolor{black}{\tiny\ttfamily\strut .\textcolor{gray}{$\hookleftarrow$}}}\hskip 0pt plus 1pt\relax\allowbreak{}\colorbox[HTML]{FFFFFF}{\textcolor{black}{\tiny\ttfamily\strut 3}}\hskip 0pt plus 1pt\relax\allowbreak{}\colorbox[HTML]{FFFFFF}{\textcolor{black}{\tiny\ttfamily\strut .}}\hskip 0pt plus 1pt\relax\allowbreak{}\colorbox[HTML]{FFFEFE}{\textcolor{black}{\tiny\ttfamily\strut  Drivers}}\hskip 0pt plus 1pt\relax\allowbreak{}\colorbox[HTML]{FFFFFF}{\textcolor{black}{\tiny\ttfamily\strut :}}\hskip 0pt plus 1pt\relax\allowbreak{}\colorbox[HTML]{FFFFFF}{\textcolor{black}{\tiny\ttfamily\strut  Drivers}}\hskip 0pt plus 1pt\relax\allowbreak{}\colorbox[HTML]{F6F6FF}{\textcolor{black}{\tiny\ttfamily\strut  are}}\hskip 0pt plus 1pt\relax\allowbreak{}\colorbox[HTML]{FDFDFF}{\textcolor{black}{\tiny\ttfamily\strut  software}}\hskip 0pt plus 1pt\relax\allowbreak{}\colorbox[HTML]{FFFEFE}{\textcolor{black}{\tiny\ttfamily\strut  that}}\hskip 0pt plus 1pt\relax\allowbreak{}\colorbox[HTML]{FBFBFF}{\textcolor{black}{\tiny\ttfamily\strut  enable}}\hskip 0pt plus 1pt\relax\allowbreak{}\colorbox[HTML]{FFFCFC}{\textcolor{black}{\tiny\ttfamily\strut  hardware}}\hskip 0pt plus 1pt\relax\allowbreak{}\colorbox[HTML]{FFFFFF}{\textcolor{black}{\tiny\ttfamily\strut  components}}\hskip 0pt plus 1pt\relax\allowbreak{}\colorbox[HTML]{FFFFFF}{\textcolor{black}{\tiny\ttfamily\strut  to}}\hskip 0pt plus 1pt\relax\allowbreak{}\colorbox[HTML]{FFFFFF}{\textcolor{black}{\tiny\ttfamily\strut  communicate}}\hskip 0pt plus 1pt\relax\allowbreak{}\colorbox[HTML]{FFFCFC}{\textcolor{black}{\tiny\ttfamily\strut  with}}\hskip 0pt plus 1pt\relax\allowbreak{}\colorbox[HTML]{FFFFFF}{\textcolor{black}{\tiny\ttfamily\strut  the}}\hskip 0pt plus 1pt\relax\allowbreak{}\colorbox[HTML]{FFE0E0}{\textcolor{black}{\tiny\ttfamily\strut  operating}}\hskip 0pt plus 1pt\relax\allowbreak{}\colorbox[HTML]{FFFFFF}{\textcolor{black}{\tiny\ttfamily\strut  system}}\hskip 0pt plus 1pt\relax\allowbreak{}\colorbox[HTML]{FAFAFF}{\textcolor{black}{\tiny\ttfamily\strut .}}\hskip 0pt plus 1pt\relax\allowbreak{}\colorbox[HTML]{FFF1F1}{\textcolor{black}{\tiny\ttfamily\strut  Without}}\hskip 0pt plus 1pt\relax\allowbreak{}\colorbox[HTML]{F0F0FF}{\textcolor{black}{\tiny\ttfamily\strut  drivers}}\hskip 0pt plus 1pt\relax\allowbreak{}\colorbox[HTML]{FFFFFF}{\textcolor{black}{\tiny\ttfamily\strut ,}}\hskip 0pt plus 1pt\relax\allowbreak{}\colorbox[HTML]{FFF4F4}{\textcolor{black}{\tiny\ttfamily\strut  the}}\hskip 0pt plus 1pt\relax\allowbreak{}\colorbox[HTML]{FFF9F9}{\textcolor{black}{\tiny\ttfamily\strut  hardware}}\hskip 0pt plus 1pt\relax\allowbreak{}\colorbox[HTML]{FFEFEF}{\textcolor{black}{\tiny\ttfamily\strut  components}}\hskip 0pt plus 1pt\relax\allowbreak{}\colorbox[HTML]{FEFEFF}{\textcolor{black}{\tiny\ttfamily\strut  may}}\hskip 0pt plus 1pt\relax\allowbreak{}\colorbox[HTML]{FFFFFF}{\textcolor{black}{\tiny\ttfamily\strut  not}}\hskip 0pt plus 1pt\relax\allowbreak{}\colorbox[HTML]{FFFFFF}{\textcolor{black}{\tiny\ttfamily\strut  function}}\hskip 0pt plus 1pt\relax\allowbreak{}\colorbox[HTML]{F9F9FF}{\textcolor{black}{\tiny\ttfamily\strut  properly}}\hskip 0pt plus 1pt\relax\allowbreak{}\colorbox[HTML]{FFFAFA}{\textcolor{black}{\tiny\ttfamily\strut .\textcolor{gray}{$\hookleftarrow$}}}\hskip 0pt plus 1pt\relax\allowbreak{}\colorbox[HTML]{FFF6F6}{\textcolor{black}{\tiny\ttfamily\strut 4}}\hskip 0pt plus 1pt\relax\allowbreak{}\colorbox[HTML]{FFFFFF}{\textcolor{black}{\tiny\ttfamily\strut .}}\hskip 0pt plus 1pt\relax\allowbreak{}\colorbox[HTML]{F8F8FF}{\textcolor{black}{\tiny\ttfamily\strut  Ant}}\hskip 0pt plus 1pt\relax\allowbreak{}\colorbox[HTML]{FFFFFF}{\textcolor{black}{\tiny\ttfamily\strut ivirus}}\hskip 0pt plus 1pt\relax\allowbreak{}\colorbox[HTML]{FFFEFE}{\textcolor{black}{\tiny\ttfamily\strut  Software}}\hskip 0pt plus 1pt\relax\allowbreak{}\colorbox[HTML]{FFFFFF}{\textcolor{black}{\tiny\ttfamily\strut :}}\hskip 0pt plus 1pt\relax\allowbreak{}\colorbox[HTML]{FFFFFF}{\textcolor{black}{\tiny\ttfamily\strut  Ant}}\hskip 0pt plus 1pt\relax\allowbreak{}\colorbox[HTML]{FFFFFF}{\textcolor{black}{\tiny\ttfamily\strut ivirus}}\hskip 0pt plus 1pt\relax\allowbreak{}\colorbox[HTML]{FFFFFF}{\textcolor{black}{\tiny\ttfamily\strut  software}}\hskip 0pt plus 1pt\relax\allowbreak{}\colorbox[HTML]{FFF3F3}{\textcolor{black}{\tiny\ttfamily\strut  helps}}\hskip 0pt plus 1pt\relax\allowbreak{}\colorbox[HTML]{FFFFFF}{\textcolor{black}{\tiny\ttfamily\strut  protect}}\hskip 0pt plus 1pt\relax\allowbreak{}\colorbox[HTML]{FFFFFF}{\textcolor{black}{\tiny\ttfamily\strut  the}}\hskip 0pt plus 1pt\relax\allowbreak{}\colorbox[HTML]{FFFFFF}{\textcolor{black}{\tiny\ttfamily\strut  computer}}\hskip 0pt plus 1pt\relax\allowbreak{}\colorbox[HTML]{FFFFFF}{\textcolor{black}{\tiny\ttfamily\strut  from}}\hskip 0pt plus 1pt\relax\allowbreak{}\colorbox[HTML]{FFFFFF}{\textcolor{black}{\tiny\ttfamily\strut  malware}}\hskip 0pt plus 1pt\relax\allowbreak{}\colorbox[HTML]{FFFEFE}{\textcolor{black}{\tiny\ttfamily\strut  and}}\hskip 0pt plus 1pt\relax\allowbreak{}\colorbox[HTML]{F6F6FF}{\textcolor{black}{\tiny\ttfamily\strut  other}}\hskip 0pt plus 1pt\relax\allowbreak{}\colorbox[HTML]{FBFBFF}{\textcolor{black}{\tiny\ttfamily\strut  security}}\hskip 0pt plus 1pt\relax\allowbreak{}\colorbox[HTML]{FFFFFF}{\textcolor{black}{\tiny\ttfamily\strut  threats}}\hskip 0pt plus 1pt\relax\allowbreak{}\colorbox[HTML]{BCBCFF}{\textcolor{black}{\tiny\ttfamily\strut .\textcolor{gray}{$\hookleftarrow$}}}\hskip 0pt plus 1pt\relax\allowbreak{}\colorbox[HTML]{FFFAFA}{\textcolor{black}{\tiny\ttfamily\strut 5}}\hskip 0pt plus 1pt\relax\allowbreak{}\colorbox[HTML]{FFFFFF}{\textcolor{black}{\tiny\ttfamily\strut .}}\hskip 0pt plus 1pt\relax\allowbreak{}\colorbox[HTML]{DFDFFF}{\textcolor{black}{\tiny\ttfamily\strut  Software}}\hskip 0pt plus 1pt\relax\allowbreak{}\colorbox[HTML]{FFFCFC}{\textcolor{black}{\tiny\ttfamily\strut  Updates}}\hskip 0pt plus 1pt\relax\allowbreak{}\colorbox[HTML]{FFFFFF}{\textcolor{black}{\tiny\ttfamily\strut :}}\hskip 0pt plus 1pt\relax\allowbreak{}\colorbox[HTML]{FFFFFF}{\textcolor{black}{\tiny\ttfamily\strut  Regular}}\hskip 0pt plus 1pt\relax\allowbreak{}\colorbox[HTML]{FFEAEA}{\textcolor{black}{\tiny\ttfamily\strut  software}}\hskip 0pt plus 1pt\relax\allowbreak{}\colorbox[HTML]{FFFFFF}{\textcolor{black}{\tiny\ttfamily\strut  updates}}\hskip 0pt plus 1pt\relax\allowbreak{}\colorbox[HTML]{FFF8F8}{\textcolor{black}{\tiny\ttfamily\strut  ensure}}\hskip 0pt plus 1pt\relax\allowbreak{}\colorbox[HTML]{E5E5FF}{\textcolor{black}{\tiny\ttfamily\strut  the}}\hskip 0pt plus 1pt\relax\allowbreak{}\colorbox[HTML]{FDFDFF}{\textcolor{black}{\tiny\ttfamily\strut  computer}}\hskip 0pt plus 1pt\relax\allowbreak{}\colorbox[HTML]{FFCDCD}{\textcolor{black}{\tiny\ttfamily\strut  remains}}\hskip 0pt plus 1pt\relax\allowbreak{}\colorbox[HTML]{FEFEFF}{\textcolor{black}{\tiny\ttfamily\strut  secure}}\hskip 0pt plus 1pt\relax\allowbreak{}\colorbox[HTML]{FFFEFE}{\textcolor{black}{\tiny\ttfamily\strut  and}}\hskip 0pt plus 1pt\relax\allowbreak{}\colorbox[HTML]{E9E9FF}{\textcolor{black}{\tiny\ttfamily\strut  runs}}\hskip 0pt plus 1pt\relax\allowbreak{}\colorbox[HTML]{FFFEFE}{\textcolor{black}{\tiny\ttfamily\strut  at}}\hskip 0pt plus 1pt\relax\allowbreak{}\colorbox[HTML]{FFE7E7}{\textcolor{black}{\tiny\ttfamily\strut  its}}\hskip 0pt plus 1pt\relax\allowbreak{}\colorbox[HTML]{DFDFFF}{\textcolor{black}{\tiny\ttfamily\strut  best}}\hskip 0pt plus 1pt\relax\allowbreak{}\colorbox[HTML]{FFE8E8}{\textcolor{black}{\tiny\ttfamily\strut  performance}}\hskip 0pt plus 1pt\relax\allowbreak{}\colorbox[HTML]{FFD1D1}{\textcolor{black}{\tiny\ttfamily\strut .\textcolor{gray}{$\hookleftarrow$}}}\hskip 0pt plus 1pt\relax\allowbreak{}\colorbox[HTML]{FFE2E2}{\textcolor{black}{\tiny\ttfamily\strut The}}\hskip 0pt plus 1pt\relax\allowbreak{}\colorbox[HTML]{FEFEFF}{\textcolor{black}{\tiny\ttfamily\strut  inter}}\hskip 0pt plus 1pt\relax\allowbreak{}\colorbox[HTML]{FFFFFF}{\textcolor{black}{\tiny\ttfamily\strut dependence}}\hskip 0pt plus 1pt\relax\allowbreak{}\colorbox[HTML]{FFFEFE}{\textcolor{black}{\tiny\ttfamily\strut  and}}\hskip 0pt plus 1pt\relax\allowbreak{}\colorbox[HTML]{FFFFFF}{\textcolor{black}{\tiny\ttfamily\strut  inter}}\hskip 0pt plus 1pt\relax\allowbreak{}\colorbox[HTML]{FFFFFF}{\textcolor{black}{\tiny\ttfamily\strut connection}}\hskip 0pt plus 1pt\relax\allowbreak{}\colorbox[HTML]{FFFAFA}{\textcolor{black}{\tiny\ttfamily\strut  between}}\hskip 0pt plus 1pt\relax\allowbreak{}\colorbox[HTML]{FFF8F8}{\textcolor{black}{\tiny\ttfamily\strut  these}}\hskip 0pt plus 1pt\relax\allowbreak{}\colorbox[HTML]{FFDEDE}{\textcolor{black}{\tiny\ttfamily\strut  hardware}}\hskip 0pt plus 1pt\relax\allowbreak{}\colorbox[HTML]{FFFFFF}{\textcolor{black}{\tiny\ttfamily\strut  and}}\hskip 0pt plus 1pt\relax\allowbreak{}\colorbox[HTML]{FFFFFF}{\textcolor{black}{\tiny\ttfamily\strut  software}}\hskip 0pt plus 1pt\relax\allowbreak{}\colorbox[HTML]{FFFFFF}{\textcolor{black}{\tiny\ttfamily\strut  components}}\hskip 0pt plus 1pt\relax\allowbreak{}\colorbox[HTML]{FBFBFF}{\textcolor{black}{\tiny\ttfamily\strut  are}}\hskip 0pt plus 1pt\relax\allowbreak{}\colorbox[HTML]{8E8EFF}{\textcolor{black}{\tiny\ttfamily\strut  critical}}\hskip 0pt plus 1pt\relax\allowbreak{}\colorbox[HTML]{FFFBFB}{\textcolor{black}{\tiny\ttfamily\strut  for}}\hskip 0pt plus 1pt\relax\allowbreak{}\colorbox[HTML]{FFF9F9}{\textcolor{black}{\tiny\ttfamily\strut  the}}\hskip 0pt plus 1pt\relax\allowbreak{}\colorbox[HTML]{ECECFF}{\textcolor{black}{\tiny\ttfamily\strut  computer}}\hskip 0pt plus 1pt\relax\allowbreak{}\colorbox[HTML]{F7F7FF}{\textcolor{black}{\tiny\ttfamily\strut 's}}\hskip 0pt plus 1pt\relax\allowbreak{}\colorbox[HTML]{FFFFFF}{\textcolor{black}{\tiny\ttfamily\strut  optimal}}\hskip 0pt plus 1pt\relax\allowbreak{}\colorbox[HTML]{E7E7FF}{\textcolor{black}{\tiny\ttfamily\strut  performance}}\hskip 0pt plus 1pt\relax\allowbreak{}\colorbox[HTML]{E2E2FF}{\textcolor{black}{\tiny\ttfamily\strut  and}}\hskip 0pt plus 1pt\relax\allowbreak{}\colorbox[HTML]{FFFFFF}{\textcolor{black}{\tiny\ttfamily\strut  functionality}}\hskip 0pt plus 1pt\relax\allowbreak{}\colorbox[HTML]{FFFCFC}{\textcolor{black}{\tiny\ttfamily\strut .}}\hskip 0pt plus 1pt\relax\allowbreak{}\colorbox[HTML]{F7F7FF}{\textcolor{black}{\tiny\ttfamily\strut  The}}\hskip 0pt plus 1pt\relax\allowbreak{}\colorbox[HTML]{FFFCFC}{\textcolor{black}{\tiny\ttfamily\strut  CPU}}\hskip 0pt plus 1pt\relax\allowbreak{}\colorbox[HTML]{FEFEFF}{\textcolor{black}{\tiny\ttfamily\strut  and}}\hskip 0pt plus 1pt\relax\allowbreak{}\colorbox[HTML]{FCFCFF}{\textcolor{black}{\tiny\ttfamily\strut  RAM}}\hskip 0pt plus 1pt\relax\allowbreak{}\colorbox[HTML]{F8F8FF}{\textcolor{black}{\tiny\ttfamily\strut  work}}\hskip 0pt plus 1pt\relax\allowbreak{}\colorbox[HTML]{D6D6FF}{\textcolor{black}{\tiny\ttfamily\strut  together}}\hskip 0pt plus 1pt\relax\allowbreak{}\colorbox[HTML]{F1F1FF}{\textcolor{black}{\tiny\ttfamily\strut  to}}\hskip 0pt plus 1pt\relax\allowbreak{}\colorbox[HTML]{FFFEFE}{\textcolor{black}{\tiny\ttfamily\strut  process}}\hskip 0pt plus 1pt\relax\allowbreak{}\colorbox[HTML]{FFFCFC}{\textcolor{black}{\tiny\ttfamily\strut  data}}\hskip 0pt plus 1pt\relax\allowbreak{}\colorbox[HTML]{EBEBFF}{\textcolor{black}{\tiny\ttfamily\strut ,}}\hskip 0pt plus 1pt\relax\allowbreak{}\colorbox[HTML]{D5D5FF}{\textcolor{black}{\tiny\ttfamily\strut  while}}\hskip 0pt plus 1pt\relax\allowbreak{}\colorbox[HTML]{FFFFFF}{\textcolor{black}{\tiny\ttfamily\strut  the}}\hskip 0pt plus 1pt\relax\allowbreak{}\colorbox[HTML]{FFFDFD}{\textcolor{black}{\tiny\ttfamily\strut  GPU}}\hskip 0pt plus 1pt\relax\allowbreak{}\colorbox[HTML]{FBFBFF}{\textcolor{black}{\tiny\ttfamily\strut  and}}\hskip 0pt plus 1pt\relax\allowbreak{}\colorbox[HTML]{EAEAFF}{\textcolor{black}{\tiny\ttfamily\strut  motherboard}}\hskip 0pt plus 1pt\relax\allowbreak{}\colorbox[HTML]{FCFCFF}{\textcolor{black}{\tiny\ttfamily\strut  work}}\hskip 0pt plus 1pt\relax\allowbreak{}\colorbox[HTML]{DDDDFF}{\textcolor{black}{\tiny\ttfamily\strut  together}}\hskip 0pt plus 1pt\relax\allowbreak{}\colorbox[HTML]{FFFFFF}{\textcolor{black}{\tiny\ttfamily\strut  to}}\hskip 0pt plus 1pt\relax\allowbreak{}\colorbox[HTML]{EBEBFF}{\textcolor{black}{\tiny\ttfamily\strut  render}}\hskip 0pt plus 1pt\relax\allowbreak{}\colorbox[HTML]{F5F5FF}{\textcolor{black}{\tiny\ttfamily\strut  graphics}}\hskip 0pt plus 1pt\relax\allowbreak{}\colorbox[HTML]{EFEFFF}{\textcolor{black}{\tiny\ttfamily\strut .}}\hskip 0pt plus 1pt\relax\allowbreak{}\colorbox[HTML]{F6F6FF}{\textcolor{black}{\tiny\ttfamily\strut  The}}\hskip 0pt plus 1pt\relax\allowbreak{}\colorbox[HTML]{FFECEC}{\textcolor{black}{\tiny\ttfamily\strut  operating}}\hskip 0pt plus 1pt\relax\allowbreak{}\colorbox[HTML]{FFFFFF}{\textcolor{black}{\tiny\ttfamily\strut  system}}\hskip 0pt plus 1pt\relax\allowbreak{}\colorbox[HTML]{F6F6FF}{\textcolor{black}{\tiny\ttfamily\strut  and}}\hskip 0pt plus 1pt\relax\allowbreak{}\colorbox[HTML]{D7D7FF}{\textcolor{black}{\tiny\ttfamily\strut  applications}}\hskip 0pt plus 1pt\relax\allowbreak{}\colorbox[HTML]{E5E5FF}{\textcolor{black}{\tiny\ttfamily\strut  work}}\hskip 0pt plus 1pt\relax\allowbreak{}\colorbox[HTML]{D2D2FF}{\textcolor{black}{\tiny\ttfamily\strut  together}}\hskip 0pt plus 1pt\relax\allowbreak{}\colorbox[HTML]{FFFFFF}{\textcolor{black}{\tiny\ttfamily\strut  to}}\hskip 0pt plus 1pt\relax\allowbreak{}\colorbox[HTML]{F9F9FF}{\textcolor{black}{\tiny\ttfamily\strut  provide}}\hskip 0pt plus 1pt\relax\allowbreak{}\colorbox[HTML]{FFF7F7}{\textcolor{black}{\tiny\ttfamily\strut  a}}\hskip 0pt plus 1pt\relax\allowbreak{}\colorbox[HTML]{ECECFF}{\textcolor{black}{\tiny\ttfamily\strut  user}}\hskip 0pt plus 1pt\relax\allowbreak{}\colorbox[HTML]{FCFCFF}{\textcolor{black}{\tiny\ttfamily\strut  interface}}\hskip 0pt plus 1pt\relax\allowbreak{}\colorbox[HTML]{FFDDDD}{\textcolor{black}{\tiny\ttfamily\strut ,}}\hskip 0pt plus 1pt\relax\allowbreak{}\colorbox[HTML]{F0F0FF}{\textcolor{black}{\tiny\ttfamily\strut  while}}\hskip 0pt plus 1pt\relax\allowbreak{}\colorbox[HTML]{FFEAEA}{\textcolor{black}{\tiny\ttfamily\strut  the}}\hskip 0pt plus 1pt\relax\allowbreak{}\colorbox[HTML]{FEFEFF}{\textcolor{black}{\tiny\ttfamily\strut  drivers}}\hskip 0pt plus 1pt\relax\allowbreak{}\colorbox[HTML]{E6E6FF}{\textcolor{black}{\tiny\ttfamily\strut  enable}}\hskip 0pt plus 1pt\relax\allowbreak{}\colorbox[HTML]{D6D6FF}{\textcolor{black}{\tiny\ttfamily\strut  hardware}}\hskip 0pt plus 1pt\relax\allowbreak{}\colorbox[HTML]{FFFBFB}{\textcolor{black}{\tiny\ttfamily\strut  components}}\hskip 0pt plus 1pt\relax\allowbreak{}\colorbox[HTML]{FFFFFF}{\textcolor{black}{\tiny\ttfamily\strut  to}}\hskip 0pt plus 1pt\relax\allowbreak{}\colorbox[HTML]{FFEEEE}{\textcolor{black}{\tiny\ttfamily\strut  communicate}}\hskip 0pt plus 1pt\relax\allowbreak{}\colorbox[HTML]{FDFDFF}{\textcolor{black}{\tiny\ttfamily\strut  with}}\hskip 0pt plus 1pt\relax\allowbreak{}\colorbox[HTML]{FFFFFF}{\textcolor{black}{\tiny\ttfamily\strut  the}}\hskip 0pt plus 1pt\relax\allowbreak{}\colorbox[HTML]{FDFDFF}{\textcolor{black}{\tiny\ttfamily\strut  operating}}\hskip 0pt plus 1pt\relax\allowbreak{}\colorbox[HTML]{FFFFFF}{\textcolor{black}{\tiny\ttfamily\strut  system}}\hskip 0pt plus 1pt\relax\allowbreak{}\colorbox[HTML]{FFF8F8}{\textcolor{black}{\tiny\ttfamily\strut .}}\hskip 0pt plus 1pt\relax\allowbreak{}\colorbox[HTML]{FFCACA}{\textcolor{black}{\tiny\ttfamily\strut  The}}\hskip 0pt plus 1pt\relax\allowbreak{}\colorbox[HTML]{B5B5FF}{\textcolor{black}{\tiny\ttfamily\strut  antivirus}}\hskip 0pt plus 1pt\relax\allowbreak{}\colorbox[HTML]{FFFFFF}{\textcolor{black}{\tiny\ttfamily\strut  software}}\hskip 0pt plus 1pt\relax\allowbreak{}\colorbox[HTML]{F6F6FF}{\textcolor{black}{\tiny\ttfamily\strut  and}}\hskip 0pt plus 1pt\relax\allowbreak{}\colorbox[HTML]{FFFFFF}{\textcolor{black}{\tiny\ttfamily\strut  software}}\hskip 0pt plus 1pt\relax\allowbreak{}\colorbox[HTML]{FFFFFF}{\textcolor{black}{\tiny\ttfamily\strut  updates}}\hskip 0pt plus 1pt\relax\allowbreak{}\colorbox[HTML]{F1F1FF}{\textcolor{black}{\tiny\ttfamily\strut  ensure}}\hskip 0pt plus 1pt\relax\allowbreak{}\colorbox[HTML]{FCFCFF}{\textcolor{black}{\tiny\ttfamily\strut  the}}\hskip 0pt plus 1pt\relax\allowbreak{}\colorbox[HTML]{FAFAFF}{\textcolor{black}{\tiny\ttfamily\strut  computer}}\hskip 0pt plus 1pt\relax\allowbreak{}\colorbox[HTML]{FFFFFF}{\textcolor{black}{\tiny\ttfamily\strut  remains}}\hskip 0pt plus 1pt\relax\allowbreak{}\colorbox[HTML]{FFFFFF}{\textcolor{black}{\tiny\ttfamily\strut  secure}}\hskip 0pt plus 1pt\relax\allowbreak{}\colorbox[HTML]{FEFEFF}{\textcolor{black}{\tiny\ttfamily\strut  and}}\hskip 0pt plus 1pt\relax\allowbreak{}\colorbox[HTML]{F3F3FF}{\textcolor{black}{\tiny\ttfamily\strut  runs}}\hskip 0pt plus 1pt\relax\allowbreak{}\colorbox[HTML]{FFF8F8}{\textcolor{black}{\tiny\ttfamily\strut  at}}\hskip 0pt plus 1pt\relax\allowbreak{}\colorbox[HTML]{FEFEFF}{\textcolor{black}{\tiny\ttfamily\strut  its}}\hskip 0pt plus 1pt\relax\allowbreak{}\colorbox[HTML]{FFFFFF}{\textcolor{black}{\tiny\ttfamily\strut  best}}\hskip 0pt plus 1pt\relax\allowbreak{}\colorbox[HTML]{FFF7F7}{\textcolor{black}{\tiny\ttfamily\strut  performance}}\hskip 0pt plus 1pt\relax\allowbreak{}\colorbox[HTML]{FFCDCD}{\textcolor{black}{\tiny\ttfamily\strut .}}\hskip 0pt plus 1pt\relax\allowbreak{}\colorbox[HTML]{FFC5C5}{\textcolor{black}{\tiny\ttfamily\strut  The}}\hskip 0pt plus 1pt\relax\allowbreak{}\colorbox[HTML]{D5D5FF}{\textcolor{black}{\tiny\ttfamily\strut  inter}}\hskip 0pt plus 1pt\relax\allowbreak{}\colorbox[HTML]{FFEFEF}{\textcolor{black}{\tiny\ttfamily\strut dependence}}\hskip 0pt plus 1pt\relax\allowbreak{}\colorbox[HTML]{E8E8FF}{\textcolor{black}{\tiny\ttfamily\strut  and}}\hskip 0pt plus 1pt\relax\allowbreak{}\colorbox[HTML]{FFFFFF}{\textcolor{black}{\tiny\ttfamily\strut  inter}}\hskip 0pt plus 1pt\relax\allowbreak{}\colorbox[HTML]{FFFFFF}{\textcolor{black}{\tiny\ttfamily\strut connection}}\hskip 0pt plus 1pt\relax\allowbreak{}\colorbox[HTML]{F5F5FF}{\textcolor{black}{\tiny\ttfamily\strut  between}}\hskip 0pt plus 1pt\relax\allowbreak{}\colorbox[HTML]{F7F7FF}{\textcolor{black}{\tiny\ttfamily\strut  these}}\hskip 0pt plus 1pt\relax\allowbreak{}\colorbox[HTML]{FFFEFE}{\textcolor{black}{\tiny\ttfamily\strut  components}}\hskip 0pt plus 1pt\relax\allowbreak{}\colorbox[HTML]{FF8686}{\textcolor{black}{\tiny\ttfamily\strut  can}}\hskip 0pt plus 1pt\relax\allowbreak{}\colorbox[HTML]{FFE1E1}{\textcolor{black}{\tiny\ttfamily\strut  be}}\hskip 0pt plus 1pt\relax\allowbreak{}\colorbox[HTML]{FF7777}{\textcolor{black}{\tiny\ttfamily\strut  metaphor}}\hskip 0pt plus 1pt\relax\allowbreak{}\colorbox[HTML]{FFFFFF}{\textcolor{black}{\tiny\ttfamily\strut ically}}\hskip 0pt plus 1pt\relax\allowbreak{}\colorbox[HTML]{FFB6B6}{\textcolor{black}{\tiny\ttfamily\strut  compared}}\hskip 0pt plus 1pt\relax\allowbreak{}\colorbox[HTML]{FFFFFF}{\textcolor{black}{\tiny\ttfamily\strut  to}}\hskip 0pt plus 1pt\relax\allowbreak{}\colorbox[HTML]{FEFEFF}{\textcolor{black}{\tiny\ttfamily\strut  a}}\hskip 0pt plus 1pt\relax\allowbreak{}\colorbox[HTML]{DEDEFF}{\textcolor{black}{\tiny\ttfamily\strut  complex}}\hskip 0pt plus 1pt\relax\allowbreak{}\colorbox[HTML]{FFF3F3}{\textcolor{black}{\tiny\ttfamily\strut  ecosystem}}\hskip 0pt plus 1pt\relax\allowbreak{}\colorbox[HTML]{FFF7F7}{\textcolor{black}{\tiny\ttfamily\strut ,}}\hskip 0pt plus 1pt\relax\allowbreak{}\colorbox[HTML]{FFFFFF}{\textcolor{black}{\tiny\ttfamily\strut  where}}\hskip 0pt plus 1pt\relax\allowbreak{}\colorbox[HTML]{FFFFFF}{\textcolor{black}{\tiny\ttfamily\strut  each}}\hskip 0pt plus 1pt\relax\allowbreak{}\colorbox[HTML]{FFF8F8}{\textcolor{black}{\tiny\ttfamily\strut  component}}\hskip 0pt plus 1pt\relax\allowbreak{}\colorbox[HTML]{FFF5F5}{\textcolor{black}{\tiny\ttfamily\strut  plays}}\hskip 0pt plus 1pt\relax\allowbreak{}\colorbox[HTML]{FFFFFF}{\textcolor{black}{\tiny\ttfamily\strut  a}}\hskip 0pt plus 1pt\relax\allowbreak{}\colorbox[HTML]{FFF9F9}{\textcolor{black}{\tiny\ttfamily\strut  vital}}\hskip 0pt plus 1pt\relax\allowbreak{}\colorbox[HTML]{FFFFFF}{\textcolor{black}{\tiny\ttfamily\strut  role}}\hskip 0pt plus 1pt\relax\allowbreak{}\colorbox[HTML]{FFFFFF}{\textcolor{black}{\tiny\ttfamily\strut  in}}\hskip 0pt plus 1pt\relax\allowbreak{}\colorbox[HTML]{ECECFF}{\textcolor{black}{\tiny\ttfamily\strut  the}}\hskip 0pt plus 1pt\relax\allowbreak{}\colorbox[HTML]{FFFEFE}{\textcolor{black}{\tiny\ttfamily\strut  overall}}\hskip 0pt plus 1pt\relax\allowbreak{}\colorbox[HTML]{FFF5F5}{\textcolor{black}{\tiny\ttfamily\strut  functionality}}\hskip 0pt plus 1pt\relax\allowbreak{}\colorbox[HTML]{FFFCFC}{\textcolor{black}{\tiny\ttfamily\strut  of}}\hskip 0pt plus 1pt\relax\allowbreak{}\colorbox[HTML]{FFFFFF}{\textcolor{black}{\tiny\ttfamily\strut  the}}\hskip 0pt plus 1pt\relax\allowbreak{}\colorbox[HTML]{FFFCFC}{\textcolor{black}{\tiny\ttfamily\strut  computer}}\hskip 0pt plus 1pt\relax\allowbreak{}\colorbox[HTML]{FFC3C3}{\textcolor{black}{\tiny\ttfamily\strut .}}\hskip 0pt plus 1pt\relax\allowbreak{}\colorbox[HTML]{6B6BFF}{\textcolor{white}{\tiny\ttfamily\strut \textless{}\textbar{}im\_end\textbar{}\textgreater{}}}\hskip 0pt plus 1pt\relax\allowbreak{}
\par\medskip
\noindent\textbf{\textcolor[HTML]{2257bf}{LA}} {\scriptsize $|\delta|_{\max}{=}10.6$}: \\[2pt]
\sloppy
\colorbox[HTML]{FFB6B6}{\textcolor{black}{\tiny\ttfamily\strut Hardware}}\hskip 0pt plus 1pt\relax\allowbreak{}\colorbox[HTML]{FFE1E1}{\textcolor{black}{\tiny\ttfamily\strut  Components}}\hskip 0pt plus 1pt\relax\allowbreak{}\colorbox[HTML]{FFFBFB}{\textcolor{black}{\tiny\ttfamily\strut :\textcolor{gray}{$\hookleftarrow$}}}\hskip 0pt plus 1pt\relax\allowbreak{}\colorbox[HTML]{FFFFFF}{\textcolor{black}{\tiny\ttfamily\strut 1}}\hskip 0pt plus 1pt\relax\allowbreak{}\colorbox[HTML]{FFFFFF}{\textcolor{black}{\tiny\ttfamily\strut .}}\hskip 0pt plus 1pt\relax\allowbreak{}\colorbox[HTML]{C6C6FF}{\textcolor{black}{\tiny\ttfamily\strut  Central}}\hskip 0pt plus 1pt\relax\allowbreak{}\colorbox[HTML]{FFFFFF}{\textcolor{black}{\tiny\ttfamily\strut  Processing}}\hskip 0pt plus 1pt\relax\allowbreak{}\colorbox[HTML]{FFFFFF}{\textcolor{black}{\tiny\ttfamily\strut  Unit}}\hskip 0pt plus 1pt\relax\allowbreak{}\colorbox[HTML]{FFFFFF}{\textcolor{black}{\tiny\ttfamily\strut  (}}\hskip 0pt plus 1pt\relax\allowbreak{}\colorbox[HTML]{FFFFFF}{\textcolor{black}{\tiny\ttfamily\strut CPU}}\hskip 0pt plus 1pt\relax\allowbreak{}\colorbox[HTML]{FFFEFE}{\textcolor{black}{\tiny\ttfamily\strut ):}}\hskip 0pt plus 1pt\relax\allowbreak{}\colorbox[HTML]{F2F2FF}{\textcolor{black}{\tiny\ttfamily\strut  The}}\hskip 0pt plus 1pt\relax\allowbreak{}\colorbox[HTML]{F5F5FF}{\textcolor{black}{\tiny\ttfamily\strut  CPU}}\hskip 0pt plus 1pt\relax\allowbreak{}\colorbox[HTML]{FFDCDC}{\textcolor{black}{\tiny\ttfamily\strut ,}}\hskip 0pt plus 1pt\relax\allowbreak{}\colorbox[HTML]{CCCCFF}{\textcolor{black}{\tiny\ttfamily\strut  or}}\hskip 0pt plus 1pt\relax\allowbreak{}\colorbox[HTML]{FFEBEB}{\textcolor{black}{\tiny\ttfamily\strut  the}}\hskip 0pt plus 1pt\relax\allowbreak{}\colorbox[HTML]{FFFBFB}{\textcolor{black}{\tiny\ttfamily\strut  brain}}\hskip 0pt plus 1pt\relax\allowbreak{}\colorbox[HTML]{FFFFFF}{\textcolor{black}{\tiny\ttfamily\strut  of}}\hskip 0pt plus 1pt\relax\allowbreak{}\colorbox[HTML]{FFFFFF}{\textcolor{black}{\tiny\ttfamily\strut  the}}\hskip 0pt plus 1pt\relax\allowbreak{}\colorbox[HTML]{FFFFFF}{\textcolor{black}{\tiny\ttfamily\strut  computer}}\hskip 0pt plus 1pt\relax\allowbreak{}\colorbox[HTML]{FFFFFF}{\textcolor{black}{\tiny\ttfamily\strut ,}}\hskip 0pt plus 1pt\relax\allowbreak{}\colorbox[HTML]{D8D8FF}{\textcolor{black}{\tiny\ttfamily\strut  is}}\hskip 0pt plus 1pt\relax\allowbreak{}\colorbox[HTML]{D5D5FF}{\textcolor{black}{\tiny\ttfamily\strut  the}}\hskip 0pt plus 1pt\relax\allowbreak{}\colorbox[HTML]{E7E7FF}{\textcolor{black}{\tiny\ttfamily\strut  component}}\hskip 0pt plus 1pt\relax\allowbreak{}\colorbox[HTML]{FFF8F8}{\textcolor{black}{\tiny\ttfamily\strut  that}}\hskip 0pt plus 1pt\relax\allowbreak{}\colorbox[HTML]{FFFAFA}{\textcolor{black}{\tiny\ttfamily\strut  executes}}\hskip 0pt plus 1pt\relax\allowbreak{}\colorbox[HTML]{FFFEFE}{\textcolor{black}{\tiny\ttfamily\strut  instructions}}\hskip 0pt plus 1pt\relax\allowbreak{}\colorbox[HTML]{FFF7F7}{\textcolor{black}{\tiny\ttfamily\strut ,}}\hskip 0pt plus 1pt\relax\allowbreak{}\colorbox[HTML]{C0C0FF}{\textcolor{black}{\tiny\ttfamily\strut  performs}}\hskip 0pt plus 1pt\relax\allowbreak{}\colorbox[HTML]{FEFEFF}{\textcolor{black}{\tiny\ttfamily\strut  calculations}}\hskip 0pt plus 1pt\relax\allowbreak{}\colorbox[HTML]{FFFDFD}{\textcolor{black}{\tiny\ttfamily\strut ,}}\hskip 0pt plus 1pt\relax\allowbreak{}\colorbox[HTML]{FFFFFF}{\textcolor{black}{\tiny\ttfamily\strut  and}}\hskip 0pt plus 1pt\relax\allowbreak{}\colorbox[HTML]{EDEDFF}{\textcolor{black}{\tiny\ttfamily\strut  controls}}\hskip 0pt plus 1pt\relax\allowbreak{}\colorbox[HTML]{FBFBFF}{\textcolor{black}{\tiny\ttfamily\strut  the}}\hskip 0pt plus 1pt\relax\allowbreak{}\colorbox[HTML]{FDFDFF}{\textcolor{black}{\tiny\ttfamily\strut  flow}}\hskip 0pt plus 1pt\relax\allowbreak{}\colorbox[HTML]{FFFFFF}{\textcolor{black}{\tiny\ttfamily\strut  of}}\hskip 0pt plus 1pt\relax\allowbreak{}\colorbox[HTML]{FEFEFF}{\textcolor{black}{\tiny\ttfamily\strut  data}}\hskip 0pt plus 1pt\relax\allowbreak{}\colorbox[HTML]{CECEFF}{\textcolor{black}{\tiny\ttfamily\strut  between}}\hskip 0pt plus 1pt\relax\allowbreak{}\colorbox[HTML]{F3F3FF}{\textcolor{black}{\tiny\ttfamily\strut  different}}\hskip 0pt plus 1pt\relax\allowbreak{}\colorbox[HTML]{FFF6F6}{\textcolor{black}{\tiny\ttfamily\strut  parts}}\hskip 0pt plus 1pt\relax\allowbreak{}\colorbox[HTML]{FFFFFF}{\textcolor{black}{\tiny\ttfamily\strut  of}}\hskip 0pt plus 1pt\relax\allowbreak{}\colorbox[HTML]{FFFFFF}{\textcolor{black}{\tiny\ttfamily\strut  the}}\hskip 0pt plus 1pt\relax\allowbreak{}\colorbox[HTML]{D1D1FF}{\textcolor{black}{\tiny\ttfamily\strut  computer}}\hskip 0pt plus 1pt\relax\allowbreak{}\colorbox[HTML]{FFE6E6}{\textcolor{black}{\tiny\ttfamily\strut .\textcolor{gray}{$\hookleftarrow$}}}\hskip 0pt plus 1pt\relax\allowbreak{}\colorbox[HTML]{FFFFFF}{\textcolor{black}{\tiny\ttfamily\strut 2}}\hskip 0pt plus 1pt\relax\allowbreak{}\colorbox[HTML]{FFFFFF}{\textcolor{black}{\tiny\ttfamily\strut .}}\hskip 0pt plus 1pt\relax\allowbreak{}\colorbox[HTML]{FFFFFF}{\textcolor{black}{\tiny\ttfamily\strut  Random}}\hskip 0pt plus 1pt\relax\allowbreak{}\colorbox[HTML]{FFFFFF}{\textcolor{black}{\tiny\ttfamily\strut  Access}}\hskip 0pt plus 1pt\relax\allowbreak{}\colorbox[HTML]{FFFFFF}{\textcolor{black}{\tiny\ttfamily\strut  Memory}}\hskip 0pt plus 1pt\relax\allowbreak{}\colorbox[HTML]{FFFFFF}{\textcolor{black}{\tiny\ttfamily\strut  (}}\hskip 0pt plus 1pt\relax\allowbreak{}\colorbox[HTML]{FFFFFF}{\textcolor{black}{\tiny\ttfamily\strut RAM}}\hskip 0pt plus 1pt\relax\allowbreak{}\colorbox[HTML]{FFFFFF}{\textcolor{black}{\tiny\ttfamily\strut ):}}\hskip 0pt plus 1pt\relax\allowbreak{}\colorbox[HTML]{FFFEFE}{\textcolor{black}{\tiny\ttfamily\strut  RAM}}\hskip 0pt plus 1pt\relax\allowbreak{}\colorbox[HTML]{FFECEC}{\textcolor{black}{\tiny\ttfamily\strut ,}}\hskip 0pt plus 1pt\relax\allowbreak{}\colorbox[HTML]{9292FF}{\textcolor{black}{\tiny\ttfamily\strut  or}}\hskip 0pt plus 1pt\relax\allowbreak{}\colorbox[HTML]{FFEEEE}{\textcolor{black}{\tiny\ttfamily\strut  the}}\hskip 0pt plus 1pt\relax\allowbreak{}\colorbox[HTML]{FFFBFB}{\textcolor{black}{\tiny\ttfamily\strut  working}}\hskip 0pt plus 1pt\relax\allowbreak{}\colorbox[HTML]{FFFFFF}{\textcolor{black}{\tiny\ttfamily\strut  memory}}\hskip 0pt plus 1pt\relax\allowbreak{}\colorbox[HTML]{FFEAEA}{\textcolor{black}{\tiny\ttfamily\strut ,}}\hskip 0pt plus 1pt\relax\allowbreak{}\colorbox[HTML]{E0E0FF}{\textcolor{black}{\tiny\ttfamily\strut  is}}\hskip 0pt plus 1pt\relax\allowbreak{}\colorbox[HTML]{FFA7A7}{\textcolor{black}{\tiny\ttfamily\strut  essential}}\hskip 0pt plus 1pt\relax\allowbreak{}\colorbox[HTML]{FFFFFF}{\textcolor{black}{\tiny\ttfamily\strut  for}}\hskip 0pt plus 1pt\relax\allowbreak{}\colorbox[HTML]{FEFEFF}{\textcolor{black}{\tiny\ttfamily\strut  storing}}\hskip 0pt plus 1pt\relax\allowbreak{}\colorbox[HTML]{F2F2FF}{\textcolor{black}{\tiny\ttfamily\strut  data}}\hskip 0pt plus 1pt\relax\allowbreak{}\colorbox[HTML]{FFDCDC}{\textcolor{black}{\tiny\ttfamily\strut  temporarily}}\hskip 0pt plus 1pt\relax\allowbreak{}\colorbox[HTML]{E3E3FF}{\textcolor{black}{\tiny\ttfamily\strut  while}}\hskip 0pt plus 1pt\relax\allowbreak{}\colorbox[HTML]{FFFFFF}{\textcolor{black}{\tiny\ttfamily\strut  the}}\hskip 0pt plus 1pt\relax\allowbreak{}\colorbox[HTML]{CBCBFF}{\textcolor{black}{\tiny\ttfamily\strut  computer}}\hskip 0pt plus 1pt\relax\allowbreak{}\colorbox[HTML]{F7F7FF}{\textcolor{black}{\tiny\ttfamily\strut  is}}\hskip 0pt plus 1pt\relax\allowbreak{}\colorbox[HTML]{F5F5FF}{\textcolor{black}{\tiny\ttfamily\strut  in}}\hskip 0pt plus 1pt\relax\allowbreak{}\colorbox[HTML]{E7E7FF}{\textcolor{black}{\tiny\ttfamily\strut  use}}\hskip 0pt plus 1pt\relax\allowbreak{}\colorbox[HTML]{5656FF}{\textcolor{white}{\tiny\ttfamily\strut .}}\hskip 0pt plus 1pt\relax\allowbreak{}\colorbox[HTML]{A2A2FF}{\textcolor{black}{\tiny\ttfamily\strut  A}}\hskip 0pt plus 1pt\relax\allowbreak{}\colorbox[HTML]{FFEEEE}{\textcolor{black}{\tiny\ttfamily\strut  larger}}\hskip 0pt plus 1pt\relax\allowbreak{}\colorbox[HTML]{B6B6FF}{\textcolor{black}{\tiny\ttfamily\strut  amount}}\hskip 0pt plus 1pt\relax\allowbreak{}\colorbox[HTML]{FEFEFF}{\textcolor{black}{\tiny\ttfamily\strut  of}}\hskip 0pt plus 1pt\relax\allowbreak{}\colorbox[HTML]{FFFFFF}{\textcolor{black}{\tiny\ttfamily\strut  RAM}}\hskip 0pt plus 1pt\relax\allowbreak{}\colorbox[HTML]{E2E2FF}{\textcolor{black}{\tiny\ttfamily\strut  can}}\hskip 0pt plus 1pt\relax\allowbreak{}\colorbox[HTML]{DADAFF}{\textcolor{black}{\tiny\ttfamily\strut  significantly}}\hskip 0pt plus 1pt\relax\allowbreak{}\colorbox[HTML]{FFC2C2}{\textcolor{black}{\tiny\ttfamily\strut  enhance}}\hskip 0pt plus 1pt\relax\allowbreak{}\colorbox[HTML]{D6D6FF}{\textcolor{black}{\tiny\ttfamily\strut  the}}\hskip 0pt plus 1pt\relax\allowbreak{}\colorbox[HTML]{FFFCFC}{\textcolor{black}{\tiny\ttfamily\strut  computer}}\hskip 0pt plus 1pt\relax\allowbreak{}\colorbox[HTML]{FDFDFF}{\textcolor{black}{\tiny\ttfamily\strut 's}}\hskip 0pt plus 1pt\relax\allowbreak{}\colorbox[HTML]{B6B6FF}{\textcolor{black}{\tiny\ttfamily\strut  performance}}\hskip 0pt plus 1pt\relax\allowbreak{}\colorbox[HTML]{FCFCFF}{\textcolor{black}{\tiny\ttfamily\strut  by}}\hskip 0pt plus 1pt\relax\allowbreak{}\colorbox[HTML]{FBFBFF}{\textcolor{black}{\tiny\ttfamily\strut  allowing}}\hskip 0pt plus 1pt\relax\allowbreak{}\colorbox[HTML]{F6F6FF}{\textcolor{black}{\tiny\ttfamily\strut  it}}\hskip 0pt plus 1pt\relax\allowbreak{}\colorbox[HTML]{FFFFFF}{\textcolor{black}{\tiny\ttfamily\strut  to}}\hskip 0pt plus 1pt\relax\allowbreak{}\colorbox[HTML]{FEFEFF}{\textcolor{black}{\tiny\ttfamily\strut  handle}}\hskip 0pt plus 1pt\relax\allowbreak{}\colorbox[HTML]{FFFFFF}{\textcolor{black}{\tiny\ttfamily\strut  more}}\hskip 0pt plus 1pt\relax\allowbreak{}\colorbox[HTML]{DEDEFF}{\textcolor{black}{\tiny\ttfamily\strut  data}}\hskip 0pt plus 1pt\relax\allowbreak{}\colorbox[HTML]{E5E5FF}{\textcolor{black}{\tiny\ttfamily\strut  at}}\hskip 0pt plus 1pt\relax\allowbreak{}\colorbox[HTML]{FFFEFE}{\textcolor{black}{\tiny\ttfamily\strut  once}}\hskip 0pt plus 1pt\relax\allowbreak{}\colorbox[HTML]{FDFDFF}{\textcolor{black}{\tiny\ttfamily\strut .\textcolor{gray}{$\hookleftarrow$}}}\hskip 0pt plus 1pt\relax\allowbreak{}\colorbox[HTML]{FFFFFF}{\textcolor{black}{\tiny\ttfamily\strut 3}}\hskip 0pt plus 1pt\relax\allowbreak{}\colorbox[HTML]{FFFFFF}{\textcolor{black}{\tiny\ttfamily\strut .}}\hskip 0pt plus 1pt\relax\allowbreak{}\colorbox[HTML]{FFFFFF}{\textcolor{black}{\tiny\ttfamily\strut  Hard}}\hskip 0pt plus 1pt\relax\allowbreak{}\colorbox[HTML]{FFFEFE}{\textcolor{black}{\tiny\ttfamily\strut  Disk}}\hskip 0pt plus 1pt\relax\allowbreak{}\colorbox[HTML]{FFFFFF}{\textcolor{black}{\tiny\ttfamily\strut  Drive}}\hskip 0pt plus 1pt\relax\allowbreak{}\colorbox[HTML]{FFFFFF}{\textcolor{black}{\tiny\ttfamily\strut  (}}\hskip 0pt plus 1pt\relax\allowbreak{}\colorbox[HTML]{FFFFFF}{\textcolor{black}{\tiny\ttfamily\strut H}}\hskip 0pt plus 1pt\relax\allowbreak{}\colorbox[HTML]{FFFFFF}{\textcolor{black}{\tiny\ttfamily\strut DD}}\hskip 0pt plus 1pt\relax\allowbreak{}\colorbox[HTML]{FEFEFF}{\textcolor{black}{\tiny\ttfamily\strut )}}\hskip 0pt plus 1pt\relax\allowbreak{}\colorbox[HTML]{FAFAFF}{\textcolor{black}{\tiny\ttfamily\strut  or}}\hskip 0pt plus 1pt\relax\allowbreak{}\colorbox[HTML]{FFFFFF}{\textcolor{black}{\tiny\ttfamily\strut  Solid}}\hskip 0pt plus 1pt\relax\allowbreak{}\colorbox[HTML]{FFFFFF}{\textcolor{black}{\tiny\ttfamily\strut  State}}\hskip 0pt plus 1pt\relax\allowbreak{}\colorbox[HTML]{FFFFFF}{\textcolor{black}{\tiny\ttfamily\strut  Drive}}\hskip 0pt plus 1pt\relax\allowbreak{}\colorbox[HTML]{FFFFFF}{\textcolor{black}{\tiny\ttfamily\strut  (}}\hskip 0pt plus 1pt\relax\allowbreak{}\colorbox[HTML]{FFFFFF}{\textcolor{black}{\tiny\ttfamily\strut SS}}\hskip 0pt plus 1pt\relax\allowbreak{}\colorbox[HTML]{FFFFFF}{\textcolor{black}{\tiny\ttfamily\strut D}}\hskip 0pt plus 1pt\relax\allowbreak{}\colorbox[HTML]{FFFFFF}{\textcolor{black}{\tiny\ttfamily\strut ):}}\hskip 0pt plus 1pt\relax\allowbreak{}\colorbox[HTML]{B0B0FF}{\textcolor{black}{\tiny\ttfamily\strut  The}}\hskip 0pt plus 1pt\relax\allowbreak{}\colorbox[HTML]{FFF5F5}{\textcolor{black}{\tiny\ttfamily\strut  HDD}}\hskip 0pt plus 1pt\relax\allowbreak{}\colorbox[HTML]{FFFFFF}{\textcolor{black}{\tiny\ttfamily\strut  or}}\hskip 0pt plus 1pt\relax\allowbreak{}\colorbox[HTML]{FFFFFF}{\textcolor{black}{\tiny\ttfamily\strut  SSD}}\hskip 0pt plus 1pt\relax\allowbreak{}\colorbox[HTML]{C0C0FF}{\textcolor{black}{\tiny\ttfamily\strut  is}}\hskip 0pt plus 1pt\relax\allowbreak{}\colorbox[HTML]{B8B8FF}{\textcolor{black}{\tiny\ttfamily\strut  used}}\hskip 0pt plus 1pt\relax\allowbreak{}\colorbox[HTML]{DFDFFF}{\textcolor{black}{\tiny\ttfamily\strut  to}}\hskip 0pt plus 1pt\relax\allowbreak{}\colorbox[HTML]{FFFFFF}{\textcolor{black}{\tiny\ttfamily\strut  store}}\hskip 0pt plus 1pt\relax\allowbreak{}\colorbox[HTML]{FFEAEA}{\textcolor{black}{\tiny\ttfamily\strut  the}}\hskip 0pt plus 1pt\relax\allowbreak{}\colorbox[HTML]{FFFEFE}{\textcolor{black}{\tiny\ttfamily\strut  computer}}\hskip 0pt plus 1pt\relax\allowbreak{}\colorbox[HTML]{FEFEFF}{\textcolor{black}{\tiny\ttfamily\strut 's}}\hskip 0pt plus 1pt\relax\allowbreak{}\colorbox[HTML]{FCFCFF}{\textcolor{black}{\tiny\ttfamily\strut  operating}}\hskip 0pt plus 1pt\relax\allowbreak{}\colorbox[HTML]{FFFFFF}{\textcolor{black}{\tiny\ttfamily\strut  system}}\hskip 0pt plus 1pt\relax\allowbreak{}\colorbox[HTML]{FFFFFF}{\textcolor{black}{\tiny\ttfamily\strut ,}}\hskip 0pt plus 1pt\relax\allowbreak{}\colorbox[HTML]{FFFCFC}{\textcolor{black}{\tiny\ttfamily\strut  applications}}\hskip 0pt plus 1pt\relax\allowbreak{}\colorbox[HTML]{FFFFFF}{\textcolor{black}{\tiny\ttfamily\strut ,}}\hskip 0pt plus 1pt\relax\allowbreak{}\colorbox[HTML]{FFFEFE}{\textcolor{black}{\tiny\ttfamily\strut  and}}\hskip 0pt plus 1pt\relax\allowbreak{}\colorbox[HTML]{FFFBFB}{\textcolor{black}{\tiny\ttfamily\strut  user}}\hskip 0pt plus 1pt\relax\allowbreak{}\colorbox[HTML]{FEFEFF}{\textcolor{black}{\tiny\ttfamily\strut  data}}\hskip 0pt plus 1pt\relax\allowbreak{}\colorbox[HTML]{F5F5FF}{\textcolor{black}{\tiny\ttfamily\strut .}}\hskip 0pt plus 1pt\relax\allowbreak{}\colorbox[HTML]{E7E7FF}{\textcolor{black}{\tiny\ttfamily\strut  A}}\hskip 0pt plus 1pt\relax\allowbreak{}\colorbox[HTML]{FFF9F9}{\textcolor{black}{\tiny\ttfamily\strut  faster}}\hskip 0pt plus 1pt\relax\allowbreak{}\colorbox[HTML]{FFF7F7}{\textcolor{black}{\tiny\ttfamily\strut  drive}}\hskip 0pt plus 1pt\relax\allowbreak{}\colorbox[HTML]{FFFFFF}{\textcolor{black}{\tiny\ttfamily\strut  can}}\hskip 0pt plus 1pt\relax\allowbreak{}\colorbox[HTML]{F3F3FF}{\textcolor{black}{\tiny\ttfamily\strut  improve}}\hskip 0pt plus 1pt\relax\allowbreak{}\colorbox[HTML]{EDEDFF}{\textcolor{black}{\tiny\ttfamily\strut  the}}\hskip 0pt plus 1pt\relax\allowbreak{}\colorbox[HTML]{EDEDFF}{\textcolor{black}{\tiny\ttfamily\strut  computer}}\hskip 0pt plus 1pt\relax\allowbreak{}\colorbox[HTML]{FEFEFF}{\textcolor{black}{\tiny\ttfamily\strut 's}}\hskip 0pt plus 1pt\relax\allowbreak{}\colorbox[HTML]{FAFAFF}{\textcolor{black}{\tiny\ttfamily\strut  overall}}\hskip 0pt plus 1pt\relax\allowbreak{}\colorbox[HTML]{CACAFF}{\textcolor{black}{\tiny\ttfamily\strut  performance}}\hskip 0pt plus 1pt\relax\allowbreak{}\colorbox[HTML]{BDBDFF}{\textcolor{black}{\tiny\ttfamily\strut .\textcolor{gray}{$\hookleftarrow$}}}\hskip 0pt plus 1pt\relax\allowbreak{}\colorbox[HTML]{FDFDFF}{\textcolor{black}{\tiny\ttfamily\strut 4}}\hskip 0pt plus 1pt\relax\allowbreak{}\colorbox[HTML]{FFFFFF}{\textcolor{black}{\tiny\ttfamily\strut .}}\hskip 0pt plus 1pt\relax\allowbreak{}\colorbox[HTML]{FAFAFF}{\textcolor{black}{\tiny\ttfamily\strut  Graphics}}\hskip 0pt plus 1pt\relax\allowbreak{}\colorbox[HTML]{FFFFFF}{\textcolor{black}{\tiny\ttfamily\strut  Processing}}\hskip 0pt plus 1pt\relax\allowbreak{}\colorbox[HTML]{FFFFFF}{\textcolor{black}{\tiny\ttfamily\strut  Unit}}\hskip 0pt plus 1pt\relax\allowbreak{}\colorbox[HTML]{FFFFFF}{\textcolor{black}{\tiny\ttfamily\strut  (}}\hskip 0pt plus 1pt\relax\allowbreak{}\colorbox[HTML]{FFFFFF}{\textcolor{black}{\tiny\ttfamily\strut GPU}}\hskip 0pt plus 1pt\relax\allowbreak{}\colorbox[HTML]{FFFFFF}{\textcolor{black}{\tiny\ttfamily\strut ):}}\hskip 0pt plus 1pt\relax\allowbreak{}\colorbox[HTML]{FFFFFF}{\textcolor{black}{\tiny\ttfamily\strut  The}}\hskip 0pt plus 1pt\relax\allowbreak{}\colorbox[HTML]{FFFFFF}{\textcolor{black}{\tiny\ttfamily\strut  GPU}}\hskip 0pt plus 1pt\relax\allowbreak{}\colorbox[HTML]{E3E3FF}{\textcolor{black}{\tiny\ttfamily\strut  is}}\hskip 0pt plus 1pt\relax\allowbreak{}\colorbox[HTML]{E1E1FF}{\textcolor{black}{\tiny\ttfamily\strut  responsible}}\hskip 0pt plus 1pt\relax\allowbreak{}\colorbox[HTML]{FFFFFF}{\textcolor{black}{\tiny\ttfamily\strut  for}}\hskip 0pt plus 1pt\relax\allowbreak{}\colorbox[HTML]{FEFEFF}{\textcolor{black}{\tiny\ttfamily\strut  rendering}}\hskip 0pt plus 1pt\relax\allowbreak{}\colorbox[HTML]{FFE6E6}{\textcolor{black}{\tiny\ttfamily\strut  graphical}}\hskip 0pt plus 1pt\relax\allowbreak{}\colorbox[HTML]{FEFEFF}{\textcolor{black}{\tiny\ttfamily\strut  content}}\hskip 0pt plus 1pt\relax\allowbreak{}\colorbox[HTML]{FFFEFE}{\textcolor{black}{\tiny\ttfamily\strut  and}}\hskip 0pt plus 1pt\relax\allowbreak{}\colorbox[HTML]{FAFAFF}{\textcolor{black}{\tiny\ttfamily\strut  is}}\hskip 0pt plus 1pt\relax\allowbreak{}\colorbox[HTML]{FFD9D9}{\textcolor{black}{\tiny\ttfamily\strut  crucial}}\hskip 0pt plus 1pt\relax\allowbreak{}\colorbox[HTML]{FFFFFF}{\textcolor{black}{\tiny\ttfamily\strut  for}}\hskip 0pt plus 1pt\relax\allowbreak{}\colorbox[HTML]{E9E9FF}{\textcolor{black}{\tiny\ttfamily\strut  computer}}\hskip 0pt plus 1pt\relax\allowbreak{}\colorbox[HTML]{FFEDED}{\textcolor{black}{\tiny\ttfamily\strut -generated}}\hskip 0pt plus 1pt\relax\allowbreak{}\colorbox[HTML]{DADAFF}{\textcolor{black}{\tiny\ttfamily\strut  graphics}}\hskip 0pt plus 1pt\relax\allowbreak{}\colorbox[HTML]{FFE1E1}{\textcolor{black}{\tiny\ttfamily\strut  and}}\hskip 0pt plus 1pt\relax\allowbreak{}\colorbox[HTML]{FFFFFF}{\textcolor{black}{\tiny\ttfamily\strut  video}}\hskip 0pt plus 1pt\relax\allowbreak{}\colorbox[HTML]{CECEFF}{\textcolor{black}{\tiny\ttfamily\strut  games}}\hskip 0pt plus 1pt\relax\allowbreak{}\colorbox[HTML]{FFF6F6}{\textcolor{black}{\tiny\ttfamily\strut .}}\hskip 0pt plus 1pt\relax\allowbreak{}\colorbox[HTML]{D4D4FF}{\textcolor{black}{\tiny\ttfamily\strut  A}}\hskip 0pt plus 1pt\relax\allowbreak{}\colorbox[HTML]{FFE4E4}{\textcolor{black}{\tiny\ttfamily\strut  strong}}\hskip 0pt plus 1pt\relax\allowbreak{}\colorbox[HTML]{FFFFFF}{\textcolor{black}{\tiny\ttfamily\strut  GPU}}\hskip 0pt plus 1pt\relax\allowbreak{}\colorbox[HTML]{FDFDFF}{\textcolor{black}{\tiny\ttfamily\strut  can}}\hskip 0pt plus 1pt\relax\allowbreak{}\colorbox[HTML]{FFE9E9}{\textcolor{black}{\tiny\ttfamily\strut  greatly}}\hskip 0pt plus 1pt\relax\allowbreak{}\colorbox[HTML]{FEFEFF}{\textcolor{black}{\tiny\ttfamily\strut  enhance}}\hskip 0pt plus 1pt\relax\allowbreak{}\colorbox[HTML]{EBEBFF}{\textcolor{black}{\tiny\ttfamily\strut  the}}\hskip 0pt plus 1pt\relax\allowbreak{}\colorbox[HTML]{D1D1FF}{\textcolor{black}{\tiny\ttfamily\strut  computer}}\hskip 0pt plus 1pt\relax\allowbreak{}\colorbox[HTML]{FFFFFF}{\textcolor{black}{\tiny\ttfamily\strut 's}}\hskip 0pt plus 1pt\relax\allowbreak{}\colorbox[HTML]{C8C8FF}{\textcolor{black}{\tiny\ttfamily\strut  gaming}}\hskip 0pt plus 1pt\relax\allowbreak{}\colorbox[HTML]{E9E9FF}{\textcolor{black}{\tiny\ttfamily\strut  performance}}\hskip 0pt plus 1pt\relax\allowbreak{}\colorbox[HTML]{D2D2FF}{\textcolor{black}{\tiny\ttfamily\strut .\textcolor{gray}{$\hookleftarrow$}}}\hskip 0pt plus 1pt\relax\allowbreak{}\colorbox[HTML]{FEFEFF}{\textcolor{black}{\tiny\ttfamily\strut 5}}\hskip 0pt plus 1pt\relax\allowbreak{}\colorbox[HTML]{FFFFFF}{\textcolor{black}{\tiny\ttfamily\strut .}}\hskip 0pt plus 1pt\relax\allowbreak{}\colorbox[HTML]{F6F6FF}{\textcolor{black}{\tiny\ttfamily\strut  Mother}}\hskip 0pt plus 1pt\relax\allowbreak{}\colorbox[HTML]{FFFFFF}{\textcolor{black}{\tiny\ttfamily\strut board}}\hskip 0pt plus 1pt\relax\allowbreak{}\colorbox[HTML]{FFFFFF}{\textcolor{black}{\tiny\ttfamily\strut :}}\hskip 0pt plus 1pt\relax\allowbreak{}\colorbox[HTML]{F8F8FF}{\textcolor{black}{\tiny\ttfamily\strut  The}}\hskip 0pt plus 1pt\relax\allowbreak{}\colorbox[HTML]{FFFFFF}{\textcolor{black}{\tiny\ttfamily\strut  motherboard}}\hskip 0pt plus 1pt\relax\allowbreak{}\colorbox[HTML]{D7D7FF}{\textcolor{black}{\tiny\ttfamily\strut  is}}\hskip 0pt plus 1pt\relax\allowbreak{}\colorbox[HTML]{F9F9FF}{\textcolor{black}{\tiny\ttfamily\strut  the}}\hskip 0pt plus 1pt\relax\allowbreak{}\colorbox[HTML]{FFF7F7}{\textcolor{black}{\tiny\ttfamily\strut  backbone}}\hskip 0pt plus 1pt\relax\allowbreak{}\colorbox[HTML]{FBFBFF}{\textcolor{black}{\tiny\ttfamily\strut  of}}\hskip 0pt plus 1pt\relax\allowbreak{}\colorbox[HTML]{FEFEFF}{\textcolor{black}{\tiny\ttfamily\strut  the}}\hskip 0pt plus 1pt\relax\allowbreak{}\colorbox[HTML]{FFFFFF}{\textcolor{black}{\tiny\ttfamily\strut  computer}}\hskip 0pt plus 1pt\relax\allowbreak{}\colorbox[HTML]{FFF7F7}{\textcolor{black}{\tiny\ttfamily\strut ,}}\hskip 0pt plus 1pt\relax\allowbreak{}\colorbox[HTML]{F7F7FF}{\textcolor{black}{\tiny\ttfamily\strut  connecting}}\hskip 0pt plus 1pt\relax\allowbreak{}\colorbox[HTML]{FAFAFF}{\textcolor{black}{\tiny\ttfamily\strut  all}}\hskip 0pt plus 1pt\relax\allowbreak{}\colorbox[HTML]{EBEBFF}{\textcolor{black}{\tiny\ttfamily\strut  the}}\hskip 0pt plus 1pt\relax\allowbreak{}\colorbox[HTML]{FEFEFF}{\textcolor{black}{\tiny\ttfamily\strut  hardware}}\hskip 0pt plus 1pt\relax\allowbreak{}\colorbox[HTML]{FFFFFF}{\textcolor{black}{\tiny\ttfamily\strut  components}}\hskip 0pt plus 1pt\relax\allowbreak{}\colorbox[HTML]{FFD2D2}{\textcolor{black}{\tiny\ttfamily\strut .}}\hskip 0pt plus 1pt\relax\allowbreak{}\colorbox[HTML]{FAFAFF}{\textcolor{black}{\tiny\ttfamily\strut  It}}\hskip 0pt plus 1pt\relax\allowbreak{}\colorbox[HTML]{FFE5E5}{\textcolor{black}{\tiny\ttfamily\strut  facilitates}}\hskip 0pt plus 1pt\relax\allowbreak{}\colorbox[HTML]{F4F4FF}{\textcolor{black}{\tiny\ttfamily\strut  communication}}\hskip 0pt plus 1pt\relax\allowbreak{}\colorbox[HTML]{F0F0FF}{\textcolor{black}{\tiny\ttfamily\strut  between}}\hskip 0pt plus 1pt\relax\allowbreak{}\colorbox[HTML]{FFFFFF}{\textcolor{black}{\tiny\ttfamily\strut  the}}\hskip 0pt plus 1pt\relax\allowbreak{}\colorbox[HTML]{FFFFFF}{\textcolor{black}{\tiny\ttfamily\strut  CPU}}\hskip 0pt plus 1pt\relax\allowbreak{}\colorbox[HTML]{FEFEFF}{\textcolor{black}{\tiny\ttfamily\strut ,}}\hskip 0pt plus 1pt\relax\allowbreak{}\colorbox[HTML]{FEFEFF}{\textcolor{black}{\tiny\ttfamily\strut  RAM}}\hskip 0pt plus 1pt\relax\allowbreak{}\colorbox[HTML]{FFFFFF}{\textcolor{black}{\tiny\ttfamily\strut ,}}\hskip 0pt plus 1pt\relax\allowbreak{}\colorbox[HTML]{FFEFEF}{\textcolor{black}{\tiny\ttfamily\strut  and}}\hskip 0pt plus 1pt\relax\allowbreak{}\colorbox[HTML]{FDFDFF}{\textcolor{black}{\tiny\ttfamily\strut  other}}\hskip 0pt plus 1pt\relax\allowbreak{}\colorbox[HTML]{FFE8E8}{\textcolor{black}{\tiny\ttfamily\strut  peripherals}}\hskip 0pt plus 1pt\relax\allowbreak{}\colorbox[HTML]{9191FF}{\textcolor{black}{\tiny\ttfamily\strut .\textcolor{gray}{$\hookleftarrow$}}}\hskip 0pt plus 1pt\relax\allowbreak{}\colorbox[HTML]{FFFBFB}{\textcolor{black}{\tiny\ttfamily\strut Software}}\hskip 0pt plus 1pt\relax\allowbreak{}\colorbox[HTML]{FFFFFF}{\textcolor{black}{\tiny\ttfamily\strut  Components}}\hskip 0pt plus 1pt\relax\allowbreak{}\colorbox[HTML]{FFFFFF}{\textcolor{black}{\tiny\ttfamily\strut :\textcolor{gray}{$\hookleftarrow$}}}\hskip 0pt plus 1pt\relax\allowbreak{}\colorbox[HTML]{FEFEFF}{\textcolor{black}{\tiny\ttfamily\strut 1}}\hskip 0pt plus 1pt\relax\allowbreak{}\colorbox[HTML]{FFFFFF}{\textcolor{black}{\tiny\ttfamily\strut .}}\hskip 0pt plus 1pt\relax\allowbreak{}\colorbox[HTML]{FFFFFF}{\textcolor{black}{\tiny\ttfamily\strut  Operating}}\hskip 0pt plus 1pt\relax\allowbreak{}\colorbox[HTML]{FFFFFF}{\textcolor{black}{\tiny\ttfamily\strut  System}}\hskip 0pt plus 1pt\relax\allowbreak{}\colorbox[HTML]{FFFFFF}{\textcolor{black}{\tiny\ttfamily\strut  (}}\hskip 0pt plus 1pt\relax\allowbreak{}\colorbox[HTML]{FFFFFF}{\textcolor{black}{\tiny\ttfamily\strut OS}}\hskip 0pt plus 1pt\relax\allowbreak{}\colorbox[HTML]{FFFFFF}{\textcolor{black}{\tiny\ttfamily\strut ):}}\hskip 0pt plus 1pt\relax\allowbreak{}\colorbox[HTML]{FFFFFF}{\textcolor{black}{\tiny\ttfamily\strut  The}}\hskip 0pt plus 1pt\relax\allowbreak{}\colorbox[HTML]{FFFFFF}{\textcolor{black}{\tiny\ttfamily\strut  OS}}\hskip 0pt plus 1pt\relax\allowbreak{}\colorbox[HTML]{FFF4F4}{\textcolor{black}{\tiny\ttfamily\strut ,}}\hskip 0pt plus 1pt\relax\allowbreak{}\colorbox[HTML]{D5D5FF}{\textcolor{black}{\tiny\ttfamily\strut  such}}\hskip 0pt plus 1pt\relax\allowbreak{}\colorbox[HTML]{FFFFFF}{\textcolor{black}{\tiny\ttfamily\strut  as}}\hskip 0pt plus 1pt\relax\allowbreak{}\colorbox[HTML]{FFFFFF}{\textcolor{black}{\tiny\ttfamily\strut  Windows}}\hskip 0pt plus 1pt\relax\allowbreak{}\colorbox[HTML]{C1C1FF}{\textcolor{black}{\tiny\ttfamily\strut ,}}\hskip 0pt plus 1pt\relax\allowbreak{}\colorbox[HTML]{FEFEFF}{\textcolor{black}{\tiny\ttfamily\strut  macOS}}\hskip 0pt plus 1pt\relax\allowbreak{}\colorbox[HTML]{FFFFFF}{\textcolor{black}{\tiny\ttfamily\strut ,}}\hskip 0pt plus 1pt\relax\allowbreak{}\colorbox[HTML]{FFFFFF}{\textcolor{black}{\tiny\ttfamily\strut  or}}\hskip 0pt plus 1pt\relax\allowbreak{}\colorbox[HTML]{FFFFFF}{\textcolor{black}{\tiny\ttfamily\strut  Linux}}\hskip 0pt plus 1pt\relax\allowbreak{}\colorbox[HTML]{FFFFFF}{\textcolor{black}{\tiny\ttfamily\strut ,}}\hskip 0pt plus 1pt\relax\allowbreak{}\colorbox[HTML]{E5E5FF}{\textcolor{black}{\tiny\ttfamily\strut  is}}\hskip 0pt plus 1pt\relax\allowbreak{}\colorbox[HTML]{FFF5F5}{\textcolor{black}{\tiny\ttfamily\strut  the}}\hskip 0pt plus 1pt\relax\allowbreak{}\colorbox[HTML]{FBFBFF}{\textcolor{black}{\tiny\ttfamily\strut  software}}\hskip 0pt plus 1pt\relax\allowbreak{}\colorbox[HTML]{FDFDFF}{\textcolor{black}{\tiny\ttfamily\strut  that}}\hskip 0pt plus 1pt\relax\allowbreak{}\colorbox[HTML]{FFFFFF}{\textcolor{black}{\tiny\ttfamily\strut  manages}}\hskip 0pt plus 1pt\relax\allowbreak{}\colorbox[HTML]{E8E8FF}{\textcolor{black}{\tiny\ttfamily\strut  the}}\hskip 0pt plus 1pt\relax\allowbreak{}\colorbox[HTML]{F9F9FF}{\textcolor{black}{\tiny\ttfamily\strut  computer}}\hskip 0pt plus 1pt\relax\allowbreak{}\colorbox[HTML]{FEFEFF}{\textcolor{black}{\tiny\ttfamily\strut 's}}\hskip 0pt plus 1pt\relax\allowbreak{}\colorbox[HTML]{FFFFFF}{\textcolor{black}{\tiny\ttfamily\strut  hardware}}\hskip 0pt plus 1pt\relax\allowbreak{}\colorbox[HTML]{FDFDFF}{\textcolor{black}{\tiny\ttfamily\strut  resources}}\hskip 0pt plus 1pt\relax\allowbreak{}\colorbox[HTML]{FFFEFE}{\textcolor{black}{\tiny\ttfamily\strut  and}}\hskip 0pt plus 1pt\relax\allowbreak{}\colorbox[HTML]{FEFEFF}{\textcolor{black}{\tiny\ttfamily\strut  provides}}\hskip 0pt plus 1pt\relax\allowbreak{}\colorbox[HTML]{FDFDFF}{\textcolor{black}{\tiny\ttfamily\strut  a}}\hskip 0pt plus 1pt\relax\allowbreak{}\colorbox[HTML]{F7F7FF}{\textcolor{black}{\tiny\ttfamily\strut  user}}\hskip 0pt plus 1pt\relax\allowbreak{}\colorbox[HTML]{FFFFFF}{\textcolor{black}{\tiny\ttfamily\strut  interface}}\hskip 0pt plus 1pt\relax\allowbreak{}\colorbox[HTML]{B7B7FF}{\textcolor{black}{\tiny\ttfamily\strut  for}}\hskip 0pt plus 1pt\relax\allowbreak{}\colorbox[HTML]{ECECFF}{\textcolor{black}{\tiny\ttfamily\strut  interacting}}\hskip 0pt plus 1pt\relax\allowbreak{}\colorbox[HTML]{FFFFFF}{\textcolor{black}{\tiny\ttfamily\strut  with}}\hskip 0pt plus 1pt\relax\allowbreak{}\colorbox[HTML]{FEFEFF}{\textcolor{black}{\tiny\ttfamily\strut  the}}\hskip 0pt plus 1pt\relax\allowbreak{}\colorbox[HTML]{FCFCFF}{\textcolor{black}{\tiny\ttfamily\strut  computer}}\hskip 0pt plus 1pt\relax\allowbreak{}\colorbox[HTML]{EFEFFF}{\textcolor{black}{\tiny\ttfamily\strut .\textcolor{gray}{$\hookleftarrow$}}}\hskip 0pt plus 1pt\relax\allowbreak{}\colorbox[HTML]{FFFFFF}{\textcolor{black}{\tiny\ttfamily\strut 2}}\hskip 0pt plus 1pt\relax\allowbreak{}\colorbox[HTML]{FFFFFF}{\textcolor{black}{\tiny\ttfamily\strut .}}\hskip 0pt plus 1pt\relax\allowbreak{}\colorbox[HTML]{EAEAFF}{\textcolor{black}{\tiny\ttfamily\strut  Applications}}\hskip 0pt plus 1pt\relax\allowbreak{}\colorbox[HTML]{FFFFFF}{\textcolor{black}{\tiny\ttfamily\strut :}}\hskip 0pt plus 1pt\relax\allowbreak{}\colorbox[HTML]{FFFFFF}{\textcolor{black}{\tiny\ttfamily\strut  Applications}}\hskip 0pt plus 1pt\relax\allowbreak{}\colorbox[HTML]{FCFCFF}{\textcolor{black}{\tiny\ttfamily\strut ,}}\hskip 0pt plus 1pt\relax\allowbreak{}\colorbox[HTML]{D6D6FF}{\textcolor{black}{\tiny\ttfamily\strut  such}}\hskip 0pt plus 1pt\relax\allowbreak{}\colorbox[HTML]{FFFFFF}{\textcolor{black}{\tiny\ttfamily\strut  as}}\hskip 0pt plus 1pt\relax\allowbreak{}\colorbox[HTML]{FFF9F9}{\textcolor{black}{\tiny\ttfamily\strut  word}}\hskip 0pt plus 1pt\relax\allowbreak{}\colorbox[HTML]{FFFFFF}{\textcolor{black}{\tiny\ttfamily\strut  processors}}\hskip 0pt plus 1pt\relax\allowbreak{}\colorbox[HTML]{F6F6FF}{\textcolor{black}{\tiny\ttfamily\strut ,}}\hskip 0pt plus 1pt\relax\allowbreak{}\colorbox[HTML]{F2F2FF}{\textcolor{black}{\tiny\ttfamily\strut  web}}\hskip 0pt plus 1pt\relax\allowbreak{}\colorbox[HTML]{FFFFFF}{\textcolor{black}{\tiny\ttfamily\strut  browsers}}\hskip 0pt plus 1pt\relax\allowbreak{}\colorbox[HTML]{FFFFFF}{\textcolor{black}{\tiny\ttfamily\strut ,}}\hskip 0pt plus 1pt\relax\allowbreak{}\colorbox[HTML]{FFFFFF}{\textcolor{black}{\tiny\ttfamily\strut  and}}\hskip 0pt plus 1pt\relax\allowbreak{}\colorbox[HTML]{E5E5FF}{\textcolor{black}{\tiny\ttfamily\strut  games}}\hskip 0pt plus 1pt\relax\allowbreak{}\colorbox[HTML]{FFFFFF}{\textcolor{black}{\tiny\ttfamily\strut ,}}\hskip 0pt plus 1pt\relax\allowbreak{}\colorbox[HTML]{F5F5FF}{\textcolor{black}{\tiny\ttfamily\strut  run}}\hskip 0pt plus 1pt\relax\allowbreak{}\colorbox[HTML]{FFFFFF}{\textcolor{black}{\tiny\ttfamily\strut  on}}\hskip 0pt plus 1pt\relax\allowbreak{}\colorbox[HTML]{FBFBFF}{\textcolor{black}{\tiny\ttfamily\strut  top}}\hskip 0pt plus 1pt\relax\allowbreak{}\colorbox[HTML]{FFFFFF}{\textcolor{black}{\tiny\ttfamily\strut  of}}\hskip 0pt plus 1pt\relax\allowbreak{}\colorbox[HTML]{FFFFFF}{\textcolor{black}{\tiny\ttfamily\strut  the}}\hskip 0pt plus 1pt\relax\allowbreak{}\colorbox[HTML]{FCFCFF}{\textcolor{black}{\tiny\ttfamily\strut  OS}}\hskip 0pt plus 1pt\relax\allowbreak{}\colorbox[HTML]{F8F8FF}{\textcolor{black}{\tiny\ttfamily\strut  and}}\hskip 0pt plus 1pt\relax\allowbreak{}\colorbox[HTML]{F6F6FF}{\textcolor{black}{\tiny\ttfamily\strut  provide}}\hskip 0pt plus 1pt\relax\allowbreak{}\colorbox[HTML]{FFFFFF}{\textcolor{black}{\tiny\ttfamily\strut  specific}}\hskip 0pt plus 1pt\relax\allowbreak{}\colorbox[HTML]{D6D6FF}{\textcolor{black}{\tiny\ttfamily\strut  functionality}}\hskip 0pt plus 1pt\relax\allowbreak{}\colorbox[HTML]{FEFEFF}{\textcolor{black}{\tiny\ttfamily\strut  to}}\hskip 0pt plus 1pt\relax\allowbreak{}\colorbox[HTML]{FEFEFF}{\textcolor{black}{\tiny\ttfamily\strut  the}}\hskip 0pt plus 1pt\relax\allowbreak{}\colorbox[HTML]{FFFFFF}{\textcolor{black}{\tiny\ttfamily\strut  user}}\hskip 0pt plus 1pt\relax\allowbreak{}\colorbox[HTML]{F3F3FF}{\textcolor{black}{\tiny\ttfamily\strut .\textcolor{gray}{$\hookleftarrow$}}}\hskip 0pt plus 1pt\relax\allowbreak{}\colorbox[HTML]{FFFCFC}{\textcolor{black}{\tiny\ttfamily\strut 3}}\hskip 0pt plus 1pt\relax\allowbreak{}\colorbox[HTML]{FFFFFF}{\textcolor{black}{\tiny\ttfamily\strut .}}\hskip 0pt plus 1pt\relax\allowbreak{}\colorbox[HTML]{FFFFFF}{\textcolor{black}{\tiny\ttfamily\strut  Drivers}}\hskip 0pt plus 1pt\relax\allowbreak{}\colorbox[HTML]{FFFFFF}{\textcolor{black}{\tiny\ttfamily\strut :}}\hskip 0pt plus 1pt\relax\allowbreak{}\colorbox[HTML]{FFFFFF}{\textcolor{black}{\tiny\ttfamily\strut  Drivers}}\hskip 0pt plus 1pt\relax\allowbreak{}\colorbox[HTML]{F6F6FF}{\textcolor{black}{\tiny\ttfamily\strut  are}}\hskip 0pt plus 1pt\relax\allowbreak{}\colorbox[HTML]{FDFDFF}{\textcolor{black}{\tiny\ttfamily\strut  software}}\hskip 0pt plus 1pt\relax\allowbreak{}\colorbox[HTML]{FFEDED}{\textcolor{black}{\tiny\ttfamily\strut  that}}\hskip 0pt plus 1pt\relax\allowbreak{}\colorbox[HTML]{FFFFFF}{\textcolor{black}{\tiny\ttfamily\strut  enable}}\hskip 0pt plus 1pt\relax\allowbreak{}\colorbox[HTML]{FFECEC}{\textcolor{black}{\tiny\ttfamily\strut  hardware}}\hskip 0pt plus 1pt\relax\allowbreak{}\colorbox[HTML]{FFFFFF}{\textcolor{black}{\tiny\ttfamily\strut  components}}\hskip 0pt plus 1pt\relax\allowbreak{}\colorbox[HTML]{FFFFFF}{\textcolor{black}{\tiny\ttfamily\strut  to}}\hskip 0pt plus 1pt\relax\allowbreak{}\colorbox[HTML]{FDFDFF}{\textcolor{black}{\tiny\ttfamily\strut  communicate}}\hskip 0pt plus 1pt\relax\allowbreak{}\colorbox[HTML]{FDFDFF}{\textcolor{black}{\tiny\ttfamily\strut  with}}\hskip 0pt plus 1pt\relax\allowbreak{}\colorbox[HTML]{FFFFFF}{\textcolor{black}{\tiny\ttfamily\strut  the}}\hskip 0pt plus 1pt\relax\allowbreak{}\colorbox[HTML]{FFD3D3}{\textcolor{black}{\tiny\ttfamily\strut  operating}}\hskip 0pt plus 1pt\relax\allowbreak{}\colorbox[HTML]{FFFFFF}{\textcolor{black}{\tiny\ttfamily\strut  system}}\hskip 0pt plus 1pt\relax\allowbreak{}\colorbox[HTML]{FFFDFD}{\textcolor{black}{\tiny\ttfamily\strut .}}\hskip 0pt plus 1pt\relax\allowbreak{}\colorbox[HTML]{FDFDFF}{\textcolor{black}{\tiny\ttfamily\strut  Without}}\hskip 0pt plus 1pt\relax\allowbreak{}\colorbox[HTML]{EFEFFF}{\textcolor{black}{\tiny\ttfamily\strut  drivers}}\hskip 0pt plus 1pt\relax\allowbreak{}\colorbox[HTML]{FFFFFF}{\textcolor{black}{\tiny\ttfamily\strut ,}}\hskip 0pt plus 1pt\relax\allowbreak{}\colorbox[HTML]{FBFBFF}{\textcolor{black}{\tiny\ttfamily\strut  the}}\hskip 0pt plus 1pt\relax\allowbreak{}\colorbox[HTML]{FFFCFC}{\textcolor{black}{\tiny\ttfamily\strut  hardware}}\hskip 0pt plus 1pt\relax\allowbreak{}\colorbox[HTML]{EFEFFF}{\textcolor{black}{\tiny\ttfamily\strut  components}}\hskip 0pt plus 1pt\relax\allowbreak{}\colorbox[HTML]{CCCCFF}{\textcolor{black}{\tiny\ttfamily\strut  may}}\hskip 0pt plus 1pt\relax\allowbreak{}\colorbox[HTML]{FEFEFF}{\textcolor{black}{\tiny\ttfamily\strut  not}}\hskip 0pt plus 1pt\relax\allowbreak{}\colorbox[HTML]{FFFEFE}{\textcolor{black}{\tiny\ttfamily\strut  function}}\hskip 0pt plus 1pt\relax\allowbreak{}\colorbox[HTML]{FEFEFF}{\textcolor{black}{\tiny\ttfamily\strut  properly}}\hskip 0pt plus 1pt\relax\allowbreak{}\colorbox[HTML]{FCFCFF}{\textcolor{black}{\tiny\ttfamily\strut .\textcolor{gray}{$\hookleftarrow$}}}\hskip 0pt plus 1pt\relax\allowbreak{}\colorbox[HTML]{FFD4D4}{\textcolor{black}{\tiny\ttfamily\strut 4}}\hskip 0pt plus 1pt\relax\allowbreak{}\colorbox[HTML]{FFFFFF}{\textcolor{black}{\tiny\ttfamily\strut .}}\hskip 0pt plus 1pt\relax\allowbreak{}\colorbox[HTML]{E6E6FF}{\textcolor{black}{\tiny\ttfamily\strut  Ant}}\hskip 0pt plus 1pt\relax\allowbreak{}\colorbox[HTML]{FFFFFF}{\textcolor{black}{\tiny\ttfamily\strut ivirus}}\hskip 0pt plus 1pt\relax\allowbreak{}\colorbox[HTML]{FFEDED}{\textcolor{black}{\tiny\ttfamily\strut  Software}}\hskip 0pt plus 1pt\relax\allowbreak{}\colorbox[HTML]{FFFFFF}{\textcolor{black}{\tiny\ttfamily\strut :}}\hskip 0pt plus 1pt\relax\allowbreak{}\colorbox[HTML]{FFFFFF}{\textcolor{black}{\tiny\ttfamily\strut  Ant}}\hskip 0pt plus 1pt\relax\allowbreak{}\colorbox[HTML]{FFFFFF}{\textcolor{black}{\tiny\ttfamily\strut ivirus}}\hskip 0pt plus 1pt\relax\allowbreak{}\colorbox[HTML]{FFFFFF}{\textcolor{black}{\tiny\ttfamily\strut  software}}\hskip 0pt plus 1pt\relax\allowbreak{}\colorbox[HTML]{C5C5FF}{\textcolor{black}{\tiny\ttfamily\strut  helps}}\hskip 0pt plus 1pt\relax\allowbreak{}\colorbox[HTML]{FFFFFF}{\textcolor{black}{\tiny\ttfamily\strut  protect}}\hskip 0pt plus 1pt\relax\allowbreak{}\colorbox[HTML]{FFFFFF}{\textcolor{black}{\tiny\ttfamily\strut  the}}\hskip 0pt plus 1pt\relax\allowbreak{}\colorbox[HTML]{FFFFFF}{\textcolor{black}{\tiny\ttfamily\strut  computer}}\hskip 0pt plus 1pt\relax\allowbreak{}\colorbox[HTML]{FFFFFF}{\textcolor{black}{\tiny\ttfamily\strut  from}}\hskip 0pt plus 1pt\relax\allowbreak{}\colorbox[HTML]{FDFDFF}{\textcolor{black}{\tiny\ttfamily\strut  malware}}\hskip 0pt plus 1pt\relax\allowbreak{}\colorbox[HTML]{FFFFFF}{\textcolor{black}{\tiny\ttfamily\strut  and}}\hskip 0pt plus 1pt\relax\allowbreak{}\colorbox[HTML]{E0E0FF}{\textcolor{black}{\tiny\ttfamily\strut  other}}\hskip 0pt plus 1pt\relax\allowbreak{}\colorbox[HTML]{FDFDFF}{\textcolor{black}{\tiny\ttfamily\strut  security}}\hskip 0pt plus 1pt\relax\allowbreak{}\colorbox[HTML]{FFFFFF}{\textcolor{black}{\tiny\ttfamily\strut  threats}}\hskip 0pt plus 1pt\relax\allowbreak{}\colorbox[HTML]{9B9BFF}{\textcolor{black}{\tiny\ttfamily\strut .\textcolor{gray}{$\hookleftarrow$}}}\hskip 0pt plus 1pt\relax\allowbreak{}\colorbox[HTML]{FF8888}{\textcolor{black}{\tiny\ttfamily\strut 5}}\hskip 0pt plus 1pt\relax\allowbreak{}\colorbox[HTML]{FFFFFF}{\textcolor{black}{\tiny\ttfamily\strut .}}\hskip 0pt plus 1pt\relax\allowbreak{}\colorbox[HTML]{E3E3FF}{\textcolor{black}{\tiny\ttfamily\strut  Software}}\hskip 0pt plus 1pt\relax\allowbreak{}\colorbox[HTML]{FEFEFF}{\textcolor{black}{\tiny\ttfamily\strut  Updates}}\hskip 0pt plus 1pt\relax\allowbreak{}\colorbox[HTML]{FFFFFF}{\textcolor{black}{\tiny\ttfamily\strut :}}\hskip 0pt plus 1pt\relax\allowbreak{}\colorbox[HTML]{FFFAFA}{\textcolor{black}{\tiny\ttfamily\strut  Regular}}\hskip 0pt plus 1pt\relax\allowbreak{}\colorbox[HTML]{F3F3FF}{\textcolor{black}{\tiny\ttfamily\strut  software}}\hskip 0pt plus 1pt\relax\allowbreak{}\colorbox[HTML]{FFFFFF}{\textcolor{black}{\tiny\ttfamily\strut  updates}}\hskip 0pt plus 1pt\relax\allowbreak{}\colorbox[HTML]{FFEEEE}{\textcolor{black}{\tiny\ttfamily\strut  ensure}}\hskip 0pt plus 1pt\relax\allowbreak{}\colorbox[HTML]{FFD6D6}{\textcolor{black}{\tiny\ttfamily\strut  the}}\hskip 0pt plus 1pt\relax\allowbreak{}\colorbox[HTML]{F2F2FF}{\textcolor{black}{\tiny\ttfamily\strut  computer}}\hskip 0pt plus 1pt\relax\allowbreak{}\colorbox[HTML]{FFDADA}{\textcolor{black}{\tiny\ttfamily\strut  remains}}\hskip 0pt plus 1pt\relax\allowbreak{}\colorbox[HTML]{FFF6F6}{\textcolor{black}{\tiny\ttfamily\strut  secure}}\hskip 0pt plus 1pt\relax\allowbreak{}\colorbox[HTML]{FEFEFF}{\textcolor{black}{\tiny\ttfamily\strut  and}}\hskip 0pt plus 1pt\relax\allowbreak{}\colorbox[HTML]{F1F1FF}{\textcolor{black}{\tiny\ttfamily\strut  runs}}\hskip 0pt plus 1pt\relax\allowbreak{}\colorbox[HTML]{F8F8FF}{\textcolor{black}{\tiny\ttfamily\strut  at}}\hskip 0pt plus 1pt\relax\allowbreak{}\colorbox[HTML]{FEFEFF}{\textcolor{black}{\tiny\ttfamily\strut  its}}\hskip 0pt plus 1pt\relax\allowbreak{}\colorbox[HTML]{E5E5FF}{\textcolor{black}{\tiny\ttfamily\strut  best}}\hskip 0pt plus 1pt\relax\allowbreak{}\colorbox[HTML]{FFFDFD}{\textcolor{black}{\tiny\ttfamily\strut  performance}}\hskip 0pt plus 1pt\relax\allowbreak{}\colorbox[HTML]{A7A7FF}{\textcolor{black}{\tiny\ttfamily\strut .\textcolor{gray}{$\hookleftarrow$}}}\hskip 0pt plus 1pt\relax\allowbreak{}\colorbox[HTML]{FF5959}{\textcolor{white}{\tiny\ttfamily\strut The}}\hskip 0pt plus 1pt\relax\allowbreak{}\colorbox[HTML]{FFFFFF}{\textcolor{black}{\tiny\ttfamily\strut  inter}}\hskip 0pt plus 1pt\relax\allowbreak{}\colorbox[HTML]{FFFFFF}{\textcolor{black}{\tiny\ttfamily\strut dependence}}\hskip 0pt plus 1pt\relax\allowbreak{}\colorbox[HTML]{FFFFFF}{\textcolor{black}{\tiny\ttfamily\strut  and}}\hskip 0pt plus 1pt\relax\allowbreak{}\colorbox[HTML]{FFFFFF}{\textcolor{black}{\tiny\ttfamily\strut  inter}}\hskip 0pt plus 1pt\relax\allowbreak{}\colorbox[HTML]{FFFFFF}{\textcolor{black}{\tiny\ttfamily\strut connection}}\hskip 0pt plus 1pt\relax\allowbreak{}\colorbox[HTML]{FFFFFF}{\textcolor{black}{\tiny\ttfamily\strut  between}}\hskip 0pt plus 1pt\relax\allowbreak{}\colorbox[HTML]{FBFBFF}{\textcolor{black}{\tiny\ttfamily\strut  these}}\hskip 0pt plus 1pt\relax\allowbreak{}\colorbox[HTML]{FFDFDF}{\textcolor{black}{\tiny\ttfamily\strut  hardware}}\hskip 0pt plus 1pt\relax\allowbreak{}\colorbox[HTML]{FFFFFF}{\textcolor{black}{\tiny\ttfamily\strut  and}}\hskip 0pt plus 1pt\relax\allowbreak{}\colorbox[HTML]{FFFFFF}{\textcolor{black}{\tiny\ttfamily\strut  software}}\hskip 0pt plus 1pt\relax\allowbreak{}\colorbox[HTML]{FFFFFF}{\textcolor{black}{\tiny\ttfamily\strut  components}}\hskip 0pt plus 1pt\relax\allowbreak{}\colorbox[HTML]{FFFEFE}{\textcolor{black}{\tiny\ttfamily\strut  are}}\hskip 0pt plus 1pt\relax\allowbreak{}\colorbox[HTML]{B6B6FF}{\textcolor{black}{\tiny\ttfamily\strut  critical}}\hskip 0pt plus 1pt\relax\allowbreak{}\colorbox[HTML]{FFF9F9}{\textcolor{black}{\tiny\ttfamily\strut  for}}\hskip 0pt plus 1pt\relax\allowbreak{}\colorbox[HTML]{DDDDFF}{\textcolor{black}{\tiny\ttfamily\strut  the}}\hskip 0pt plus 1pt\relax\allowbreak{}\colorbox[HTML]{FDFDFF}{\textcolor{black}{\tiny\ttfamily\strut  computer}}\hskip 0pt plus 1pt\relax\allowbreak{}\colorbox[HTML]{FFFDFD}{\textcolor{black}{\tiny\ttfamily\strut 's}}\hskip 0pt plus 1pt\relax\allowbreak{}\colorbox[HTML]{FFFEFE}{\textcolor{black}{\tiny\ttfamily\strut  optimal}}\hskip 0pt plus 1pt\relax\allowbreak{}\colorbox[HTML]{D2D2FF}{\textcolor{black}{\tiny\ttfamily\strut  performance}}\hskip 0pt plus 1pt\relax\allowbreak{}\colorbox[HTML]{D8D8FF}{\textcolor{black}{\tiny\ttfamily\strut  and}}\hskip 0pt plus 1pt\relax\allowbreak{}\colorbox[HTML]{FFFFFF}{\textcolor{black}{\tiny\ttfamily\strut  functionality}}\hskip 0pt plus 1pt\relax\allowbreak{}\colorbox[HTML]{FFF8F8}{\textcolor{black}{\tiny\ttfamily\strut .}}\hskip 0pt plus 1pt\relax\allowbreak{}\colorbox[HTML]{FFEAEA}{\textcolor{black}{\tiny\ttfamily\strut  The}}\hskip 0pt plus 1pt\relax\allowbreak{}\colorbox[HTML]{FFFFFF}{\textcolor{black}{\tiny\ttfamily\strut  CPU}}\hskip 0pt plus 1pt\relax\allowbreak{}\colorbox[HTML]{D7D7FF}{\textcolor{black}{\tiny\ttfamily\strut  and}}\hskip 0pt plus 1pt\relax\allowbreak{}\colorbox[HTML]{F6F6FF}{\textcolor{black}{\tiny\ttfamily\strut  RAM}}\hskip 0pt plus 1pt\relax\allowbreak{}\colorbox[HTML]{E7E7FF}{\textcolor{black}{\tiny\ttfamily\strut  work}}\hskip 0pt plus 1pt\relax\allowbreak{}\colorbox[HTML]{D6D6FF}{\textcolor{black}{\tiny\ttfamily\strut  together}}\hskip 0pt plus 1pt\relax\allowbreak{}\colorbox[HTML]{ADADFF}{\textcolor{black}{\tiny\ttfamily\strut  to}}\hskip 0pt plus 1pt\relax\allowbreak{}\colorbox[HTML]{FFFBFB}{\textcolor{black}{\tiny\ttfamily\strut  process}}\hskip 0pt plus 1pt\relax\allowbreak{}\colorbox[HTML]{F6F6FF}{\textcolor{black}{\tiny\ttfamily\strut  data}}\hskip 0pt plus 1pt\relax\allowbreak{}\colorbox[HTML]{FEFEFF}{\textcolor{black}{\tiny\ttfamily\strut ,}}\hskip 0pt plus 1pt\relax\allowbreak{}\colorbox[HTML]{A1A1FF}{\textcolor{black}{\tiny\ttfamily\strut  while}}\hskip 0pt plus 1pt\relax\allowbreak{}\colorbox[HTML]{FFFFFF}{\textcolor{black}{\tiny\ttfamily\strut  the}}\hskip 0pt plus 1pt\relax\allowbreak{}\colorbox[HTML]{FFF6F6}{\textcolor{black}{\tiny\ttfamily\strut  GPU}}\hskip 0pt plus 1pt\relax\allowbreak{}\colorbox[HTML]{FFEEEE}{\textcolor{black}{\tiny\ttfamily\strut  and}}\hskip 0pt plus 1pt\relax\allowbreak{}\colorbox[HTML]{FFFEFE}{\textcolor{black}{\tiny\ttfamily\strut  motherboard}}\hskip 0pt plus 1pt\relax\allowbreak{}\colorbox[HTML]{C1C1FF}{\textcolor{black}{\tiny\ttfamily\strut  work}}\hskip 0pt plus 1pt\relax\allowbreak{}\colorbox[HTML]{AAAAFF}{\textcolor{black}{\tiny\ttfamily\strut  together}}\hskip 0pt plus 1pt\relax\allowbreak{}\colorbox[HTML]{FFFFFF}{\textcolor{black}{\tiny\ttfamily\strut  to}}\hskip 0pt plus 1pt\relax\allowbreak{}\colorbox[HTML]{EAEAFF}{\textcolor{black}{\tiny\ttfamily\strut  render}}\hskip 0pt plus 1pt\relax\allowbreak{}\colorbox[HTML]{F5F5FF}{\textcolor{black}{\tiny\ttfamily\strut  graphics}}\hskip 0pt plus 1pt\relax\allowbreak{}\colorbox[HTML]{D8D8FF}{\textcolor{black}{\tiny\ttfamily\strut .}}\hskip 0pt plus 1pt\relax\allowbreak{}\colorbox[HTML]{FAFAFF}{\textcolor{black}{\tiny\ttfamily\strut  The}}\hskip 0pt plus 1pt\relax\allowbreak{}\colorbox[HTML]{FFDADA}{\textcolor{black}{\tiny\ttfamily\strut  operating}}\hskip 0pt plus 1pt\relax\allowbreak{}\colorbox[HTML]{FFFFFF}{\textcolor{black}{\tiny\ttfamily\strut  system}}\hskip 0pt plus 1pt\relax\allowbreak{}\colorbox[HTML]{EEEEFF}{\textcolor{black}{\tiny\ttfamily\strut  and}}\hskip 0pt plus 1pt\relax\allowbreak{}\colorbox[HTML]{EDEDFF}{\textcolor{black}{\tiny\ttfamily\strut  applications}}\hskip 0pt plus 1pt\relax\allowbreak{}\colorbox[HTML]{C6C6FF}{\textcolor{black}{\tiny\ttfamily\strut  work}}\hskip 0pt plus 1pt\relax\allowbreak{}\colorbox[HTML]{DCDCFF}{\textcolor{black}{\tiny\ttfamily\strut  together}}\hskip 0pt plus 1pt\relax\allowbreak{}\colorbox[HTML]{FFFFFF}{\textcolor{black}{\tiny\ttfamily\strut  to}}\hskip 0pt plus 1pt\relax\allowbreak{}\colorbox[HTML]{F7F7FF}{\textcolor{black}{\tiny\ttfamily\strut  provide}}\hskip 0pt plus 1pt\relax\allowbreak{}\colorbox[HTML]{FFFBFB}{\textcolor{black}{\tiny\ttfamily\strut  a}}\hskip 0pt plus 1pt\relax\allowbreak{}\colorbox[HTML]{EFEFFF}{\textcolor{black}{\tiny\ttfamily\strut  user}}\hskip 0pt plus 1pt\relax\allowbreak{}\colorbox[HTML]{EFEFFF}{\textcolor{black}{\tiny\ttfamily\strut  interface}}\hskip 0pt plus 1pt\relax\allowbreak{}\colorbox[HTML]{FFCDCD}{\textcolor{black}{\tiny\ttfamily\strut ,}}\hskip 0pt plus 1pt\relax\allowbreak{}\colorbox[HTML]{D1D1FF}{\textcolor{black}{\tiny\ttfamily\strut  while}}\hskip 0pt plus 1pt\relax\allowbreak{}\colorbox[HTML]{F8F8FF}{\textcolor{black}{\tiny\ttfamily\strut  the}}\hskip 0pt plus 1pt\relax\allowbreak{}\colorbox[HTML]{FDFDFF}{\textcolor{black}{\tiny\ttfamily\strut  drivers}}\hskip 0pt plus 1pt\relax\allowbreak{}\colorbox[HTML]{DBDBFF}{\textcolor{black}{\tiny\ttfamily\strut  enable}}\hskip 0pt plus 1pt\relax\allowbreak{}\colorbox[HTML]{DEDEFF}{\textcolor{black}{\tiny\ttfamily\strut  hardware}}\hskip 0pt plus 1pt\relax\allowbreak{}\colorbox[HTML]{EFEFFF}{\textcolor{black}{\tiny\ttfamily\strut  components}}\hskip 0pt plus 1pt\relax\allowbreak{}\colorbox[HTML]{FFFFFF}{\textcolor{black}{\tiny\ttfamily\strut  to}}\hskip 0pt plus 1pt\relax\allowbreak{}\colorbox[HTML]{EDEDFF}{\textcolor{black}{\tiny\ttfamily\strut  communicate}}\hskip 0pt plus 1pt\relax\allowbreak{}\colorbox[HTML]{FCFCFF}{\textcolor{black}{\tiny\ttfamily\strut  with}}\hskip 0pt plus 1pt\relax\allowbreak{}\colorbox[HTML]{FFFFFF}{\textcolor{black}{\tiny\ttfamily\strut  the}}\hskip 0pt plus 1pt\relax\allowbreak{}\colorbox[HTML]{E2E2FF}{\textcolor{black}{\tiny\ttfamily\strut  operating}}\hskip 0pt plus 1pt\relax\allowbreak{}\colorbox[HTML]{FFFFFF}{\textcolor{black}{\tiny\ttfamily\strut  system}}\hskip 0pt plus 1pt\relax\allowbreak{}\colorbox[HTML]{FFFEFE}{\textcolor{black}{\tiny\ttfamily\strut .}}\hskip 0pt plus 1pt\relax\allowbreak{}\colorbox[HTML]{FFF9F9}{\textcolor{black}{\tiny\ttfamily\strut  The}}\hskip 0pt plus 1pt\relax\allowbreak{}\colorbox[HTML]{C1C1FF}{\textcolor{black}{\tiny\ttfamily\strut  antivirus}}\hskip 0pt plus 1pt\relax\allowbreak{}\colorbox[HTML]{FFFFFF}{\textcolor{black}{\tiny\ttfamily\strut  software}}\hskip 0pt plus 1pt\relax\allowbreak{}\colorbox[HTML]{FCFCFF}{\textcolor{black}{\tiny\ttfamily\strut  and}}\hskip 0pt plus 1pt\relax\allowbreak{}\colorbox[HTML]{FAFAFF}{\textcolor{black}{\tiny\ttfamily\strut  software}}\hskip 0pt plus 1pt\relax\allowbreak{}\colorbox[HTML]{FFFFFF}{\textcolor{black}{\tiny\ttfamily\strut  updates}}\hskip 0pt plus 1pt\relax\allowbreak{}\colorbox[HTML]{F9F9FF}{\textcolor{black}{\tiny\ttfamily\strut  ensure}}\hskip 0pt plus 1pt\relax\allowbreak{}\colorbox[HTML]{FBFBFF}{\textcolor{black}{\tiny\ttfamily\strut  the}}\hskip 0pt plus 1pt\relax\allowbreak{}\colorbox[HTML]{E8E8FF}{\textcolor{black}{\tiny\ttfamily\strut  computer}}\hskip 0pt plus 1pt\relax\allowbreak{}\colorbox[HTML]{FEFEFF}{\textcolor{black}{\tiny\ttfamily\strut  remains}}\hskip 0pt plus 1pt\relax\allowbreak{}\colorbox[HTML]{FEFEFF}{\textcolor{black}{\tiny\ttfamily\strut  secure}}\hskip 0pt plus 1pt\relax\allowbreak{}\colorbox[HTML]{FFFFFF}{\textcolor{black}{\tiny\ttfamily\strut  and}}\hskip 0pt plus 1pt\relax\allowbreak{}\colorbox[HTML]{DFDFFF}{\textcolor{black}{\tiny\ttfamily\strut  runs}}\hskip 0pt plus 1pt\relax\allowbreak{}\colorbox[HTML]{D0D0FF}{\textcolor{black}{\tiny\ttfamily\strut  at}}\hskip 0pt plus 1pt\relax\allowbreak{}\colorbox[HTML]{F8F8FF}{\textcolor{black}{\tiny\ttfamily\strut  its}}\hskip 0pt plus 1pt\relax\allowbreak{}\colorbox[HTML]{FDFDFF}{\textcolor{black}{\tiny\ttfamily\strut  best}}\hskip 0pt plus 1pt\relax\allowbreak{}\colorbox[HTML]{FAFAFF}{\textcolor{black}{\tiny\ttfamily\strut  performance}}\hskip 0pt plus 1pt\relax\allowbreak{}\colorbox[HTML]{FFB4B4}{\textcolor{black}{\tiny\ttfamily\strut .}}\hskip 0pt plus 1pt\relax\allowbreak{}\colorbox[HTML]{FFFEFE}{\textcolor{black}{\tiny\ttfamily\strut  The}}\hskip 0pt plus 1pt\relax\allowbreak{}\colorbox[HTML]{C3C3FF}{\textcolor{black}{\tiny\ttfamily\strut  inter}}\hskip 0pt plus 1pt\relax\allowbreak{}\colorbox[HTML]{D4D4FF}{\textcolor{black}{\tiny\ttfamily\strut dependence}}\hskip 0pt plus 1pt\relax\allowbreak{}\colorbox[HTML]{C4C4FF}{\textcolor{black}{\tiny\ttfamily\strut  and}}\hskip 0pt plus 1pt\relax\allowbreak{}\colorbox[HTML]{FFFFFF}{\textcolor{black}{\tiny\ttfamily\strut  inter}}\hskip 0pt plus 1pt\relax\allowbreak{}\colorbox[HTML]{FFFFFF}{\textcolor{black}{\tiny\ttfamily\strut connection}}\hskip 0pt plus 1pt\relax\allowbreak{}\colorbox[HTML]{D5D5FF}{\textcolor{black}{\tiny\ttfamily\strut  between}}\hskip 0pt plus 1pt\relax\allowbreak{}\colorbox[HTML]{FAFAFF}{\textcolor{black}{\tiny\ttfamily\strut  these}}\hskip 0pt plus 1pt\relax\allowbreak{}\colorbox[HTML]{FFFFFF}{\textcolor{black}{\tiny\ttfamily\strut  components}}\hskip 0pt plus 1pt\relax\allowbreak{}\colorbox[HTML]{FFE4E4}{\textcolor{black}{\tiny\ttfamily\strut  can}}\hskip 0pt plus 1pt\relax\allowbreak{}\colorbox[HTML]{FFE6E6}{\textcolor{black}{\tiny\ttfamily\strut  be}}\hskip 0pt plus 1pt\relax\allowbreak{}\colorbox[HTML]{FF4040}{\textcolor{white}{\tiny\ttfamily\strut  metaphor}}\hskip 0pt plus 1pt\relax\allowbreak{}\colorbox[HTML]{FFFFFF}{\textcolor{black}{\tiny\ttfamily\strut ically}}\hskip 0pt plus 1pt\relax\allowbreak{}\colorbox[HTML]{FFBEBE}{\textcolor{black}{\tiny\ttfamily\strut  compared}}\hskip 0pt plus 1pt\relax\allowbreak{}\colorbox[HTML]{FFFFFF}{\textcolor{black}{\tiny\ttfamily\strut  to}}\hskip 0pt plus 1pt\relax\allowbreak{}\colorbox[HTML]{F8F8FF}{\textcolor{black}{\tiny\ttfamily\strut  a}}\hskip 0pt plus 1pt\relax\allowbreak{}\colorbox[HTML]{F6F6FF}{\textcolor{black}{\tiny\ttfamily\strut  complex}}\hskip 0pt plus 1pt\relax\allowbreak{}\colorbox[HTML]{FFF8F8}{\textcolor{black}{\tiny\ttfamily\strut  ecosystem}}\hskip 0pt plus 1pt\relax\allowbreak{}\colorbox[HTML]{FFFBFB}{\textcolor{black}{\tiny\ttfamily\strut ,}}\hskip 0pt plus 1pt\relax\allowbreak{}\colorbox[HTML]{FFFFFF}{\textcolor{black}{\tiny\ttfamily\strut  where}}\hskip 0pt plus 1pt\relax\allowbreak{}\colorbox[HTML]{FFFDFD}{\textcolor{black}{\tiny\ttfamily\strut  each}}\hskip 0pt plus 1pt\relax\allowbreak{}\colorbox[HTML]{FAFAFF}{\textcolor{black}{\tiny\ttfamily\strut  component}}\hskip 0pt plus 1pt\relax\allowbreak{}\colorbox[HTML]{FFF5F5}{\textcolor{black}{\tiny\ttfamily\strut  plays}}\hskip 0pt plus 1pt\relax\allowbreak{}\colorbox[HTML]{FFFFFF}{\textcolor{black}{\tiny\ttfamily\strut  a}}\hskip 0pt plus 1pt\relax\allowbreak{}\colorbox[HTML]{FFECEC}{\textcolor{black}{\tiny\ttfamily\strut  vital}}\hskip 0pt plus 1pt\relax\allowbreak{}\colorbox[HTML]{FFFFFF}{\textcolor{black}{\tiny\ttfamily\strut  role}}\hskip 0pt plus 1pt\relax\allowbreak{}\colorbox[HTML]{FBFBFF}{\textcolor{black}{\tiny\ttfamily\strut  in}}\hskip 0pt plus 1pt\relax\allowbreak{}\colorbox[HTML]{EFEFFF}{\textcolor{black}{\tiny\ttfamily\strut  the}}\hskip 0pt plus 1pt\relax\allowbreak{}\colorbox[HTML]{FEFEFF}{\textcolor{black}{\tiny\ttfamily\strut  overall}}\hskip 0pt plus 1pt\relax\allowbreak{}\colorbox[HTML]{FFFEFE}{\textcolor{black}{\tiny\ttfamily\strut  functionality}}\hskip 0pt plus 1pt\relax\allowbreak{}\colorbox[HTML]{EFEFFF}{\textcolor{black}{\tiny\ttfamily\strut  of}}\hskip 0pt plus 1pt\relax\allowbreak{}\colorbox[HTML]{FFFFFF}{\textcolor{black}{\tiny\ttfamily\strut  the}}\hskip 0pt plus 1pt\relax\allowbreak{}\colorbox[HTML]{FFFBFB}{\textcolor{black}{\tiny\ttfamily\strut  computer}}\hskip 0pt plus 1pt\relax\allowbreak{}\colorbox[HTML]{FFFAFA}{\textcolor{black}{\tiny\ttfamily\strut .}}\hskip 0pt plus 1pt\relax\allowbreak{}\colorbox[HTML]{DFDFFF}{\textcolor{black}{\tiny\ttfamily\strut \textless{}\textbar{}im\_end\textbar{}\textgreater{}}}\hskip 0pt plus 1pt\relax\allowbreak{}
\par\medskip
\noindent\textit{\small Per-constraint attribution snippets (Section A) show that each constraint's $\delta^i_t$ peaks on tokens semantically aligned with the constraint's content. Full-response coloring (Section B) shows that LOO sum and LA produce visually distinct attribution patterns despite both being 1$\times$T signals.}

\endgroup

\bigskip

\subsection{Case 2: LA collapses to structurally meaningless tokens}
\label{app:heatmap:case2}

The second case shows what LA picks up when the constraint set lacks an obvious dominant constraint for a particular response region. Here the constraint set asks for three bullet points, the literal word \emph{barriers} to appear at least three times, and an explicit postscript starting with \emph{P.S.}\ at the end of the response. The per-constraint snippets in Section~A show that the postscript constraint's $\delta_{\text{LOO},i}$ peaks specifically on the tokens \texttt{Post} and \texttt{script} ($\delta = -6.65$ and $+8.96$) at positions $257$--$258$ in the response, the actual postscript onset. In Section~B, LA's strongest red token in the same response is the end-of-message token \texttt{<\textbar{}im\_end\textbar{}>}, a diffuse end-of-sequence marker that does not point to any specific satisfied constraint. LA's signal also lights up the very first bullet marker \texttt{*} at position $0$ and several mid-response generic words, while LOO sum's heaviest tokens, by contrast, sit in the actual P.S.\ region. The two estimators therefore concentrate their signal in nearly disjoint regions of the same response, and the difference cannot be explained by clip threshold or noise. LA does not have access to the per-constraint contrasts that would identify the postscript onset, and its single full-vs-empty contrast spreads support over the response as a whole.

\medskip

\begingroup
\noindent\textbf{\large Sample 6165} \quad $K{=}7$ \quad $T{=}330$ \quad LOO/LA top-20 Jaccard$=0.25$\par\nopagebreak\medskip\nopagebreak
\noindent\textbf{Section A: Per-constraint attribution snippets} ($\textcolor[HTML]{cc0000}{\blacksquare}{=}+\delta$, $\textcolor[HTML]{0000cc}{\blacksquare}{=}-\delta$, intensity $\propto |\delta|$, capped at $\pm8.0$)\par\nopagebreak\smallskip\nopagebreak
\setlength{\fboxsep}{0.6pt}
\setlength{\tabcolsep}{4pt}
\renewcommand{\arraystretch}{1.15}
\begin{tabular}{@{}p{0.10\textwidth}@{\hspace{4pt}}p{0.85\textwidth}@{}}
\toprule
\textbf{C3}\\[1pt]{\scriptsize $|\delta|_{\max}{=}14.9$} & \textit{\scriptsize ``Your answer must contain exactly 3 bullet points. Use the markdown bullet points such as: * This is point 1.''}\\[2pt]
\colorbox[HTML]{8B8BFF}{\textcolor{black}{\scriptsize\ttfamily\strut *}}\colorbox[HTML]{FF4040}{\textcolor{white}{\scriptsize\ttfamily\strut  Communication}}\colorbox[HTML]{FCFCFF}{\textcolor{black}{\scriptsize\ttfamily\strut  barriers}}\colorbox[HTML]{FEFEFF}{\textcolor{black}{\scriptsize\ttfamily\strut  in}}\colorbox[HTML]{FFFFFF}{\textcolor{black}{\scriptsize\ttfamily\strut  a}}\colorbox[HTML]{FEFEFF}{\textcolor{black}{\scriptsize\ttfamily\strut  workplace}}\colorbox[HTML]{FEFEFF}{\textcolor{black}{\scriptsize\ttfamily\strut  can}}\colorbox[HTML]{FFC8C8}{\textcolor{black}{\scriptsize\ttfamily\strut  manifest}}\colorbox[HTML]{F5F5FF}{\textcolor{black}{\scriptsize\ttfamily\strut  in}}{\scriptsize\textcolor{gray}{\dots}}\\[1pt]
{\scriptsize\textcolor{gray}{\dots}}\colorbox[HTML]{FFFFFF}{\textcolor{black}{\scriptsize\ttfamily\strut ings}}\colorbox[HTML]{FFFFFF}{\textcolor{black}{\scriptsize\ttfamily\strut ,}}\colorbox[HTML]{FFFFFF}{\textcolor{black}{\scriptsize\ttfamily\strut  and}}\colorbox[HTML]{FFF0F0}{\textcolor{black}{\scriptsize\ttfamily\strut  technical}}\colorbox[HTML]{FFE2E2}{\textcolor{black}{\scriptsize\ttfamily\strut  j}}\colorbox[HTML]{FFFFFF}{\textcolor{black}{\scriptsize\ttfamily\strut argon}}\colorbox[HTML]{A8A8FF}{\textcolor{black}{\scriptsize\ttfamily\strut .\textcolor{gray}{$\hookleftarrow$}}}\colorbox[HTML]{FF4040}{\textcolor{white}{\scriptsize\ttfamily\strut  }}\colorbox[HTML]{FFFFFF}{\textcolor{black}{\scriptsize\ttfamily\strut  *}}\colorbox[HTML]{FFCFCF}{\textcolor{black}{\scriptsize\ttfamily\strut  Language}}\colorbox[HTML]{EAEAFF}{\textcolor{black}{\scriptsize\ttfamily\strut  differences}}\colorbox[HTML]{E7E7FF}{\textcolor{black}{\scriptsize\ttfamily\strut  can}}\colorbox[HTML]{FFFEFE}{\textcolor{black}{\scriptsize\ttfamily\strut  arise}}\colorbox[HTML]{F3F3FF}{\textcolor{black}{\scriptsize\ttfamily\strut  due}}\colorbox[HTML]{FFFFFF}{\textcolor{black}{\scriptsize\ttfamily\strut  to}}{\scriptsize\textcolor{gray}{\dots}}\\[1pt]\\
\addlinespace[2pt]
\textbf{C6}\\[1pt]{\scriptsize $|\delta|_{\max}{=}11.6$} & \textit{\scriptsize ``In your response, the word *barriers* should appear at least 3 times.''}\\[2pt]
{\scriptsize\textcolor{gray}{\dots}}\colorbox[HTML]{EAEAFF}{\textcolor{black}{\scriptsize\ttfamily\strut .\textcolor{gray}{$\hookleftarrow$}}}\colorbox[HTML]{FFFFFF}{\textcolor{black}{\scriptsize\ttfamily\strut  }}\colorbox[HTML]{FFFFFF}{\textcolor{black}{\scriptsize\ttfamily\strut  *}}\colorbox[HTML]{B3B3FF}{\textcolor{black}{\scriptsize\ttfamily\strut  Cultural}}\colorbox[HTML]{E3E3FF}{\textcolor{black}{\scriptsize\ttfamily\strut  misunderstand}}\colorbox[HTML]{FFFFFF}{\textcolor{black}{\scriptsize\ttfamily\strut ings}}\colorbox[HTML]{E9E9FF}{\textcolor{black}{\scriptsize\ttfamily\strut  can}}\colorbox[HTML]{9292FF}{\textcolor{black}{\scriptsize\ttfamily\strut  occur}}\colorbox[HTML]{FFFFFF}{\textcolor{black}{\scriptsize\ttfamily\strut  when}}\colorbox[HTML]{E8E8FF}{\textcolor{black}{\scriptsize\ttfamily\strut  individuals}}\colorbox[HTML]{FFFFFF}{\textcolor{black}{\scriptsize\ttfamily\strut  from}}\colorbox[HTML]{FEFEFF}{\textcolor{black}{\scriptsize\ttfamily\strut  different}}\colorbox[HTML]{FFFBFB}{\textcolor{black}{\scriptsize\ttfamily\strut  backgrounds}}\colorbox[HTML]{FFEAEA}{\textcolor{black}{\scriptsize\ttfamily\strut  communicate}}\colorbox[HTML]{FAFAFF}{\textcolor{black}{\scriptsize\ttfamily\strut  with}}{\scriptsize\textcolor{gray}{\dots}}\\[1pt]
{\scriptsize\textcolor{gray}{\dots}}\colorbox[HTML]{FDFDFF}{\textcolor{black}{\scriptsize\ttfamily\strut  outcomes}}\colorbox[HTML]{E4E4FF}{\textcolor{black}{\scriptsize\ttfamily\strut  for}}\colorbox[HTML]{FFFFFF}{\textcolor{black}{\scriptsize\ttfamily\strut  their}}\colorbox[HTML]{FBFBFF}{\textcolor{black}{\scriptsize\ttfamily\strut  employees}}\colorbox[HTML]{FFFFFF}{\textcolor{black}{\scriptsize\ttfamily\strut  and}}\colorbox[HTML]{FEFEFF}{\textcolor{black}{\scriptsize\ttfamily\strut  clients}}\colorbox[HTML]{FFF7F7}{\textcolor{black}{\scriptsize\ttfamily\strut .}}\colorbox[HTML]{4040FF}{\textcolor{white}{\scriptsize\ttfamily\strut \textless{}\textbar{}im\_end\textbar{}\textgreater{}}}\\[1pt]\\
\addlinespace[2pt]
\textbf{C7}\\[1pt]{\scriptsize $|\delta|_{\max}{=}9.0$} & \textit{\scriptsize ``At the end of your response, please explicitly add a postscript starting with P.S.''}\\[2pt]
{\scriptsize\textcolor{gray}{\dots}}\colorbox[HTML]{FFFFFF}{\textcolor{black}{\scriptsize\ttfamily\strut  about}}\colorbox[HTML]{FDFDFF}{\textcolor{black}{\scriptsize\ttfamily\strut  different}}\colorbox[HTML]{FFFDFD}{\textcolor{black}{\scriptsize\ttfamily\strut  cultural}}\colorbox[HTML]{FAFAFF}{\textcolor{black}{\scriptsize\ttfamily\strut  perspectives}}\colorbox[HTML]{FFFFFF}{\textcolor{black}{\scriptsize\ttfamily\strut  and}}\colorbox[HTML]{FFFEFE}{\textcolor{black}{\scriptsize\ttfamily\strut  values}}\colorbox[HTML]{FFEDED}{\textcolor{black}{\scriptsize\ttfamily\strut .\textcolor{gray}{$\hookleftarrow$}}}\colorbox[HTML]{6161FF}{\textcolor{white}{\scriptsize\ttfamily\strut  }}\colorbox[HTML]{FFF9F9}{\textcolor{black}{\scriptsize\ttfamily\strut  *}}\colorbox[HTML]{F3F3FF}{\textcolor{black}{\scriptsize\ttfamily\strut  Technical}}\colorbox[HTML]{FFFDFD}{\textcolor{black}{\scriptsize\ttfamily\strut  j}}\colorbox[HTML]{FFFFFF}{\textcolor{black}{\scriptsize\ttfamily\strut argon}}\colorbox[HTML]{FFFCFC}{\textcolor{black}{\scriptsize\ttfamily\strut  can}}\colorbox[HTML]{FAFAFF}{\textcolor{black}{\scriptsize\ttfamily\strut  be}}\colorbox[HTML]{FDFDFF}{\textcolor{black}{\scriptsize\ttfamily\strut  a}}{\scriptsize\textcolor{gray}{\dots}}\\[1pt]
{\scriptsize\textcolor{gray}{\dots}}\colorbox[HTML]{FFFFFF}{\textcolor{black}{\scriptsize\ttfamily\strut ,}}\colorbox[HTML]{FFFFFF}{\textcolor{black}{\scriptsize\ttfamily\strut  and}}\colorbox[HTML]{FDFDFF}{\textcolor{black}{\scriptsize\ttfamily\strut  using}}\colorbox[HTML]{FFF5F5}{\textcolor{black}{\scriptsize\ttfamily\strut  appropriate}}\colorbox[HTML]{FFFEFE}{\textcolor{black}{\scriptsize\ttfamily\strut  gestures}}\colorbox[HTML]{FF9F9F}{\textcolor{black}{\scriptsize\ttfamily\strut .\textcolor{gray}{$\hookleftarrow$}\textcolor{gray}{$\hookleftarrow$}}}\colorbox[HTML]{6060FF}{\textcolor{white}{\scriptsize\ttfamily\strut Post}}\colorbox[HTML]{FF4040}{\textcolor{white}{\scriptsize\ttfamily\strut script}}\colorbox[HTML]{FFDCDC}{\textcolor{black}{\scriptsize\ttfamily\strut :\textcolor{gray}{$\hookleftarrow$}}}\colorbox[HTML]{F3F3FF}{\textcolor{black}{\scriptsize\ttfamily\strut In}}\colorbox[HTML]{FFCFCF}{\textcolor{black}{\scriptsize\ttfamily\strut  conclusion}}\colorbox[HTML]{FFFFFF}{\textcolor{black}{\scriptsize\ttfamily\strut ,}}\colorbox[HTML]{F7F7FF}{\textcolor{black}{\scriptsize\ttfamily\strut  communication}}\colorbox[HTML]{FFFDFD}{\textcolor{black}{\scriptsize\ttfamily\strut  barriers}}\colorbox[HTML]{FFF2F2}{\textcolor{black}{\scriptsize\ttfamily\strut  in}}{\scriptsize\textcolor{gray}{\dots}}\\[1pt]\\
\addlinespace[2pt]
\bottomrule
\end{tabular}
\par\medskip
\noindent\textbf{Section B: Full response} (same tokens shown twice, colored once by LOO sum, once by LA) $-$ vmax$=\pm8.0$\par\nopagebreak\smallskip\nopagebreak
\noindent\textbf{\textcolor[HTML]{c41e3a}{LOO sum}} {\scriptsize $|\delta|_{\max}{=}28.7$}: \\[2pt]
\sloppy
\colorbox[HTML]{4040FF}{\textcolor{white}{\tiny\ttfamily\strut *}}\hskip 0pt plus 1pt\relax\allowbreak{}\colorbox[HTML]{FF4040}{\textcolor{white}{\tiny\ttfamily\strut  Communication}}\hskip 0pt plus 1pt\relax\allowbreak{}\colorbox[HTML]{FFEEEE}{\textcolor{black}{\tiny\ttfamily\strut  barriers}}\hskip 0pt plus 1pt\relax\allowbreak{}\colorbox[HTML]{FFB2B2}{\textcolor{black}{\tiny\ttfamily\strut  in}}\hskip 0pt plus 1pt\relax\allowbreak{}\colorbox[HTML]{FFFCFC}{\textcolor{black}{\tiny\ttfamily\strut  a}}\hskip 0pt plus 1pt\relax\allowbreak{}\colorbox[HTML]{FFF4F4}{\textcolor{black}{\tiny\ttfamily\strut  workplace}}\hskip 0pt plus 1pt\relax\allowbreak{}\colorbox[HTML]{FFF1F1}{\textcolor{black}{\tiny\ttfamily\strut  can}}\hskip 0pt plus 1pt\relax\allowbreak{}\colorbox[HTML]{FFB6B6}{\textcolor{black}{\tiny\ttfamily\strut  manifest}}\hskip 0pt plus 1pt\relax\allowbreak{}\colorbox[HTML]{FFCCCC}{\textcolor{black}{\tiny\ttfamily\strut  in}}\hskip 0pt plus 1pt\relax\allowbreak{}\colorbox[HTML]{FFE9E9}{\textcolor{black}{\tiny\ttfamily\strut  various}}\hskip 0pt plus 1pt\relax\allowbreak{}\colorbox[HTML]{FFF5F5}{\textcolor{black}{\tiny\ttfamily\strut  ways}}\hskip 0pt plus 1pt\relax\allowbreak{}\colorbox[HTML]{FFFBFB}{\textcolor{black}{\tiny\ttfamily\strut ,}}\hskip 0pt plus 1pt\relax\allowbreak{}\colorbox[HTML]{FFEDED}{\textcolor{black}{\tiny\ttfamily\strut  including}}\hskip 0pt plus 1pt\relax\allowbreak{}\colorbox[HTML]{F2F2FF}{\textcolor{black}{\tiny\ttfamily\strut  language}}\hskip 0pt plus 1pt\relax\allowbreak{}\colorbox[HTML]{FFFDFD}{\textcolor{black}{\tiny\ttfamily\strut  differences}}\hskip 0pt plus 1pt\relax\allowbreak{}\colorbox[HTML]{F0F0FF}{\textcolor{black}{\tiny\ttfamily\strut ,}}\hskip 0pt plus 1pt\relax\allowbreak{}\colorbox[HTML]{FFA5A5}{\textcolor{black}{\tiny\ttfamily\strut  cultural}}\hskip 0pt plus 1pt\relax\allowbreak{}\colorbox[HTML]{FFFFFF}{\textcolor{black}{\tiny\ttfamily\strut  misunderstand}}\hskip 0pt plus 1pt\relax\allowbreak{}\colorbox[HTML]{FFFFFF}{\textcolor{black}{\tiny\ttfamily\strut ings}}\hskip 0pt plus 1pt\relax\allowbreak{}\colorbox[HTML]{FFFFFF}{\textcolor{black}{\tiny\ttfamily\strut ,}}\hskip 0pt plus 1pt\relax\allowbreak{}\colorbox[HTML]{FFFEFE}{\textcolor{black}{\tiny\ttfamily\strut  and}}\hskip 0pt plus 1pt\relax\allowbreak{}\colorbox[HTML]{FFFFFF}{\textcolor{black}{\tiny\ttfamily\strut  technical}}\hskip 0pt plus 1pt\relax\allowbreak{}\colorbox[HTML]{FFCACA}{\textcolor{black}{\tiny\ttfamily\strut  j}}\hskip 0pt plus 1pt\relax\allowbreak{}\colorbox[HTML]{FFFFFF}{\textcolor{black}{\tiny\ttfamily\strut argon}}\hskip 0pt plus 1pt\relax\allowbreak{}\colorbox[HTML]{A6A6FF}{\textcolor{black}{\tiny\ttfamily\strut .\textcolor{gray}{$\hookleftarrow$}}}\hskip 0pt plus 1pt\relax\allowbreak{}\colorbox[HTML]{FF4040}{\textcolor{white}{\tiny\ttfamily\strut  }}\hskip 0pt plus 1pt\relax\allowbreak{}\colorbox[HTML]{FFC5C5}{\textcolor{black}{\tiny\ttfamily\strut  *}}\hskip 0pt plus 1pt\relax\allowbreak{}\colorbox[HTML]{FFA6A6}{\textcolor{black}{\tiny\ttfamily\strut  Language}}\hskip 0pt plus 1pt\relax\allowbreak{}\colorbox[HTML]{FFFEFE}{\textcolor{black}{\tiny\ttfamily\strut  differences}}\hskip 0pt plus 1pt\relax\allowbreak{}\colorbox[HTML]{A9A9FF}{\textcolor{black}{\tiny\ttfamily\strut  can}}\hskip 0pt plus 1pt\relax\allowbreak{}\colorbox[HTML]{FFF7F7}{\textcolor{black}{\tiny\ttfamily\strut  arise}}\hskip 0pt plus 1pt\relax\allowbreak{}\colorbox[HTML]{BCBCFF}{\textcolor{black}{\tiny\ttfamily\strut  due}}\hskip 0pt plus 1pt\relax\allowbreak{}\colorbox[HTML]{FFFFFF}{\textcolor{black}{\tiny\ttfamily\strut  to}}\hskip 0pt plus 1pt\relax\allowbreak{}\colorbox[HTML]{FDFDFF}{\textcolor{black}{\tiny\ttfamily\strut  the}}\hskip 0pt plus 1pt\relax\allowbreak{}\colorbox[HTML]{F6F6FF}{\textcolor{black}{\tiny\ttfamily\strut  use}}\hskip 0pt plus 1pt\relax\allowbreak{}\colorbox[HTML]{FFFFFF}{\textcolor{black}{\tiny\ttfamily\strut  of}}\hskip 0pt plus 1pt\relax\allowbreak{}\colorbox[HTML]{FAFAFF}{\textcolor{black}{\tiny\ttfamily\strut  different}}\hskip 0pt plus 1pt\relax\allowbreak{}\colorbox[HTML]{F7F7FF}{\textcolor{black}{\tiny\ttfamily\strut  languages}}\hskip 0pt plus 1pt\relax\allowbreak{}\colorbox[HTML]{FEFEFF}{\textcolor{black}{\tiny\ttfamily\strut  or}}\hskip 0pt plus 1pt\relax\allowbreak{}\colorbox[HTML]{FFF7F7}{\textcolor{black}{\tiny\ttfamily\strut  dialect}}\hskip 0pt plus 1pt\relax\allowbreak{}\colorbox[HTML]{FFFFFF}{\textcolor{black}{\tiny\ttfamily\strut s}}\hskip 0pt plus 1pt\relax\allowbreak{}\colorbox[HTML]{FFECEC}{\textcolor{black}{\tiny\ttfamily\strut ,}}\hskip 0pt plus 1pt\relax\allowbreak{}\colorbox[HTML]{FFF2F2}{\textcolor{black}{\tiny\ttfamily\strut  which}}\hskip 0pt plus 1pt\relax\allowbreak{}\colorbox[HTML]{FFFAFA}{\textcolor{black}{\tiny\ttfamily\strut  can}}\hskip 0pt plus 1pt\relax\allowbreak{}\colorbox[HTML]{E4E4FF}{\textcolor{black}{\tiny\ttfamily\strut  lead}}\hskip 0pt plus 1pt\relax\allowbreak{}\colorbox[HTML]{FFFFFF}{\textcolor{black}{\tiny\ttfamily\strut  to}}\hskip 0pt plus 1pt\relax\allowbreak{}\colorbox[HTML]{CACAFF}{\textcolor{black}{\tiny\ttfamily\strut  mis}}\hskip 0pt plus 1pt\relax\allowbreak{}\colorbox[HTML]{FAFAFF}{\textcolor{black}{\tiny\ttfamily\strut interpret}}\hskip 0pt plus 1pt\relax\allowbreak{}\colorbox[HTML]{FFFCFC}{\textcolor{black}{\tiny\ttfamily\strut ation}}\hskip 0pt plus 1pt\relax\allowbreak{}\colorbox[HTML]{CFCFFF}{\textcolor{black}{\tiny\ttfamily\strut  of}}\hskip 0pt plus 1pt\relax\allowbreak{}\colorbox[HTML]{FFFEFE}{\textcolor{black}{\tiny\ttfamily\strut  messages}}\hskip 0pt plus 1pt\relax\allowbreak{}\colorbox[HTML]{FAFAFF}{\textcolor{black}{\tiny\ttfamily\strut .}}\hskip 0pt plus 1pt\relax\allowbreak{}\colorbox[HTML]{FF8C8C}{\textcolor{black}{\tiny\ttfamily\strut  To}}\hskip 0pt plus 1pt\relax\allowbreak{}\colorbox[HTML]{FFE2E2}{\textcolor{black}{\tiny\ttfamily\strut  overcome}}\hskip 0pt plus 1pt\relax\allowbreak{}\colorbox[HTML]{F8F8FF}{\textcolor{black}{\tiny\ttfamily\strut  this}}\hskip 0pt plus 1pt\relax\allowbreak{}\colorbox[HTML]{FF8383}{\textcolor{black}{\tiny\ttfamily\strut  barrier}}\hskip 0pt plus 1pt\relax\allowbreak{}\colorbox[HTML]{FFFEFE}{\textcolor{black}{\tiny\ttfamily\strut ,}}\hskip 0pt plus 1pt\relax\allowbreak{}\colorbox[HTML]{FFDDDD}{\textcolor{black}{\tiny\ttfamily\strut  organizations}}\hskip 0pt plus 1pt\relax\allowbreak{}\colorbox[HTML]{FDFDFF}{\textcolor{black}{\tiny\ttfamily\strut  can}}\hskip 0pt plus 1pt\relax\allowbreak{}\colorbox[HTML]{FFDEDE}{\textcolor{black}{\tiny\ttfamily\strut  implement}}\hskip 0pt plus 1pt\relax\allowbreak{}\colorbox[HTML]{FFE1E1}{\textcolor{black}{\tiny\ttfamily\strut  mult}}\hskip 0pt plus 1pt\relax\allowbreak{}\colorbox[HTML]{FFFFFF}{\textcolor{black}{\tiny\ttfamily\strut ilingual}}\hskip 0pt plus 1pt\relax\allowbreak{}\colorbox[HTML]{EFEFFF}{\textcolor{black}{\tiny\ttfamily\strut  communication}}\hskip 0pt plus 1pt\relax\allowbreak{}\colorbox[HTML]{FCFCFF}{\textcolor{black}{\tiny\ttfamily\strut  tools}}\hskip 0pt plus 1pt\relax\allowbreak{}\colorbox[HTML]{FFF4F4}{\textcolor{black}{\tiny\ttfamily\strut  or}}\hskip 0pt plus 1pt\relax\allowbreak{}\colorbox[HTML]{FBFBFF}{\textcolor{black}{\tiny\ttfamily\strut  provide}}\hskip 0pt plus 1pt\relax\allowbreak{}\colorbox[HTML]{FFF6F6}{\textcolor{black}{\tiny\ttfamily\strut  translation}}\hskip 0pt plus 1pt\relax\allowbreak{}\colorbox[HTML]{FFFFFF}{\textcolor{black}{\tiny\ttfamily\strut  services}}\hskip 0pt plus 1pt\relax\allowbreak{}\colorbox[HTML]{FFEFEF}{\textcolor{black}{\tiny\ttfamily\strut  for}}\hskip 0pt plus 1pt\relax\allowbreak{}\colorbox[HTML]{FFF9F9}{\textcolor{black}{\tiny\ttfamily\strut  employees}}\hskip 0pt plus 1pt\relax\allowbreak{}\colorbox[HTML]{FFFCFC}{\textcolor{black}{\tiny\ttfamily\strut  who}}\hskip 0pt plus 1pt\relax\allowbreak{}\colorbox[HTML]{FBFBFF}{\textcolor{black}{\tiny\ttfamily\strut  speak}}\hskip 0pt plus 1pt\relax\allowbreak{}\colorbox[HTML]{FFFFFF}{\textcolor{black}{\tiny\ttfamily\strut  different}}\hskip 0pt plus 1pt\relax\allowbreak{}\colorbox[HTML]{FFFFFF}{\textcolor{black}{\tiny\ttfamily\strut  languages}}\hskip 0pt plus 1pt\relax\allowbreak{}\colorbox[HTML]{C9C9FF}{\textcolor{black}{\tiny\ttfamily\strut .\textcolor{gray}{$\hookleftarrow$}}}\hskip 0pt plus 1pt\relax\allowbreak{}\colorbox[HTML]{FFFFFF}{\textcolor{black}{\tiny\ttfamily\strut  }}\hskip 0pt plus 1pt\relax\allowbreak{}\colorbox[HTML]{FFC9C9}{\textcolor{black}{\tiny\ttfamily\strut  *}}\hskip 0pt plus 1pt\relax\allowbreak{}\colorbox[HTML]{FFE0E0}{\textcolor{black}{\tiny\ttfamily\strut  Cultural}}\hskip 0pt plus 1pt\relax\allowbreak{}\colorbox[HTML]{F4F4FF}{\textcolor{black}{\tiny\ttfamily\strut  misunderstand}}\hskip 0pt plus 1pt\relax\allowbreak{}\colorbox[HTML]{FFFFFF}{\textcolor{black}{\tiny\ttfamily\strut ings}}\hskip 0pt plus 1pt\relax\allowbreak{}\colorbox[HTML]{D5D5FF}{\textcolor{black}{\tiny\ttfamily\strut  can}}\hskip 0pt plus 1pt\relax\allowbreak{}\colorbox[HTML]{5454FF}{\textcolor{white}{\tiny\ttfamily\strut  occur}}\hskip 0pt plus 1pt\relax\allowbreak{}\colorbox[HTML]{FFFDFD}{\textcolor{black}{\tiny\ttfamily\strut  when}}\hskip 0pt plus 1pt\relax\allowbreak{}\colorbox[HTML]{FFF2F2}{\textcolor{black}{\tiny\ttfamily\strut  individuals}}\hskip 0pt plus 1pt\relax\allowbreak{}\colorbox[HTML]{FFFFFF}{\textcolor{black}{\tiny\ttfamily\strut  from}}\hskip 0pt plus 1pt\relax\allowbreak{}\colorbox[HTML]{F6F6FF}{\textcolor{black}{\tiny\ttfamily\strut  different}}\hskip 0pt plus 1pt\relax\allowbreak{}\colorbox[HTML]{E4E4FF}{\textcolor{black}{\tiny\ttfamily\strut  backgrounds}}\hskip 0pt plus 1pt\relax\allowbreak{}\colorbox[HTML]{FFDFDF}{\textcolor{black}{\tiny\ttfamily\strut  communicate}}\hskip 0pt plus 1pt\relax\allowbreak{}\colorbox[HTML]{FFFBFB}{\textcolor{black}{\tiny\ttfamily\strut  with}}\hskip 0pt plus 1pt\relax\allowbreak{}\colorbox[HTML]{FFFCFC}{\textcolor{black}{\tiny\ttfamily\strut  each}}\hskip 0pt plus 1pt\relax\allowbreak{}\colorbox[HTML]{FFFFFF}{\textcolor{black}{\tiny\ttfamily\strut  other}}\hskip 0pt plus 1pt\relax\allowbreak{}\colorbox[HTML]{EDEDFF}{\textcolor{black}{\tiny\ttfamily\strut .}}\hskip 0pt plus 1pt\relax\allowbreak{}\colorbox[HTML]{FFD3D3}{\textcolor{black}{\tiny\ttfamily\strut  To}}\hskip 0pt plus 1pt\relax\allowbreak{}\colorbox[HTML]{FFFCFC}{\textcolor{black}{\tiny\ttfamily\strut  address}}\hskip 0pt plus 1pt\relax\allowbreak{}\colorbox[HTML]{E3E3FF}{\textcolor{black}{\tiny\ttfamily\strut  this}}\hskip 0pt plus 1pt\relax\allowbreak{}\colorbox[HTML]{FFEAEA}{\textcolor{black}{\tiny\ttfamily\strut ,}}\hskip 0pt plus 1pt\relax\allowbreak{}\colorbox[HTML]{FFD4D4}{\textcolor{black}{\tiny\ttfamily\strut  it}}\hskip 0pt plus 1pt\relax\allowbreak{}\colorbox[HTML]{FDFDFF}{\textcolor{black}{\tiny\ttfamily\strut  is}}\hskip 0pt plus 1pt\relax\allowbreak{}\colorbox[HTML]{F8F8FF}{\textcolor{black}{\tiny\ttfamily\strut  essential}}\hskip 0pt plus 1pt\relax\allowbreak{}\colorbox[HTML]{FCFCFF}{\textcolor{black}{\tiny\ttfamily\strut  to}}\hskip 0pt plus 1pt\relax\allowbreak{}\colorbox[HTML]{FFFFFF}{\textcolor{black}{\tiny\ttfamily\strut  foster}}\hskip 0pt plus 1pt\relax\allowbreak{}\colorbox[HTML]{FFFDFD}{\textcolor{black}{\tiny\ttfamily\strut  an}}\hskip 0pt plus 1pt\relax\allowbreak{}\colorbox[HTML]{FDFDFF}{\textcolor{black}{\tiny\ttfamily\strut  inclusive}}\hskip 0pt plus 1pt\relax\allowbreak{}\colorbox[HTML]{FFEFEF}{\textcolor{black}{\tiny\ttfamily\strut  and}}\hskip 0pt plus 1pt\relax\allowbreak{}\colorbox[HTML]{EBEBFF}{\textcolor{black}{\tiny\ttfamily\strut  respectful}}\hskip 0pt plus 1pt\relax\allowbreak{}\colorbox[HTML]{FBFBFF}{\textcolor{black}{\tiny\ttfamily\strut  workplace}}\hskip 0pt plus 1pt\relax\allowbreak{}\colorbox[HTML]{FAFAFF}{\textcolor{black}{\tiny\ttfamily\strut  culture}}\hskip 0pt plus 1pt\relax\allowbreak{}\colorbox[HTML]{F7F7FF}{\textcolor{black}{\tiny\ttfamily\strut  that}}\hskip 0pt plus 1pt\relax\allowbreak{}\colorbox[HTML]{FFE6E6}{\textcolor{black}{\tiny\ttfamily\strut  values}}\hskip 0pt plus 1pt\relax\allowbreak{}\colorbox[HTML]{FEFEFF}{\textcolor{black}{\tiny\ttfamily\strut  diversity}}\hskip 0pt plus 1pt\relax\allowbreak{}\colorbox[HTML]{D8D8FF}{\textcolor{black}{\tiny\ttfamily\strut  and}}\hskip 0pt plus 1pt\relax\allowbreak{}\colorbox[HTML]{FFFAFA}{\textcolor{black}{\tiny\ttfamily\strut  promotes}}\hskip 0pt plus 1pt\relax\allowbreak{}\colorbox[HTML]{F9F9FF}{\textcolor{black}{\tiny\ttfamily\strut  cross}}\hskip 0pt plus 1pt\relax\allowbreak{}\colorbox[HTML]{FFFFFF}{\textcolor{black}{\tiny\ttfamily\strut -cultural}}\hskip 0pt plus 1pt\relax\allowbreak{}\colorbox[HTML]{FDFDFF}{\textcolor{black}{\tiny\ttfamily\strut  understanding}}\hskip 0pt plus 1pt\relax\allowbreak{}\colorbox[HTML]{FFFAFA}{\textcolor{black}{\tiny\ttfamily\strut .}}\hskip 0pt plus 1pt\relax\allowbreak{}\colorbox[HTML]{FF9898}{\textcolor{black}{\tiny\ttfamily\strut  This}}\hskip 0pt plus 1pt\relax\allowbreak{}\colorbox[HTML]{FFFEFE}{\textcolor{black}{\tiny\ttfamily\strut  can}}\hskip 0pt plus 1pt\relax\allowbreak{}\colorbox[HTML]{FFFEFE}{\textcolor{black}{\tiny\ttfamily\strut  be}}\hskip 0pt plus 1pt\relax\allowbreak{}\colorbox[HTML]{FFFDFD}{\textcolor{black}{\tiny\ttfamily\strut  achieved}}\hskip 0pt plus 1pt\relax\allowbreak{}\colorbox[HTML]{F8F8FF}{\textcolor{black}{\tiny\ttfamily\strut  through}}\hskip 0pt plus 1pt\relax\allowbreak{}\colorbox[HTML]{FFEFEF}{\textcolor{black}{\tiny\ttfamily\strut  training}}\hskip 0pt plus 1pt\relax\allowbreak{}\colorbox[HTML]{FFF5F5}{\textcolor{black}{\tiny\ttfamily\strut  programs}}\hskip 0pt plus 1pt\relax\allowbreak{}\colorbox[HTML]{FFFAFA}{\textcolor{black}{\tiny\ttfamily\strut  that}}\hskip 0pt plus 1pt\relax\allowbreak{}\colorbox[HTML]{FCFCFF}{\textcolor{black}{\tiny\ttfamily\strut  teach}}\hskip 0pt plus 1pt\relax\allowbreak{}\colorbox[HTML]{FFFCFC}{\textcolor{black}{\tiny\ttfamily\strut  employees}}\hskip 0pt plus 1pt\relax\allowbreak{}\colorbox[HTML]{FFFFFF}{\textcolor{black}{\tiny\ttfamily\strut  about}}\hskip 0pt plus 1pt\relax\allowbreak{}\colorbox[HTML]{F0F0FF}{\textcolor{black}{\tiny\ttfamily\strut  different}}\hskip 0pt plus 1pt\relax\allowbreak{}\colorbox[HTML]{FFFCFC}{\textcolor{black}{\tiny\ttfamily\strut  cultural}}\hskip 0pt plus 1pt\relax\allowbreak{}\colorbox[HTML]{E0E0FF}{\textcolor{black}{\tiny\ttfamily\strut  perspectives}}\hskip 0pt plus 1pt\relax\allowbreak{}\colorbox[HTML]{FFFEFE}{\textcolor{black}{\tiny\ttfamily\strut  and}}\hskip 0pt plus 1pt\relax\allowbreak{}\colorbox[HTML]{F0F0FF}{\textcolor{black}{\tiny\ttfamily\strut  values}}\hskip 0pt plus 1pt\relax\allowbreak{}\colorbox[HTML]{BCBCFF}{\textcolor{black}{\tiny\ttfamily\strut .\textcolor{gray}{$\hookleftarrow$}}}\hskip 0pt plus 1pt\relax\allowbreak{}\colorbox[HTML]{4040FF}{\textcolor{white}{\tiny\ttfamily\strut  }}\hskip 0pt plus 1pt\relax\allowbreak{}\colorbox[HTML]{A5A5FF}{\textcolor{black}{\tiny\ttfamily\strut  *}}\hskip 0pt plus 1pt\relax\allowbreak{}\colorbox[HTML]{FFF4F4}{\textcolor{black}{\tiny\ttfamily\strut  Technical}}\hskip 0pt plus 1pt\relax\allowbreak{}\colorbox[HTML]{FFE2E2}{\textcolor{black}{\tiny\ttfamily\strut  j}}\hskip 0pt plus 1pt\relax\allowbreak{}\colorbox[HTML]{FFFFFF}{\textcolor{black}{\tiny\ttfamily\strut argon}}\hskip 0pt plus 1pt\relax\allowbreak{}\colorbox[HTML]{FFFAFA}{\textcolor{black}{\tiny\ttfamily\strut  can}}\hskip 0pt plus 1pt\relax\allowbreak{}\colorbox[HTML]{FFFFFF}{\textcolor{black}{\tiny\ttfamily\strut  be}}\hskip 0pt plus 1pt\relax\allowbreak{}\colorbox[HTML]{FFF8F8}{\textcolor{black}{\tiny\ttfamily\strut  a}}\hskip 0pt plus 1pt\relax\allowbreak{}\colorbox[HTML]{FFF8F8}{\textcolor{black}{\tiny\ttfamily\strut  significant}}\hskip 0pt plus 1pt\relax\allowbreak{}\colorbox[HTML]{FFEFEF}{\textcolor{black}{\tiny\ttfamily\strut  barrier}}\hskip 0pt plus 1pt\relax\allowbreak{}\colorbox[HTML]{FFEDED}{\textcolor{black}{\tiny\ttfamily\strut  in}}\hskip 0pt plus 1pt\relax\allowbreak{}\colorbox[HTML]{FFF4F4}{\textcolor{black}{\tiny\ttfamily\strut  professional}}\hskip 0pt plus 1pt\relax\allowbreak{}\colorbox[HTML]{EFEFFF}{\textcolor{black}{\tiny\ttfamily\strut  communication}}\hskip 0pt plus 1pt\relax\allowbreak{}\colorbox[HTML]{FFFBFB}{\textcolor{black}{\tiny\ttfamily\strut ,}}\hskip 0pt plus 1pt\relax\allowbreak{}\colorbox[HTML]{F8F8FF}{\textcolor{black}{\tiny\ttfamily\strut  especially}}\hskip 0pt plus 1pt\relax\allowbreak{}\colorbox[HTML]{F7F7FF}{\textcolor{black}{\tiny\ttfamily\strut  when}}\hskip 0pt plus 1pt\relax\allowbreak{}\colorbox[HTML]{FEFEFF}{\textcolor{black}{\tiny\ttfamily\strut  working}}\hskip 0pt plus 1pt\relax\allowbreak{}\colorbox[HTML]{FFF0F0}{\textcolor{black}{\tiny\ttfamily\strut  in}}\hskip 0pt plus 1pt\relax\allowbreak{}\colorbox[HTML]{FFF2F2}{\textcolor{black}{\tiny\ttfamily\strut  a}}\hskip 0pt plus 1pt\relax\allowbreak{}\colorbox[HTML]{FFF0F0}{\textcolor{black}{\tiny\ttfamily\strut  team}}\hskip 0pt plus 1pt\relax\allowbreak{}\colorbox[HTML]{FFFFFF}{\textcolor{black}{\tiny\ttfamily\strut  with}}\hskip 0pt plus 1pt\relax\allowbreak{}\colorbox[HTML]{F6F6FF}{\textcolor{black}{\tiny\ttfamily\strut  members}}\hskip 0pt plus 1pt\relax\allowbreak{}\colorbox[HTML]{FFFFFF}{\textcolor{black}{\tiny\ttfamily\strut  from}}\hskip 0pt plus 1pt\relax\allowbreak{}\colorbox[HTML]{FEFEFF}{\textcolor{black}{\tiny\ttfamily\strut  diverse}}\hskip 0pt plus 1pt\relax\allowbreak{}\colorbox[HTML]{FFF4F4}{\textcolor{black}{\tiny\ttfamily\strut  backgrounds}}\hskip 0pt plus 1pt\relax\allowbreak{}\colorbox[HTML]{F5F5FF}{\textcolor{black}{\tiny\ttfamily\strut  or}}\hskip 0pt plus 1pt\relax\allowbreak{}\colorbox[HTML]{FFE9E9}{\textcolor{black}{\tiny\ttfamily\strut  in}}\hskip 0pt plus 1pt\relax\allowbreak{}\colorbox[HTML]{FFE9E9}{\textcolor{black}{\tiny\ttfamily\strut  an}}\hskip 0pt plus 1pt\relax\allowbreak{}\colorbox[HTML]{FFFAFA}{\textcolor{black}{\tiny\ttfamily\strut  industry}}\hskip 0pt plus 1pt\relax\allowbreak{}\colorbox[HTML]{FFFDFD}{\textcolor{black}{\tiny\ttfamily\strut  with}}\hskip 0pt plus 1pt\relax\allowbreak{}\colorbox[HTML]{FFFBFB}{\textcolor{black}{\tiny\ttfamily\strut  specialized}}\hskip 0pt plus 1pt\relax\allowbreak{}\colorbox[HTML]{FFFDFD}{\textcolor{black}{\tiny\ttfamily\strut  terminology}}\hskip 0pt plus 1pt\relax\allowbreak{}\colorbox[HTML]{FEFEFF}{\textcolor{black}{\tiny\ttfamily\strut .}}\hskip 0pt plus 1pt\relax\allowbreak{}\colorbox[HTML]{FFBFBF}{\textcolor{black}{\tiny\ttfamily\strut  To}}\hskip 0pt plus 1pt\relax\allowbreak{}\colorbox[HTML]{FFFBFB}{\textcolor{black}{\tiny\ttfamily\strut  overcome}}\hskip 0pt plus 1pt\relax\allowbreak{}\colorbox[HTML]{FFFEFE}{\textcolor{black}{\tiny\ttfamily\strut  this}}\hskip 0pt plus 1pt\relax\allowbreak{}\colorbox[HTML]{FFC8C8}{\textcolor{black}{\tiny\ttfamily\strut  barrier}}\hskip 0pt plus 1pt\relax\allowbreak{}\colorbox[HTML]{FFFFFF}{\textcolor{black}{\tiny\ttfamily\strut ,}}\hskip 0pt plus 1pt\relax\allowbreak{}\colorbox[HTML]{FFBFBF}{\textcolor{black}{\tiny\ttfamily\strut  organizations}}\hskip 0pt plus 1pt\relax\allowbreak{}\colorbox[HTML]{FEFEFF}{\textcolor{black}{\tiny\ttfamily\strut  should}}\hskip 0pt plus 1pt\relax\allowbreak{}\colorbox[HTML]{FFF0F0}{\textcolor{black}{\tiny\ttfamily\strut  establish}}\hskip 0pt plus 1pt\relax\allowbreak{}\colorbox[HTML]{FBFBFF}{\textcolor{black}{\tiny\ttfamily\strut  clear}}\hskip 0pt plus 1pt\relax\allowbreak{}\colorbox[HTML]{FFF3F3}{\textcolor{black}{\tiny\ttfamily\strut  communication}}\hskip 0pt plus 1pt\relax\allowbreak{}\colorbox[HTML]{F8F8FF}{\textcolor{black}{\tiny\ttfamily\strut  guidelines}}\hskip 0pt plus 1pt\relax\allowbreak{}\colorbox[HTML]{FCFCFF}{\textcolor{black}{\tiny\ttfamily\strut  that}}\hskip 0pt plus 1pt\relax\allowbreak{}\colorbox[HTML]{FFEDED}{\textcolor{black}{\tiny\ttfamily\strut  define}}\hskip 0pt plus 1pt\relax\allowbreak{}\colorbox[HTML]{FFFCFC}{\textcolor{black}{\tiny\ttfamily\strut  technical}}\hskip 0pt plus 1pt\relax\allowbreak{}\colorbox[HTML]{FFFFFF}{\textcolor{black}{\tiny\ttfamily\strut  terms}}\hskip 0pt plus 1pt\relax\allowbreak{}\colorbox[HTML]{FFFFFF}{\textcolor{black}{\tiny\ttfamily\strut  and}}\hskip 0pt plus 1pt\relax\allowbreak{}\colorbox[HTML]{FFF7F7}{\textcolor{black}{\tiny\ttfamily\strut  ensure}}\hskip 0pt plus 1pt\relax\allowbreak{}\colorbox[HTML]{FAFAFF}{\textcolor{black}{\tiny\ttfamily\strut  that}}\hskip 0pt plus 1pt\relax\allowbreak{}\colorbox[HTML]{FFFCFC}{\textcolor{black}{\tiny\ttfamily\strut  all}}\hskip 0pt plus 1pt\relax\allowbreak{}\colorbox[HTML]{FEFEFF}{\textcolor{black}{\tiny\ttfamily\strut  employees}}\hskip 0pt plus 1pt\relax\allowbreak{}\colorbox[HTML]{FEFEFF}{\textcolor{black}{\tiny\ttfamily\strut  are}}\hskip 0pt plus 1pt\relax\allowbreak{}\colorbox[HTML]{FFFFFF}{\textcolor{black}{\tiny\ttfamily\strut  familiar}}\hskip 0pt plus 1pt\relax\allowbreak{}\colorbox[HTML]{FFFFFF}{\textcolor{black}{\tiny\ttfamily\strut  with}}\hskip 0pt plus 1pt\relax\allowbreak{}\colorbox[HTML]{FBFBFF}{\textcolor{black}{\tiny\ttfamily\strut  them}}\hskip 0pt plus 1pt\relax\allowbreak{}\colorbox[HTML]{FFC9C9}{\textcolor{black}{\tiny\ttfamily\strut .\textcolor{gray}{$\hookleftarrow$}}}\hskip 0pt plus 1pt\relax\allowbreak{}\colorbox[HTML]{FF4F4F}{\textcolor{white}{\tiny\ttfamily\strut  }}\hskip 0pt plus 1pt\relax\allowbreak{}\colorbox[HTML]{FF4242}{\textcolor{white}{\tiny\ttfamily\strut  *}}\hskip 0pt plus 1pt\relax\allowbreak{}\colorbox[HTML]{FFE6E6}{\textcolor{black}{\tiny\ttfamily\strut  Additionally}}\hskip 0pt plus 1pt\relax\allowbreak{}\colorbox[HTML]{FFFFFF}{\textcolor{black}{\tiny\ttfamily\strut ,}}\hskip 0pt plus 1pt\relax\allowbreak{}\colorbox[HTML]{FFD5D5}{\textcolor{black}{\tiny\ttfamily\strut  non}}\hskip 0pt plus 1pt\relax\allowbreak{}\colorbox[HTML]{FFFFFF}{\textcolor{black}{\tiny\ttfamily\strut -ver}}\hskip 0pt plus 1pt\relax\allowbreak{}\colorbox[HTML]{FFFFFF}{\textcolor{black}{\tiny\ttfamily\strut bal}}\hskip 0pt plus 1pt\relax\allowbreak{}\colorbox[HTML]{FFF6F6}{\textcolor{black}{\tiny\ttfamily\strut  communication}}\hskip 0pt plus 1pt\relax\allowbreak{}\colorbox[HTML]{FFD7D7}{\textcolor{black}{\tiny\ttfamily\strut  can}}\hskip 0pt plus 1pt\relax\allowbreak{}\colorbox[HTML]{F9F9FF}{\textcolor{black}{\tiny\ttfamily\strut  also}}\hskip 0pt plus 1pt\relax\allowbreak{}\colorbox[HTML]{FFF5F5}{\textcolor{black}{\tiny\ttfamily\strut  be}}\hskip 0pt plus 1pt\relax\allowbreak{}\colorbox[HTML]{FFFFFF}{\textcolor{black}{\tiny\ttfamily\strut  a}}\hskip 0pt plus 1pt\relax\allowbreak{}\colorbox[HTML]{FFC5C5}{\textcolor{black}{\tiny\ttfamily\strut  barrier}}\hskip 0pt plus 1pt\relax\allowbreak{}\colorbox[HTML]{FFF2F2}{\textcolor{black}{\tiny\ttfamily\strut  in}}\hskip 0pt plus 1pt\relax\allowbreak{}\colorbox[HTML]{E8E8FF}{\textcolor{black}{\tiny\ttfamily\strut  the}}\hskip 0pt plus 1pt\relax\allowbreak{}\colorbox[HTML]{FFE7E7}{\textcolor{black}{\tiny\ttfamily\strut  workplace}}\hskip 0pt plus 1pt\relax\allowbreak{}\colorbox[HTML]{FFF7F7}{\textcolor{black}{\tiny\ttfamily\strut .}}\hskip 0pt plus 1pt\relax\allowbreak{}\colorbox[HTML]{FFCACA}{\textcolor{black}{\tiny\ttfamily\strut  This}}\hskip 0pt plus 1pt\relax\allowbreak{}\colorbox[HTML]{FFF1F1}{\textcolor{black}{\tiny\ttfamily\strut  can}}\hskip 0pt plus 1pt\relax\allowbreak{}\colorbox[HTML]{FFE2E2}{\textcolor{black}{\tiny\ttfamily\strut  include}}\hskip 0pt plus 1pt\relax\allowbreak{}\colorbox[HTML]{FAFAFF}{\textcolor{black}{\tiny\ttfamily\strut  poor}}\hskip 0pt plus 1pt\relax\allowbreak{}\colorbox[HTML]{FFF7F7}{\textcolor{black}{\tiny\ttfamily\strut  body}}\hskip 0pt plus 1pt\relax\allowbreak{}\colorbox[HTML]{FFFFFF}{\textcolor{black}{\tiny\ttfamily\strut  language}}\hskip 0pt plus 1pt\relax\allowbreak{}\colorbox[HTML]{FFFCFC}{\textcolor{black}{\tiny\ttfamily\strut ,}}\hskip 0pt plus 1pt\relax\allowbreak{}\colorbox[HTML]{FFF2F2}{\textcolor{black}{\tiny\ttfamily\strut  lack}}\hskip 0pt plus 1pt\relax\allowbreak{}\colorbox[HTML]{FFFFFF}{\textcolor{black}{\tiny\ttfamily\strut  of}}\hskip 0pt plus 1pt\relax\allowbreak{}\colorbox[HTML]{FFFFFF}{\textcolor{black}{\tiny\ttfamily\strut  eye}}\hskip 0pt plus 1pt\relax\allowbreak{}\colorbox[HTML]{FFFFFF}{\textcolor{black}{\tiny\ttfamily\strut  contact}}\hskip 0pt plus 1pt\relax\allowbreak{}\colorbox[HTML]{FFFFFF}{\textcolor{black}{\tiny\ttfamily\strut ,}}\hskip 0pt plus 1pt\relax\allowbreak{}\colorbox[HTML]{FFFFFF}{\textcolor{black}{\tiny\ttfamily\strut  or}}\hskip 0pt plus 1pt\relax\allowbreak{}\colorbox[HTML]{FEFEFF}{\textcolor{black}{\tiny\ttfamily\strut  inappropriate}}\hskip 0pt plus 1pt\relax\allowbreak{}\colorbox[HTML]{FBFBFF}{\textcolor{black}{\tiny\ttfamily\strut  gestures}}\hskip 0pt plus 1pt\relax\allowbreak{}\colorbox[HTML]{FFF9F9}{\textcolor{black}{\tiny\ttfamily\strut .}}\hskip 0pt plus 1pt\relax\allowbreak{}\colorbox[HTML]{FFD5D5}{\textcolor{black}{\tiny\ttfamily\strut  To}}\hskip 0pt plus 1pt\relax\allowbreak{}\colorbox[HTML]{FFD3D3}{\textcolor{black}{\tiny\ttfamily\strut  overcome}}\hskip 0pt plus 1pt\relax\allowbreak{}\colorbox[HTML]{F0F0FF}{\textcolor{black}{\tiny\ttfamily\strut  this}}\hskip 0pt plus 1pt\relax\allowbreak{}\colorbox[HTML]{FFC9C9}{\textcolor{black}{\tiny\ttfamily\strut  barrier}}\hskip 0pt plus 1pt\relax\allowbreak{}\colorbox[HTML]{FFFFFF}{\textcolor{black}{\tiny\ttfamily\strut ,}}\hskip 0pt plus 1pt\relax\allowbreak{}\colorbox[HTML]{FFF4F4}{\textcolor{black}{\tiny\ttfamily\strut  employees}}\hskip 0pt plus 1pt\relax\allowbreak{}\colorbox[HTML]{FFFBFB}{\textcolor{black}{\tiny\ttfamily\strut  should}}\hskip 0pt plus 1pt\relax\allowbreak{}\colorbox[HTML]{FEFEFF}{\textcolor{black}{\tiny\ttfamily\strut  be}}\hskip 0pt plus 1pt\relax\allowbreak{}\colorbox[HTML]{EEEEFF}{\textcolor{black}{\tiny\ttfamily\strut  trained}}\hskip 0pt plus 1pt\relax\allowbreak{}\colorbox[HTML]{FFF1F1}{\textcolor{black}{\tiny\ttfamily\strut  on}}\hskip 0pt plus 1pt\relax\allowbreak{}\colorbox[HTML]{FFFFFF}{\textcolor{black}{\tiny\ttfamily\strut  effective}}\hskip 0pt plus 1pt\relax\allowbreak{}\colorbox[HTML]{FDFDFF}{\textcolor{black}{\tiny\ttfamily\strut  non}}\hskip 0pt plus 1pt\relax\allowbreak{}\colorbox[HTML]{FFFFFF}{\textcolor{black}{\tiny\ttfamily\strut -ver}}\hskip 0pt plus 1pt\relax\allowbreak{}\colorbox[HTML]{FFFFFF}{\textcolor{black}{\tiny\ttfamily\strut bal}}\hskip 0pt plus 1pt\relax\allowbreak{}\colorbox[HTML]{FFFFFF}{\textcolor{black}{\tiny\ttfamily\strut  communication}}\hskip 0pt plus 1pt\relax\allowbreak{}\colorbox[HTML]{FFF3F3}{\textcolor{black}{\tiny\ttfamily\strut  techniques}}\hskip 0pt plus 1pt\relax\allowbreak{}\colorbox[HTML]{FFF0F0}{\textcolor{black}{\tiny\ttfamily\strut ,}}\hskip 0pt plus 1pt\relax\allowbreak{}\colorbox[HTML]{FFE0E0}{\textcolor{black}{\tiny\ttfamily\strut  such}}\hskip 0pt plus 1pt\relax\allowbreak{}\colorbox[HTML]{FFFFFF}{\textcolor{black}{\tiny\ttfamily\strut  as}}\hskip 0pt plus 1pt\relax\allowbreak{}\colorbox[HTML]{FFFFFF}{\textcolor{black}{\tiny\ttfamily\strut  maintaining}}\hskip 0pt plus 1pt\relax\allowbreak{}\colorbox[HTML]{FBFBFF}{\textcolor{black}{\tiny\ttfamily\strut  eye}}\hskip 0pt plus 1pt\relax\allowbreak{}\colorbox[HTML]{FFFFFF}{\textcolor{black}{\tiny\ttfamily\strut  contact}}\hskip 0pt plus 1pt\relax\allowbreak{}\colorbox[HTML]{EFEFFF}{\textcolor{black}{\tiny\ttfamily\strut ,}}\hskip 0pt plus 1pt\relax\allowbreak{}\colorbox[HTML]{FFFFFF}{\textcolor{black}{\tiny\ttfamily\strut  using}}\hskip 0pt plus 1pt\relax\allowbreak{}\colorbox[HTML]{FFFFFF}{\textcolor{black}{\tiny\ttfamily\strut  appropriate}}\hskip 0pt plus 1pt\relax\allowbreak{}\colorbox[HTML]{EBEBFF}{\textcolor{black}{\tiny\ttfamily\strut  body}}\hskip 0pt plus 1pt\relax\allowbreak{}\colorbox[HTML]{FFFFFF}{\textcolor{black}{\tiny\ttfamily\strut  language}}\hskip 0pt plus 1pt\relax\allowbreak{}\colorbox[HTML]{FFFFFF}{\textcolor{black}{\tiny\ttfamily\strut ,}}\hskip 0pt plus 1pt\relax\allowbreak{}\colorbox[HTML]{FFFFFF}{\textcolor{black}{\tiny\ttfamily\strut  and}}\hskip 0pt plus 1pt\relax\allowbreak{}\colorbox[HTML]{FFFEFE}{\textcolor{black}{\tiny\ttfamily\strut  using}}\hskip 0pt plus 1pt\relax\allowbreak{}\colorbox[HTML]{FFF3F3}{\textcolor{black}{\tiny\ttfamily\strut  appropriate}}\hskip 0pt plus 1pt\relax\allowbreak{}\colorbox[HTML]{FFF7F7}{\textcolor{black}{\tiny\ttfamily\strut  gestures}}\hskip 0pt plus 1pt\relax\allowbreak{}\colorbox[HTML]{FFDCDC}{\textcolor{black}{\tiny\ttfamily\strut .\textcolor{gray}{$\hookleftarrow$}\textcolor{gray}{$\hookleftarrow$}}}\hskip 0pt plus 1pt\relax\allowbreak{}\colorbox[HTML]{9E9EFF}{\textcolor{black}{\tiny\ttfamily\strut Post}}\hskip 0pt plus 1pt\relax\allowbreak{}\colorbox[HTML]{FF8F8F}{\textcolor{black}{\tiny\ttfamily\strut script}}\hskip 0pt plus 1pt\relax\allowbreak{}\colorbox[HTML]{FF5151}{\textcolor{white}{\tiny\ttfamily\strut :\textcolor{gray}{$\hookleftarrow$}}}\hskip 0pt plus 1pt\relax\allowbreak{}\colorbox[HTML]{FFC3C3}{\textcolor{black}{\tiny\ttfamily\strut In}}\hskip 0pt plus 1pt\relax\allowbreak{}\colorbox[HTML]{B5B5FF}{\textcolor{black}{\tiny\ttfamily\strut  conclusion}}\hskip 0pt plus 1pt\relax\allowbreak{}\colorbox[HTML]{FFFFFF}{\textcolor{black}{\tiny\ttfamily\strut ,}}\hskip 0pt plus 1pt\relax\allowbreak{}\colorbox[HTML]{FFDCDC}{\textcolor{black}{\tiny\ttfamily\strut  communication}}\hskip 0pt plus 1pt\relax\allowbreak{}\colorbox[HTML]{FFFDFD}{\textcolor{black}{\tiny\ttfamily\strut  barriers}}\hskip 0pt plus 1pt\relax\allowbreak{}\colorbox[HTML]{FFF8F8}{\textcolor{black}{\tiny\ttfamily\strut  in}}\hskip 0pt plus 1pt\relax\allowbreak{}\colorbox[HTML]{FFFDFD}{\textcolor{black}{\tiny\ttfamily\strut  the}}\hskip 0pt plus 1pt\relax\allowbreak{}\colorbox[HTML]{FFFEFE}{\textcolor{black}{\tiny\ttfamily\strut  workplace}}\hskip 0pt plus 1pt\relax\allowbreak{}\colorbox[HTML]{FFFBFB}{\textcolor{black}{\tiny\ttfamily\strut  can}}\hskip 0pt plus 1pt\relax\allowbreak{}\colorbox[HTML]{FFE8E8}{\textcolor{black}{\tiny\ttfamily\strut  significantly}}\hskip 0pt plus 1pt\relax\allowbreak{}\colorbox[HTML]{FFFCFC}{\textcolor{black}{\tiny\ttfamily\strut  impact}}\hskip 0pt plus 1pt\relax\allowbreak{}\colorbox[HTML]{FEFEFF}{\textcolor{black}{\tiny\ttfamily\strut  productivity}}\hskip 0pt plus 1pt\relax\allowbreak{}\colorbox[HTML]{FEFEFF}{\textcolor{black}{\tiny\ttfamily\strut  and}}\hskip 0pt plus 1pt\relax\allowbreak{}\colorbox[HTML]{F8F8FF}{\textcolor{black}{\tiny\ttfamily\strut  collaboration}}\hskip 0pt plus 1pt\relax\allowbreak{}\colorbox[HTML]{FBFBFF}{\textcolor{black}{\tiny\ttfamily\strut .}}\hskip 0pt plus 1pt\relax\allowbreak{}\colorbox[HTML]{FFD8D8}{\textcolor{black}{\tiny\ttfamily\strut  To}}\hskip 0pt plus 1pt\relax\allowbreak{}\colorbox[HTML]{FFDCDC}{\textcolor{black}{\tiny\ttfamily\strut  overcome}}\hskip 0pt plus 1pt\relax\allowbreak{}\colorbox[HTML]{FFFBFB}{\textcolor{black}{\tiny\ttfamily\strut  these}}\hskip 0pt plus 1pt\relax\allowbreak{}\colorbox[HTML]{FFF1F1}{\textcolor{black}{\tiny\ttfamily\strut  barriers}}\hskip 0pt plus 1pt\relax\allowbreak{}\colorbox[HTML]{FFFFFF}{\textcolor{black}{\tiny\ttfamily\strut ,}}\hskip 0pt plus 1pt\relax\allowbreak{}\colorbox[HTML]{F2F2FF}{\textcolor{black}{\tiny\ttfamily\strut  organizations}}\hskip 0pt plus 1pt\relax\allowbreak{}\colorbox[HTML]{F1F1FF}{\textcolor{black}{\tiny\ttfamily\strut  should}}\hskip 0pt plus 1pt\relax\allowbreak{}\colorbox[HTML]{FFFEFE}{\textcolor{black}{\tiny\ttfamily\strut  prioritize}}\hskip 0pt plus 1pt\relax\allowbreak{}\colorbox[HTML]{FCFCFF}{\textcolor{black}{\tiny\ttfamily\strut  creating}}\hskip 0pt plus 1pt\relax\allowbreak{}\colorbox[HTML]{FFFFFF}{\textcolor{black}{\tiny\ttfamily\strut  a}}\hskip 0pt plus 1pt\relax\allowbreak{}\colorbox[HTML]{FFF8F8}{\textcolor{black}{\tiny\ttfamily\strut  diverse}}\hskip 0pt plus 1pt\relax\allowbreak{}\colorbox[HTML]{FFFEFE}{\textcolor{black}{\tiny\ttfamily\strut  and}}\hskip 0pt plus 1pt\relax\allowbreak{}\colorbox[HTML]{FFFFFF}{\textcolor{black}{\tiny\ttfamily\strut  inclusive}}\hskip 0pt plus 1pt\relax\allowbreak{}\colorbox[HTML]{DADAFF}{\textcolor{black}{\tiny\ttfamily\strut  workplace}}\hskip 0pt plus 1pt\relax\allowbreak{}\colorbox[HTML]{FFFAFA}{\textcolor{black}{\tiny\ttfamily\strut  culture}}\hskip 0pt plus 1pt\relax\allowbreak{}\colorbox[HTML]{FFF7F7}{\textcolor{black}{\tiny\ttfamily\strut ,}}\hskip 0pt plus 1pt\relax\allowbreak{}\colorbox[HTML]{FCFCFF}{\textcolor{black}{\tiny\ttfamily\strut  establishing}}\hskip 0pt plus 1pt\relax\allowbreak{}\colorbox[HTML]{FFFFFF}{\textcolor{black}{\tiny\ttfamily\strut  clear}}\hskip 0pt plus 1pt\relax\allowbreak{}\colorbox[HTML]{FFFFFF}{\textcolor{black}{\tiny\ttfamily\strut  communication}}\hskip 0pt plus 1pt\relax\allowbreak{}\colorbox[HTML]{FFFFFF}{\textcolor{black}{\tiny\ttfamily\strut  guidelines}}\hskip 0pt plus 1pt\relax\allowbreak{}\colorbox[HTML]{FFFFFF}{\textcolor{black}{\tiny\ttfamily\strut ,}}\hskip 0pt plus 1pt\relax\allowbreak{}\colorbox[HTML]{FFFDFD}{\textcolor{black}{\tiny\ttfamily\strut  and}}\hskip 0pt plus 1pt\relax\allowbreak{}\colorbox[HTML]{FDFDFF}{\textcolor{black}{\tiny\ttfamily\strut  providing}}\hskip 0pt plus 1pt\relax\allowbreak{}\colorbox[HTML]{FFFEFE}{\textcolor{black}{\tiny\ttfamily\strut  training}}\hskip 0pt plus 1pt\relax\allowbreak{}\colorbox[HTML]{F8F8FF}{\textcolor{black}{\tiny\ttfamily\strut  on}}\hskip 0pt plus 1pt\relax\allowbreak{}\colorbox[HTML]{FEFEFF}{\textcolor{black}{\tiny\ttfamily\strut  effective}}\hskip 0pt plus 1pt\relax\allowbreak{}\colorbox[HTML]{FFFEFE}{\textcolor{black}{\tiny\ttfamily\strut  communication}}\hskip 0pt plus 1pt\relax\allowbreak{}\colorbox[HTML]{FFF6F6}{\textcolor{black}{\tiny\ttfamily\strut  techniques}}\hskip 0pt plus 1pt\relax\allowbreak{}\colorbox[HTML]{FFFEFE}{\textcolor{black}{\tiny\ttfamily\strut .}}\hskip 0pt plus 1pt\relax\allowbreak{}\colorbox[HTML]{FFB5B5}{\textcolor{black}{\tiny\ttfamily\strut  By}}\hskip 0pt plus 1pt\relax\allowbreak{}\colorbox[HTML]{FFECEC}{\textcolor{black}{\tiny\ttfamily\strut  addressing}}\hskip 0pt plus 1pt\relax\allowbreak{}\colorbox[HTML]{F8F8FF}{\textcolor{black}{\tiny\ttfamily\strut  these}}\hskip 0pt plus 1pt\relax\allowbreak{}\colorbox[HTML]{FFD0D0}{\textcolor{black}{\tiny\ttfamily\strut  barriers}}\hskip 0pt plus 1pt\relax\allowbreak{}\colorbox[HTML]{FFFCFC}{\textcolor{black}{\tiny\ttfamily\strut ,}}\hskip 0pt plus 1pt\relax\allowbreak{}\colorbox[HTML]{FFFAFA}{\textcolor{black}{\tiny\ttfamily\strut  organizations}}\hskip 0pt plus 1pt\relax\allowbreak{}\colorbox[HTML]{FFFFFF}{\textcolor{black}{\tiny\ttfamily\strut  can}}\hskip 0pt plus 1pt\relax\allowbreak{}\colorbox[HTML]{FFF9F9}{\textcolor{black}{\tiny\ttfamily\strut  enhance}}\hskip 0pt plus 1pt\relax\allowbreak{}\colorbox[HTML]{FFF8F8}{\textcolor{black}{\tiny\ttfamily\strut  their}}\hskip 0pt plus 1pt\relax\allowbreak{}\colorbox[HTML]{EEEEFF}{\textcolor{black}{\tiny\ttfamily\strut  communication}}\hskip 0pt plus 1pt\relax\allowbreak{}\colorbox[HTML]{ECECFF}{\textcolor{black}{\tiny\ttfamily\strut  and}}\hskip 0pt plus 1pt\relax\allowbreak{}\colorbox[HTML]{FFF6F6}{\textcolor{black}{\tiny\ttfamily\strut  work}}\hskip 0pt plus 1pt\relax\allowbreak{}\colorbox[HTML]{FFF6F6}{\textcolor{black}{\tiny\ttfamily\strut  efficiency}}\hskip 0pt plus 1pt\relax\allowbreak{}\colorbox[HTML]{FFFFFF}{\textcolor{black}{\tiny\ttfamily\strut ,}}\hskip 0pt plus 1pt\relax\allowbreak{}\colorbox[HTML]{FFEEEE}{\textcolor{black}{\tiny\ttfamily\strut  ultimately}}\hskip 0pt plus 1pt\relax\allowbreak{}\colorbox[HTML]{FFFFFF}{\textcolor{black}{\tiny\ttfamily\strut  leading}}\hskip 0pt plus 1pt\relax\allowbreak{}\colorbox[HTML]{FFFFFF}{\textcolor{black}{\tiny\ttfamily\strut  to}}\hskip 0pt plus 1pt\relax\allowbreak{}\colorbox[HTML]{FFF9F9}{\textcolor{black}{\tiny\ttfamily\strut  better}}\hskip 0pt plus 1pt\relax\allowbreak{}\colorbox[HTML]{F7F7FF}{\textcolor{black}{\tiny\ttfamily\strut  outcomes}}\hskip 0pt plus 1pt\relax\allowbreak{}\colorbox[HTML]{ECECFF}{\textcolor{black}{\tiny\ttfamily\strut  for}}\hskip 0pt plus 1pt\relax\allowbreak{}\colorbox[HTML]{FFEEEE}{\textcolor{black}{\tiny\ttfamily\strut  their}}\hskip 0pt plus 1pt\relax\allowbreak{}\colorbox[HTML]{F2F2FF}{\textcolor{black}{\tiny\ttfamily\strut  employees}}\hskip 0pt plus 1pt\relax\allowbreak{}\colorbox[HTML]{FFFFFF}{\textcolor{black}{\tiny\ttfamily\strut  and}}\hskip 0pt plus 1pt\relax\allowbreak{}\colorbox[HTML]{FFF1F1}{\textcolor{black}{\tiny\ttfamily\strut  clients}}\hskip 0pt plus 1pt\relax\allowbreak{}\colorbox[HTML]{FFDBDB}{\textcolor{black}{\tiny\ttfamily\strut .}}\hskip 0pt plus 1pt\relax\allowbreak{}\colorbox[HTML]{4040FF}{\textcolor{white}{\tiny\ttfamily\strut \textless{}\textbar{}im\_end\textbar{}\textgreater{}}}\hskip 0pt plus 1pt\relax\allowbreak{}
\par\medskip
\noindent\textbf{\textcolor[HTML]{2257bf}{LA}} {\scriptsize $|\delta|_{\max}{=}20.9$}: \\[2pt]
\sloppy
\colorbox[HTML]{FF4040}{\textcolor{white}{\tiny\ttfamily\strut *}}\hskip 0pt plus 1pt\relax\allowbreak{}\colorbox[HTML]{FFC5C5}{\textcolor{black}{\tiny\ttfamily\strut  Communication}}\hskip 0pt plus 1pt\relax\allowbreak{}\colorbox[HTML]{FCFCFF}{\textcolor{black}{\tiny\ttfamily\strut  barriers}}\hskip 0pt plus 1pt\relax\allowbreak{}\colorbox[HTML]{F4F4FF}{\textcolor{black}{\tiny\ttfamily\strut  in}}\hskip 0pt plus 1pt\relax\allowbreak{}\colorbox[HTML]{FFEBEB}{\textcolor{black}{\tiny\ttfamily\strut  a}}\hskip 0pt plus 1pt\relax\allowbreak{}\colorbox[HTML]{FEFEFF}{\textcolor{black}{\tiny\ttfamily\strut  workplace}}\hskip 0pt plus 1pt\relax\allowbreak{}\colorbox[HTML]{FEFEFF}{\textcolor{black}{\tiny\ttfamily\strut  can}}\hskip 0pt plus 1pt\relax\allowbreak{}\colorbox[HTML]{FFC8C8}{\textcolor{black}{\tiny\ttfamily\strut  manifest}}\hskip 0pt plus 1pt\relax\allowbreak{}\colorbox[HTML]{F5F5FF}{\textcolor{black}{\tiny\ttfamily\strut  in}}\hskip 0pt plus 1pt\relax\allowbreak{}\colorbox[HTML]{FAFAFF}{\textcolor{black}{\tiny\ttfamily\strut  various}}\hskip 0pt plus 1pt\relax\allowbreak{}\colorbox[HTML]{F7F7FF}{\textcolor{black}{\tiny\ttfamily\strut  ways}}\hskip 0pt plus 1pt\relax\allowbreak{}\colorbox[HTML]{FFD8D8}{\textcolor{black}{\tiny\ttfamily\strut ,}}\hskip 0pt plus 1pt\relax\allowbreak{}\colorbox[HTML]{FFBDBD}{\textcolor{black}{\tiny\ttfamily\strut  including}}\hskip 0pt plus 1pt\relax\allowbreak{}\colorbox[HTML]{DADAFF}{\textcolor{black}{\tiny\ttfamily\strut  language}}\hskip 0pt plus 1pt\relax\allowbreak{}\colorbox[HTML]{FFFDFD}{\textcolor{black}{\tiny\ttfamily\strut  differences}}\hskip 0pt plus 1pt\relax\allowbreak{}\colorbox[HTML]{DADAFF}{\textcolor{black}{\tiny\ttfamily\strut ,}}\hskip 0pt plus 1pt\relax\allowbreak{}\colorbox[HTML]{DEDEFF}{\textcolor{black}{\tiny\ttfamily\strut  cultural}}\hskip 0pt plus 1pt\relax\allowbreak{}\colorbox[HTML]{FFFEFE}{\textcolor{black}{\tiny\ttfamily\strut  misunderstand}}\hskip 0pt plus 1pt\relax\allowbreak{}\colorbox[HTML]{FFFFFF}{\textcolor{black}{\tiny\ttfamily\strut ings}}\hskip 0pt plus 1pt\relax\allowbreak{}\colorbox[HTML]{FFFFFF}{\textcolor{black}{\tiny\ttfamily\strut ,}}\hskip 0pt plus 1pt\relax\allowbreak{}\colorbox[HTML]{FFA4A4}{\textcolor{black}{\tiny\ttfamily\strut  and}}\hskip 0pt plus 1pt\relax\allowbreak{}\colorbox[HTML]{FFFBFB}{\textcolor{black}{\tiny\ttfamily\strut  technical}}\hskip 0pt plus 1pt\relax\allowbreak{}\colorbox[HTML]{FFE8E8}{\textcolor{black}{\tiny\ttfamily\strut  j}}\hskip 0pt plus 1pt\relax\allowbreak{}\colorbox[HTML]{FFFFFF}{\textcolor{black}{\tiny\ttfamily\strut argon}}\hskip 0pt plus 1pt\relax\allowbreak{}\colorbox[HTML]{6363FF}{\textcolor{white}{\tiny\ttfamily\strut .\textcolor{gray}{$\hookleftarrow$}}}\hskip 0pt plus 1pt\relax\allowbreak{}\colorbox[HTML]{FCFCFF}{\textcolor{black}{\tiny\ttfamily\strut  }}\hskip 0pt plus 1pt\relax\allowbreak{}\colorbox[HTML]{FFFFFF}{\textcolor{black}{\tiny\ttfamily\strut  *}}\hskip 0pt plus 1pt\relax\allowbreak{}\colorbox[HTML]{7F7FFF}{\textcolor{black}{\tiny\ttfamily\strut  Language}}\hskip 0pt plus 1pt\relax\allowbreak{}\colorbox[HTML]{DDDDFF}{\textcolor{black}{\tiny\ttfamily\strut  differences}}\hskip 0pt plus 1pt\relax\allowbreak{}\colorbox[HTML]{C3C3FF}{\textcolor{black}{\tiny\ttfamily\strut  can}}\hskip 0pt plus 1pt\relax\allowbreak{}\colorbox[HTML]{7D7DFF}{\textcolor{white}{\tiny\ttfamily\strut  arise}}\hskip 0pt plus 1pt\relax\allowbreak{}\colorbox[HTML]{CECEFF}{\textcolor{black}{\tiny\ttfamily\strut  due}}\hskip 0pt plus 1pt\relax\allowbreak{}\colorbox[HTML]{FFFFFF}{\textcolor{black}{\tiny\ttfamily\strut  to}}\hskip 0pt plus 1pt\relax\allowbreak{}\colorbox[HTML]{FFFCFC}{\textcolor{black}{\tiny\ttfamily\strut  the}}\hskip 0pt plus 1pt\relax\allowbreak{}\colorbox[HTML]{C9C9FF}{\textcolor{black}{\tiny\ttfamily\strut  use}}\hskip 0pt plus 1pt\relax\allowbreak{}\colorbox[HTML]{FFFFFF}{\textcolor{black}{\tiny\ttfamily\strut  of}}\hskip 0pt plus 1pt\relax\allowbreak{}\colorbox[HTML]{D4D4FF}{\textcolor{black}{\tiny\ttfamily\strut  different}}\hskip 0pt plus 1pt\relax\allowbreak{}\colorbox[HTML]{C0C0FF}{\textcolor{black}{\tiny\ttfamily\strut  languages}}\hskip 0pt plus 1pt\relax\allowbreak{}\colorbox[HTML]{FFFDFD}{\textcolor{black}{\tiny\ttfamily\strut  or}}\hskip 0pt plus 1pt\relax\allowbreak{}\colorbox[HTML]{FBFBFF}{\textcolor{black}{\tiny\ttfamily\strut  dialect}}\hskip 0pt plus 1pt\relax\allowbreak{}\colorbox[HTML]{FFFFFF}{\textcolor{black}{\tiny\ttfamily\strut s}}\hskip 0pt plus 1pt\relax\allowbreak{}\colorbox[HTML]{FFEDED}{\textcolor{black}{\tiny\ttfamily\strut ,}}\hskip 0pt plus 1pt\relax\allowbreak{}\colorbox[HTML]{FDFDFF}{\textcolor{black}{\tiny\ttfamily\strut  which}}\hskip 0pt plus 1pt\relax\allowbreak{}\colorbox[HTML]{FDFDFF}{\textcolor{black}{\tiny\ttfamily\strut  can}}\hskip 0pt plus 1pt\relax\allowbreak{}\colorbox[HTML]{E4E4FF}{\textcolor{black}{\tiny\ttfamily\strut  lead}}\hskip 0pt plus 1pt\relax\allowbreak{}\colorbox[HTML]{FFFFFF}{\textcolor{black}{\tiny\ttfamily\strut  to}}\hskip 0pt plus 1pt\relax\allowbreak{}\colorbox[HTML]{D9D9FF}{\textcolor{black}{\tiny\ttfamily\strut  mis}}\hskip 0pt plus 1pt\relax\allowbreak{}\colorbox[HTML]{FCFCFF}{\textcolor{black}{\tiny\ttfamily\strut interpret}}\hskip 0pt plus 1pt\relax\allowbreak{}\colorbox[HTML]{E1E1FF}{\textcolor{black}{\tiny\ttfamily\strut ation}}\hskip 0pt plus 1pt\relax\allowbreak{}\colorbox[HTML]{D4D4FF}{\textcolor{black}{\tiny\ttfamily\strut  of}}\hskip 0pt plus 1pt\relax\allowbreak{}\colorbox[HTML]{FFFFFF}{\textcolor{black}{\tiny\ttfamily\strut  messages}}\hskip 0pt plus 1pt\relax\allowbreak{}\colorbox[HTML]{FAFAFF}{\textcolor{black}{\tiny\ttfamily\strut .}}\hskip 0pt plus 1pt\relax\allowbreak{}\colorbox[HTML]{C9C9FF}{\textcolor{black}{\tiny\ttfamily\strut  To}}\hskip 0pt plus 1pt\relax\allowbreak{}\colorbox[HTML]{FCFCFF}{\textcolor{black}{\tiny\ttfamily\strut  overcome}}\hskip 0pt plus 1pt\relax\allowbreak{}\colorbox[HTML]{F0F0FF}{\textcolor{black}{\tiny\ttfamily\strut  this}}\hskip 0pt plus 1pt\relax\allowbreak{}\colorbox[HTML]{FFFDFD}{\textcolor{black}{\tiny\ttfamily\strut  barrier}}\hskip 0pt plus 1pt\relax\allowbreak{}\colorbox[HTML]{FFFFFF}{\textcolor{black}{\tiny\ttfamily\strut ,}}\hskip 0pt plus 1pt\relax\allowbreak{}\colorbox[HTML]{AEAEFF}{\textcolor{black}{\tiny\ttfamily\strut  organizations}}\hskip 0pt plus 1pt\relax\allowbreak{}\colorbox[HTML]{FAFAFF}{\textcolor{black}{\tiny\ttfamily\strut  can}}\hskip 0pt plus 1pt\relax\allowbreak{}\colorbox[HTML]{FFBEBE}{\textcolor{black}{\tiny\ttfamily\strut  implement}}\hskip 0pt plus 1pt\relax\allowbreak{}\colorbox[HTML]{FFFFFF}{\textcolor{black}{\tiny\ttfamily\strut  mult}}\hskip 0pt plus 1pt\relax\allowbreak{}\colorbox[HTML]{FFFFFF}{\textcolor{black}{\tiny\ttfamily\strut ilingual}}\hskip 0pt plus 1pt\relax\allowbreak{}\colorbox[HTML]{F9F9FF}{\textcolor{black}{\tiny\ttfamily\strut  communication}}\hskip 0pt plus 1pt\relax\allowbreak{}\colorbox[HTML]{FFFBFB}{\textcolor{black}{\tiny\ttfamily\strut  tools}}\hskip 0pt plus 1pt\relax\allowbreak{}\colorbox[HTML]{FFFCFC}{\textcolor{black}{\tiny\ttfamily\strut  or}}\hskip 0pt plus 1pt\relax\allowbreak{}\colorbox[HTML]{FAFAFF}{\textcolor{black}{\tiny\ttfamily\strut  provide}}\hskip 0pt plus 1pt\relax\allowbreak{}\colorbox[HTML]{FFF1F1}{\textcolor{black}{\tiny\ttfamily\strut  translation}}\hskip 0pt plus 1pt\relax\allowbreak{}\colorbox[HTML]{FFFFFF}{\textcolor{black}{\tiny\ttfamily\strut  services}}\hskip 0pt plus 1pt\relax\allowbreak{}\colorbox[HTML]{FFFEFE}{\textcolor{black}{\tiny\ttfamily\strut  for}}\hskip 0pt plus 1pt\relax\allowbreak{}\colorbox[HTML]{CECEFF}{\textcolor{black}{\tiny\ttfamily\strut  employees}}\hskip 0pt plus 1pt\relax\allowbreak{}\colorbox[HTML]{E9E9FF}{\textcolor{black}{\tiny\ttfamily\strut  who}}\hskip 0pt plus 1pt\relax\allowbreak{}\colorbox[HTML]{FAFAFF}{\textcolor{black}{\tiny\ttfamily\strut  speak}}\hskip 0pt plus 1pt\relax\allowbreak{}\colorbox[HTML]{FFFFFF}{\textcolor{black}{\tiny\ttfamily\strut  different}}\hskip 0pt plus 1pt\relax\allowbreak{}\colorbox[HTML]{FFFFFF}{\textcolor{black}{\tiny\ttfamily\strut  languages}}\hskip 0pt plus 1pt\relax\allowbreak{}\colorbox[HTML]{8E8EFF}{\textcolor{black}{\tiny\ttfamily\strut .\textcolor{gray}{$\hookleftarrow$}}}\hskip 0pt plus 1pt\relax\allowbreak{}\colorbox[HTML]{FFFFFF}{\textcolor{black}{\tiny\ttfamily\strut  }}\hskip 0pt plus 1pt\relax\allowbreak{}\colorbox[HTML]{FEFEFF}{\textcolor{black}{\tiny\ttfamily\strut  *}}\hskip 0pt plus 1pt\relax\allowbreak{}\colorbox[HTML]{8A8AFF}{\textcolor{black}{\tiny\ttfamily\strut  Cultural}}\hskip 0pt plus 1pt\relax\allowbreak{}\colorbox[HTML]{E3E3FF}{\textcolor{black}{\tiny\ttfamily\strut  misunderstand}}\hskip 0pt plus 1pt\relax\allowbreak{}\colorbox[HTML]{FFFFFF}{\textcolor{black}{\tiny\ttfamily\strut ings}}\hskip 0pt plus 1pt\relax\allowbreak{}\colorbox[HTML]{D5D5FF}{\textcolor{black}{\tiny\ttfamily\strut  can}}\hskip 0pt plus 1pt\relax\allowbreak{}\colorbox[HTML]{7272FF}{\textcolor{white}{\tiny\ttfamily\strut  occur}}\hskip 0pt plus 1pt\relax\allowbreak{}\colorbox[HTML]{FFFFFF}{\textcolor{black}{\tiny\ttfamily\strut  when}}\hskip 0pt plus 1pt\relax\allowbreak{}\colorbox[HTML]{F2F2FF}{\textcolor{black}{\tiny\ttfamily\strut  individuals}}\hskip 0pt plus 1pt\relax\allowbreak{}\colorbox[HTML]{FFFFFF}{\textcolor{black}{\tiny\ttfamily\strut  from}}\hskip 0pt plus 1pt\relax\allowbreak{}\colorbox[HTML]{F5F5FF}{\textcolor{black}{\tiny\ttfamily\strut  different}}\hskip 0pt plus 1pt\relax\allowbreak{}\colorbox[HTML]{FFEBEB}{\textcolor{black}{\tiny\ttfamily\strut  backgrounds}}\hskip 0pt plus 1pt\relax\allowbreak{}\colorbox[HTML]{F6F6FF}{\textcolor{black}{\tiny\ttfamily\strut  communicate}}\hskip 0pt plus 1pt\relax\allowbreak{}\colorbox[HTML]{D1D1FF}{\textcolor{black}{\tiny\ttfamily\strut  with}}\hskip 0pt plus 1pt\relax\allowbreak{}\colorbox[HTML]{FFFDFD}{\textcolor{black}{\tiny\ttfamily\strut  each}}\hskip 0pt plus 1pt\relax\allowbreak{}\colorbox[HTML]{FFFFFF}{\textcolor{black}{\tiny\ttfamily\strut  other}}\hskip 0pt plus 1pt\relax\allowbreak{}\colorbox[HTML]{EAEAFF}{\textcolor{black}{\tiny\ttfamily\strut .}}\hskip 0pt plus 1pt\relax\allowbreak{}\colorbox[HTML]{D6D6FF}{\textcolor{black}{\tiny\ttfamily\strut  To}}\hskip 0pt plus 1pt\relax\allowbreak{}\colorbox[HTML]{FFFAFA}{\textcolor{black}{\tiny\ttfamily\strut  address}}\hskip 0pt plus 1pt\relax\allowbreak{}\colorbox[HTML]{E3E3FF}{\textcolor{black}{\tiny\ttfamily\strut  this}}\hskip 0pt plus 1pt\relax\allowbreak{}\colorbox[HTML]{FEFEFF}{\textcolor{black}{\tiny\ttfamily\strut ,}}\hskip 0pt plus 1pt\relax\allowbreak{}\colorbox[HTML]{E9E9FF}{\textcolor{black}{\tiny\ttfamily\strut  it}}\hskip 0pt plus 1pt\relax\allowbreak{}\colorbox[HTML]{FFF6F6}{\textcolor{black}{\tiny\ttfamily\strut  is}}\hskip 0pt plus 1pt\relax\allowbreak{}\colorbox[HTML]{F6F6FF}{\textcolor{black}{\tiny\ttfamily\strut  essential}}\hskip 0pt plus 1pt\relax\allowbreak{}\colorbox[HTML]{FDFDFF}{\textcolor{black}{\tiny\ttfamily\strut  to}}\hskip 0pt plus 1pt\relax\allowbreak{}\colorbox[HTML]{FFFFFF}{\textcolor{black}{\tiny\ttfamily\strut  foster}}\hskip 0pt plus 1pt\relax\allowbreak{}\colorbox[HTML]{FFFBFB}{\textcolor{black}{\tiny\ttfamily\strut  an}}\hskip 0pt plus 1pt\relax\allowbreak{}\colorbox[HTML]{F9F9FF}{\textcolor{black}{\tiny\ttfamily\strut  inclusive}}\hskip 0pt plus 1pt\relax\allowbreak{}\colorbox[HTML]{CBCBFF}{\textcolor{black}{\tiny\ttfamily\strut  and}}\hskip 0pt plus 1pt\relax\allowbreak{}\colorbox[HTML]{E1E1FF}{\textcolor{black}{\tiny\ttfamily\strut  respectful}}\hskip 0pt plus 1pt\relax\allowbreak{}\colorbox[HTML]{EBEBFF}{\textcolor{black}{\tiny\ttfamily\strut  workplace}}\hskip 0pt plus 1pt\relax\allowbreak{}\colorbox[HTML]{FAFAFF}{\textcolor{black}{\tiny\ttfamily\strut  culture}}\hskip 0pt plus 1pt\relax\allowbreak{}\colorbox[HTML]{D5D5FF}{\textcolor{black}{\tiny\ttfamily\strut  that}}\hskip 0pt plus 1pt\relax\allowbreak{}\colorbox[HTML]{EBEBFF}{\textcolor{black}{\tiny\ttfamily\strut  values}}\hskip 0pt plus 1pt\relax\allowbreak{}\colorbox[HTML]{FCFCFF}{\textcolor{black}{\tiny\ttfamily\strut  diversity}}\hskip 0pt plus 1pt\relax\allowbreak{}\colorbox[HTML]{D9D9FF}{\textcolor{black}{\tiny\ttfamily\strut  and}}\hskip 0pt plus 1pt\relax\allowbreak{}\colorbox[HTML]{FCFCFF}{\textcolor{black}{\tiny\ttfamily\strut  promotes}}\hskip 0pt plus 1pt\relax\allowbreak{}\colorbox[HTML]{E4E4FF}{\textcolor{black}{\tiny\ttfamily\strut  cross}}\hskip 0pt plus 1pt\relax\allowbreak{}\colorbox[HTML]{FFFFFF}{\textcolor{black}{\tiny\ttfamily\strut -cultural}}\hskip 0pt plus 1pt\relax\allowbreak{}\colorbox[HTML]{F6F6FF}{\textcolor{black}{\tiny\ttfamily\strut  understanding}}\hskip 0pt plus 1pt\relax\allowbreak{}\colorbox[HTML]{FFF9F9}{\textcolor{black}{\tiny\ttfamily\strut .}}\hskip 0pt plus 1pt\relax\allowbreak{}\colorbox[HTML]{FEFEFF}{\textcolor{black}{\tiny\ttfamily\strut  This}}\hskip 0pt plus 1pt\relax\allowbreak{}\colorbox[HTML]{FFFFFF}{\textcolor{black}{\tiny\ttfamily\strut  can}}\hskip 0pt plus 1pt\relax\allowbreak{}\colorbox[HTML]{FFFDFD}{\textcolor{black}{\tiny\ttfamily\strut  be}}\hskip 0pt plus 1pt\relax\allowbreak{}\colorbox[HTML]{FFFFFF}{\textcolor{black}{\tiny\ttfamily\strut  achieved}}\hskip 0pt plus 1pt\relax\allowbreak{}\colorbox[HTML]{F8F8FF}{\textcolor{black}{\tiny\ttfamily\strut  through}}\hskip 0pt plus 1pt\relax\allowbreak{}\colorbox[HTML]{FFFDFD}{\textcolor{black}{\tiny\ttfamily\strut  training}}\hskip 0pt plus 1pt\relax\allowbreak{}\colorbox[HTML]{EFEFFF}{\textcolor{black}{\tiny\ttfamily\strut  programs}}\hskip 0pt plus 1pt\relax\allowbreak{}\colorbox[HTML]{FFF2F2}{\textcolor{black}{\tiny\ttfamily\strut  that}}\hskip 0pt plus 1pt\relax\allowbreak{}\colorbox[HTML]{EAEAFF}{\textcolor{black}{\tiny\ttfamily\strut  teach}}\hskip 0pt plus 1pt\relax\allowbreak{}\colorbox[HTML]{F6F6FF}{\textcolor{black}{\tiny\ttfamily\strut  employees}}\hskip 0pt plus 1pt\relax\allowbreak{}\colorbox[HTML]{FEFEFF}{\textcolor{black}{\tiny\ttfamily\strut  about}}\hskip 0pt plus 1pt\relax\allowbreak{}\colorbox[HTML]{F7F7FF}{\textcolor{black}{\tiny\ttfamily\strut  different}}\hskip 0pt plus 1pt\relax\allowbreak{}\colorbox[HTML]{FFF9F9}{\textcolor{black}{\tiny\ttfamily\strut  cultural}}\hskip 0pt plus 1pt\relax\allowbreak{}\colorbox[HTML]{F6F6FF}{\textcolor{black}{\tiny\ttfamily\strut  perspectives}}\hskip 0pt plus 1pt\relax\allowbreak{}\colorbox[HTML]{FFFFFF}{\textcolor{black}{\tiny\ttfamily\strut  and}}\hskip 0pt plus 1pt\relax\allowbreak{}\colorbox[HTML]{C0C0FF}{\textcolor{black}{\tiny\ttfamily\strut  values}}\hskip 0pt plus 1pt\relax\allowbreak{}\colorbox[HTML]{5E5EFF}{\textcolor{white}{\tiny\ttfamily\strut .\textcolor{gray}{$\hookleftarrow$}}}\hskip 0pt plus 1pt\relax\allowbreak{}\colorbox[HTML]{4040FF}{\textcolor{white}{\tiny\ttfamily\strut  }}\hskip 0pt plus 1pt\relax\allowbreak{}\colorbox[HTML]{8484FF}{\textcolor{black}{\tiny\ttfamily\strut  *}}\hskip 0pt plus 1pt\relax\allowbreak{}\colorbox[HTML]{6D6DFF}{\textcolor{white}{\tiny\ttfamily\strut  Technical}}\hskip 0pt plus 1pt\relax\allowbreak{}\colorbox[HTML]{FBFBFF}{\textcolor{black}{\tiny\ttfamily\strut  j}}\hskip 0pt plus 1pt\relax\allowbreak{}\colorbox[HTML]{FFFFFF}{\textcolor{black}{\tiny\ttfamily\strut argon}}\hskip 0pt plus 1pt\relax\allowbreak{}\colorbox[HTML]{F8F8FF}{\textcolor{black}{\tiny\ttfamily\strut  can}}\hskip 0pt plus 1pt\relax\allowbreak{}\colorbox[HTML]{EEEEFF}{\textcolor{black}{\tiny\ttfamily\strut  be}}\hskip 0pt plus 1pt\relax\allowbreak{}\colorbox[HTML]{FCFCFF}{\textcolor{black}{\tiny\ttfamily\strut  a}}\hskip 0pt plus 1pt\relax\allowbreak{}\colorbox[HTML]{FFEEEE}{\textcolor{black}{\tiny\ttfamily\strut  significant}}\hskip 0pt plus 1pt\relax\allowbreak{}\colorbox[HTML]{FFFFFF}{\textcolor{black}{\tiny\ttfamily\strut  barrier}}\hskip 0pt plus 1pt\relax\allowbreak{}\colorbox[HTML]{FFF3F3}{\textcolor{black}{\tiny\ttfamily\strut  in}}\hskip 0pt plus 1pt\relax\allowbreak{}\colorbox[HTML]{FFC5C5}{\textcolor{black}{\tiny\ttfamily\strut  professional}}\hskip 0pt plus 1pt\relax\allowbreak{}\colorbox[HTML]{F9F9FF}{\textcolor{black}{\tiny\ttfamily\strut  communication}}\hskip 0pt plus 1pt\relax\allowbreak{}\colorbox[HTML]{FFFFFF}{\textcolor{black}{\tiny\ttfamily\strut ,}}\hskip 0pt plus 1pt\relax\allowbreak{}\colorbox[HTML]{FBFBFF}{\textcolor{black}{\tiny\ttfamily\strut  especially}}\hskip 0pt plus 1pt\relax\allowbreak{}\colorbox[HTML]{F4F4FF}{\textcolor{black}{\tiny\ttfamily\strut  when}}\hskip 0pt plus 1pt\relax\allowbreak{}\colorbox[HTML]{D0D0FF}{\textcolor{black}{\tiny\ttfamily\strut  working}}\hskip 0pt plus 1pt\relax\allowbreak{}\colorbox[HTML]{FFECEC}{\textcolor{black}{\tiny\ttfamily\strut  in}}\hskip 0pt plus 1pt\relax\allowbreak{}\colorbox[HTML]{D7D7FF}{\textcolor{black}{\tiny\ttfamily\strut  a}}\hskip 0pt plus 1pt\relax\allowbreak{}\colorbox[HTML]{FFEFEF}{\textcolor{black}{\tiny\ttfamily\strut  team}}\hskip 0pt plus 1pt\relax\allowbreak{}\colorbox[HTML]{FFFFFF}{\textcolor{black}{\tiny\ttfamily\strut  with}}\hskip 0pt plus 1pt\relax\allowbreak{}\colorbox[HTML]{F1F1FF}{\textcolor{black}{\tiny\ttfamily\strut  members}}\hskip 0pt plus 1pt\relax\allowbreak{}\colorbox[HTML]{FFFEFE}{\textcolor{black}{\tiny\ttfamily\strut  from}}\hskip 0pt plus 1pt\relax\allowbreak{}\colorbox[HTML]{FFEBEB}{\textcolor{black}{\tiny\ttfamily\strut  diverse}}\hskip 0pt plus 1pt\relax\allowbreak{}\colorbox[HTML]{D9D9FF}{\textcolor{black}{\tiny\ttfamily\strut  backgrounds}}\hskip 0pt plus 1pt\relax\allowbreak{}\colorbox[HTML]{FBFBFF}{\textcolor{black}{\tiny\ttfamily\strut  or}}\hskip 0pt plus 1pt\relax\allowbreak{}\colorbox[HTML]{FAFAFF}{\textcolor{black}{\tiny\ttfamily\strut  in}}\hskip 0pt plus 1pt\relax\allowbreak{}\colorbox[HTML]{F5F5FF}{\textcolor{black}{\tiny\ttfamily\strut  an}}\hskip 0pt plus 1pt\relax\allowbreak{}\colorbox[HTML]{FDFDFF}{\textcolor{black}{\tiny\ttfamily\strut  industry}}\hskip 0pt plus 1pt\relax\allowbreak{}\colorbox[HTML]{FEFEFF}{\textcolor{black}{\tiny\ttfamily\strut  with}}\hskip 0pt plus 1pt\relax\allowbreak{}\colorbox[HTML]{FFFCFC}{\textcolor{black}{\tiny\ttfamily\strut  specialized}}\hskip 0pt plus 1pt\relax\allowbreak{}\colorbox[HTML]{FBFBFF}{\textcolor{black}{\tiny\ttfamily\strut  terminology}}\hskip 0pt plus 1pt\relax\allowbreak{}\colorbox[HTML]{FFFFFF}{\textcolor{black}{\tiny\ttfamily\strut .}}\hskip 0pt plus 1pt\relax\allowbreak{}\colorbox[HTML]{ECECFF}{\textcolor{black}{\tiny\ttfamily\strut  To}}\hskip 0pt plus 1pt\relax\allowbreak{}\colorbox[HTML]{DADAFF}{\textcolor{black}{\tiny\ttfamily\strut  overcome}}\hskip 0pt plus 1pt\relax\allowbreak{}\colorbox[HTML]{FCFCFF}{\textcolor{black}{\tiny\ttfamily\strut  this}}\hskip 0pt plus 1pt\relax\allowbreak{}\colorbox[HTML]{FCFCFF}{\textcolor{black}{\tiny\ttfamily\strut  barrier}}\hskip 0pt plus 1pt\relax\allowbreak{}\colorbox[HTML]{FFFFFF}{\textcolor{black}{\tiny\ttfamily\strut ,}}\hskip 0pt plus 1pt\relax\allowbreak{}\colorbox[HTML]{8484FF}{\textcolor{black}{\tiny\ttfamily\strut  organizations}}\hskip 0pt plus 1pt\relax\allowbreak{}\colorbox[HTML]{FFC1C1}{\textcolor{black}{\tiny\ttfamily\strut  should}}\hskip 0pt plus 1pt\relax\allowbreak{}\colorbox[HTML]{FBFBFF}{\textcolor{black}{\tiny\ttfamily\strut  establish}}\hskip 0pt plus 1pt\relax\allowbreak{}\colorbox[HTML]{FEFEFF}{\textcolor{black}{\tiny\ttfamily\strut  clear}}\hskip 0pt plus 1pt\relax\allowbreak{}\colorbox[HTML]{F4F4FF}{\textcolor{black}{\tiny\ttfamily\strut  communication}}\hskip 0pt plus 1pt\relax\allowbreak{}\colorbox[HTML]{FBFBFF}{\textcolor{black}{\tiny\ttfamily\strut  guidelines}}\hskip 0pt plus 1pt\relax\allowbreak{}\colorbox[HTML]{FFF2F2}{\textcolor{black}{\tiny\ttfamily\strut  that}}\hskip 0pt plus 1pt\relax\allowbreak{}\colorbox[HTML]{F1F1FF}{\textcolor{black}{\tiny\ttfamily\strut  define}}\hskip 0pt plus 1pt\relax\allowbreak{}\colorbox[HTML]{EAEAFF}{\textcolor{black}{\tiny\ttfamily\strut  technical}}\hskip 0pt plus 1pt\relax\allowbreak{}\colorbox[HTML]{FFFFFF}{\textcolor{black}{\tiny\ttfamily\strut  terms}}\hskip 0pt plus 1pt\relax\allowbreak{}\colorbox[HTML]{FEFEFF}{\textcolor{black}{\tiny\ttfamily\strut  and}}\hskip 0pt plus 1pt\relax\allowbreak{}\colorbox[HTML]{FFECEC}{\textcolor{black}{\tiny\ttfamily\strut  ensure}}\hskip 0pt plus 1pt\relax\allowbreak{}\colorbox[HTML]{F8F8FF}{\textcolor{black}{\tiny\ttfamily\strut  that}}\hskip 0pt plus 1pt\relax\allowbreak{}\colorbox[HTML]{FFF9F9}{\textcolor{black}{\tiny\ttfamily\strut  all}}\hskip 0pt plus 1pt\relax\allowbreak{}\colorbox[HTML]{FFFAFA}{\textcolor{black}{\tiny\ttfamily\strut  employees}}\hskip 0pt plus 1pt\relax\allowbreak{}\colorbox[HTML]{FFFEFE}{\textcolor{black}{\tiny\ttfamily\strut  are}}\hskip 0pt plus 1pt\relax\allowbreak{}\colorbox[HTML]{F8F8FF}{\textcolor{black}{\tiny\ttfamily\strut  familiar}}\hskip 0pt plus 1pt\relax\allowbreak{}\colorbox[HTML]{FFFFFF}{\textcolor{black}{\tiny\ttfamily\strut  with}}\hskip 0pt plus 1pt\relax\allowbreak{}\colorbox[HTML]{F2F2FF}{\textcolor{black}{\tiny\ttfamily\strut  them}}\hskip 0pt plus 1pt\relax\allowbreak{}\colorbox[HTML]{FFF3F3}{\textcolor{black}{\tiny\ttfamily\strut .\textcolor{gray}{$\hookleftarrow$}}}\hskip 0pt plus 1pt\relax\allowbreak{}\colorbox[HTML]{ECECFF}{\textcolor{black}{\tiny\ttfamily\strut  }}\hskip 0pt plus 1pt\relax\allowbreak{}\colorbox[HTML]{F9F9FF}{\textcolor{black}{\tiny\ttfamily\strut  *}}\hskip 0pt plus 1pt\relax\allowbreak{}\colorbox[HTML]{CFCFFF}{\textcolor{black}{\tiny\ttfamily\strut  Additionally}}\hskip 0pt plus 1pt\relax\allowbreak{}\colorbox[HTML]{FFFFFF}{\textcolor{black}{\tiny\ttfamily\strut ,}}\hskip 0pt plus 1pt\relax\allowbreak{}\colorbox[HTML]{6A6AFF}{\textcolor{white}{\tiny\ttfamily\strut  non}}\hskip 0pt plus 1pt\relax\allowbreak{}\colorbox[HTML]{FFFBFB}{\textcolor{black}{\tiny\ttfamily\strut -ver}}\hskip 0pt plus 1pt\relax\allowbreak{}\colorbox[HTML]{FFFFFF}{\textcolor{black}{\tiny\ttfamily\strut bal}}\hskip 0pt plus 1pt\relax\allowbreak{}\colorbox[HTML]{FFF9F9}{\textcolor{black}{\tiny\ttfamily\strut  communication}}\hskip 0pt plus 1pt\relax\allowbreak{}\colorbox[HTML]{FFFDFD}{\textcolor{black}{\tiny\ttfamily\strut  can}}\hskip 0pt plus 1pt\relax\allowbreak{}\colorbox[HTML]{FBFBFF}{\textcolor{black}{\tiny\ttfamily\strut  also}}\hskip 0pt plus 1pt\relax\allowbreak{}\colorbox[HTML]{F2F2FF}{\textcolor{black}{\tiny\ttfamily\strut  be}}\hskip 0pt plus 1pt\relax\allowbreak{}\colorbox[HTML]{FFFFFF}{\textcolor{black}{\tiny\ttfamily\strut  a}}\hskip 0pt plus 1pt\relax\allowbreak{}\colorbox[HTML]{FCFCFF}{\textcolor{black}{\tiny\ttfamily\strut  barrier}}\hskip 0pt plus 1pt\relax\allowbreak{}\colorbox[HTML]{DADAFF}{\textcolor{black}{\tiny\ttfamily\strut  in}}\hskip 0pt plus 1pt\relax\allowbreak{}\colorbox[HTML]{E8E8FF}{\textcolor{black}{\tiny\ttfamily\strut  the}}\hskip 0pt plus 1pt\relax\allowbreak{}\colorbox[HTML]{FBFBFF}{\textcolor{black}{\tiny\ttfamily\strut  workplace}}\hskip 0pt plus 1pt\relax\allowbreak{}\colorbox[HTML]{FFFEFE}{\textcolor{black}{\tiny\ttfamily\strut .}}\hskip 0pt plus 1pt\relax\allowbreak{}\colorbox[HTML]{D4D4FF}{\textcolor{black}{\tiny\ttfamily\strut  This}}\hskip 0pt plus 1pt\relax\allowbreak{}\colorbox[HTML]{F3F3FF}{\textcolor{black}{\tiny\ttfamily\strut  can}}\hskip 0pt plus 1pt\relax\allowbreak{}\colorbox[HTML]{FCFCFF}{\textcolor{black}{\tiny\ttfamily\strut  include}}\hskip 0pt plus 1pt\relax\allowbreak{}\colorbox[HTML]{FFF1F1}{\textcolor{black}{\tiny\ttfamily\strut  poor}}\hskip 0pt plus 1pt\relax\allowbreak{}\colorbox[HTML]{F9F9FF}{\textcolor{black}{\tiny\ttfamily\strut  body}}\hskip 0pt plus 1pt\relax\allowbreak{}\colorbox[HTML]{FFFFFF}{\textcolor{black}{\tiny\ttfamily\strut  language}}\hskip 0pt plus 1pt\relax\allowbreak{}\colorbox[HTML]{FDFDFF}{\textcolor{black}{\tiny\ttfamily\strut ,}}\hskip 0pt plus 1pt\relax\allowbreak{}\colorbox[HTML]{FEFEFF}{\textcolor{black}{\tiny\ttfamily\strut  lack}}\hskip 0pt plus 1pt\relax\allowbreak{}\colorbox[HTML]{FFFFFF}{\textcolor{black}{\tiny\ttfamily\strut  of}}\hskip 0pt plus 1pt\relax\allowbreak{}\colorbox[HTML]{FFFFFF}{\textcolor{black}{\tiny\ttfamily\strut  eye}}\hskip 0pt plus 1pt\relax\allowbreak{}\colorbox[HTML]{FFFFFF}{\textcolor{black}{\tiny\ttfamily\strut  contact}}\hskip 0pt plus 1pt\relax\allowbreak{}\colorbox[HTML]{FFFFFF}{\textcolor{black}{\tiny\ttfamily\strut ,}}\hskip 0pt plus 1pt\relax\allowbreak{}\colorbox[HTML]{FFFBFB}{\textcolor{black}{\tiny\ttfamily\strut  or}}\hskip 0pt plus 1pt\relax\allowbreak{}\colorbox[HTML]{FFFDFD}{\textcolor{black}{\tiny\ttfamily\strut  inappropriate}}\hskip 0pt plus 1pt\relax\allowbreak{}\colorbox[HTML]{FFF8F8}{\textcolor{black}{\tiny\ttfamily\strut  gestures}}\hskip 0pt plus 1pt\relax\allowbreak{}\colorbox[HTML]{EFEFFF}{\textcolor{black}{\tiny\ttfamily\strut .}}\hskip 0pt plus 1pt\relax\allowbreak{}\colorbox[HTML]{ECECFF}{\textcolor{black}{\tiny\ttfamily\strut  To}}\hskip 0pt plus 1pt\relax\allowbreak{}\colorbox[HTML]{ECECFF}{\textcolor{black}{\tiny\ttfamily\strut  overcome}}\hskip 0pt plus 1pt\relax\allowbreak{}\colorbox[HTML]{E7E7FF}{\textcolor{black}{\tiny\ttfamily\strut  this}}\hskip 0pt plus 1pt\relax\allowbreak{}\colorbox[HTML]{D7D7FF}{\textcolor{black}{\tiny\ttfamily\strut  barrier}}\hskip 0pt plus 1pt\relax\allowbreak{}\colorbox[HTML]{FFFFFF}{\textcolor{black}{\tiny\ttfamily\strut ,}}\hskip 0pt plus 1pt\relax\allowbreak{}\colorbox[HTML]{E3E3FF}{\textcolor{black}{\tiny\ttfamily\strut  employees}}\hskip 0pt plus 1pt\relax\allowbreak{}\colorbox[HTML]{FFF7F7}{\textcolor{black}{\tiny\ttfamily\strut  should}}\hskip 0pt plus 1pt\relax\allowbreak{}\colorbox[HTML]{FEFEFF}{\textcolor{black}{\tiny\ttfamily\strut  be}}\hskip 0pt plus 1pt\relax\allowbreak{}\colorbox[HTML]{E3E3FF}{\textcolor{black}{\tiny\ttfamily\strut  trained}}\hskip 0pt plus 1pt\relax\allowbreak{}\colorbox[HTML]{F8F8FF}{\textcolor{black}{\tiny\ttfamily\strut  on}}\hskip 0pt plus 1pt\relax\allowbreak{}\colorbox[HTML]{FFFFFF}{\textcolor{black}{\tiny\ttfamily\strut  effective}}\hskip 0pt plus 1pt\relax\allowbreak{}\colorbox[HTML]{FEFEFF}{\textcolor{black}{\tiny\ttfamily\strut  non}}\hskip 0pt plus 1pt\relax\allowbreak{}\colorbox[HTML]{FFFFFF}{\textcolor{black}{\tiny\ttfamily\strut -ver}}\hskip 0pt plus 1pt\relax\allowbreak{}\colorbox[HTML]{FFFFFF}{\textcolor{black}{\tiny\ttfamily\strut bal}}\hskip 0pt plus 1pt\relax\allowbreak{}\colorbox[HTML]{FFFFFF}{\textcolor{black}{\tiny\ttfamily\strut  communication}}\hskip 0pt plus 1pt\relax\allowbreak{}\colorbox[HTML]{FFFBFB}{\textcolor{black}{\tiny\ttfamily\strut  techniques}}\hskip 0pt plus 1pt\relax\allowbreak{}\colorbox[HTML]{FFFEFE}{\textcolor{black}{\tiny\ttfamily\strut ,}}\hskip 0pt plus 1pt\relax\allowbreak{}\colorbox[HTML]{FBFBFF}{\textcolor{black}{\tiny\ttfamily\strut  such}}\hskip 0pt plus 1pt\relax\allowbreak{}\colorbox[HTML]{FFFFFF}{\textcolor{black}{\tiny\ttfamily\strut  as}}\hskip 0pt plus 1pt\relax\allowbreak{}\colorbox[HTML]{FFFEFE}{\textcolor{black}{\tiny\ttfamily\strut  maintaining}}\hskip 0pt plus 1pt\relax\allowbreak{}\colorbox[HTML]{F9F9FF}{\textcolor{black}{\tiny\ttfamily\strut  eye}}\hskip 0pt plus 1pt\relax\allowbreak{}\colorbox[HTML]{FFFFFF}{\textcolor{black}{\tiny\ttfamily\strut  contact}}\hskip 0pt plus 1pt\relax\allowbreak{}\colorbox[HTML]{ECECFF}{\textcolor{black}{\tiny\ttfamily\strut ,}}\hskip 0pt plus 1pt\relax\allowbreak{}\colorbox[HTML]{FEFEFF}{\textcolor{black}{\tiny\ttfamily\strut  using}}\hskip 0pt plus 1pt\relax\allowbreak{}\colorbox[HTML]{E7E7FF}{\textcolor{black}{\tiny\ttfamily\strut  appropriate}}\hskip 0pt plus 1pt\relax\allowbreak{}\colorbox[HTML]{F8F8FF}{\textcolor{black}{\tiny\ttfamily\strut  body}}\hskip 0pt plus 1pt\relax\allowbreak{}\colorbox[HTML]{FFFFFF}{\textcolor{black}{\tiny\ttfamily\strut  language}}\hskip 0pt plus 1pt\relax\allowbreak{}\colorbox[HTML]{FFFFFF}{\textcolor{black}{\tiny\ttfamily\strut ,}}\hskip 0pt plus 1pt\relax\allowbreak{}\colorbox[HTML]{FFFFFF}{\textcolor{black}{\tiny\ttfamily\strut  and}}\hskip 0pt plus 1pt\relax\allowbreak{}\colorbox[HTML]{D3D3FF}{\textcolor{black}{\tiny\ttfamily\strut  using}}\hskip 0pt plus 1pt\relax\allowbreak{}\colorbox[HTML]{CDCDFF}{\textcolor{black}{\tiny\ttfamily\strut  appropriate}}\hskip 0pt plus 1pt\relax\allowbreak{}\colorbox[HTML]{FDFDFF}{\textcolor{black}{\tiny\ttfamily\strut  gestures}}\hskip 0pt plus 1pt\relax\allowbreak{}\colorbox[HTML]{E5E5FF}{\textcolor{black}{\tiny\ttfamily\strut .\textcolor{gray}{$\hookleftarrow$}\textcolor{gray}{$\hookleftarrow$}}}\hskip 0pt plus 1pt\relax\allowbreak{}\colorbox[HTML]{FFA8A8}{\textcolor{black}{\tiny\ttfamily\strut Post}}\hskip 0pt plus 1pt\relax\allowbreak{}\colorbox[HTML]{FFF8F8}{\textcolor{black}{\tiny\ttfamily\strut script}}\hskip 0pt plus 1pt\relax\allowbreak{}\colorbox[HTML]{E3E3FF}{\textcolor{black}{\tiny\ttfamily\strut :\textcolor{gray}{$\hookleftarrow$}}}\hskip 0pt plus 1pt\relax\allowbreak{}\colorbox[HTML]{F7F7FF}{\textcolor{black}{\tiny\ttfamily\strut In}}\hskip 0pt plus 1pt\relax\allowbreak{}\colorbox[HTML]{D5D5FF}{\textcolor{black}{\tiny\ttfamily\strut  conclusion}}\hskip 0pt plus 1pt\relax\allowbreak{}\colorbox[HTML]{FFFFFF}{\textcolor{black}{\tiny\ttfamily\strut ,}}\hskip 0pt plus 1pt\relax\allowbreak{}\colorbox[HTML]{8787FF}{\textcolor{black}{\tiny\ttfamily\strut  communication}}\hskip 0pt plus 1pt\relax\allowbreak{}\colorbox[HTML]{FDFDFF}{\textcolor{black}{\tiny\ttfamily\strut  barriers}}\hskip 0pt plus 1pt\relax\allowbreak{}\colorbox[HTML]{D4D4FF}{\textcolor{black}{\tiny\ttfamily\strut  in}}\hskip 0pt plus 1pt\relax\allowbreak{}\colorbox[HTML]{FBFBFF}{\textcolor{black}{\tiny\ttfamily\strut  the}}\hskip 0pt plus 1pt\relax\allowbreak{}\colorbox[HTML]{FEFEFF}{\textcolor{black}{\tiny\ttfamily\strut  workplace}}\hskip 0pt plus 1pt\relax\allowbreak{}\colorbox[HTML]{F4F4FF}{\textcolor{black}{\tiny\ttfamily\strut  can}}\hskip 0pt plus 1pt\relax\allowbreak{}\colorbox[HTML]{FFE1E1}{\textcolor{black}{\tiny\ttfamily\strut  significantly}}\hskip 0pt plus 1pt\relax\allowbreak{}\colorbox[HTML]{F3F3FF}{\textcolor{black}{\tiny\ttfamily\strut  impact}}\hskip 0pt plus 1pt\relax\allowbreak{}\colorbox[HTML]{FFFFFF}{\textcolor{black}{\tiny\ttfamily\strut  productivity}}\hskip 0pt plus 1pt\relax\allowbreak{}\colorbox[HTML]{FFF5F5}{\textcolor{black}{\tiny\ttfamily\strut  and}}\hskip 0pt plus 1pt\relax\allowbreak{}\colorbox[HTML]{F8F8FF}{\textcolor{black}{\tiny\ttfamily\strut  collaboration}}\hskip 0pt plus 1pt\relax\allowbreak{}\colorbox[HTML]{FEFEFF}{\textcolor{black}{\tiny\ttfamily\strut .}}\hskip 0pt plus 1pt\relax\allowbreak{}\colorbox[HTML]{C4C4FF}{\textcolor{black}{\tiny\ttfamily\strut  To}}\hskip 0pt plus 1pt\relax\allowbreak{}\colorbox[HTML]{BABAFF}{\textcolor{black}{\tiny\ttfamily\strut  overcome}}\hskip 0pt plus 1pt\relax\allowbreak{}\colorbox[HTML]{FDFDFF}{\textcolor{black}{\tiny\ttfamily\strut  these}}\hskip 0pt plus 1pt\relax\allowbreak{}\colorbox[HTML]{F8F8FF}{\textcolor{black}{\tiny\ttfamily\strut  barriers}}\hskip 0pt plus 1pt\relax\allowbreak{}\colorbox[HTML]{FFFFFF}{\textcolor{black}{\tiny\ttfamily\strut ,}}\hskip 0pt plus 1pt\relax\allowbreak{}\colorbox[HTML]{B3B3FF}{\textcolor{black}{\tiny\ttfamily\strut  organizations}}\hskip 0pt plus 1pt\relax\allowbreak{}\colorbox[HTML]{DFDFFF}{\textcolor{black}{\tiny\ttfamily\strut  should}}\hskip 0pt plus 1pt\relax\allowbreak{}\colorbox[HTML]{FAFAFF}{\textcolor{black}{\tiny\ttfamily\strut  prioritize}}\hskip 0pt plus 1pt\relax\allowbreak{}\colorbox[HTML]{F7F7FF}{\textcolor{black}{\tiny\ttfamily\strut  creating}}\hskip 0pt plus 1pt\relax\allowbreak{}\colorbox[HTML]{FCFCFF}{\textcolor{black}{\tiny\ttfamily\strut  a}}\hskip 0pt plus 1pt\relax\allowbreak{}\colorbox[HTML]{E6E6FF}{\textcolor{black}{\tiny\ttfamily\strut  diverse}}\hskip 0pt plus 1pt\relax\allowbreak{}\colorbox[HTML]{FBFBFF}{\textcolor{black}{\tiny\ttfamily\strut  and}}\hskip 0pt plus 1pt\relax\allowbreak{}\colorbox[HTML]{FFFFFF}{\textcolor{black}{\tiny\ttfamily\strut  inclusive}}\hskip 0pt plus 1pt\relax\allowbreak{}\colorbox[HTML]{9B9BFF}{\textcolor{black}{\tiny\ttfamily\strut  workplace}}\hskip 0pt plus 1pt\relax\allowbreak{}\colorbox[HTML]{F5F5FF}{\textcolor{black}{\tiny\ttfamily\strut  culture}}\hskip 0pt plus 1pt\relax\allowbreak{}\colorbox[HTML]{FCFCFF}{\textcolor{black}{\tiny\ttfamily\strut ,}}\hskip 0pt plus 1pt\relax\allowbreak{}\colorbox[HTML]{D6D6FF}{\textcolor{black}{\tiny\ttfamily\strut  establishing}}\hskip 0pt plus 1pt\relax\allowbreak{}\colorbox[HTML]{FFFFFF}{\textcolor{black}{\tiny\ttfamily\strut  clear}}\hskip 0pt plus 1pt\relax\allowbreak{}\colorbox[HTML]{FEFEFF}{\textcolor{black}{\tiny\ttfamily\strut  communication}}\hskip 0pt plus 1pt\relax\allowbreak{}\colorbox[HTML]{FEFEFF}{\textcolor{black}{\tiny\ttfamily\strut  guidelines}}\hskip 0pt plus 1pt\relax\allowbreak{}\colorbox[HTML]{FFFFFF}{\textcolor{black}{\tiny\ttfamily\strut ,}}\hskip 0pt plus 1pt\relax\allowbreak{}\colorbox[HTML]{FFF7F7}{\textcolor{black}{\tiny\ttfamily\strut  and}}\hskip 0pt plus 1pt\relax\allowbreak{}\colorbox[HTML]{FDFDFF}{\textcolor{black}{\tiny\ttfamily\strut  providing}}\hskip 0pt plus 1pt\relax\allowbreak{}\colorbox[HTML]{F5F5FF}{\textcolor{black}{\tiny\ttfamily\strut  training}}\hskip 0pt plus 1pt\relax\allowbreak{}\colorbox[HTML]{FFEAEA}{\textcolor{black}{\tiny\ttfamily\strut  on}}\hskip 0pt plus 1pt\relax\allowbreak{}\colorbox[HTML]{FBFBFF}{\textcolor{black}{\tiny\ttfamily\strut  effective}}\hskip 0pt plus 1pt\relax\allowbreak{}\colorbox[HTML]{FFFFFF}{\textcolor{black}{\tiny\ttfamily\strut  communication}}\hskip 0pt plus 1pt\relax\allowbreak{}\colorbox[HTML]{F7F7FF}{\textcolor{black}{\tiny\ttfamily\strut  techniques}}\hskip 0pt plus 1pt\relax\allowbreak{}\colorbox[HTML]{FFFFFF}{\textcolor{black}{\tiny\ttfamily\strut .}}\hskip 0pt plus 1pt\relax\allowbreak{}\colorbox[HTML]{F4F4FF}{\textcolor{black}{\tiny\ttfamily\strut  By}}\hskip 0pt plus 1pt\relax\allowbreak{}\colorbox[HTML]{FFF4F4}{\textcolor{black}{\tiny\ttfamily\strut  addressing}}\hskip 0pt plus 1pt\relax\allowbreak{}\colorbox[HTML]{F4F4FF}{\textcolor{black}{\tiny\ttfamily\strut  these}}\hskip 0pt plus 1pt\relax\allowbreak{}\colorbox[HTML]{F5F5FF}{\textcolor{black}{\tiny\ttfamily\strut  barriers}}\hskip 0pt plus 1pt\relax\allowbreak{}\colorbox[HTML]{FDFDFF}{\textcolor{black}{\tiny\ttfamily\strut ,}}\hskip 0pt plus 1pt\relax\allowbreak{}\colorbox[HTML]{8E8EFF}{\textcolor{black}{\tiny\ttfamily\strut  organizations}}\hskip 0pt plus 1pt\relax\allowbreak{}\colorbox[HTML]{FFFFFF}{\textcolor{black}{\tiny\ttfamily\strut  can}}\hskip 0pt plus 1pt\relax\allowbreak{}\colorbox[HTML]{FFC0C0}{\textcolor{black}{\tiny\ttfamily\strut  enhance}}\hskip 0pt plus 1pt\relax\allowbreak{}\colorbox[HTML]{FFD9D9}{\textcolor{black}{\tiny\ttfamily\strut  their}}\hskip 0pt plus 1pt\relax\allowbreak{}\colorbox[HTML]{F0F0FF}{\textcolor{black}{\tiny\ttfamily\strut  communication}}\hskip 0pt plus 1pt\relax\allowbreak{}\colorbox[HTML]{E4E4FF}{\textcolor{black}{\tiny\ttfamily\strut  and}}\hskip 0pt plus 1pt\relax\allowbreak{}\colorbox[HTML]{FFD9D9}{\textcolor{black}{\tiny\ttfamily\strut  work}}\hskip 0pt plus 1pt\relax\allowbreak{}\colorbox[HTML]{FFDADA}{\textcolor{black}{\tiny\ttfamily\strut  efficiency}}\hskip 0pt plus 1pt\relax\allowbreak{}\colorbox[HTML]{E9E9FF}{\textcolor{black}{\tiny\ttfamily\strut ,}}\hskip 0pt plus 1pt\relax\allowbreak{}\colorbox[HTML]{FBFBFF}{\textcolor{black}{\tiny\ttfamily\strut  ultimately}}\hskip 0pt plus 1pt\relax\allowbreak{}\colorbox[HTML]{FEFEFF}{\textcolor{black}{\tiny\ttfamily\strut  leading}}\hskip 0pt plus 1pt\relax\allowbreak{}\colorbox[HTML]{FFFFFF}{\textcolor{black}{\tiny\ttfamily\strut  to}}\hskip 0pt plus 1pt\relax\allowbreak{}\colorbox[HTML]{FFFDFD}{\textcolor{black}{\tiny\ttfamily\strut  better}}\hskip 0pt plus 1pt\relax\allowbreak{}\colorbox[HTML]{FEFEFF}{\textcolor{black}{\tiny\ttfamily\strut  outcomes}}\hskip 0pt plus 1pt\relax\allowbreak{}\colorbox[HTML]{DBDBFF}{\textcolor{black}{\tiny\ttfamily\strut  for}}\hskip 0pt plus 1pt\relax\allowbreak{}\colorbox[HTML]{D8D8FF}{\textcolor{black}{\tiny\ttfamily\strut  their}}\hskip 0pt plus 1pt\relax\allowbreak{}\colorbox[HTML]{F2F2FF}{\textcolor{black}{\tiny\ttfamily\strut  employees}}\hskip 0pt plus 1pt\relax\allowbreak{}\colorbox[HTML]{FFFFFF}{\textcolor{black}{\tiny\ttfamily\strut  and}}\hskip 0pt plus 1pt\relax\allowbreak{}\colorbox[HTML]{FFE0E0}{\textcolor{black}{\tiny\ttfamily\strut  clients}}\hskip 0pt plus 1pt\relax\allowbreak{}\colorbox[HTML]{FDFDFF}{\textcolor{black}{\tiny\ttfamily\strut .}}\hskip 0pt plus 1pt\relax\allowbreak{}\colorbox[HTML]{FF4040}{\textcolor{white}{\tiny\ttfamily\strut \textless{}\textbar{}im\_end\textbar{}\textgreater{}}}\hskip 0pt plus 1pt\relax\allowbreak{}
\par\medskip
\noindent\textit{\small Per-constraint attribution snippets (Section A) show that each constraint's $\delta^i_t$ peaks on tokens semantically aligned with the constraint's content. Full-response coloring (Section B) shows that LOO sum and LA produce visually distinct attribution patterns despite both being 1$\times$T signals.}

\endgroup

\bigskip

\subsection{Case 3: Constraint identity at a shared salient token}
\label{app:heatmap:case3}

The third case shows how the $K \times T$ LOO matrix retains constraint identity at a token that is salient under both aggregate signals. The constraint set requires a Markdown table with no more than three rows, but the response contains no table. Ablating the table constraint produces its largest absolute shift on the token \texttt{These} at position $259$ ($\delta=-21.3$). Both the LOO sum and LA assign a strong shift to this token in Section~B. The per-constraint LOO view associates the shift with the table constraint, while LA provides the full-versus-empty aggregate.

\medskip

\begingroup
\noindent\textbf{\large Sample 9169} \quad $K{=}5$ \quad $T{=}308$ \quad LOO/LA top-20 Jaccard$=0.21$\par\nopagebreak\medskip\nopagebreak
\noindent\textbf{Section A: Per-constraint attribution snippets} ($\textcolor[HTML]{cc0000}{\blacksquare}{=}+\delta$, $\textcolor[HTML]{0000cc}{\blacksquare}{=}-\delta$, intensity $\propto |\delta|$, capped at $\pm8.0$)\par\nopagebreak\smallskip\nopagebreak
\setlength{\fboxsep}{0.6pt}
\setlength{\tabcolsep}{4pt}
\renewcommand{\arraystretch}{1.15}
\begin{tabular}{@{}p{0.10\textwidth}@{\hspace{4pt}}p{0.85\textwidth}@{}}
\toprule
\textbf{C1}\\[1pt]{\scriptsize $|\delta|_{\max}{=}7.0$} & \textit{\scriptsize ``When crafting your response, ensure that it does not exceed 300 words''}\\[2pt]
{\scriptsize\textcolor{gray}{\dots}}\colorbox[HTML]{F1F1FF}{\textcolor{black}{\scriptsize\ttfamily\strut  a}}\colorbox[HTML]{FCFCFF}{\textcolor{black}{\scriptsize\ttfamily\strut  significant}}\colorbox[HTML]{FFFAFA}{\textcolor{black}{\scriptsize\ttfamily\strut  contribution}}\colorbox[HTML]{FDFDFF}{\textcolor{black}{\scriptsize\ttfamily\strut  to}}\colorbox[HTML]{FDFDFF}{\textcolor{black}{\scriptsize\ttfamily\strut  the}}\colorbox[HTML]{FFFFFF}{\textcolor{black}{\scriptsize\ttfamily\strut  field}}\colorbox[HTML]{FFFFFF}{\textcolor{black}{\scriptsize\ttfamily\strut .}}\colorbox[HTML]{B9B9FF}{\textcolor{black}{\scriptsize\ttfamily\strut  His}}\colorbox[HTML]{FFFEFE}{\textcolor{black}{\scriptsize\ttfamily\strut  research}}\colorbox[HTML]{FFF4F4}{\textcolor{black}{\scriptsize\ttfamily\strut  emphasizes}}\colorbox[HTML]{FFFFFF}{\textcolor{black}{\scriptsize\ttfamily\strut  the}}\colorbox[HTML]{FFF5F5}{\textcolor{black}{\scriptsize\ttfamily\strut  concept}}\colorbox[HTML]{FFFFFF}{\textcolor{black}{\scriptsize\ttfamily\strut  of}}\colorbox[HTML]{FFFFFF}{\textcolor{black}{\scriptsize\ttfamily\strut  mutually}}\colorbox[HTML]{FFFFFF}{\textcolor{black}{\scriptsize\ttfamily\strut  assured}}{\scriptsize\textcolor{gray}{\dots}}\\[1pt]
{\scriptsize\textcolor{gray}{\dots}}\colorbox[HTML]{FFFFFF}{\textcolor{black}{\scriptsize\ttfamily\strut  reform}}\colorbox[HTML]{FFFFFF}{\textcolor{black}{\scriptsize\ttfamily\strut ,}}\colorbox[HTML]{FFFFFF}{\textcolor{black}{\scriptsize\ttfamily\strut  and}}\colorbox[HTML]{FFFFFF}{\textcolor{black}{\scriptsize\ttfamily\strut  international}}\colorbox[HTML]{FFFFFF}{\textcolor{black}{\scriptsize\ttfamily\strut  trade}}\colorbox[HTML]{FDFDFF}{\textcolor{black}{\scriptsize\ttfamily\strut  politics}}\colorbox[HTML]{C8C8FF}{\textcolor{black}{\scriptsize\ttfamily\strut .}}\colorbox[HTML]{5757FF}{\textcolor{white}{\scriptsize\ttfamily\strut  Their}}\colorbox[HTML]{FAFAFF}{\textcolor{black}{\scriptsize\ttfamily\strut  research}}\colorbox[HTML]{EFEFFF}{\textcolor{black}{\scriptsize\ttfamily\strut  underscores}}\colorbox[HTML]{FFFFFF}{\textcolor{black}{\scriptsize\ttfamily\strut  the}}\colorbox[HTML]{FEFEFF}{\textcolor{black}{\scriptsize\ttfamily\strut  importance}}\colorbox[HTML]{FFFFFF}{\textcolor{black}{\scriptsize\ttfamily\strut  of}}\colorbox[HTML]{FBFBFF}{\textcolor{black}{\scriptsize\ttfamily\strut  understanding}}\colorbox[HTML]{F0F0FF}{\textcolor{black}{\scriptsize\ttfamily\strut  the}}{\scriptsize\textcolor{gray}{\dots}}\\[1pt]\\
\addlinespace[2pt]
\textbf{C3}\\[1pt]{\scriptsize $|\delta|_{\max}{=}8.8$} & \textit{\scriptsize ``Additionally, the response should be written in English with each word capitalized.''}\\[2pt]
{\scriptsize\textcolor{gray}{\dots}}\colorbox[HTML]{FFEFEF}{\textcolor{black}{\scriptsize\ttfamily\strut \textcolor{gray}{$\hookleftarrow$}\textcolor{gray}{$\hookleftarrow$}}}\colorbox[HTML]{A7A7FF}{\textcolor{black}{\scriptsize\ttfamily\strut Dr}}\colorbox[HTML]{FFFFFF}{\textcolor{black}{\scriptsize\ttfamily\strut .}}\colorbox[HTML]{CDCDFF}{\textcolor{black}{\scriptsize\ttfamily\strut  Don}}\colorbox[HTML]{FEFEFF}{\textcolor{black}{\scriptsize\ttfamily\strut  Cas}}\colorbox[HTML]{FFFFFF}{\textcolor{black}{\scriptsize\ttfamily\strut ler}}\colorbox[HTML]{A2A2FF}{\textcolor{black}{\scriptsize\ttfamily\strut 's}}\colorbox[HTML]{6E6EFF}{\textcolor{white}{\scriptsize\ttfamily\strut  work}}\colorbox[HTML]{A3A3FF}{\textcolor{black}{\scriptsize\ttfamily\strut  in}}\colorbox[HTML]{FFEEEE}{\textcolor{black}{\scriptsize\ttfamily\strut  nuclear}}\colorbox[HTML]{FFFFFF}{\textcolor{black}{\scriptsize\ttfamily\strut  disarm}}\colorbox[HTML]{FFFFFF}{\textcolor{black}{\scriptsize\ttfamily\strut ament}}\colorbox[HTML]{FDFDFF}{\textcolor{black}{\scriptsize\ttfamily\strut  is}}\colorbox[HTML]{FFF7F7}{\textcolor{black}{\scriptsize\ttfamily\strut  fascinating}}\colorbox[HTML]{FAFAFF}{\textcolor{black}{\scriptsize\ttfamily\strut  and}}{\scriptsize\textcolor{gray}{\dots}}\\[1pt]
{\scriptsize\textcolor{gray}{\dots}}\colorbox[HTML]{FFFFFF}{\textcolor{black}{\scriptsize\ttfamily\strut  War}}\colorbox[HTML]{FFFFFF}{\textcolor{black}{\scriptsize\ttfamily\strut .\textcolor{gray}{$\hookleftarrow$}\textcolor{gray}{$\hookleftarrow$}}}\colorbox[HTML]{6161FF}{\textcolor{white}{\scriptsize\ttfamily\strut On}}\colorbox[HTML]{BCBCFF}{\textcolor{black}{\scriptsize\ttfamily\strut  the}}\colorbox[HTML]{FBFBFF}{\textcolor{black}{\scriptsize\ttfamily\strut  other}}\colorbox[HTML]{FFFFFF}{\textcolor{black}{\scriptsize\ttfamily\strut  hand}}\colorbox[HTML]{FFFFFF}{\textcolor{black}{\scriptsize\ttfamily\strut ,}}\colorbox[HTML]{4040FF}{\textcolor{white}{\scriptsize\ttfamily\strut  exploring}}\colorbox[HTML]{FEFEFF}{\textcolor{black}{\scriptsize\ttfamily\strut  other}}\colorbox[HTML]{FFFDFD}{\textcolor{black}{\scriptsize\ttfamily\strut  professors}}\colorbox[HTML]{FFF9F9}{\textcolor{black}{\scriptsize\ttfamily\strut '}}\colorbox[HTML]{FFFEFE}{\textcolor{black}{\scriptsize\ttfamily\strut  work}}\colorbox[HTML]{F8F8FF}{\textcolor{black}{\scriptsize\ttfamily\strut  in}}\colorbox[HTML]{FCFCFF}{\textcolor{black}{\scriptsize\ttfamily\strut  the}}\colorbox[HTML]{FFFDFD}{\textcolor{black}{\scriptsize\ttfamily\strut  areas}}{\scriptsize\textcolor{gray}{\dots}}\\[1pt]\\
\addlinespace[2pt]
\textbf{C5}\\[1pt]{\scriptsize $|\delta|_{\max}{=}21.3$} & \textit{\scriptsize ``and a table with no more than 3 rows, clearly integrating these elements into your answer.''}\\[2pt]
{\scriptsize\textcolor{gray}{\dots}}\colorbox[HTML]{FFFFFF}{\textcolor{black}{\scriptsize\ttfamily\strut  States}}\colorbox[HTML]{FFFFFF}{\textcolor{black}{\scriptsize\ttfamily\strut  in}}\colorbox[HTML]{FBFBFF}{\textcolor{black}{\scriptsize\ttfamily\strut  shaping}}\colorbox[HTML]{FDFDFF}{\textcolor{black}{\scriptsize\ttfamily\strut  global}}\colorbox[HTML]{FCFCFF}{\textcolor{black}{\scriptsize\ttfamily\strut  economic}}\colorbox[HTML]{FFFFFF}{\textcolor{black}{\scriptsize\ttfamily\strut  policies}}\colorbox[HTML]{FFF9F9}{\textcolor{black}{\scriptsize\ttfamily\strut .\textcolor{gray}{$\hookleftarrow$}\textcolor{gray}{$\hookleftarrow$}}}\colorbox[HTML]{4040FF}{\textcolor{white}{\scriptsize\ttfamily\strut These}}\colorbox[HTML]{E9E9FF}{\textcolor{black}{\scriptsize\ttfamily\strut  professors}}\colorbox[HTML]{FFC0C0}{\textcolor{black}{\scriptsize\ttfamily\strut ,}}\colorbox[HTML]{E2E2FF}{\textcolor{black}{\scriptsize\ttfamily\strut  along}}\colorbox[HTML]{FFFFFF}{\textcolor{black}{\scriptsize\ttfamily\strut  with}}\colorbox[HTML]{FFFFFF}{\textcolor{black}{\scriptsize\ttfamily\strut  Dr}}\colorbox[HTML]{FFFFFF}{\textcolor{black}{\scriptsize\ttfamily\strut .}}\colorbox[HTML]{FDFDFF}{\textcolor{black}{\scriptsize\ttfamily\strut  Cas}}{\scriptsize\textcolor{gray}{\dots}}\\[1pt]
{\scriptsize\textcolor{gray}{\dots}}\colorbox[HTML]{FFFFFF}{\textcolor{black}{\scriptsize\ttfamily\strut  to}}\colorbox[HTML]{FDFDFF}{\textcolor{black}{\scriptsize\ttfamily\strut  inform}}\colorbox[HTML]{FFFCFC}{\textcolor{black}{\scriptsize\ttfamily\strut  future}}\colorbox[HTML]{FFFFFF}{\textcolor{black}{\scriptsize\ttfamily\strut  policies}}\colorbox[HTML]{EFEFFF}{\textcolor{black}{\scriptsize\ttfamily\strut  and}}\colorbox[HTML]{F8F8FF}{\textcolor{black}{\scriptsize\ttfamily\strut  actions}}\colorbox[HTML]{6C6CFF}{\textcolor{white}{\scriptsize\ttfamily\strut .}}\colorbox[HTML]{4040FF}{\textcolor{white}{\scriptsize\ttfamily\strut \textless{}\textbar{}im\_end\textbar{}\textgreater{}}}\\[1pt]\\
\addlinespace[2pt]
\bottomrule
\end{tabular}
\par\medskip
\noindent\textbf{Section B: Full response} (same tokens shown twice, colored once by LOO sum, once by LA) $-$ vmax$=\pm8.0$\par\nopagebreak\smallskip\nopagebreak
\noindent\textbf{\textcolor[HTML]{c41e3a}{LOO sum}} {\scriptsize $|\delta|_{\max}{=}25.1$}: \\[2pt]
\sloppy
\colorbox[HTML]{FFA0A0}{\textcolor{black}{\tiny\ttfamily\strut \#}}\hskip 0pt plus 1pt\relax\allowbreak{}\colorbox[HTML]{FCFCFF}{\textcolor{black}{\tiny\ttfamily\strut  Nuclear}}\hskip 0pt plus 1pt\relax\allowbreak{}\colorbox[HTML]{FFF8F8}{\textcolor{black}{\tiny\ttfamily\strut  Dis}}\hskip 0pt plus 1pt\relax\allowbreak{}\colorbox[HTML]{FFFFFF}{\textcolor{black}{\tiny\ttfamily\strut arm}}\hskip 0pt plus 1pt\relax\allowbreak{}\colorbox[HTML]{FFFFFF}{\textcolor{black}{\tiny\ttfamily\strut ament}}\hskip 0pt plus 1pt\relax\allowbreak{}\colorbox[HTML]{DFDFFF}{\textcolor{black}{\tiny\ttfamily\strut  Research}}\hskip 0pt plus 1pt\relax\allowbreak{}\colorbox[HTML]{FFBDBD}{\textcolor{black}{\tiny\ttfamily\strut \textcolor{gray}{$\hookleftarrow$}\textcolor{gray}{$\hookleftarrow$}}}\hskip 0pt plus 1pt\relax\allowbreak{}\colorbox[HTML]{9393FF}{\textcolor{black}{\tiny\ttfamily\strut Dr}}\hskip 0pt plus 1pt\relax\allowbreak{}\colorbox[HTML]{FFFFFF}{\textcolor{black}{\tiny\ttfamily\strut .}}\hskip 0pt plus 1pt\relax\allowbreak{}\colorbox[HTML]{FFFDFD}{\textcolor{black}{\tiny\ttfamily\strut  Don}}\hskip 0pt plus 1pt\relax\allowbreak{}\colorbox[HTML]{FF8585}{\textcolor{black}{\tiny\ttfamily\strut  Cas}}\hskip 0pt plus 1pt\relax\allowbreak{}\colorbox[HTML]{FFFFFF}{\textcolor{black}{\tiny\ttfamily\strut ler}}\hskip 0pt plus 1pt\relax\allowbreak{}\colorbox[HTML]{B6B6FF}{\textcolor{black}{\tiny\ttfamily\strut 's}}\hskip 0pt plus 1pt\relax\allowbreak{}\colorbox[HTML]{C1C1FF}{\textcolor{black}{\tiny\ttfamily\strut  work}}\hskip 0pt plus 1pt\relax\allowbreak{}\colorbox[HTML]{F7F7FF}{\textcolor{black}{\tiny\ttfamily\strut  in}}\hskip 0pt plus 1pt\relax\allowbreak{}\colorbox[HTML]{FFC1C1}{\textcolor{black}{\tiny\ttfamily\strut  nuclear}}\hskip 0pt plus 1pt\relax\allowbreak{}\colorbox[HTML]{FFFFFF}{\textcolor{black}{\tiny\ttfamily\strut  disarm}}\hskip 0pt plus 1pt\relax\allowbreak{}\colorbox[HTML]{FFFFFF}{\textcolor{black}{\tiny\ttfamily\strut ament}}\hskip 0pt plus 1pt\relax\allowbreak{}\colorbox[HTML]{FFFEFE}{\textcolor{black}{\tiny\ttfamily\strut  is}}\hskip 0pt plus 1pt\relax\allowbreak{}\colorbox[HTML]{FFCFCF}{\textcolor{black}{\tiny\ttfamily\strut  fascinating}}\hskip 0pt plus 1pt\relax\allowbreak{}\colorbox[HTML]{FFF7F7}{\textcolor{black}{\tiny\ttfamily\strut  and}}\hskip 0pt plus 1pt\relax\allowbreak{}\colorbox[HTML]{DADAFF}{\textcolor{black}{\tiny\ttfamily\strut  has}}\hskip 0pt plus 1pt\relax\allowbreak{}\colorbox[HTML]{FCFCFF}{\textcolor{black}{\tiny\ttfamily\strut  been}}\hskip 0pt plus 1pt\relax\allowbreak{}\colorbox[HTML]{FCFCFF}{\textcolor{black}{\tiny\ttfamily\strut  a}}\hskip 0pt plus 1pt\relax\allowbreak{}\colorbox[HTML]{FAFAFF}{\textcolor{black}{\tiny\ttfamily\strut  significant}}\hskip 0pt plus 1pt\relax\allowbreak{}\colorbox[HTML]{FFF0F0}{\textcolor{black}{\tiny\ttfamily\strut  contribution}}\hskip 0pt plus 1pt\relax\allowbreak{}\colorbox[HTML]{FCFCFF}{\textcolor{black}{\tiny\ttfamily\strut  to}}\hskip 0pt plus 1pt\relax\allowbreak{}\colorbox[HTML]{FEFEFF}{\textcolor{black}{\tiny\ttfamily\strut  the}}\hskip 0pt plus 1pt\relax\allowbreak{}\colorbox[HTML]{FFFEFE}{\textcolor{black}{\tiny\ttfamily\strut  field}}\hskip 0pt plus 1pt\relax\allowbreak{}\colorbox[HTML]{FFFFFF}{\textcolor{black}{\tiny\ttfamily\strut .}}\hskip 0pt plus 1pt\relax\allowbreak{}\colorbox[HTML]{5B5BFF}{\textcolor{white}{\tiny\ttfamily\strut  His}}\hskip 0pt plus 1pt\relax\allowbreak{}\colorbox[HTML]{F8F8FF}{\textcolor{black}{\tiny\ttfamily\strut  research}}\hskip 0pt plus 1pt\relax\allowbreak{}\colorbox[HTML]{FFDFDF}{\textcolor{black}{\tiny\ttfamily\strut  emphasizes}}\hskip 0pt plus 1pt\relax\allowbreak{}\colorbox[HTML]{FFFFFF}{\textcolor{black}{\tiny\ttfamily\strut  the}}\hskip 0pt plus 1pt\relax\allowbreak{}\colorbox[HTML]{FF9797}{\textcolor{black}{\tiny\ttfamily\strut  concept}}\hskip 0pt plus 1pt\relax\allowbreak{}\colorbox[HTML]{FEFEFF}{\textcolor{black}{\tiny\ttfamily\strut  of}}\hskip 0pt plus 1pt\relax\allowbreak{}\colorbox[HTML]{FFFFFF}{\textcolor{black}{\tiny\ttfamily\strut  mutually}}\hskip 0pt plus 1pt\relax\allowbreak{}\colorbox[HTML]{FFFFFF}{\textcolor{black}{\tiny\ttfamily\strut  assured}}\hskip 0pt plus 1pt\relax\allowbreak{}\colorbox[HTML]{FFFFFF}{\textcolor{black}{\tiny\ttfamily\strut  destruction}}\hskip 0pt plus 1pt\relax\allowbreak{}\colorbox[HTML]{F2F2FF}{\textcolor{black}{\tiny\ttfamily\strut ,}}\hskip 0pt plus 1pt\relax\allowbreak{}\colorbox[HTML]{FDFDFF}{\textcolor{black}{\tiny\ttfamily\strut  which}}\hskip 0pt plus 1pt\relax\allowbreak{}\colorbox[HTML]{FFF1F1}{\textcolor{black}{\tiny\ttfamily\strut  serves}}\hskip 0pt plus 1pt\relax\allowbreak{}\colorbox[HTML]{FFFFFF}{\textcolor{black}{\tiny\ttfamily\strut  as}}\hskip 0pt plus 1pt\relax\allowbreak{}\colorbox[HTML]{FFFFFF}{\textcolor{black}{\tiny\ttfamily\strut  a}}\hskip 0pt plus 1pt\relax\allowbreak{}\colorbox[HTML]{E7E7FF}{\textcolor{black}{\tiny\ttfamily\strut  powerful}}\hskip 0pt plus 1pt\relax\allowbreak{}\colorbox[HTML]{FFFFFF}{\textcolor{black}{\tiny\ttfamily\strut  deterrent}}\hskip 0pt plus 1pt\relax\allowbreak{}\colorbox[HTML]{F2F2FF}{\textcolor{black}{\tiny\ttfamily\strut  against}}\hskip 0pt plus 1pt\relax\allowbreak{}\colorbox[HTML]{FFEDED}{\textcolor{black}{\tiny\ttfamily\strut  armed}}\hskip 0pt plus 1pt\relax\allowbreak{}\colorbox[HTML]{FFFCFC}{\textcolor{black}{\tiny\ttfamily\strut  conflict}}\hskip 0pt plus 1pt\relax\allowbreak{}\colorbox[HTML]{F6F6FF}{\textcolor{black}{\tiny\ttfamily\strut .}}\hskip 0pt plus 1pt\relax\allowbreak{}\colorbox[HTML]{A0A0FF}{\textcolor{black}{\tiny\ttfamily\strut  Cas}}\hskip 0pt plus 1pt\relax\allowbreak{}\colorbox[HTML]{FFFFFF}{\textcolor{black}{\tiny\ttfamily\strut ler}}\hskip 0pt plus 1pt\relax\allowbreak{}\colorbox[HTML]{FEFEFF}{\textcolor{black}{\tiny\ttfamily\strut 's}}\hskip 0pt plus 1pt\relax\allowbreak{}\colorbox[HTML]{FFB2B2}{\textcolor{black}{\tiny\ttfamily\strut  essay}}\hskip 0pt plus 1pt\relax\allowbreak{}\colorbox[HTML]{FFF7F7}{\textcolor{black}{\tiny\ttfamily\strut  in}}\hskip 0pt plus 1pt\relax\allowbreak{}\colorbox[HTML]{6D6DFF}{\textcolor{white}{\tiny\ttfamily\strut  AP}}\hskip 0pt plus 1pt\relax\allowbreak{}\colorbox[HTML]{FFE5E5}{\textcolor{black}{\tiny\ttfamily\strut  World}}\hskip 0pt plus 1pt\relax\allowbreak{}\colorbox[HTML]{FFFFFF}{\textcolor{black}{\tiny\ttfamily\strut  History}}\hskip 0pt plus 1pt\relax\allowbreak{}\colorbox[HTML]{E6E6FF}{\textcolor{black}{\tiny\ttfamily\strut  del}}\hskip 0pt plus 1pt\relax\allowbreak{}\colorbox[HTML]{FFF1F1}{\textcolor{black}{\tiny\ttfamily\strut ved}}\hskip 0pt plus 1pt\relax\allowbreak{}\colorbox[HTML]{FEFEFF}{\textcolor{black}{\tiny\ttfamily\strut  into}}\hskip 0pt plus 1pt\relax\allowbreak{}\colorbox[HTML]{D5D5FF}{\textcolor{black}{\tiny\ttfamily\strut  the}}\hskip 0pt plus 1pt\relax\allowbreak{}\colorbox[HTML]{9D9DFF}{\textcolor{black}{\tiny\ttfamily\strut  Cold}}\hskip 0pt plus 1pt\relax\allowbreak{}\colorbox[HTML]{FFFFFF}{\textcolor{black}{\tiny\ttfamily\strut  War}}\hskip 0pt plus 1pt\relax\allowbreak{}\colorbox[HTML]{FFF5F5}{\textcolor{black}{\tiny\ttfamily\strut  era}}\hskip 0pt plus 1pt\relax\allowbreak{}\colorbox[HTML]{F8F8FF}{\textcolor{black}{\tiny\ttfamily\strut ,}}\hskip 0pt plus 1pt\relax\allowbreak{}\colorbox[HTML]{F0F0FF}{\textcolor{black}{\tiny\ttfamily\strut  where}}\hskip 0pt plus 1pt\relax\allowbreak{}\colorbox[HTML]{D2D2FF}{\textcolor{black}{\tiny\ttfamily\strut  the}}\hskip 0pt plus 1pt\relax\allowbreak{}\colorbox[HTML]{F8F8FF}{\textcolor{black}{\tiny\ttfamily\strut  fear}}\hskip 0pt plus 1pt\relax\allowbreak{}\colorbox[HTML]{FFFFFF}{\textcolor{black}{\tiny\ttfamily\strut  of}}\hskip 0pt plus 1pt\relax\allowbreak{}\colorbox[HTML]{F6F6FF}{\textcolor{black}{\tiny\ttfamily\strut  nuclear}}\hskip 0pt plus 1pt\relax\allowbreak{}\colorbox[HTML]{FAFAFF}{\textcolor{black}{\tiny\ttfamily\strut  ann}}\hskip 0pt plus 1pt\relax\allowbreak{}\colorbox[HTML]{FFFFFF}{\textcolor{black}{\tiny\ttfamily\strut ihilation}}\hskip 0pt plus 1pt\relax\allowbreak{}\colorbox[HTML]{EAEAFF}{\textcolor{black}{\tiny\ttfamily\strut  shaped}}\hskip 0pt plus 1pt\relax\allowbreak{}\colorbox[HTML]{F5F5FF}{\textcolor{black}{\tiny\ttfamily\strut  global}}\hskip 0pt plus 1pt\relax\allowbreak{}\colorbox[HTML]{FEFEFF}{\textcolor{black}{\tiny\ttfamily\strut  politics}}\hskip 0pt plus 1pt\relax\allowbreak{}\colorbox[HTML]{EBEBFF}{\textcolor{black}{\tiny\ttfamily\strut .}}\hskip 0pt plus 1pt\relax\allowbreak{}\colorbox[HTML]{8E8EFF}{\textcolor{black}{\tiny\ttfamily\strut  This}}\hskip 0pt plus 1pt\relax\allowbreak{}\colorbox[HTML]{EBEBFF}{\textcolor{black}{\tiny\ttfamily\strut  perspective}}\hskip 0pt plus 1pt\relax\allowbreak{}\colorbox[HTML]{D8D8FF}{\textcolor{black}{\tiny\ttfamily\strut  on}}\hskip 0pt plus 1pt\relax\allowbreak{}\colorbox[HTML]{FFEAEA}{\textcolor{black}{\tiny\ttfamily\strut  nuclear}}\hskip 0pt plus 1pt\relax\allowbreak{}\colorbox[HTML]{F8F8FF}{\textcolor{black}{\tiny\ttfamily\strut  deter}}\hskip 0pt plus 1pt\relax\allowbreak{}\colorbox[HTML]{FFFFFF}{\textcolor{black}{\tiny\ttfamily\strut rence}}\hskip 0pt plus 1pt\relax\allowbreak{}\colorbox[HTML]{F5F5FF}{\textcolor{black}{\tiny\ttfamily\strut  is}}\hskip 0pt plus 1pt\relax\allowbreak{}\colorbox[HTML]{EBEBFF}{\textcolor{black}{\tiny\ttfamily\strut  crucial}}\hskip 0pt plus 1pt\relax\allowbreak{}\colorbox[HTML]{FBFBFF}{\textcolor{black}{\tiny\ttfamily\strut  for}}\hskip 0pt plus 1pt\relax\allowbreak{}\colorbox[HTML]{FFFFFF}{\textcolor{black}{\tiny\ttfamily\strut  understanding}}\hskip 0pt plus 1pt\relax\allowbreak{}\colorbox[HTML]{EDEDFF}{\textcolor{black}{\tiny\ttfamily\strut  the}}\hskip 0pt plus 1pt\relax\allowbreak{}\colorbox[HTML]{FDFDFF}{\textcolor{black}{\tiny\ttfamily\strut  dynamics}}\hskip 0pt plus 1pt\relax\allowbreak{}\colorbox[HTML]{FFFFFF}{\textcolor{black}{\tiny\ttfamily\strut  of}}\hskip 0pt plus 1pt\relax\allowbreak{}\colorbox[HTML]{FFFFFF}{\textcolor{black}{\tiny\ttfamily\strut  international}}\hskip 0pt plus 1pt\relax\allowbreak{}\colorbox[HTML]{FFF9F9}{\textcolor{black}{\tiny\ttfamily\strut  relations}}\hskip 0pt plus 1pt\relax\allowbreak{}\colorbox[HTML]{D6D6FF}{\textcolor{black}{\tiny\ttfamily\strut  during}}\hskip 0pt plus 1pt\relax\allowbreak{}\colorbox[HTML]{FEFEFF}{\textcolor{black}{\tiny\ttfamily\strut  the}}\hskip 0pt plus 1pt\relax\allowbreak{}\colorbox[HTML]{FEFEFF}{\textcolor{black}{\tiny\ttfamily\strut  Cold}}\hskip 0pt plus 1pt\relax\allowbreak{}\colorbox[HTML]{FFFFFF}{\textcolor{black}{\tiny\ttfamily\strut  War}}\hskip 0pt plus 1pt\relax\allowbreak{}\colorbox[HTML]{FFFAFA}{\textcolor{black}{\tiny\ttfamily\strut .\textcolor{gray}{$\hookleftarrow$}\textcolor{gray}{$\hookleftarrow$}}}\hskip 0pt plus 1pt\relax\allowbreak{}\colorbox[HTML]{C5C5FF}{\textcolor{black}{\tiny\ttfamily\strut On}}\hskip 0pt plus 1pt\relax\allowbreak{}\colorbox[HTML]{FFC3C3}{\textcolor{black}{\tiny\ttfamily\strut  the}}\hskip 0pt plus 1pt\relax\allowbreak{}\colorbox[HTML]{FAFAFF}{\textcolor{black}{\tiny\ttfamily\strut  other}}\hskip 0pt plus 1pt\relax\allowbreak{}\colorbox[HTML]{FFFFFF}{\textcolor{black}{\tiny\ttfamily\strut  hand}}\hskip 0pt plus 1pt\relax\allowbreak{}\colorbox[HTML]{FFFFFF}{\textcolor{black}{\tiny\ttfamily\strut ,}}\hskip 0pt plus 1pt\relax\allowbreak{}\colorbox[HTML]{4040FF}{\textcolor{white}{\tiny\ttfamily\strut  exploring}}\hskip 0pt plus 1pt\relax\allowbreak{}\colorbox[HTML]{FFDBDB}{\textcolor{black}{\tiny\ttfamily\strut  other}}\hskip 0pt plus 1pt\relax\allowbreak{}\colorbox[HTML]{FFFEFE}{\textcolor{black}{\tiny\ttfamily\strut  professors}}\hskip 0pt plus 1pt\relax\allowbreak{}\colorbox[HTML]{FFE7E7}{\textcolor{black}{\tiny\ttfamily\strut '}}\hskip 0pt plus 1pt\relax\allowbreak{}\colorbox[HTML]{FFF0F0}{\textcolor{black}{\tiny\ttfamily\strut  work}}\hskip 0pt plus 1pt\relax\allowbreak{}\colorbox[HTML]{FFCDCD}{\textcolor{black}{\tiny\ttfamily\strut  in}}\hskip 0pt plus 1pt\relax\allowbreak{}\colorbox[HTML]{F0F0FF}{\textcolor{black}{\tiny\ttfamily\strut  the}}\hskip 0pt plus 1pt\relax\allowbreak{}\colorbox[HTML]{FFFDFD}{\textcolor{black}{\tiny\ttfamily\strut  areas}}\hskip 0pt plus 1pt\relax\allowbreak{}\colorbox[HTML]{FFFFFF}{\textcolor{black}{\tiny\ttfamily\strut  of}}\hskip 0pt plus 1pt\relax\allowbreak{}\colorbox[HTML]{FFFFFF}{\textcolor{black}{\tiny\ttfamily\strut  land}}\hskip 0pt plus 1pt\relax\allowbreak{}\colorbox[HTML]{FFFFFF}{\textcolor{black}{\tiny\ttfamily\strut  reform}}\hskip 0pt plus 1pt\relax\allowbreak{}\colorbox[HTML]{FF9B9B}{\textcolor{black}{\tiny\ttfamily\strut  or}}\hskip 0pt plus 1pt\relax\allowbreak{}\colorbox[HTML]{FFFFFF}{\textcolor{black}{\tiny\ttfamily\strut  international}}\hskip 0pt plus 1pt\relax\allowbreak{}\colorbox[HTML]{FFFFFF}{\textcolor{black}{\tiny\ttfamily\strut  trade}}\hskip 0pt plus 1pt\relax\allowbreak{}\colorbox[HTML]{FFFFFF}{\textcolor{black}{\tiny\ttfamily\strut  politics}}\hskip 0pt plus 1pt\relax\allowbreak{}\colorbox[HTML]{FFD0D0}{\textcolor{black}{\tiny\ttfamily\strut  can}}\hskip 0pt plus 1pt\relax\allowbreak{}\colorbox[HTML]{FFFFFF}{\textcolor{black}{\tiny\ttfamily\strut  provide}}\hskip 0pt plus 1pt\relax\allowbreak{}\colorbox[HTML]{FFF4F4}{\textcolor{black}{\tiny\ttfamily\strut  a}}\hskip 0pt plus 1pt\relax\allowbreak{}\colorbox[HTML]{FFF8F8}{\textcolor{black}{\tiny\ttfamily\strut  broader}}\hskip 0pt plus 1pt\relax\allowbreak{}\colorbox[HTML]{FFF3F3}{\textcolor{black}{\tiny\ttfamily\strut  perspective}}\hskip 0pt plus 1pt\relax\allowbreak{}\colorbox[HTML]{FFFCFC}{\textcolor{black}{\tiny\ttfamily\strut  on}}\hskip 0pt plus 1pt\relax\allowbreak{}\colorbox[HTML]{E1E1FF}{\textcolor{black}{\tiny\ttfamily\strut  U}}\hskip 0pt plus 1pt\relax\allowbreak{}\colorbox[HTML]{E6E6FF}{\textcolor{black}{\tiny\ttfamily\strut .S}}\hskip 0pt plus 1pt\relax\allowbreak{}\colorbox[HTML]{FFFFFF}{\textcolor{black}{\tiny\ttfamily\strut .}}\hskip 0pt plus 1pt\relax\allowbreak{}\colorbox[HTML]{FFD4D4}{\textcolor{black}{\tiny\ttfamily\strut  institutions}}\hskip 0pt plus 1pt\relax\allowbreak{}\colorbox[HTML]{FFFEFE}{\textcolor{black}{\tiny\ttfamily\strut  and}}\hskip 0pt plus 1pt\relax\allowbreak{}\colorbox[HTML]{FFD2D2}{\textcolor{black}{\tiny\ttfamily\strut  policies}}\hskip 0pt plus 1pt\relax\allowbreak{}\colorbox[HTML]{E0E0FF}{\textcolor{black}{\tiny\ttfamily\strut .}}\hskip 0pt plus 1pt\relax\allowbreak{}\colorbox[HTML]{ABABFF}{\textcolor{black}{\tiny\ttfamily\strut  Prof}}\hskip 0pt plus 1pt\relax\allowbreak{}\colorbox[HTML]{FF7171}{\textcolor{black}{\tiny\ttfamily\strut ess}}\hskip 0pt plus 1pt\relax\allowbreak{}\colorbox[HTML]{FFFFFF}{\textcolor{black}{\tiny\ttfamily\strut ors}}\hskip 0pt plus 1pt\relax\allowbreak{}\colorbox[HTML]{FFEEEE}{\textcolor{black}{\tiny\ttfamily\strut  from}}\hskip 0pt plus 1pt\relax\allowbreak{}\colorbox[HTML]{E3E3FF}{\textcolor{black}{\tiny\ttfamily\strut  the}}\hskip 0pt plus 1pt\relax\allowbreak{}\colorbox[HTML]{9999FF}{\textcolor{black}{\tiny\ttfamily\strut  University}}\hskip 0pt plus 1pt\relax\allowbreak{}\colorbox[HTML]{E7E7FF}{\textcolor{black}{\tiny\ttfamily\strut  of}}\hskip 0pt plus 1pt\relax\allowbreak{}\colorbox[HTML]{FFFFFF}{\textcolor{black}{\tiny\ttfamily\strut  Illinois}}\hskip 0pt plus 1pt\relax\allowbreak{}\colorbox[HTML]{FFB9B9}{\textcolor{black}{\tiny\ttfamily\strut  at}}\hskip 0pt plus 1pt\relax\allowbreak{}\colorbox[HTML]{FFFFFF}{\textcolor{black}{\tiny\ttfamily\strut  Urb}}\hskip 0pt plus 1pt\relax\allowbreak{}\colorbox[HTML]{FFFFFF}{\textcolor{black}{\tiny\ttfamily\strut ana}}\hskip 0pt plus 1pt\relax\allowbreak{}\colorbox[HTML]{FFFFFF}{\textcolor{black}{\tiny\ttfamily\strut -Ch}}\hskip 0pt plus 1pt\relax\allowbreak{}\colorbox[HTML]{FFFFFF}{\textcolor{black}{\tiny\ttfamily\strut ampaign}}\hskip 0pt plus 1pt\relax\allowbreak{}\colorbox[HTML]{F1F1FF}{\textcolor{black}{\tiny\ttfamily\strut  (}}\hskip 0pt plus 1pt\relax\allowbreak{}\colorbox[HTML]{FEFEFF}{\textcolor{black}{\tiny\ttfamily\strut UI}}\hskip 0pt plus 1pt\relax\allowbreak{}\colorbox[HTML]{FFFFFF}{\textcolor{black}{\tiny\ttfamily\strut UC}}\hskip 0pt plus 1pt\relax\allowbreak{}\colorbox[HTML]{FFFFFF}{\textcolor{black}{\tiny\ttfamily\strut )}}\hskip 0pt plus 1pt\relax\allowbreak{}\colorbox[HTML]{FFFDFD}{\textcolor{black}{\tiny\ttfamily\strut  have}}\hskip 0pt plus 1pt\relax\allowbreak{}\colorbox[HTML]{FFFEFE}{\textcolor{black}{\tiny\ttfamily\strut  made}}\hskip 0pt plus 1pt\relax\allowbreak{}\colorbox[HTML]{FFF8F8}{\textcolor{black}{\tiny\ttfamily\strut  substantial}}\hskip 0pt plus 1pt\relax\allowbreak{}\colorbox[HTML]{FFFFFF}{\textcolor{black}{\tiny\ttfamily\strut  contributions}}\hskip 0pt plus 1pt\relax\allowbreak{}\colorbox[HTML]{F8F8FF}{\textcolor{black}{\tiny\ttfamily\strut  to}}\hskip 0pt plus 1pt\relax\allowbreak{}\colorbox[HTML]{FFFFFF}{\textcolor{black}{\tiny\ttfamily\strut  these}}\hskip 0pt plus 1pt\relax\allowbreak{}\colorbox[HTML]{FEFEFF}{\textcolor{black}{\tiny\ttfamily\strut  fields}}\hskip 0pt plus 1pt\relax\allowbreak{}\colorbox[HTML]{7A7AFF}{\textcolor{white}{\tiny\ttfamily\strut .}}\hskip 0pt plus 1pt\relax\allowbreak{}\colorbox[HTML]{FFBCBC}{\textcolor{black}{\tiny\ttfamily\strut  For}}\hskip 0pt plus 1pt\relax\allowbreak{}\colorbox[HTML]{FFFDFD}{\textcolor{black}{\tiny\ttfamily\strut  instance}}\hskip 0pt plus 1pt\relax\allowbreak{}\colorbox[HTML]{FFFFFF}{\textcolor{black}{\tiny\ttfamily\strut ,}}\hskip 0pt plus 1pt\relax\allowbreak{}\colorbox[HTML]{FFD7D7}{\textcolor{black}{\tiny\ttfamily\strut  Prof}}\hskip 0pt plus 1pt\relax\allowbreak{}\colorbox[HTML]{FFFAFA}{\textcolor{black}{\tiny\ttfamily\strut .}}\hskip 0pt plus 1pt\relax\allowbreak{}\colorbox[HTML]{FFC9C9}{\textcolor{black}{\tiny\ttfamily\strut  John}}\hskip 0pt plus 1pt\relax\allowbreak{}\colorbox[HTML]{FFCBCB}{\textcolor{black}{\tiny\ttfamily\strut  Doe}}\hskip 0pt plus 1pt\relax\allowbreak{}\colorbox[HTML]{DEDEFF}{\textcolor{black}{\tiny\ttfamily\strut ,}}\hskip 0pt plus 1pt\relax\allowbreak{}\colorbox[HTML]{FFF9F9}{\textcolor{black}{\tiny\ttfamily\strut  an}}\hskip 0pt plus 1pt\relax\allowbreak{}\colorbox[HTML]{FEFEFF}{\textcolor{black}{\tiny\ttfamily\strut  expert}}\hskip 0pt plus 1pt\relax\allowbreak{}\colorbox[HTML]{FFFFFF}{\textcolor{black}{\tiny\ttfamily\strut  in}}\hskip 0pt plus 1pt\relax\allowbreak{}\colorbox[HTML]{FFEBEB}{\textcolor{black}{\tiny\ttfamily\strut  land}}\hskip 0pt plus 1pt\relax\allowbreak{}\colorbox[HTML]{FFFFFF}{\textcolor{black}{\tiny\ttfamily\strut  reform}}\hskip 0pt plus 1pt\relax\allowbreak{}\colorbox[HTML]{FFFFFF}{\textcolor{black}{\tiny\ttfamily\strut ,}}\hskip 0pt plus 1pt\relax\allowbreak{}\colorbox[HTML]{ECECFF}{\textcolor{black}{\tiny\ttfamily\strut  has}}\hskip 0pt plus 1pt\relax\allowbreak{}\colorbox[HTML]{FFFEFE}{\textcolor{black}{\tiny\ttfamily\strut  published}}\hskip 0pt plus 1pt\relax\allowbreak{}\colorbox[HTML]{FFFEFE}{\textcolor{black}{\tiny\ttfamily\strut  extensively}}\hskip 0pt plus 1pt\relax\allowbreak{}\colorbox[HTML]{FFFFFF}{\textcolor{black}{\tiny\ttfamily\strut  on}}\hskip 0pt plus 1pt\relax\allowbreak{}\colorbox[HTML]{FEFEFF}{\textcolor{black}{\tiny\ttfamily\strut  the}}\hskip 0pt plus 1pt\relax\allowbreak{}\colorbox[HTML]{FFFCFC}{\textcolor{black}{\tiny\ttfamily\strut  history}}\hskip 0pt plus 1pt\relax\allowbreak{}\colorbox[HTML]{FFFFFF}{\textcolor{black}{\tiny\ttfamily\strut  and}}\hskip 0pt plus 1pt\relax\allowbreak{}\colorbox[HTML]{FFFDFD}{\textcolor{black}{\tiny\ttfamily\strut  impacts}}\hskip 0pt plus 1pt\relax\allowbreak{}\colorbox[HTML]{FFFFFF}{\textcolor{black}{\tiny\ttfamily\strut  of}}\hskip 0pt plus 1pt\relax\allowbreak{}\colorbox[HTML]{FFFEFE}{\textcolor{black}{\tiny\ttfamily\strut  land}}\hskip 0pt plus 1pt\relax\allowbreak{}\colorbox[HTML]{FFF6F6}{\textcolor{black}{\tiny\ttfamily\strut  reforms}}\hskip 0pt plus 1pt\relax\allowbreak{}\colorbox[HTML]{FFFFFF}{\textcolor{black}{\tiny\ttfamily\strut  in}}\hskip 0pt plus 1pt\relax\allowbreak{}\colorbox[HTML]{FFF8F8}{\textcolor{black}{\tiny\ttfamily\strut  developing}}\hskip 0pt plus 1pt\relax\allowbreak{}\colorbox[HTML]{FFFAFA}{\textcolor{black}{\tiny\ttfamily\strut  countries}}\hskip 0pt plus 1pt\relax\allowbreak{}\colorbox[HTML]{B8B8FF}{\textcolor{black}{\tiny\ttfamily\strut .}}\hskip 0pt plus 1pt\relax\allowbreak{}\colorbox[HTML]{B4B4FF}{\textcolor{black}{\tiny\ttfamily\strut  His}}\hskip 0pt plus 1pt\relax\allowbreak{}\colorbox[HTML]{F8F8FF}{\textcolor{black}{\tiny\ttfamily\strut  research}}\hskip 0pt plus 1pt\relax\allowbreak{}\colorbox[HTML]{FDFDFF}{\textcolor{black}{\tiny\ttfamily\strut  highlights}}\hskip 0pt plus 1pt\relax\allowbreak{}\colorbox[HTML]{FFFCFC}{\textcolor{black}{\tiny\ttfamily\strut  the}}\hskip 0pt plus 1pt\relax\allowbreak{}\colorbox[HTML]{F3F3FF}{\textcolor{black}{\tiny\ttfamily\strut  role}}\hskip 0pt plus 1pt\relax\allowbreak{}\colorbox[HTML]{FFFFFF}{\textcolor{black}{\tiny\ttfamily\strut  of}}\hskip 0pt plus 1pt\relax\allowbreak{}\colorbox[HTML]{FEFEFF}{\textcolor{black}{\tiny\ttfamily\strut  land}}\hskip 0pt plus 1pt\relax\allowbreak{}\colorbox[HTML]{FFF1F1}{\textcolor{black}{\tiny\ttfamily\strut  reform}}\hskip 0pt plus 1pt\relax\allowbreak{}\colorbox[HTML]{FFFEFE}{\textcolor{black}{\tiny\ttfamily\strut  in}}\hskip 0pt plus 1pt\relax\allowbreak{}\colorbox[HTML]{FFE9E9}{\textcolor{black}{\tiny\ttfamily\strut  economic}}\hskip 0pt plus 1pt\relax\allowbreak{}\colorbox[HTML]{FEFEFF}{\textcolor{black}{\tiny\ttfamily\strut  development}}\hskip 0pt plus 1pt\relax\allowbreak{}\colorbox[HTML]{FFFFFF}{\textcolor{black}{\tiny\ttfamily\strut  and}}\hskip 0pt plus 1pt\relax\allowbreak{}\colorbox[HTML]{FFFFFF}{\textcolor{black}{\tiny\ttfamily\strut  social}}\hskip 0pt plus 1pt\relax\allowbreak{}\colorbox[HTML]{FFF4F4}{\textcolor{black}{\tiny\ttfamily\strut  justice}}\hskip 0pt plus 1pt\relax\allowbreak{}\colorbox[HTML]{FFFCFC}{\textcolor{black}{\tiny\ttfamily\strut .\textcolor{gray}{$\hookleftarrow$}\textcolor{gray}{$\hookleftarrow$}}}\hskip 0pt plus 1pt\relax\allowbreak{}\colorbox[HTML]{8989FF}{\textcolor{black}{\tiny\ttfamily\strut Prof}}\hskip 0pt plus 1pt\relax\allowbreak{}\colorbox[HTML]{FFFFFF}{\textcolor{black}{\tiny\ttfamily\strut .}}\hskip 0pt plus 1pt\relax\allowbreak{}\colorbox[HTML]{FFFBFB}{\textcolor{black}{\tiny\ttfamily\strut  Jane}}\hskip 0pt plus 1pt\relax\allowbreak{}\colorbox[HTML]{FFFFFF}{\textcolor{black}{\tiny\ttfamily\strut  Smith}}\hskip 0pt plus 1pt\relax\allowbreak{}\colorbox[HTML]{FFFCFC}{\textcolor{black}{\tiny\ttfamily\strut ,}}\hskip 0pt plus 1pt\relax\allowbreak{}\colorbox[HTML]{FFF0F0}{\textcolor{black}{\tiny\ttfamily\strut  an}}\hskip 0pt plus 1pt\relax\allowbreak{}\colorbox[HTML]{FFF6F6}{\textcolor{black}{\tiny\ttfamily\strut  authority}}\hskip 0pt plus 1pt\relax\allowbreak{}\colorbox[HTML]{FFF2F2}{\textcolor{black}{\tiny\ttfamily\strut  on}}\hskip 0pt plus 1pt\relax\allowbreak{}\colorbox[HTML]{FFFFFF}{\textcolor{black}{\tiny\ttfamily\strut  international}}\hskip 0pt plus 1pt\relax\allowbreak{}\colorbox[HTML]{FFFFFF}{\textcolor{black}{\tiny\ttfamily\strut  trade}}\hskip 0pt plus 1pt\relax\allowbreak{}\colorbox[HTML]{FFFFFF}{\textcolor{black}{\tiny\ttfamily\strut  politics}}\hskip 0pt plus 1pt\relax\allowbreak{}\colorbox[HTML]{FFFFFF}{\textcolor{black}{\tiny\ttfamily\strut ,}}\hskip 0pt plus 1pt\relax\allowbreak{}\colorbox[HTML]{FBFBFF}{\textcolor{black}{\tiny\ttfamily\strut  has}}\hskip 0pt plus 1pt\relax\allowbreak{}\colorbox[HTML]{FFF7F7}{\textcolor{black}{\tiny\ttfamily\strut  also}}\hskip 0pt plus 1pt\relax\allowbreak{}\colorbox[HTML]{FFECEC}{\textcolor{black}{\tiny\ttfamily\strut  contributed}}\hskip 0pt plus 1pt\relax\allowbreak{}\colorbox[HTML]{FDFDFF}{\textcolor{black}{\tiny\ttfamily\strut  significantly}}\hskip 0pt plus 1pt\relax\allowbreak{}\colorbox[HTML]{F4F4FF}{\textcolor{black}{\tiny\ttfamily\strut  to}}\hskip 0pt plus 1pt\relax\allowbreak{}\colorbox[HTML]{FFF6F6}{\textcolor{black}{\tiny\ttfamily\strut  the}}\hskip 0pt plus 1pt\relax\allowbreak{}\colorbox[HTML]{FFFBFB}{\textcolor{black}{\tiny\ttfamily\strut  academic}}\hskip 0pt plus 1pt\relax\allowbreak{}\colorbox[HTML]{FFFFFF}{\textcolor{black}{\tiny\ttfamily\strut  discourse}}\hskip 0pt plus 1pt\relax\allowbreak{}\colorbox[HTML]{FFF1F1}{\textcolor{black}{\tiny\ttfamily\strut  on}}\hskip 0pt plus 1pt\relax\allowbreak{}\colorbox[HTML]{FFFCFC}{\textcolor{black}{\tiny\ttfamily\strut  global}}\hskip 0pt plus 1pt\relax\allowbreak{}\colorbox[HTML]{F3F3FF}{\textcolor{black}{\tiny\ttfamily\strut  economic}}\hskip 0pt plus 1pt\relax\allowbreak{}\colorbox[HTML]{EFEFFF}{\textcolor{black}{\tiny\ttfamily\strut  systems}}\hskip 0pt plus 1pt\relax\allowbreak{}\colorbox[HTML]{FFF3F3}{\textcolor{black}{\tiny\ttfamily\strut  and}}\hskip 0pt plus 1pt\relax\allowbreak{}\colorbox[HTML]{FFFEFE}{\textcolor{black}{\tiny\ttfamily\strut  international}}\hskip 0pt plus 1pt\relax\allowbreak{}\colorbox[HTML]{FFFFFF}{\textcolor{black}{\tiny\ttfamily\strut  trade}}\hskip 0pt plus 1pt\relax\allowbreak{}\colorbox[HTML]{EFEFFF}{\textcolor{black}{\tiny\ttfamily\strut  policies}}\hskip 0pt plus 1pt\relax\allowbreak{}\colorbox[HTML]{FFFFFF}{\textcolor{black}{\tiny\ttfamily\strut .}}\hskip 0pt plus 1pt\relax\allowbreak{}\colorbox[HTML]{FFFFFF}{\textcolor{black}{\tiny\ttfamily\strut  Her}}\hskip 0pt plus 1pt\relax\allowbreak{}\colorbox[HTML]{FBFBFF}{\textcolor{black}{\tiny\ttfamily\strut  work}}\hskip 0pt plus 1pt\relax\allowbreak{}\colorbox[HTML]{FFF8F8}{\textcolor{black}{\tiny\ttfamily\strut  often}}\hskip 0pt plus 1pt\relax\allowbreak{}\colorbox[HTML]{FFFFFF}{\textcolor{black}{\tiny\ttfamily\strut  focuses}}\hskip 0pt plus 1pt\relax\allowbreak{}\colorbox[HTML]{FFFFFF}{\textcolor{black}{\tiny\ttfamily\strut  on}}\hskip 0pt plus 1pt\relax\allowbreak{}\colorbox[HTML]{FFFEFE}{\textcolor{black}{\tiny\ttfamily\strut  the}}\hskip 0pt plus 1pt\relax\allowbreak{}\colorbox[HTML]{DDDDFF}{\textcolor{black}{\tiny\ttfamily\strut  historical}}\hskip 0pt plus 1pt\relax\allowbreak{}\colorbox[HTML]{F0F0FF}{\textcolor{black}{\tiny\ttfamily\strut  context}}\hskip 0pt plus 1pt\relax\allowbreak{}\colorbox[HTML]{E6E6FF}{\textcolor{black}{\tiny\ttfamily\strut  of}}\hskip 0pt plus 1pt\relax\allowbreak{}\colorbox[HTML]{FFF9F9}{\textcolor{black}{\tiny\ttfamily\strut  international}}\hskip 0pt plus 1pt\relax\allowbreak{}\colorbox[HTML]{FFFFFF}{\textcolor{black}{\tiny\ttfamily\strut  trade}}\hskip 0pt plus 1pt\relax\allowbreak{}\colorbox[HTML]{FDFDFF}{\textcolor{black}{\tiny\ttfamily\strut  agreements}}\hskip 0pt plus 1pt\relax\allowbreak{}\colorbox[HTML]{C3C3FF}{\textcolor{black}{\tiny\ttfamily\strut ,}}\hskip 0pt plus 1pt\relax\allowbreak{}\colorbox[HTML]{F2F2FF}{\textcolor{black}{\tiny\ttfamily\strut  the}}\hskip 0pt plus 1pt\relax\allowbreak{}\colorbox[HTML]{FFFEFE}{\textcolor{black}{\tiny\ttfamily\strut  impact}}\hskip 0pt plus 1pt\relax\allowbreak{}\colorbox[HTML]{FAFAFF}{\textcolor{black}{\tiny\ttfamily\strut  of}}\hskip 0pt plus 1pt\relax\allowbreak{}\colorbox[HTML]{F3F3FF}{\textcolor{black}{\tiny\ttfamily\strut  globalization}}\hskip 0pt plus 1pt\relax\allowbreak{}\colorbox[HTML]{FFFDFD}{\textcolor{black}{\tiny\ttfamily\strut ,}}\hskip 0pt plus 1pt\relax\allowbreak{}\colorbox[HTML]{FFFFFF}{\textcolor{black}{\tiny\ttfamily\strut  and}}\hskip 0pt plus 1pt\relax\allowbreak{}\colorbox[HTML]{FFFEFE}{\textcolor{black}{\tiny\ttfamily\strut  the}}\hskip 0pt plus 1pt\relax\allowbreak{}\colorbox[HTML]{FFFDFD}{\textcolor{black}{\tiny\ttfamily\strut  role}}\hskip 0pt plus 1pt\relax\allowbreak{}\colorbox[HTML]{FFFFFF}{\textcolor{black}{\tiny\ttfamily\strut  of}}\hskip 0pt plus 1pt\relax\allowbreak{}\colorbox[HTML]{FFEFEF}{\textcolor{black}{\tiny\ttfamily\strut  the}}\hskip 0pt plus 1pt\relax\allowbreak{}\colorbox[HTML]{CDCDFF}{\textcolor{black}{\tiny\ttfamily\strut  United}}\hskip 0pt plus 1pt\relax\allowbreak{}\colorbox[HTML]{FFFFFF}{\textcolor{black}{\tiny\ttfamily\strut  States}}\hskip 0pt plus 1pt\relax\allowbreak{}\colorbox[HTML]{FFFFFF}{\textcolor{black}{\tiny\ttfamily\strut  in}}\hskip 0pt plus 1pt\relax\allowbreak{}\colorbox[HTML]{FEFEFF}{\textcolor{black}{\tiny\ttfamily\strut  shaping}}\hskip 0pt plus 1pt\relax\allowbreak{}\colorbox[HTML]{FBFBFF}{\textcolor{black}{\tiny\ttfamily\strut  global}}\hskip 0pt plus 1pt\relax\allowbreak{}\colorbox[HTML]{F8F8FF}{\textcolor{black}{\tiny\ttfamily\strut  economic}}\hskip 0pt plus 1pt\relax\allowbreak{}\colorbox[HTML]{FFFCFC}{\textcolor{black}{\tiny\ttfamily\strut  policies}}\hskip 0pt plus 1pt\relax\allowbreak{}\colorbox[HTML]{FFF9F9}{\textcolor{black}{\tiny\ttfamily\strut .\textcolor{gray}{$\hookleftarrow$}\textcolor{gray}{$\hookleftarrow$}}}\hskip 0pt plus 1pt\relax\allowbreak{}\colorbox[HTML]{4040FF}{\textcolor{white}{\tiny\ttfamily\strut These}}\hskip 0pt plus 1pt\relax\allowbreak{}\colorbox[HTML]{B8B8FF}{\textcolor{black}{\tiny\ttfamily\strut  professors}}\hskip 0pt plus 1pt\relax\allowbreak{}\colorbox[HTML]{FF8C8C}{\textcolor{black}{\tiny\ttfamily\strut ,}}\hskip 0pt plus 1pt\relax\allowbreak{}\colorbox[HTML]{B4B4FF}{\textcolor{black}{\tiny\ttfamily\strut  along}}\hskip 0pt plus 1pt\relax\allowbreak{}\colorbox[HTML]{FEFEFF}{\textcolor{black}{\tiny\ttfamily\strut  with}}\hskip 0pt plus 1pt\relax\allowbreak{}\colorbox[HTML]{ADADFF}{\textcolor{black}{\tiny\ttfamily\strut  Dr}}\hskip 0pt plus 1pt\relax\allowbreak{}\colorbox[HTML]{FFFFFF}{\textcolor{black}{\tiny\ttfamily\strut .}}\hskip 0pt plus 1pt\relax\allowbreak{}\colorbox[HTML]{EDEDFF}{\textcolor{black}{\tiny\ttfamily\strut  Cas}}\hskip 0pt plus 1pt\relax\allowbreak{}\colorbox[HTML]{FFFFFF}{\textcolor{black}{\tiny\ttfamily\strut ler}}\hskip 0pt plus 1pt\relax\allowbreak{}\colorbox[HTML]{FFFFFF}{\textcolor{black}{\tiny\ttfamily\strut ,}}\hskip 0pt plus 1pt\relax\allowbreak{}\colorbox[HTML]{FCFCFF}{\textcolor{black}{\tiny\ttfamily\strut  offer}}\hskip 0pt plus 1pt\relax\allowbreak{}\colorbox[HTML]{DFDFFF}{\textcolor{black}{\tiny\ttfamily\strut  valuable}}\hskip 0pt plus 1pt\relax\allowbreak{}\colorbox[HTML]{FFFFFF}{\textcolor{black}{\tiny\ttfamily\strut  insights}}\hskip 0pt plus 1pt\relax\allowbreak{}\colorbox[HTML]{FFFFFF}{\textcolor{black}{\tiny\ttfamily\strut  into}}\hskip 0pt plus 1pt\relax\allowbreak{}\colorbox[HTML]{FFE5E5}{\textcolor{black}{\tiny\ttfamily\strut  the}}\hskip 0pt plus 1pt\relax\allowbreak{}\colorbox[HTML]{FFEAEA}{\textcolor{black}{\tiny\ttfamily\strut  historical}}\hskip 0pt plus 1pt\relax\allowbreak{}\colorbox[HTML]{FFFEFE}{\textcolor{black}{\tiny\ttfamily\strut  and}}\hskip 0pt plus 1pt\relax\allowbreak{}\colorbox[HTML]{FDFDFF}{\textcolor{black}{\tiny\ttfamily\strut  contemporary}}\hskip 0pt plus 1pt\relax\allowbreak{}\colorbox[HTML]{FFF8F8}{\textcolor{black}{\tiny\ttfamily\strut  aspects}}\hskip 0pt plus 1pt\relax\allowbreak{}\colorbox[HTML]{FFFFFF}{\textcolor{black}{\tiny\ttfamily\strut  of}}\hskip 0pt plus 1pt\relax\allowbreak{}\colorbox[HTML]{FFF1F1}{\textcolor{black}{\tiny\ttfamily\strut  nuclear}}\hskip 0pt plus 1pt\relax\allowbreak{}\colorbox[HTML]{FFFFFF}{\textcolor{black}{\tiny\ttfamily\strut  disarm}}\hskip 0pt plus 1pt\relax\allowbreak{}\colorbox[HTML]{FFFFFF}{\textcolor{black}{\tiny\ttfamily\strut ament}}\hskip 0pt plus 1pt\relax\allowbreak{}\colorbox[HTML]{FDFDFF}{\textcolor{black}{\tiny\ttfamily\strut ,}}\hskip 0pt plus 1pt\relax\allowbreak{}\colorbox[HTML]{FFFFFF}{\textcolor{black}{\tiny\ttfamily\strut  land}}\hskip 0pt plus 1pt\relax\allowbreak{}\colorbox[HTML]{FFFFFF}{\textcolor{black}{\tiny\ttfamily\strut  reform}}\hskip 0pt plus 1pt\relax\allowbreak{}\colorbox[HTML]{FFFFFF}{\textcolor{black}{\tiny\ttfamily\strut ,}}\hskip 0pt plus 1pt\relax\allowbreak{}\colorbox[HTML]{FFFFFF}{\textcolor{black}{\tiny\ttfamily\strut  and}}\hskip 0pt plus 1pt\relax\allowbreak{}\colorbox[HTML]{FFFFFF}{\textcolor{black}{\tiny\ttfamily\strut  international}}\hskip 0pt plus 1pt\relax\allowbreak{}\colorbox[HTML]{FFFFFF}{\textcolor{black}{\tiny\ttfamily\strut  trade}}\hskip 0pt plus 1pt\relax\allowbreak{}\colorbox[HTML]{FEFEFF}{\textcolor{black}{\tiny\ttfamily\strut  politics}}\hskip 0pt plus 1pt\relax\allowbreak{}\colorbox[HTML]{5353FF}{\textcolor{white}{\tiny\ttfamily\strut .}}\hskip 0pt plus 1pt\relax\allowbreak{}\colorbox[HTML]{4040FF}{\textcolor{white}{\tiny\ttfamily\strut  Their}}\hskip 0pt plus 1pt\relax\allowbreak{}\colorbox[HTML]{F7F7FF}{\textcolor{black}{\tiny\ttfamily\strut  research}}\hskip 0pt plus 1pt\relax\allowbreak{}\colorbox[HTML]{D1D1FF}{\textcolor{black}{\tiny\ttfamily\strut  underscores}}\hskip 0pt plus 1pt\relax\allowbreak{}\colorbox[HTML]{FFFFFF}{\textcolor{black}{\tiny\ttfamily\strut  the}}\hskip 0pt plus 1pt\relax\allowbreak{}\colorbox[HTML]{FFFEFE}{\textcolor{black}{\tiny\ttfamily\strut  importance}}\hskip 0pt plus 1pt\relax\allowbreak{}\colorbox[HTML]{FFFFFF}{\textcolor{black}{\tiny\ttfamily\strut  of}}\hskip 0pt plus 1pt\relax\allowbreak{}\colorbox[HTML]{FBFBFF}{\textcolor{black}{\tiny\ttfamily\strut  understanding}}\hskip 0pt plus 1pt\relax\allowbreak{}\colorbox[HTML]{FBFBFF}{\textcolor{black}{\tiny\ttfamily\strut  the}}\hskip 0pt plus 1pt\relax\allowbreak{}\colorbox[HTML]{FFD8D8}{\textcolor{black}{\tiny\ttfamily\strut  past}}\hskip 0pt plus 1pt\relax\allowbreak{}\colorbox[HTML]{FFFFFF}{\textcolor{black}{\tiny\ttfamily\strut  to}}\hskip 0pt plus 1pt\relax\allowbreak{}\colorbox[HTML]{F5F5FF}{\textcolor{black}{\tiny\ttfamily\strut  inform}}\hskip 0pt plus 1pt\relax\allowbreak{}\colorbox[HTML]{FFF8F8}{\textcolor{black}{\tiny\ttfamily\strut  future}}\hskip 0pt plus 1pt\relax\allowbreak{}\colorbox[HTML]{FFFCFC}{\textcolor{black}{\tiny\ttfamily\strut  policies}}\hskip 0pt plus 1pt\relax\allowbreak{}\colorbox[HTML]{F6F6FF}{\textcolor{black}{\tiny\ttfamily\strut  and}}\hskip 0pt plus 1pt\relax\allowbreak{}\colorbox[HTML]{F8F8FF}{\textcolor{black}{\tiny\ttfamily\strut  actions}}\hskip 0pt plus 1pt\relax\allowbreak{}\colorbox[HTML]{8A8AFF}{\textcolor{black}{\tiny\ttfamily\strut .}}\hskip 0pt plus 1pt\relax\allowbreak{}\colorbox[HTML]{4040FF}{\textcolor{white}{\tiny\ttfamily\strut \textless{}\textbar{}im\_end\textbar{}\textgreater{}}}\hskip 0pt plus 1pt\relax\allowbreak{}
\par\medskip
\noindent\textbf{\textcolor[HTML]{2257bf}{LA}} {\scriptsize $|\delta|_{\max}{=}9.9$}: \\[2pt]
\sloppy
\colorbox[HTML]{FF4040}{\textcolor{white}{\tiny\ttfamily\strut \#}}\hskip 0pt plus 1pt\relax\allowbreak{}\colorbox[HTML]{FF7F7F}{\textcolor{black}{\tiny\ttfamily\strut  Nuclear}}\hskip 0pt plus 1pt\relax\allowbreak{}\colorbox[HTML]{FFFFFF}{\textcolor{black}{\tiny\ttfamily\strut  Dis}}\hskip 0pt plus 1pt\relax\allowbreak{}\colorbox[HTML]{FFFFFF}{\textcolor{black}{\tiny\ttfamily\strut arm}}\hskip 0pt plus 1pt\relax\allowbreak{}\colorbox[HTML]{FFFFFF}{\textcolor{black}{\tiny\ttfamily\strut ament}}\hskip 0pt plus 1pt\relax\allowbreak{}\colorbox[HTML]{EDEDFF}{\textcolor{black}{\tiny\ttfamily\strut  Research}}\hskip 0pt plus 1pt\relax\allowbreak{}\colorbox[HTML]{FFA3A3}{\textcolor{black}{\tiny\ttfamily\strut \textcolor{gray}{$\hookleftarrow$}\textcolor{gray}{$\hookleftarrow$}}}\hskip 0pt plus 1pt\relax\allowbreak{}\colorbox[HTML]{4040FF}{\textcolor{white}{\tiny\ttfamily\strut Dr}}\hskip 0pt plus 1pt\relax\allowbreak{}\colorbox[HTML]{FFFFFF}{\textcolor{black}{\tiny\ttfamily\strut .}}\hskip 0pt plus 1pt\relax\allowbreak{}\colorbox[HTML]{CDCDFF}{\textcolor{black}{\tiny\ttfamily\strut  Don}}\hskip 0pt plus 1pt\relax\allowbreak{}\colorbox[HTML]{FEFEFF}{\textcolor{black}{\tiny\ttfamily\strut  Cas}}\hskip 0pt plus 1pt\relax\allowbreak{}\colorbox[HTML]{FFFFFF}{\textcolor{black}{\tiny\ttfamily\strut ler}}\hskip 0pt plus 1pt\relax\allowbreak{}\colorbox[HTML]{A1A1FF}{\textcolor{black}{\tiny\ttfamily\strut 's}}\hskip 0pt plus 1pt\relax\allowbreak{}\colorbox[HTML]{6D6DFF}{\textcolor{white}{\tiny\ttfamily\strut  work}}\hskip 0pt plus 1pt\relax\allowbreak{}\colorbox[HTML]{ADADFF}{\textcolor{black}{\tiny\ttfamily\strut  in}}\hskip 0pt plus 1pt\relax\allowbreak{}\colorbox[HTML]{FFFCFC}{\textcolor{black}{\tiny\ttfamily\strut  nuclear}}\hskip 0pt plus 1pt\relax\allowbreak{}\colorbox[HTML]{FFFFFF}{\textcolor{black}{\tiny\ttfamily\strut  disarm}}\hskip 0pt plus 1pt\relax\allowbreak{}\colorbox[HTML]{FFFFFF}{\textcolor{black}{\tiny\ttfamily\strut ament}}\hskip 0pt plus 1pt\relax\allowbreak{}\colorbox[HTML]{FBFBFF}{\textcolor{black}{\tiny\ttfamily\strut  is}}\hskip 0pt plus 1pt\relax\allowbreak{}\colorbox[HTML]{F0F0FF}{\textcolor{black}{\tiny\ttfamily\strut  fascinating}}\hskip 0pt plus 1pt\relax\allowbreak{}\colorbox[HTML]{9090FF}{\textcolor{black}{\tiny\ttfamily\strut  and}}\hskip 0pt plus 1pt\relax\allowbreak{}\colorbox[HTML]{E9E9FF}{\textcolor{black}{\tiny\ttfamily\strut  has}}\hskip 0pt plus 1pt\relax\allowbreak{}\colorbox[HTML]{FFE6E6}{\textcolor{black}{\tiny\ttfamily\strut  been}}\hskip 0pt plus 1pt\relax\allowbreak{}\colorbox[HTML]{FFFEFE}{\textcolor{black}{\tiny\ttfamily\strut  a}}\hskip 0pt plus 1pt\relax\allowbreak{}\colorbox[HTML]{F7F7FF}{\textcolor{black}{\tiny\ttfamily\strut  significant}}\hskip 0pt plus 1pt\relax\allowbreak{}\colorbox[HTML]{FFE3E3}{\textcolor{black}{\tiny\ttfamily\strut  contribution}}\hskip 0pt plus 1pt\relax\allowbreak{}\colorbox[HTML]{FDFDFF}{\textcolor{black}{\tiny\ttfamily\strut  to}}\hskip 0pt plus 1pt\relax\allowbreak{}\colorbox[HTML]{FCFCFF}{\textcolor{black}{\tiny\ttfamily\strut  the}}\hskip 0pt plus 1pt\relax\allowbreak{}\colorbox[HTML]{FFFEFE}{\textcolor{black}{\tiny\ttfamily\strut  field}}\hskip 0pt plus 1pt\relax\allowbreak{}\colorbox[HTML]{FFFDFD}{\textcolor{black}{\tiny\ttfamily\strut .}}\hskip 0pt plus 1pt\relax\allowbreak{}\colorbox[HTML]{9191FF}{\textcolor{black}{\tiny\ttfamily\strut  His}}\hskip 0pt plus 1pt\relax\allowbreak{}\colorbox[HTML]{F4F4FF}{\textcolor{black}{\tiny\ttfamily\strut  research}}\hskip 0pt plus 1pt\relax\allowbreak{}\colorbox[HTML]{FFA9A9}{\textcolor{black}{\tiny\ttfamily\strut  emphasizes}}\hskip 0pt plus 1pt\relax\allowbreak{}\colorbox[HTML]{FFFFFF}{\textcolor{black}{\tiny\ttfamily\strut  the}}\hskip 0pt plus 1pt\relax\allowbreak{}\colorbox[HTML]{FFC8C8}{\textcolor{black}{\tiny\ttfamily\strut  concept}}\hskip 0pt plus 1pt\relax\allowbreak{}\colorbox[HTML]{FFFFFF}{\textcolor{black}{\tiny\ttfamily\strut  of}}\hskip 0pt plus 1pt\relax\allowbreak{}\colorbox[HTML]{FFFEFE}{\textcolor{black}{\tiny\ttfamily\strut  mutually}}\hskip 0pt plus 1pt\relax\allowbreak{}\colorbox[HTML]{FFFFFF}{\textcolor{black}{\tiny\ttfamily\strut  assured}}\hskip 0pt plus 1pt\relax\allowbreak{}\colorbox[HTML]{FFFFFF}{\textcolor{black}{\tiny\ttfamily\strut  destruction}}\hskip 0pt plus 1pt\relax\allowbreak{}\colorbox[HTML]{FF7878}{\textcolor{black}{\tiny\ttfamily\strut ,}}\hskip 0pt plus 1pt\relax\allowbreak{}\colorbox[HTML]{FFFFFF}{\textcolor{black}{\tiny\ttfamily\strut  which}}\hskip 0pt plus 1pt\relax\allowbreak{}\colorbox[HTML]{FFF6F6}{\textcolor{black}{\tiny\ttfamily\strut  serves}}\hskip 0pt plus 1pt\relax\allowbreak{}\colorbox[HTML]{FFFFFF}{\textcolor{black}{\tiny\ttfamily\strut  as}}\hskip 0pt plus 1pt\relax\allowbreak{}\colorbox[HTML]{FFFFFF}{\textcolor{black}{\tiny\ttfamily\strut  a}}\hskip 0pt plus 1pt\relax\allowbreak{}\colorbox[HTML]{F4F4FF}{\textcolor{black}{\tiny\ttfamily\strut  powerful}}\hskip 0pt plus 1pt\relax\allowbreak{}\colorbox[HTML]{FFFFFF}{\textcolor{black}{\tiny\ttfamily\strut  deterrent}}\hskip 0pt plus 1pt\relax\allowbreak{}\colorbox[HTML]{FDFDFF}{\textcolor{black}{\tiny\ttfamily\strut  against}}\hskip 0pt plus 1pt\relax\allowbreak{}\colorbox[HTML]{FFE7E7}{\textcolor{black}{\tiny\ttfamily\strut  armed}}\hskip 0pt plus 1pt\relax\allowbreak{}\colorbox[HTML]{FFFFFF}{\textcolor{black}{\tiny\ttfamily\strut  conflict}}\hskip 0pt plus 1pt\relax\allowbreak{}\colorbox[HTML]{F8F8FF}{\textcolor{black}{\tiny\ttfamily\strut .}}\hskip 0pt plus 1pt\relax\allowbreak{}\colorbox[HTML]{FFE8E8}{\textcolor{black}{\tiny\ttfamily\strut  Cas}}\hskip 0pt plus 1pt\relax\allowbreak{}\colorbox[HTML]{FFFFFF}{\textcolor{black}{\tiny\ttfamily\strut ler}}\hskip 0pt plus 1pt\relax\allowbreak{}\colorbox[HTML]{FFFFFF}{\textcolor{black}{\tiny\ttfamily\strut 's}}\hskip 0pt plus 1pt\relax\allowbreak{}\colorbox[HTML]{FFB1B1}{\textcolor{black}{\tiny\ttfamily\strut  essay}}\hskip 0pt plus 1pt\relax\allowbreak{}\colorbox[HTML]{FFEBEB}{\textcolor{black}{\tiny\ttfamily\strut  in}}\hskip 0pt plus 1pt\relax\allowbreak{}\colorbox[HTML]{C5C5FF}{\textcolor{black}{\tiny\ttfamily\strut  AP}}\hskip 0pt plus 1pt\relax\allowbreak{}\colorbox[HTML]{FCFCFF}{\textcolor{black}{\tiny\ttfamily\strut  World}}\hskip 0pt plus 1pt\relax\allowbreak{}\colorbox[HTML]{FFFFFF}{\textcolor{black}{\tiny\ttfamily\strut  History}}\hskip 0pt plus 1pt\relax\allowbreak{}\colorbox[HTML]{FFF9F9}{\textcolor{black}{\tiny\ttfamily\strut  del}}\hskip 0pt plus 1pt\relax\allowbreak{}\colorbox[HTML]{FFFCFC}{\textcolor{black}{\tiny\ttfamily\strut ved}}\hskip 0pt plus 1pt\relax\allowbreak{}\colorbox[HTML]{FDFDFF}{\textcolor{black}{\tiny\ttfamily\strut  into}}\hskip 0pt plus 1pt\relax\allowbreak{}\colorbox[HTML]{E6E6FF}{\textcolor{black}{\tiny\ttfamily\strut  the}}\hskip 0pt plus 1pt\relax\allowbreak{}\colorbox[HTML]{9F9FFF}{\textcolor{black}{\tiny\ttfamily\strut  Cold}}\hskip 0pt plus 1pt\relax\allowbreak{}\colorbox[HTML]{FFFFFF}{\textcolor{black}{\tiny\ttfamily\strut  War}}\hskip 0pt plus 1pt\relax\allowbreak{}\colorbox[HTML]{FFF6F6}{\textcolor{black}{\tiny\ttfamily\strut  era}}\hskip 0pt plus 1pt\relax\allowbreak{}\colorbox[HTML]{FFFAFA}{\textcolor{black}{\tiny\ttfamily\strut ,}}\hskip 0pt plus 1pt\relax\allowbreak{}\colorbox[HTML]{F5F5FF}{\textcolor{black}{\tiny\ttfamily\strut  where}}\hskip 0pt plus 1pt\relax\allowbreak{}\colorbox[HTML]{C9C9FF}{\textcolor{black}{\tiny\ttfamily\strut  the}}\hskip 0pt plus 1pt\relax\allowbreak{}\colorbox[HTML]{FFFCFC}{\textcolor{black}{\tiny\ttfamily\strut  fear}}\hskip 0pt plus 1pt\relax\allowbreak{}\colorbox[HTML]{FFFFFF}{\textcolor{black}{\tiny\ttfamily\strut  of}}\hskip 0pt plus 1pt\relax\allowbreak{}\colorbox[HTML]{FBFBFF}{\textcolor{black}{\tiny\ttfamily\strut  nuclear}}\hskip 0pt plus 1pt\relax\allowbreak{}\colorbox[HTML]{FDFDFF}{\textcolor{black}{\tiny\ttfamily\strut  ann}}\hskip 0pt plus 1pt\relax\allowbreak{}\colorbox[HTML]{FFFFFF}{\textcolor{black}{\tiny\ttfamily\strut ihilation}}\hskip 0pt plus 1pt\relax\allowbreak{}\colorbox[HTML]{FFF5F5}{\textcolor{black}{\tiny\ttfamily\strut  shaped}}\hskip 0pt plus 1pt\relax\allowbreak{}\colorbox[HTML]{FFFCFC}{\textcolor{black}{\tiny\ttfamily\strut  global}}\hskip 0pt plus 1pt\relax\allowbreak{}\colorbox[HTML]{FFFDFD}{\textcolor{black}{\tiny\ttfamily\strut  politics}}\hskip 0pt plus 1pt\relax\allowbreak{}\colorbox[HTML]{C1C1FF}{\textcolor{black}{\tiny\ttfamily\strut .}}\hskip 0pt plus 1pt\relax\allowbreak{}\colorbox[HTML]{9898FF}{\textcolor{black}{\tiny\ttfamily\strut  This}}\hskip 0pt plus 1pt\relax\allowbreak{}\colorbox[HTML]{FFE9E9}{\textcolor{black}{\tiny\ttfamily\strut  perspective}}\hskip 0pt plus 1pt\relax\allowbreak{}\colorbox[HTML]{F4F4FF}{\textcolor{black}{\tiny\ttfamily\strut  on}}\hskip 0pt plus 1pt\relax\allowbreak{}\colorbox[HTML]{FFEFEF}{\textcolor{black}{\tiny\ttfamily\strut  nuclear}}\hskip 0pt plus 1pt\relax\allowbreak{}\colorbox[HTML]{F9F9FF}{\textcolor{black}{\tiny\ttfamily\strut  deter}}\hskip 0pt plus 1pt\relax\allowbreak{}\colorbox[HTML]{FFFFFF}{\textcolor{black}{\tiny\ttfamily\strut rence}}\hskip 0pt plus 1pt\relax\allowbreak{}\colorbox[HTML]{ECECFF}{\textcolor{black}{\tiny\ttfamily\strut  is}}\hskip 0pt plus 1pt\relax\allowbreak{}\colorbox[HTML]{EAEAFF}{\textcolor{black}{\tiny\ttfamily\strut  crucial}}\hskip 0pt plus 1pt\relax\allowbreak{}\colorbox[HTML]{FFF7F7}{\textcolor{black}{\tiny\ttfamily\strut  for}}\hskip 0pt plus 1pt\relax\allowbreak{}\colorbox[HTML]{FFFFFF}{\textcolor{black}{\tiny\ttfamily\strut  understanding}}\hskip 0pt plus 1pt\relax\allowbreak{}\colorbox[HTML]{E8E8FF}{\textcolor{black}{\tiny\ttfamily\strut  the}}\hskip 0pt plus 1pt\relax\allowbreak{}\colorbox[HTML]{FFFEFE}{\textcolor{black}{\tiny\ttfamily\strut  dynamics}}\hskip 0pt plus 1pt\relax\allowbreak{}\colorbox[HTML]{FFFFFF}{\textcolor{black}{\tiny\ttfamily\strut  of}}\hskip 0pt plus 1pt\relax\allowbreak{}\colorbox[HTML]{FFFFFF}{\textcolor{black}{\tiny\ttfamily\strut  international}}\hskip 0pt plus 1pt\relax\allowbreak{}\colorbox[HTML]{FFFFFF}{\textcolor{black}{\tiny\ttfamily\strut  relations}}\hskip 0pt plus 1pt\relax\allowbreak{}\colorbox[HTML]{DDDDFF}{\textcolor{black}{\tiny\ttfamily\strut  during}}\hskip 0pt plus 1pt\relax\allowbreak{}\colorbox[HTML]{F8F8FF}{\textcolor{black}{\tiny\ttfamily\strut  the}}\hskip 0pt plus 1pt\relax\allowbreak{}\colorbox[HTML]{FEFEFF}{\textcolor{black}{\tiny\ttfamily\strut  Cold}}\hskip 0pt plus 1pt\relax\allowbreak{}\colorbox[HTML]{FFFFFF}{\textcolor{black}{\tiny\ttfamily\strut  War}}\hskip 0pt plus 1pt\relax\allowbreak{}\colorbox[HTML]{FFF1F1}{\textcolor{black}{\tiny\ttfamily\strut .\textcolor{gray}{$\hookleftarrow$}\textcolor{gray}{$\hookleftarrow$}}}\hskip 0pt plus 1pt\relax\allowbreak{}\colorbox[HTML]{4040FF}{\textcolor{white}{\tiny\ttfamily\strut On}}\hskip 0pt plus 1pt\relax\allowbreak{}\colorbox[HTML]{B3B3FF}{\textcolor{black}{\tiny\ttfamily\strut  the}}\hskip 0pt plus 1pt\relax\allowbreak{}\colorbox[HTML]{F8F8FF}{\textcolor{black}{\tiny\ttfamily\strut  other}}\hskip 0pt plus 1pt\relax\allowbreak{}\colorbox[HTML]{FFFFFF}{\textcolor{black}{\tiny\ttfamily\strut  hand}}\hskip 0pt plus 1pt\relax\allowbreak{}\colorbox[HTML]{FFFFFF}{\textcolor{black}{\tiny\ttfamily\strut ,}}\hskip 0pt plus 1pt\relax\allowbreak{}\colorbox[HTML]{4040FF}{\textcolor{white}{\tiny\ttfamily\strut  exploring}}\hskip 0pt plus 1pt\relax\allowbreak{}\colorbox[HTML]{FBFBFF}{\textcolor{black}{\tiny\ttfamily\strut  other}}\hskip 0pt plus 1pt\relax\allowbreak{}\colorbox[HTML]{F2F2FF}{\textcolor{black}{\tiny\ttfamily\strut  professors}}\hskip 0pt plus 1pt\relax\allowbreak{}\colorbox[HTML]{FFE8E8}{\textcolor{black}{\tiny\ttfamily\strut '}}\hskip 0pt plus 1pt\relax\allowbreak{}\colorbox[HTML]{FDFDFF}{\textcolor{black}{\tiny\ttfamily\strut  work}}\hskip 0pt plus 1pt\relax\allowbreak{}\colorbox[HTML]{FFF0F0}{\textcolor{black}{\tiny\ttfamily\strut  in}}\hskip 0pt plus 1pt\relax\allowbreak{}\colorbox[HTML]{F0F0FF}{\textcolor{black}{\tiny\ttfamily\strut  the}}\hskip 0pt plus 1pt\relax\allowbreak{}\colorbox[HTML]{FFFFFF}{\textcolor{black}{\tiny\ttfamily\strut  areas}}\hskip 0pt plus 1pt\relax\allowbreak{}\colorbox[HTML]{FFFFFF}{\textcolor{black}{\tiny\ttfamily\strut  of}}\hskip 0pt plus 1pt\relax\allowbreak{}\colorbox[HTML]{FFFFFF}{\textcolor{black}{\tiny\ttfamily\strut  land}}\hskip 0pt plus 1pt\relax\allowbreak{}\colorbox[HTML]{FFFFFF}{\textcolor{black}{\tiny\ttfamily\strut  reform}}\hskip 0pt plus 1pt\relax\allowbreak{}\colorbox[HTML]{FFB6B6}{\textcolor{black}{\tiny\ttfamily\strut  or}}\hskip 0pt plus 1pt\relax\allowbreak{}\colorbox[HTML]{FFFFFF}{\textcolor{black}{\tiny\ttfamily\strut  international}}\hskip 0pt plus 1pt\relax\allowbreak{}\colorbox[HTML]{FFFFFF}{\textcolor{black}{\tiny\ttfamily\strut  trade}}\hskip 0pt plus 1pt\relax\allowbreak{}\colorbox[HTML]{FFFFFF}{\textcolor{black}{\tiny\ttfamily\strut  politics}}\hskip 0pt plus 1pt\relax\allowbreak{}\colorbox[HTML]{FFEBEB}{\textcolor{black}{\tiny\ttfamily\strut  can}}\hskip 0pt plus 1pt\relax\allowbreak{}\colorbox[HTML]{FEFEFF}{\textcolor{black}{\tiny\ttfamily\strut  provide}}\hskip 0pt plus 1pt\relax\allowbreak{}\colorbox[HTML]{FFFFFF}{\textcolor{black}{\tiny\ttfamily\strut  a}}\hskip 0pt plus 1pt\relax\allowbreak{}\colorbox[HTML]{FFFBFB}{\textcolor{black}{\tiny\ttfamily\strut  broader}}\hskip 0pt plus 1pt\relax\allowbreak{}\colorbox[HTML]{FFF3F3}{\textcolor{black}{\tiny\ttfamily\strut  perspective}}\hskip 0pt plus 1pt\relax\allowbreak{}\colorbox[HTML]{F9F9FF}{\textcolor{black}{\tiny\ttfamily\strut  on}}\hskip 0pt plus 1pt\relax\allowbreak{}\colorbox[HTML]{D7D7FF}{\textcolor{black}{\tiny\ttfamily\strut  U}}\hskip 0pt plus 1pt\relax\allowbreak{}\colorbox[HTML]{BBBBFF}{\textcolor{black}{\tiny\ttfamily\strut .S}}\hskip 0pt plus 1pt\relax\allowbreak{}\colorbox[HTML]{FFFFFF}{\textcolor{black}{\tiny\ttfamily\strut .}}\hskip 0pt plus 1pt\relax\allowbreak{}\colorbox[HTML]{FFF5F5}{\textcolor{black}{\tiny\ttfamily\strut  institutions}}\hskip 0pt plus 1pt\relax\allowbreak{}\colorbox[HTML]{F5F5FF}{\textcolor{black}{\tiny\ttfamily\strut  and}}\hskip 0pt plus 1pt\relax\allowbreak{}\colorbox[HTML]{FFD1D1}{\textcolor{black}{\tiny\ttfamily\strut  policies}}\hskip 0pt plus 1pt\relax\allowbreak{}\colorbox[HTML]{EDEDFF}{\textcolor{black}{\tiny\ttfamily\strut .}}\hskip 0pt plus 1pt\relax\allowbreak{}\colorbox[HTML]{FFC9C9}{\textcolor{black}{\tiny\ttfamily\strut  Prof}}\hskip 0pt plus 1pt\relax\allowbreak{}\colorbox[HTML]{F4F4FF}{\textcolor{black}{\tiny\ttfamily\strut ess}}\hskip 0pt plus 1pt\relax\allowbreak{}\colorbox[HTML]{FFFFFF}{\textcolor{black}{\tiny\ttfamily\strut ors}}\hskip 0pt plus 1pt\relax\allowbreak{}\colorbox[HTML]{9D9DFF}{\textcolor{black}{\tiny\ttfamily\strut  from}}\hskip 0pt plus 1pt\relax\allowbreak{}\colorbox[HTML]{E8E8FF}{\textcolor{black}{\tiny\ttfamily\strut  the}}\hskip 0pt plus 1pt\relax\allowbreak{}\colorbox[HTML]{D4D4FF}{\textcolor{black}{\tiny\ttfamily\strut  University}}\hskip 0pt plus 1pt\relax\allowbreak{}\colorbox[HTML]{7878FF}{\textcolor{white}{\tiny\ttfamily\strut  of}}\hskip 0pt plus 1pt\relax\allowbreak{}\colorbox[HTML]{FFFFFF}{\textcolor{black}{\tiny\ttfamily\strut  Illinois}}\hskip 0pt plus 1pt\relax\allowbreak{}\colorbox[HTML]{E8E8FF}{\textcolor{black}{\tiny\ttfamily\strut  at}}\hskip 0pt plus 1pt\relax\allowbreak{}\colorbox[HTML]{FFFFFF}{\textcolor{black}{\tiny\ttfamily\strut  Urb}}\hskip 0pt plus 1pt\relax\allowbreak{}\colorbox[HTML]{FFFFFF}{\textcolor{black}{\tiny\ttfamily\strut ana}}\hskip 0pt plus 1pt\relax\allowbreak{}\colorbox[HTML]{FFFFFF}{\textcolor{black}{\tiny\ttfamily\strut -Ch}}\hskip 0pt plus 1pt\relax\allowbreak{}\colorbox[HTML]{FFFFFF}{\textcolor{black}{\tiny\ttfamily\strut ampaign}}\hskip 0pt plus 1pt\relax\allowbreak{}\colorbox[HTML]{F6F6FF}{\textcolor{black}{\tiny\ttfamily\strut  (}}\hskip 0pt plus 1pt\relax\allowbreak{}\colorbox[HTML]{FEFEFF}{\textcolor{black}{\tiny\ttfamily\strut UI}}\hskip 0pt plus 1pt\relax\allowbreak{}\colorbox[HTML]{FFFFFF}{\textcolor{black}{\tiny\ttfamily\strut UC}}\hskip 0pt plus 1pt\relax\allowbreak{}\colorbox[HTML]{FFFFFF}{\textcolor{black}{\tiny\ttfamily\strut )}}\hskip 0pt plus 1pt\relax\allowbreak{}\colorbox[HTML]{FFFFFF}{\textcolor{black}{\tiny\ttfamily\strut  have}}\hskip 0pt plus 1pt\relax\allowbreak{}\colorbox[HTML]{EFEFFF}{\textcolor{black}{\tiny\ttfamily\strut  made}}\hskip 0pt plus 1pt\relax\allowbreak{}\colorbox[HTML]{FFFEFE}{\textcolor{black}{\tiny\ttfamily\strut  substantial}}\hskip 0pt plus 1pt\relax\allowbreak{}\colorbox[HTML]{FFFFFF}{\textcolor{black}{\tiny\ttfamily\strut  contributions}}\hskip 0pt plus 1pt\relax\allowbreak{}\colorbox[HTML]{F4F4FF}{\textcolor{black}{\tiny\ttfamily\strut  to}}\hskip 0pt plus 1pt\relax\allowbreak{}\colorbox[HTML]{FFFFFF}{\textcolor{black}{\tiny\ttfamily\strut  these}}\hskip 0pt plus 1pt\relax\allowbreak{}\colorbox[HTML]{FFFCFC}{\textcolor{black}{\tiny\ttfamily\strut  fields}}\hskip 0pt plus 1pt\relax\allowbreak{}\colorbox[HTML]{ABABFF}{\textcolor{black}{\tiny\ttfamily\strut .}}\hskip 0pt plus 1pt\relax\allowbreak{}\colorbox[HTML]{FF8E8E}{\textcolor{black}{\tiny\ttfamily\strut  For}}\hskip 0pt plus 1pt\relax\allowbreak{}\colorbox[HTML]{FFFEFE}{\textcolor{black}{\tiny\ttfamily\strut  instance}}\hskip 0pt plus 1pt\relax\allowbreak{}\colorbox[HTML]{FFFEFE}{\textcolor{black}{\tiny\ttfamily\strut ,}}\hskip 0pt plus 1pt\relax\allowbreak{}\colorbox[HTML]{F4F4FF}{\textcolor{black}{\tiny\ttfamily\strut  Prof}}\hskip 0pt plus 1pt\relax\allowbreak{}\colorbox[HTML]{FFC8C8}{\textcolor{black}{\tiny\ttfamily\strut .}}\hskip 0pt plus 1pt\relax\allowbreak{}\colorbox[HTML]{FFEEEE}{\textcolor{black}{\tiny\ttfamily\strut  John}}\hskip 0pt plus 1pt\relax\allowbreak{}\colorbox[HTML]{FF9E9E}{\textcolor{black}{\tiny\ttfamily\strut  Doe}}\hskip 0pt plus 1pt\relax\allowbreak{}\colorbox[HTML]{F4F4FF}{\textcolor{black}{\tiny\ttfamily\strut ,}}\hskip 0pt plus 1pt\relax\allowbreak{}\colorbox[HTML]{FFF4F4}{\textcolor{black}{\tiny\ttfamily\strut  an}}\hskip 0pt plus 1pt\relax\allowbreak{}\colorbox[HTML]{FEFEFF}{\textcolor{black}{\tiny\ttfamily\strut  expert}}\hskip 0pt plus 1pt\relax\allowbreak{}\colorbox[HTML]{FFFFFF}{\textcolor{black}{\tiny\ttfamily\strut  in}}\hskip 0pt plus 1pt\relax\allowbreak{}\colorbox[HTML]{FFFEFE}{\textcolor{black}{\tiny\ttfamily\strut  land}}\hskip 0pt plus 1pt\relax\allowbreak{}\colorbox[HTML]{FFFFFF}{\textcolor{black}{\tiny\ttfamily\strut  reform}}\hskip 0pt plus 1pt\relax\allowbreak{}\colorbox[HTML]{FFFFFF}{\textcolor{black}{\tiny\ttfamily\strut ,}}\hskip 0pt plus 1pt\relax\allowbreak{}\colorbox[HTML]{FF8B8B}{\textcolor{black}{\tiny\ttfamily\strut  has}}\hskip 0pt plus 1pt\relax\allowbreak{}\colorbox[HTML]{FFF9F9}{\textcolor{black}{\tiny\ttfamily\strut  published}}\hskip 0pt plus 1pt\relax\allowbreak{}\colorbox[HTML]{FEFEFF}{\textcolor{black}{\tiny\ttfamily\strut  extensively}}\hskip 0pt plus 1pt\relax\allowbreak{}\colorbox[HTML]{FFFFFF}{\textcolor{black}{\tiny\ttfamily\strut  on}}\hskip 0pt plus 1pt\relax\allowbreak{}\colorbox[HTML]{FCFCFF}{\textcolor{black}{\tiny\ttfamily\strut  the}}\hskip 0pt plus 1pt\relax\allowbreak{}\colorbox[HTML]{D2D2FF}{\textcolor{black}{\tiny\ttfamily\strut  history}}\hskip 0pt plus 1pt\relax\allowbreak{}\colorbox[HTML]{FFFFFF}{\textcolor{black}{\tiny\ttfamily\strut  and}}\hskip 0pt plus 1pt\relax\allowbreak{}\colorbox[HTML]{FFF4F4}{\textcolor{black}{\tiny\ttfamily\strut  impacts}}\hskip 0pt plus 1pt\relax\allowbreak{}\colorbox[HTML]{FFFFFF}{\textcolor{black}{\tiny\ttfamily\strut  of}}\hskip 0pt plus 1pt\relax\allowbreak{}\colorbox[HTML]{FDFDFF}{\textcolor{black}{\tiny\ttfamily\strut  land}}\hskip 0pt plus 1pt\relax\allowbreak{}\colorbox[HTML]{F2F2FF}{\textcolor{black}{\tiny\ttfamily\strut  reforms}}\hskip 0pt plus 1pt\relax\allowbreak{}\colorbox[HTML]{FFFFFF}{\textcolor{black}{\tiny\ttfamily\strut  in}}\hskip 0pt plus 1pt\relax\allowbreak{}\colorbox[HTML]{FFE9E9}{\textcolor{black}{\tiny\ttfamily\strut  developing}}\hskip 0pt plus 1pt\relax\allowbreak{}\colorbox[HTML]{FEFEFF}{\textcolor{black}{\tiny\ttfamily\strut  countries}}\hskip 0pt plus 1pt\relax\allowbreak{}\colorbox[HTML]{E6E6FF}{\textcolor{black}{\tiny\ttfamily\strut .}}\hskip 0pt plus 1pt\relax\allowbreak{}\colorbox[HTML]{FFB2B2}{\textcolor{black}{\tiny\ttfamily\strut  His}}\hskip 0pt plus 1pt\relax\allowbreak{}\colorbox[HTML]{FFEFEF}{\textcolor{black}{\tiny\ttfamily\strut  research}}\hskip 0pt plus 1pt\relax\allowbreak{}\colorbox[HTML]{FFD1D1}{\textcolor{black}{\tiny\ttfamily\strut  highlights}}\hskip 0pt plus 1pt\relax\allowbreak{}\colorbox[HTML]{FFFFFF}{\textcolor{black}{\tiny\ttfamily\strut  the}}\hskip 0pt plus 1pt\relax\allowbreak{}\colorbox[HTML]{FFFFFF}{\textcolor{black}{\tiny\ttfamily\strut  role}}\hskip 0pt plus 1pt\relax\allowbreak{}\colorbox[HTML]{FFFFFF}{\textcolor{black}{\tiny\ttfamily\strut  of}}\hskip 0pt plus 1pt\relax\allowbreak{}\colorbox[HTML]{F3F3FF}{\textcolor{black}{\tiny\ttfamily\strut  land}}\hskip 0pt plus 1pt\relax\allowbreak{}\colorbox[HTML]{ECECFF}{\textcolor{black}{\tiny\ttfamily\strut  reform}}\hskip 0pt plus 1pt\relax\allowbreak{}\colorbox[HTML]{FFFEFE}{\textcolor{black}{\tiny\ttfamily\strut  in}}\hskip 0pt plus 1pt\relax\allowbreak{}\colorbox[HTML]{FFF8F8}{\textcolor{black}{\tiny\ttfamily\strut  economic}}\hskip 0pt plus 1pt\relax\allowbreak{}\colorbox[HTML]{FFFFFF}{\textcolor{black}{\tiny\ttfamily\strut  development}}\hskip 0pt plus 1pt\relax\allowbreak{}\colorbox[HTML]{FFF6F6}{\textcolor{black}{\tiny\ttfamily\strut  and}}\hskip 0pt plus 1pt\relax\allowbreak{}\colorbox[HTML]{FFFEFE}{\textcolor{black}{\tiny\ttfamily\strut  social}}\hskip 0pt plus 1pt\relax\allowbreak{}\colorbox[HTML]{FFFAFA}{\textcolor{black}{\tiny\ttfamily\strut  justice}}\hskip 0pt plus 1pt\relax\allowbreak{}\colorbox[HTML]{FFF7F7}{\textcolor{black}{\tiny\ttfamily\strut .\textcolor{gray}{$\hookleftarrow$}\textcolor{gray}{$\hookleftarrow$}}}\hskip 0pt plus 1pt\relax\allowbreak{}\colorbox[HTML]{FFF4F4}{\textcolor{black}{\tiny\ttfamily\strut Prof}}\hskip 0pt plus 1pt\relax\allowbreak{}\colorbox[HTML]{FFFFFF}{\textcolor{black}{\tiny\ttfamily\strut .}}\hskip 0pt plus 1pt\relax\allowbreak{}\colorbox[HTML]{FDFDFF}{\textcolor{black}{\tiny\ttfamily\strut  Jane}}\hskip 0pt plus 1pt\relax\allowbreak{}\colorbox[HTML]{FFFFFF}{\textcolor{black}{\tiny\ttfamily\strut  Smith}}\hskip 0pt plus 1pt\relax\allowbreak{}\colorbox[HTML]{F7F7FF}{\textcolor{black}{\tiny\ttfamily\strut ,}}\hskip 0pt plus 1pt\relax\allowbreak{}\colorbox[HTML]{F1F1FF}{\textcolor{black}{\tiny\ttfamily\strut  an}}\hskip 0pt plus 1pt\relax\allowbreak{}\colorbox[HTML]{FFDDDD}{\textcolor{black}{\tiny\ttfamily\strut  authority}}\hskip 0pt plus 1pt\relax\allowbreak{}\colorbox[HTML]{FFFFFF}{\textcolor{black}{\tiny\ttfamily\strut  on}}\hskip 0pt plus 1pt\relax\allowbreak{}\colorbox[HTML]{FFFFFF}{\textcolor{black}{\tiny\ttfamily\strut  international}}\hskip 0pt plus 1pt\relax\allowbreak{}\colorbox[HTML]{FFFFFF}{\textcolor{black}{\tiny\ttfamily\strut  trade}}\hskip 0pt plus 1pt\relax\allowbreak{}\colorbox[HTML]{FFFFFF}{\textcolor{black}{\tiny\ttfamily\strut  politics}}\hskip 0pt plus 1pt\relax\allowbreak{}\colorbox[HTML]{FFFFFF}{\textcolor{black}{\tiny\ttfamily\strut ,}}\hskip 0pt plus 1pt\relax\allowbreak{}\colorbox[HTML]{FCFCFF}{\textcolor{black}{\tiny\ttfamily\strut  has}}\hskip 0pt plus 1pt\relax\allowbreak{}\colorbox[HTML]{D7D7FF}{\textcolor{black}{\tiny\ttfamily\strut  also}}\hskip 0pt plus 1pt\relax\allowbreak{}\colorbox[HTML]{FFF2F2}{\textcolor{black}{\tiny\ttfamily\strut  contributed}}\hskip 0pt plus 1pt\relax\allowbreak{}\colorbox[HTML]{FDFDFF}{\textcolor{black}{\tiny\ttfamily\strut  significantly}}\hskip 0pt plus 1pt\relax\allowbreak{}\colorbox[HTML]{EBEBFF}{\textcolor{black}{\tiny\ttfamily\strut  to}}\hskip 0pt plus 1pt\relax\allowbreak{}\colorbox[HTML]{FFFBFB}{\textcolor{black}{\tiny\ttfamily\strut  the}}\hskip 0pt plus 1pt\relax\allowbreak{}\colorbox[HTML]{FFFAFA}{\textcolor{black}{\tiny\ttfamily\strut  academic}}\hskip 0pt plus 1pt\relax\allowbreak{}\colorbox[HTML]{FFFFFF}{\textcolor{black}{\tiny\ttfamily\strut  discourse}}\hskip 0pt plus 1pt\relax\allowbreak{}\colorbox[HTML]{DBDBFF}{\textcolor{black}{\tiny\ttfamily\strut  on}}\hskip 0pt plus 1pt\relax\allowbreak{}\colorbox[HTML]{FCFCFF}{\textcolor{black}{\tiny\ttfamily\strut  global}}\hskip 0pt plus 1pt\relax\allowbreak{}\colorbox[HTML]{F2F2FF}{\textcolor{black}{\tiny\ttfamily\strut  economic}}\hskip 0pt plus 1pt\relax\allowbreak{}\colorbox[HTML]{ECECFF}{\textcolor{black}{\tiny\ttfamily\strut  systems}}\hskip 0pt plus 1pt\relax\allowbreak{}\colorbox[HTML]{DDDDFF}{\textcolor{black}{\tiny\ttfamily\strut  and}}\hskip 0pt plus 1pt\relax\allowbreak{}\colorbox[HTML]{EDEDFF}{\textcolor{black}{\tiny\ttfamily\strut  international}}\hskip 0pt plus 1pt\relax\allowbreak{}\colorbox[HTML]{FDFDFF}{\textcolor{black}{\tiny\ttfamily\strut  trade}}\hskip 0pt plus 1pt\relax\allowbreak{}\colorbox[HTML]{F2F2FF}{\textcolor{black}{\tiny\ttfamily\strut  policies}}\hskip 0pt plus 1pt\relax\allowbreak{}\colorbox[HTML]{FFFFFF}{\textcolor{black}{\tiny\ttfamily\strut .}}\hskip 0pt plus 1pt\relax\allowbreak{}\colorbox[HTML]{FFFFFF}{\textcolor{black}{\tiny\ttfamily\strut  Her}}\hskip 0pt plus 1pt\relax\allowbreak{}\colorbox[HTML]{FDFDFF}{\textcolor{black}{\tiny\ttfamily\strut  work}}\hskip 0pt plus 1pt\relax\allowbreak{}\colorbox[HTML]{D6D6FF}{\textcolor{black}{\tiny\ttfamily\strut  often}}\hskip 0pt plus 1pt\relax\allowbreak{}\colorbox[HTML]{FFFEFE}{\textcolor{black}{\tiny\ttfamily\strut  focuses}}\hskip 0pt plus 1pt\relax\allowbreak{}\colorbox[HTML]{FFFFFF}{\textcolor{black}{\tiny\ttfamily\strut  on}}\hskip 0pt plus 1pt\relax\allowbreak{}\colorbox[HTML]{FFFDFD}{\textcolor{black}{\tiny\ttfamily\strut  the}}\hskip 0pt plus 1pt\relax\allowbreak{}\colorbox[HTML]{D8D8FF}{\textcolor{black}{\tiny\ttfamily\strut  historical}}\hskip 0pt plus 1pt\relax\allowbreak{}\colorbox[HTML]{F0F0FF}{\textcolor{black}{\tiny\ttfamily\strut  context}}\hskip 0pt plus 1pt\relax\allowbreak{}\colorbox[HTML]{E3E3FF}{\textcolor{black}{\tiny\ttfamily\strut  of}}\hskip 0pt plus 1pt\relax\allowbreak{}\colorbox[HTML]{F4F4FF}{\textcolor{black}{\tiny\ttfamily\strut  international}}\hskip 0pt plus 1pt\relax\allowbreak{}\colorbox[HTML]{FFFFFF}{\textcolor{black}{\tiny\ttfamily\strut  trade}}\hskip 0pt plus 1pt\relax\allowbreak{}\colorbox[HTML]{FEFEFF}{\textcolor{black}{\tiny\ttfamily\strut  agreements}}\hskip 0pt plus 1pt\relax\allowbreak{}\colorbox[HTML]{8E8EFF}{\textcolor{black}{\tiny\ttfamily\strut ,}}\hskip 0pt plus 1pt\relax\allowbreak{}\colorbox[HTML]{9A9AFF}{\textcolor{black}{\tiny\ttfamily\strut  the}}\hskip 0pt plus 1pt\relax\allowbreak{}\colorbox[HTML]{FAFAFF}{\textcolor{black}{\tiny\ttfamily\strut  impact}}\hskip 0pt plus 1pt\relax\allowbreak{}\colorbox[HTML]{FAFAFF}{\textcolor{black}{\tiny\ttfamily\strut  of}}\hskip 0pt plus 1pt\relax\allowbreak{}\colorbox[HTML]{FAFAFF}{\textcolor{black}{\tiny\ttfamily\strut  globalization}}\hskip 0pt plus 1pt\relax\allowbreak{}\colorbox[HTML]{FFE4E4}{\textcolor{black}{\tiny\ttfamily\strut ,}}\hskip 0pt plus 1pt\relax\allowbreak{}\colorbox[HTML]{FFFFFF}{\textcolor{black}{\tiny\ttfamily\strut  and}}\hskip 0pt plus 1pt\relax\allowbreak{}\colorbox[HTML]{FEFEFF}{\textcolor{black}{\tiny\ttfamily\strut  the}}\hskip 0pt plus 1pt\relax\allowbreak{}\colorbox[HTML]{FBFBFF}{\textcolor{black}{\tiny\ttfamily\strut  role}}\hskip 0pt plus 1pt\relax\allowbreak{}\colorbox[HTML]{FFFFFF}{\textcolor{black}{\tiny\ttfamily\strut  of}}\hskip 0pt plus 1pt\relax\allowbreak{}\colorbox[HTML]{FFF8F8}{\textcolor{black}{\tiny\ttfamily\strut  the}}\hskip 0pt plus 1pt\relax\allowbreak{}\colorbox[HTML]{E0E0FF}{\textcolor{black}{\tiny\ttfamily\strut  United}}\hskip 0pt plus 1pt\relax\allowbreak{}\colorbox[HTML]{FFFFFF}{\textcolor{black}{\tiny\ttfamily\strut  States}}\hskip 0pt plus 1pt\relax\allowbreak{}\colorbox[HTML]{FFFFFF}{\textcolor{black}{\tiny\ttfamily\strut  in}}\hskip 0pt plus 1pt\relax\allowbreak{}\colorbox[HTML]{F9F9FF}{\textcolor{black}{\tiny\ttfamily\strut  shaping}}\hskip 0pt plus 1pt\relax\allowbreak{}\colorbox[HTML]{FEFEFF}{\textcolor{black}{\tiny\ttfamily\strut  global}}\hskip 0pt plus 1pt\relax\allowbreak{}\colorbox[HTML]{F5F5FF}{\textcolor{black}{\tiny\ttfamily\strut  economic}}\hskip 0pt plus 1pt\relax\allowbreak{}\colorbox[HTML]{FFFDFD}{\textcolor{black}{\tiny\ttfamily\strut  policies}}\hskip 0pt plus 1pt\relax\allowbreak{}\colorbox[HTML]{FFFFFF}{\textcolor{black}{\tiny\ttfamily\strut .\textcolor{gray}{$\hookleftarrow$}\textcolor{gray}{$\hookleftarrow$}}}\hskip 0pt plus 1pt\relax\allowbreak{}\colorbox[HTML]{4040FF}{\textcolor{white}{\tiny\ttfamily\strut These}}\hskip 0pt plus 1pt\relax\allowbreak{}\colorbox[HTML]{DCDCFF}{\textcolor{black}{\tiny\ttfamily\strut  professors}}\hskip 0pt plus 1pt\relax\allowbreak{}\colorbox[HTML]{FFDDDD}{\textcolor{black}{\tiny\ttfamily\strut ,}}\hskip 0pt plus 1pt\relax\allowbreak{}\colorbox[HTML]{E4E4FF}{\textcolor{black}{\tiny\ttfamily\strut  along}}\hskip 0pt plus 1pt\relax\allowbreak{}\colorbox[HTML]{FFFFFF}{\textcolor{black}{\tiny\ttfamily\strut  with}}\hskip 0pt plus 1pt\relax\allowbreak{}\colorbox[HTML]{FFE4E4}{\textcolor{black}{\tiny\ttfamily\strut  Dr}}\hskip 0pt plus 1pt\relax\allowbreak{}\colorbox[HTML]{FFFFFF}{\textcolor{black}{\tiny\ttfamily\strut .}}\hskip 0pt plus 1pt\relax\allowbreak{}\colorbox[HTML]{FBFBFF}{\textcolor{black}{\tiny\ttfamily\strut  Cas}}\hskip 0pt plus 1pt\relax\allowbreak{}\colorbox[HTML]{FFFFFF}{\textcolor{black}{\tiny\ttfamily\strut ler}}\hskip 0pt plus 1pt\relax\allowbreak{}\colorbox[HTML]{FFFFFF}{\textcolor{black}{\tiny\ttfamily\strut ,}}\hskip 0pt plus 1pt\relax\allowbreak{}\colorbox[HTML]{FFFFFF}{\textcolor{black}{\tiny\ttfamily\strut  offer}}\hskip 0pt plus 1pt\relax\allowbreak{}\colorbox[HTML]{E3E3FF}{\textcolor{black}{\tiny\ttfamily\strut  valuable}}\hskip 0pt plus 1pt\relax\allowbreak{}\colorbox[HTML]{FFFFFF}{\textcolor{black}{\tiny\ttfamily\strut  insights}}\hskip 0pt plus 1pt\relax\allowbreak{}\colorbox[HTML]{FFFDFD}{\textcolor{black}{\tiny\ttfamily\strut  into}}\hskip 0pt plus 1pt\relax\allowbreak{}\colorbox[HTML]{FCFCFF}{\textcolor{black}{\tiny\ttfamily\strut  the}}\hskip 0pt plus 1pt\relax\allowbreak{}\colorbox[HTML]{EBEBFF}{\textcolor{black}{\tiny\ttfamily\strut  historical}}\hskip 0pt plus 1pt\relax\allowbreak{}\colorbox[HTML]{FFFEFE}{\textcolor{black}{\tiny\ttfamily\strut  and}}\hskip 0pt plus 1pt\relax\allowbreak{}\colorbox[HTML]{FDFDFF}{\textcolor{black}{\tiny\ttfamily\strut  contemporary}}\hskip 0pt plus 1pt\relax\allowbreak{}\colorbox[HTML]{FFFBFB}{\textcolor{black}{\tiny\ttfamily\strut  aspects}}\hskip 0pt plus 1pt\relax\allowbreak{}\colorbox[HTML]{FFFFFF}{\textcolor{black}{\tiny\ttfamily\strut  of}}\hskip 0pt plus 1pt\relax\allowbreak{}\colorbox[HTML]{C4C4FF}{\textcolor{black}{\tiny\ttfamily\strut  nuclear}}\hskip 0pt plus 1pt\relax\allowbreak{}\colorbox[HTML]{FFFFFF}{\textcolor{black}{\tiny\ttfamily\strut  disarm}}\hskip 0pt plus 1pt\relax\allowbreak{}\colorbox[HTML]{FFFFFF}{\textcolor{black}{\tiny\ttfamily\strut ament}}\hskip 0pt plus 1pt\relax\allowbreak{}\colorbox[HTML]{FFFDFD}{\textcolor{black}{\tiny\ttfamily\strut ,}}\hskip 0pt plus 1pt\relax\allowbreak{}\colorbox[HTML]{FFFFFF}{\textcolor{black}{\tiny\ttfamily\strut  land}}\hskip 0pt plus 1pt\relax\allowbreak{}\colorbox[HTML]{FFFFFF}{\textcolor{black}{\tiny\ttfamily\strut  reform}}\hskip 0pt plus 1pt\relax\allowbreak{}\colorbox[HTML]{FFFFFF}{\textcolor{black}{\tiny\ttfamily\strut ,}}\hskip 0pt plus 1pt\relax\allowbreak{}\colorbox[HTML]{FFFFFF}{\textcolor{black}{\tiny\ttfamily\strut  and}}\hskip 0pt plus 1pt\relax\allowbreak{}\colorbox[HTML]{FFFFFF}{\textcolor{black}{\tiny\ttfamily\strut  international}}\hskip 0pt plus 1pt\relax\allowbreak{}\colorbox[HTML]{FFFFFF}{\textcolor{black}{\tiny\ttfamily\strut  trade}}\hskip 0pt plus 1pt\relax\allowbreak{}\colorbox[HTML]{FFFEFE}{\textcolor{black}{\tiny\ttfamily\strut  politics}}\hskip 0pt plus 1pt\relax\allowbreak{}\colorbox[HTML]{9A9AFF}{\textcolor{black}{\tiny\ttfamily\strut .}}\hskip 0pt plus 1pt\relax\allowbreak{}\colorbox[HTML]{B1B1FF}{\textcolor{black}{\tiny\ttfamily\strut  Their}}\hskip 0pt plus 1pt\relax\allowbreak{}\colorbox[HTML]{FFFCFC}{\textcolor{black}{\tiny\ttfamily\strut  research}}\hskip 0pt plus 1pt\relax\allowbreak{}\colorbox[HTML]{FFA4A4}{\textcolor{black}{\tiny\ttfamily\strut  underscores}}\hskip 0pt plus 1pt\relax\allowbreak{}\colorbox[HTML]{FFFFFF}{\textcolor{black}{\tiny\ttfamily\strut  the}}\hskip 0pt plus 1pt\relax\allowbreak{}\colorbox[HTML]{FFE6E6}{\textcolor{black}{\tiny\ttfamily\strut  importance}}\hskip 0pt plus 1pt\relax\allowbreak{}\colorbox[HTML]{FFFFFF}{\textcolor{black}{\tiny\ttfamily\strut  of}}\hskip 0pt plus 1pt\relax\allowbreak{}\colorbox[HTML]{FDFDFF}{\textcolor{black}{\tiny\ttfamily\strut  understanding}}\hskip 0pt plus 1pt\relax\allowbreak{}\colorbox[HTML]{EBEBFF}{\textcolor{black}{\tiny\ttfamily\strut  the}}\hskip 0pt plus 1pt\relax\allowbreak{}\colorbox[HTML]{FFDCDC}{\textcolor{black}{\tiny\ttfamily\strut  past}}\hskip 0pt plus 1pt\relax\allowbreak{}\colorbox[HTML]{FFFFFF}{\textcolor{black}{\tiny\ttfamily\strut  to}}\hskip 0pt plus 1pt\relax\allowbreak{}\colorbox[HTML]{F8F8FF}{\textcolor{black}{\tiny\ttfamily\strut  inform}}\hskip 0pt plus 1pt\relax\allowbreak{}\colorbox[HTML]{FFE8E8}{\textcolor{black}{\tiny\ttfamily\strut  future}}\hskip 0pt plus 1pt\relax\allowbreak{}\colorbox[HTML]{FFFCFC}{\textcolor{black}{\tiny\ttfamily\strut  policies}}\hskip 0pt plus 1pt\relax\allowbreak{}\colorbox[HTML]{EFEFFF}{\textcolor{black}{\tiny\ttfamily\strut  and}}\hskip 0pt plus 1pt\relax\allowbreak{}\colorbox[HTML]{F1F1FF}{\textcolor{black}{\tiny\ttfamily\strut  actions}}\hskip 0pt plus 1pt\relax\allowbreak{}\colorbox[HTML]{CACAFF}{\textcolor{black}{\tiny\ttfamily\strut .}}\hskip 0pt plus 1pt\relax\allowbreak{}\colorbox[HTML]{FFF7F7}{\textcolor{black}{\tiny\ttfamily\strut \textless{}\textbar{}im\_end\textbar{}\textgreater{}}}\hskip 0pt plus 1pt\relax\allowbreak{}
\par\medskip
\noindent\textit{\small Per-constraint attribution snippets (Section A) show that each constraint's $\delta^i_t$ peaks on tokens semantically aligned with the constraint's content. Full-response coloring (Section B) shows that LOO sum and LA produce visually distinct attribution patterns despite both being 1$\times$T signals.}

\endgroup

\twocolumn

\subsection{Synthesis, selection methodology, and caveats}
\label{app:heatmap:methodology}

The three cases together illustrate a three-step ladder of LOO behavior that LA cannot replicate. In Case~1, LOO localizes constraint satisfaction at the literal token level. In Case~2, LOO localizes constraint satisfaction at the structural region level while LA collapses to a vacuous EOS token. In Case~3, LOO localizes constraint \emph{violations}, identifying the response region where the student failed to comply, a diagnostic capability LA does not have because it sees only the full-vs-empty aggregate. Each rung of the ladder exercises a different aspect of the $K\!\times\!T$ attribution matrix described in Appendix~\ref{app:mobius:kt}, namely literal keyword peaks, regional concentration, and violation-region marking.

The cases were selected from a candidate bank of $169$ HIR-16K samples that satisfied a strict filter on LOO/LA divergence (top-$20$ token Jaccard $\le 0.30$, sign disagreement $\ge 35$\%, and residual $L_1$ at least equal to the $L_1$ magnitude of the LA signal itself). From this filtered pool, $80$ samples were uniformly drawn across constraint count $K \in [5, 10]$ and inspected by four parallel LLM-based evaluation agents, with each agent scoring every candidate on (a) per-constraint semantic alignment of $\delta_{\text{LOO},i}$ peaks and (b) the degree to which LA top tokens depart from the per-constraint winners, with the latter axis weighted $4\times$. The three cases shown above were the highest-scoring representatives of three qualitatively distinct LOO-vs-LA disagreement modes. The remaining high-scoring candidates exhibit similar patterns within these modes (e.g., additional code-domain and prefix-sentence variants).

Two caveats are warranted. First, the visualization is descriptive. It illustrates the reward geometry that the estimators present to the optimizer, not the resulting PPO trajectory or final policy. Second, the strict filter selects for cases of large LOO/LA divergence, and the cases here therefore over-represent the high-disagreement tail of the HIR-16K distribution. On samples where a single constraint dominates the teacher's full-context conditioning, LA and LOO sum partially agree and the visual difference is subtler. Aggregate statistics across the full $15{,}450$-sample evaluation (Table~\ref{tab:mobius_residual}, Figure~\ref{fig:loo_decomp}) provide the corresponding non-cherry-picked summary.

\section{Aggregation Method Ablation}
\label{app:aggregation}

\method{} aggregates the per-constraint signals $\delta_i(t)$ into the token-level shaping term $\delta(t)$ via summation (Eq.~\eqref{eq:delta_agg}). Two natural alternatives are mean aggregation and top-$k$ aggregation. We clarify the relation between mean and sum below, and report an empirical ablation of top-$k{=}1$ (max) and top-$k{=}2$ against the default sum.

\paragraph{Mean aggregation is a rescaled alternative.}
Replacing the sum in Eq.~\eqref{eq:delta_agg} by the mean $\delta_{\mathrm{mean}}(t) = \frac{1}{|C|}\sum_i \delta_i(t)$ divides the unclipped shaping signal for each sample by $|C|$. Under the M\"obius decomposition in \S\ref{app:mobius:lavsloo}, this gives $\delta_{\mathrm{mean}} = \frac{1}{|C|}\sum_{\emptyset\neq T\subseteq C}|T|\,m(T)$, so mean preserves the same relative interaction-order weighting as sum within a sample. It is therefore not ruled out by the decomposition. Because $|C|$ varies across samples, however, mean also changes their relative shaping scale; a fair empirical comparison would require retuning $\lambda$ and potentially the clip threshold $c$. We did not run that additional comparison, so the empirical ablation below is limited to sum and the two top-$k$ variants.

\paragraph{Top-$k$ as a sparsified alternative.}
A natural sparsification of the sum is top-$k$ selection: at each token, keep only the $k$ largest-magnitude per-constraint shifts and discard the rest. The $k{=}1$ case is the max aggregation, $\delta(t) = \mathrm{sign}(\delta_{i^*}(t))\cdot|\delta_{i^*}(t)|$ with $i^* = \arg\max_i |\delta_i(t)|$; the $k{=}2$ case sums the two largest. Sparsified aggregation could in principle improve credit assignment by suppressing weak signal from constraints that are only marginally relevant to a given token, at the cost of discarding the multi-constraint interactions documented in \S\ref{app:mobius:emp}.

\paragraph{Empirical comparison: Sum is the strongest aggregation tested.}
Table~\ref{tab:aggregation_ablation} reports MulDimIF accuracy on Overall and Level-$4$ for Sum, Top-$k{=}2$, and Top-$k{=}1$ under both model pairs, with all other hyperparameters fixed at paper-main values ($\lambda{=}2$, $c{=}5$, $\rho{=}1.0$, $1{,}500$ verl-steps). Sum is the strongest variant on the aggregate Overall metric on both model pairs (Qwen2.5: $+1.2$ over Top-$k{=}2$, $+1.9$ over Top-$k{=}1$; Qwen3: $+2.5$ over Top-$k{=}2$, $+1.8$ over Top-$k{=}1$). On the harder Level-$4$ split, aggregation differences narrow: Top-$k{=}2$ marginally leads on Qwen2.5 ($+0.3$ over Sum), and Sum ties Top-$k{=}1$ on Qwen3 ($63.8$). The sparsified top-$k$ aggregations underperform Sum on the aggregate Overall metric, suggesting that discarding the cardinality-weighted contribution of multi-constraint interactions characterized in \S\ref{app:mobius:lavsloo} outweighs any credit-assignment gain. The summation aggregation we adopt is therefore the strongest of the three choices we tested on the aggregate metric, in addition to being the principled M\"obius weighting; it additionally requires no per-token argmax computation and incorporates contributions from all $|C|$ counterfactual teacher passes the framework already produces for each sample.

\begin{table}[t]
  \centering
  \small
  \setlength{\tabcolsep}{6pt}
  \renewcommand{\arraystretch}{1.1}
  \begin{tabular}{l cc cc}
    \toprule
     & \multicolumn{2}{c}{\textbf{Qwen2.5-1.5B}} & \multicolumn{2}{c}{\textbf{Qwen3-4B}} \\
    \cmidrule(lr){2-3} \cmidrule(lr){4-5}
    \textbf{Aggregation} & \textbf{Overall} & \textbf{Level-4} & \textbf{Overall} & \textbf{Level-4} \\
    \midrule
    \textbf{Sum}   & \textbf{72.9} & 59.4          & \textbf{76.2} & \textbf{63.8} \\
    Top-$k{=}2$              & 71.7          & \textbf{59.7} & 73.7          & 62.2          \\
    Top-$k{=}1$       & 71.0          & 57.9          & 74.4          & \textbf{63.8} \\
    \bottomrule
  \end{tabular}
  \caption{Aggregation method ablation: MulDimIF accuracy (\%) under three aggregation methods for combining the per-constraint signals $\delta_i(t)$ into the token-level shaping term $\delta(t)$. All variants share LOO counterfactual structure and identical hyperparameters ($\lambda=2$, $c=5$, $\rho=1.0$, $1{,}500$ verl-steps); only the aggregation rule differs.}
  \label{tab:aggregation_ablation}
\end{table}

\section{Sensitivity to Shaping Strength and Clip Threshold}
\label{app:lambda_clip}

\method{} introduces two hyperparameters on top of the vanilla sampled-token OPD update: the shaping strength $\lambda$, which scales the counterfactual reward, and the clip threshold $c$, which bounds the aggregate per-token counterfactual shift $\delta(t)=\sum_i\delta_i(t)$. As defined in Eq.~\eqref{eq:delta_clip}, clipping is applied after summation: $\hat{\delta}(t)=\mathrm{clip}(\delta(t),-c,c)$. Consequently, a large aggregate shift may arise either from one extreme per-constraint contrast or from the accumulation of several moderate contrasts. The bound $|\hat{\delta}(t)|\le c$ ensures that the absolute shaping contribution is at most $\lambda c$ per token. To characterize the joint behavior of $\lambda$ and $c$ around the paper-main configuration $(\lambda{=}2.0,\, c{=}5.0)$, we sweep a $14$-cell cross-shaped grid on the Qwen2.5-1.5B / Qwen2.5-7B pair: a full $c$ sweep at $\lambda{=}2.0$ across $c \in \{1, 2, 5, 10, 20\}$, a full $\lambda$ sweep at $c{=}5.0$ across $\lambda \in \{0.5, 1, 2, 4\}$, and additional corner cells. All runs share the same $1{,}500$ verl-step training horizon, training data, and PPO configuration as the main paper; only $\lambda$ and $c$ vary.

Table~\ref{tab:lambda_clip_grid} reports three validation benchmark scores (IFEval/IFBench/MulDimIF) per cell. The OPD baseline reference row and the paper-main cell $(\lambda{=}2.0,\, c{=}5.0)$ follow the main-results evaluation protocol of Sec.~\ref{sec:setup}. Within each benchmark column we mark the best score in \textbf{bold} and the second-best with an \underline{underline}.

\begin{table*}[t]
  \centering
  \small
  \setlength{\tabcolsep}{4pt}
  \renewcommand{\arraystretch}{1.15}
  \begin{tabular}{l c c c c c}
    \toprule
    & $c = 1$ & $c = 2$ & $c = 5$ & $c = 10$ & $c = 20$ \\
    \midrule
    $\lambda = 0.5$ & $71.5\,/\,24.5\,/\,69.7$ & $72.8\,/\,24.7\,/\,70.2$ & $73.7\,/\,25.8\,/\,71.1$ & --- & --- \\
    $\lambda = 1.0$ & $73.0\,/\,24.9\,/\,70.6$ & $73.7\,/\,25.3\,/\,70.9$ & $75.4\,/\,\underline{27.6}\,/\,71.4$ & --- & --- \\
    $\lambda = 2.0$ & $74.3\,/\,25.6\,/\,70.6$ & $73.6\,/\,25.8\,/\,71.3$ & $\mathbf{76.0\,/\,28.2\,/\,72.9}^{\star}$ & $75.2\,/\,25.9\,/\,\underline{71.5}$ & $\underline{75.8}\,/\,27.1\,/\,71.2$ \\
    $\lambda = 4.0$ & --- & --- & $75.7\,/\,26.2\,/\,70.9$ & $74.7\,/\,27.4\,/\,70.2$ & $73.9\,/\,25.5\,/\,69.7$ \\
    \midrule
    \emph{OPD baseline} & \multicolumn{5}{l}{$70.7\,/\,22.8\,/\,69.5$ \quad (reference; same evaluation protocol as the paper-main cell)} \\
    \bottomrule
  \end{tabular}
  \caption{IFEval strict-prompt / IFBench strict-prompt / MulDimIF overall accuracy (\%) for the $(\lambda, c)$ ablation grid on Qwen2.5. \textbf{Bold} marks the best score on each benchmark; \underline{underline} marks the second-best. The paper-main cell $(\lambda{=}2.0,\, c{=}5.0)$, marked $\star$, attains all three column-best scores.}
  \label{tab:lambda_clip_grid}
\end{table*}

\paragraph{A universal positive shift over OPD baseline.}
Every one of the $14$ cells improves on the OPD baseline reference triple $70.7\,/\,22.8\,/\,69.5$ on all three benchmarks. Even the most extreme corner $(\lambda{=}0.5, c{=}1)$ reaches $71.5\,/\,24.5\,/\,69.7$, with $+0.8$ and $+1.7$\,pp gains on IFEval and IFBench while remaining at or above the OPD baseline on MulDimIF overall accuracy; the paper-main cell reaches $76.0\,/\,28.2\,/\,72.9$. The counterfactual signal therefore provides positive return throughout the entire grid; the gain is not contingent on locating a narrow operating point.

\paragraph{Shaping strength traces a unimodal curve with empirical optimum at $\lambda{=}2$.}
Holding the clip threshold fixed at $c{=}5$, all three benchmarks trace a unimodal curve along $\lambda$. IFEval rises $73.7 \to 75.4 \to \mathbf{76.0} \to 75.7$ as $\lambda$ traverses $\{0.5, 1, 2, 4\}$; IFBench follows a more pronounced unimodality $25.8 \to 27.6 \to \mathbf{28.2} \to 26.2$, with a clear $2.0$\,pp drop at $\lambda{=}4$; MulDimIF overall accuracy likewise peaks at $\lambda{=}2$ ($71.1 \to 71.4 \to \mathbf{72.9} \to 70.9$). Because $\lambda$ scales the clipped shaping term $\hat{\delta}(t)$ while leaving the vanilla OPD term unchanged, the decline at $\lambda{=}4$ is consistent with excessive shaping strength on this model pair. Among the tested values, the grid therefore identifies $\lambda{=}2$ as the empirical optimum.

\paragraph{The default clip threshold performs best among the tested values.}
Holding $\lambda{=}2$, the paper-main threshold $c{=}5$ attains the highest score on all three benchmarks among the five tested values. Across $c \in \{1,2,5,10,20\}$, the score ranges are $2.4$\,pp on IFEval, $2.6$\,pp on IFBench, and $2.3$\,pp on MulDimIF overall. For comparison, varying $\lambda$ at $c{=}5$ produces ranges of $2.3$, $2.4$, and $2.0$\,pp on the same benchmarks. Performance therefore varies to a similar extent along both axes, and the effect of $c$ is non-monotonic over the tested values. We retain $c{=}5$ because it is the best tested threshold for this model pair.

\paragraph{The paper-main cell is the joint optimum on a benign neighborhood.}
Aggregating the two axis findings, the paper-main configuration is the only cell that attains the best score on every benchmark simultaneously. The per-benchmark runners-up are $\lambda{=}2,\, c{=}20$ on IFEval ($75.8$), $\lambda{=}1,\, c{=}5$ on IFBench ($27.6$), and $\lambda{=}2,\, c{=}10$ on MulDimIF overall accuracy ($71.5$). All three sit inside the $\lambda \in \{1, 2\} \times c \in \{5, 10, 20\}$ neighborhood that surrounds the paper-main cell. Failure modes appear only at the extreme corners (the tight-clip / weak-shaping corner that under-uses the counterfactual signal, and the high-$\lambda$ row that over-uses it relative to the baseline OPD reward), and even those corners retain positive net gain over OPD baseline on every benchmark. The paper-main configuration $(\lambda{=}2.0,\, c{=}5.0)$ emerged from this sweep as the joint per-benchmark optimum, and we retain it throughout the rest of the paper. Practitioners deploying \method{} can expect comparable behavior at any operating point within the same neighborhood without re-tuning.

\section{Constraint Subsampling: How Many Counterfactual Contexts Are Needed?}
\label{app:ratio_sweep}

\begin{table*}[!t]
  \centering
  \small
  \setlength{\tabcolsep}{4pt}
  \begin{tabular}{l c c c c}
    \toprule
    \textbf{Config} & \textbf{IFEval} & \textbf{IFBench} & \textbf{MulDimIF Overall} & \textbf{MulDimIF L4} \\
    \midrule
    OPD baseline                & $70.7$           & $22.8$           & $69.5$           & $54.9$ \\
    \method{} $\rho{=}0.10$     & $75.1\,(+4.4)$   & $25.6\,(+2.8)$   & $71.3\,(+1.8)$   & $57.3\,(+2.4)$ \\
    \method{} $\rho{=}0.25$     & $74.5\,(+3.8)$   & $26.1\,(+3.3)$   & $72.5\,(+3.0)$   & $58.4\,(+3.5)$ \\
    \method{} $\rho{=}0.50$     & $\mathbf{76.0}\,(+5.3)$ & $26.9\,(+4.1)$   & $71.6\,(+2.1)$   & $58.2\,(+3.3)$ \\
    \method{} $\rho{=}1.00$     & $\mathbf{76.0}\,(+5.3)$ & $\mathbf{28.2}\,(+5.4)$ & $\mathbf{72.9}\,(+3.4)$ & $\mathbf{59.4}\,(+4.5)$ \\
    \bottomrule
  \end{tabular}
  \caption{Cross-benchmark accuracy (\%) for different Qwen2.5 subsampling-ratio \method{} variants and the OPD baseline. The IFEval, IFBench, and MulDimIF Overall entries in the OPD baseline and $\rho{=}1.00$ rows reproduce Table~\ref{tab:main_results}; the $\rho{=}1.00$ MulDimIF L4 entry reproduces Table~\ref{tab:loo_la_difficulty} (3-sample mean under the unified evaluation protocol of \S\ref{sec:setup}). Parentheses are pp gains over the OPD baseline. Best $\rho$ per column is bolded; ties are also bolded.}
  \label{tab:ratio_sweep_xbench}
\end{table*}

\begin{figure*}[!t]
  \centering
  \includegraphics[width=\textwidth]{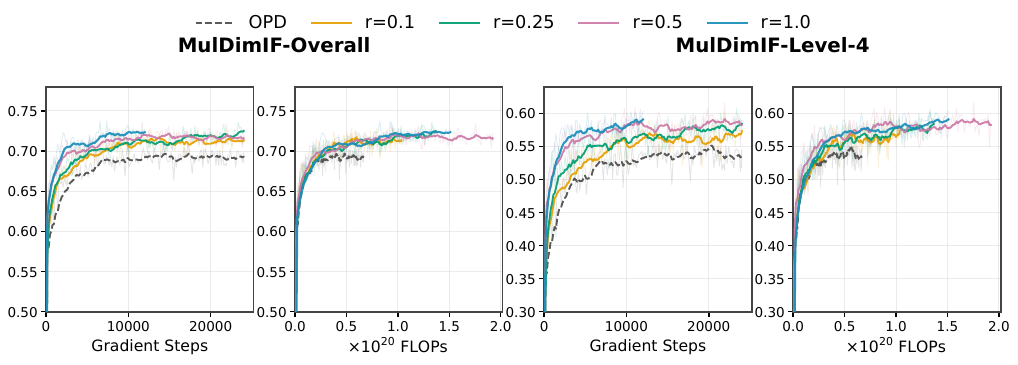}
  \caption{\textbf{Constraint subsampling ratio sweep on Qwen2.5-1.5B.} MulDimIF Overall (columns 1--2) and Level-4 (columns 3--4) accuracy for OPD baseline (gray dashed) and four \method{} configurations with $\rho \in \{0.10, 0.25, 0.50, 1.00\}$. Within each metric, the left column plots accuracy against gradient steps, the right column against cumulative training FLOPs. Curves are EMA-smoothed (weight $0.85$) with the raw trajectory shown as a faint shadow. Higher $\rho$ has higher per-step compute, so the same number of gradient steps consumes more FLOPs; conversely, at the same FLOPs budget, lower $\rho$ has executed more steps.}
  \label{fig:ratio_sweep}
\end{figure*}

The dominant per-sample cost of \method{} is the $|C|+1$ teacher forward passes needed to score the full constraint set $C$ and each leave-one-out variant. A natural way to amortize this cost is to subsample counterfactual contexts: with subsampling ratio $\rho \in (0, 1]$, we retain only $\lceil \rho \, |C| \rceil$ leave-one-out contexts per sample and rescale the summed signal by $|C| / \lceil \rho \, |C| \rceil$ to preserve its expected magnitude. The main paper reports \method{} with $\rho{=}1.00$; this appendix asks whether the dense leave-one-out scheme is necessary, or whether a sparser subsampled signal already captures the gain.

\paragraph{Setup.}
We sweep $\rho \in \{0.10,\, 0.25,\, 0.50,\, 1.00\}$ on the Qwen2.5-1.5B / Qwen2.5-7B pair, holding all other hyperparameters fixed at $\lambda{=}2.0$ and $c{=}5.0$. The vanilla sampled-token OPD baseline is included as a reference. Per-step compute grows with $\rho$ because the number of teacher forward passes per gradient step is approximately proportional to $\rho \, |C|$; the four \method{} configurations therefore have monotonically increasing per-step compute (qualitatively, $\rho{=}1.00$ costs roughly $3\times$ more per step than $\rho{=}0.10$ on the Qwen2.5-1.5B/7B pair). The $\rho < 1.0$ runs use $3{,}000$ verl-steps, whereas the default $\rho{=}1.00$ run uses $1{,}500$. Table~\ref{tab:ratio_sweep_xbench} therefore compares selected checkpoints rather than equal numbers of updates; Figure~\ref{fig:ratio_sweep} reports the complete gradient-step and FLOP trajectories.

\paragraph{Comparison of subsampling-ratio variants across benchmarks.}
Table~\ref{tab:ratio_sweep_xbench} reports the performance of four variants and OPD baseline on four instruction-following benchmarks.

\paragraph{(1) All $\rho > 0$ substantially outperform OPD baseline on every benchmark.}
Even the sparsest configuration $\rho{=}0.10$, which uses less than a full set of leave-one-out contexts per sample on average, lifts the student by $+1.8$\,pp to $+4.4$\,pp across the four benchmarks. The counterfactual signal does not require dense sampling to provide supervision beyond what full-context-only OPD extracts; a partial estimate already breaks the OPD baseline ceiling.

\paragraph{(2) The marginal value of higher $\rho$ is benchmark-stratified.}
On MulDimIF Overall the ratio sweep saturates early: $\rho{=}0.25$ already comes within $0.4$\,pp of $\rho{=}1.00$ ($72.5$ vs.\ $72.9$), and $\rho{=}0.50$ even regresses slightly ($71.6$); the L1/L2/L3 sub-levels are dominated by the student's general capability ceiling, dragging the aggregate flat. IFEval shows a similar saturation pattern, with $\rho{=}0.50$ tying $\rho{=}1.00$ at $76.0$. On the constraint-stressed benchmarks the story is different. IFBench accuracy improves monotonically from $25.6$\,(\!$\rho{=}0.10$\!) to $28.2$\,(\!$\rho{=}1.00$\!), a $+2.6$\,pp gap from the sparsest to the densest configuration, on the same scale as the $+2.8$\,pp gain from OPD baseline to $\rho{=}0.10$ on the same metric. MulDimIF Level-4 follows the same pattern, increasing from $57.3$ at $\rho{=}0.10$ to $59.4$ at $\rho{=}1.00$ (with a minor non-monotone dip at $\rho{=}0.50$); the resulting $+2.1$\,pp gap nearly matches the $+2.4$\,pp OPD baseline-to-$\rho{=}0.10$ jump.

\paragraph{(3) The benchmark stratification recapitulates the main-paper constraint-dilution story.}
The metrics where $\rho$ keeps mattering (IFBench and MulDimIF Level-4) are precisely the constraint-stressed regimes where the main paper argues full-context teacher distributions most dilute per-constraint information. The metrics where $\rho$ saturates early (MulDimIF Overall, dominated by easier sub-levels) are regimes where general instruction-following capability rather than constraint-level discrimination is the bottleneck. Denser counterfactual coverage helps exactly where per-constraint attribution matters most, which is the same physical effect that motivates the method itself, viewed through a different axis. Where the main paper varies constraint count $|C|$ (Level-1 to Level-4), this ablation varies the fraction of constraints whose contribution is explicitly attributed.

\paragraph{The compute--quality picture.}
Figure~\ref{fig:ratio_sweep} plots the five MulDimIF trajectories on both a gradient-step axis (columns 1 and 3) and an absolute-FLOPs axis (columns 2 and 4). The two views correspond to two different deployment questions. Under iso-gradient-step (the left view of each metric), higher $\rho$ trains faster per step. Under iso-FLOPs (the right view of each metric), higher $\rho$ costs more per step, so a lower $\rho$ trained for more steps occupies a different region of the curve. Reading the columns together, no \method{} configuration is Pareto-dominated by OPD baseline on any axis; among \method{} configurations, the choice of $\rho$ trades total compute against quality on the constraint-stressed slices.

\paragraph{Why the main paper uses $\rho{=}1.00$.}
The contribution claim of \method{}, namely that per-constraint counterfactual attribution surfaces supervision a single full-context teacher pass conflates, is most directly demonstrated in the constraint-stressed regimes. These are exactly the metrics on which $\rho{=}1.00$ delivers gains of $+2.6$\,pp on IFBench and $+2.1$\,pp on MulDimIF Level-4 over the $\rho{=}0.10$ configuration. On IFBench in particular, the ``densification'' gap from $\rho{=}0.10$ to $\rho{=}1.00$ is on the same scale as the ``existence'' gap from no counterfactual signal (OPD baseline) to a sparse one ($\rho{=}0.10$). Reporting \method{} at $\rho{=}1.00$ therefore characterizes the upper end of the leave-one-out estimator's quality; sparser counterfactual sampling is a viable compute-saving option for applications that do not bottleneck on constraint-level discrimination, where $\rho \in [0.25, 0.50]$ recovers most of the gain on aggregate metrics like MulDimIF Overall and IFEval at a fraction of the per-step cost.

\end{document}